\documentclass[colophon, final, english]{phduio}
\usepackage{phdstyle}   % Custom style
\usepackage{kantlipsum} % Dummy text
\usepackage{amssymb}
\usepackage{array}
\usepackage{comment}
\usepackage{multirow}
\usepackage{threeparttable,booktabs}
\usepackage{etoolbox}
\usepackage{pifont}
\usepackage{bigints}
\usepackage{booktabs}
\usepackage{wrapfig}
\usepackage{mdframed}

\usepackage{caption} % Figures caption
\newcommand{\etal}{\textit{et al}.}

\newcommand*{\upSmallFrown}{\mathbin{\raisebox{0.9ex}{$\smallfrown$}}}
\DeclareMathOperator*{\argmin}{argmin}
\DeclareUnicodeCharacter{0301}{\'{e}}

\title{Efficient Modelling Across Time of Human Actions and Interactions}
\author{Alexandros Stergiou}
\department{Department}
\faculty{Information and computing sciences}
\affiliation
{
    Optional Second Affiliation
    \and
    Optional Further Specification
}
\ISSN{TBD}           
\dissertationseries{1234}

\makeatletter
\let\blx@rerun@biber\relax
\makeatother

\begin{document}

    \frontmatter        % Folios in Roman numerals, unnumbered chapters.

    \uiotitle

    \chapter{Acknowledgements}

\vspace{3em}

Four year ago I was accepted to work under the supervision of Ronald Poppe and Remco Veltkamp on dyadic human interactions. Over the course of these years, they have given me the opportunity to develop my own research ideas throughout my PhD. I was encouraged to expand upon research problems within the scope of my project and to further grow as a researcher with independent ideas. I would like especially thank Ronald not only for his academic support and overall enthusiasm in my work, but also for his efforts and the lengths that he has gone through for creating a welcoming environment for me and helping me to get where I am right now. Your occasional barging in the office will be missed but hopefully I will be able to replicate it myself in my next steps. 

I would like to warmly thank all of my committee members for the time that they have dedicated as committee members for my thesis.

Thanks to my colleagues, friends and co-authors Georgios Kapidis, Grigorios Kalliatakis and Christos Chrysoulas for their valuable help and collaboration in our joint work on spatio-temporal feature examinations, as well as our discussions during my PhD. Thank you George for being a co-contributor to Chapter 7 and for all the great times that we had both in and outside the office. Gregory and Christos, thank you for including me in your discussions and being part of our own group back in Essex. Our discussions have given me both insights and a better understanding of research and working in academia. I am especially looking forward for our collaboration in future projects. 

Thank you to my mother Glyka and my father Elias for being by my side throughout my life and academic journey. From the first days of my bachelor degree to this day, their support has been extremely valuable. Their sacrifices and hard work has been an inspiration for me to better myself.

Last, but not least, I am thankful for to my partner Deborah for her positivity, patience and unconditional love. She has been a rock to lean on and has made me deeply grateful of unplanned meetings in unanticipated places.

    \chapter{List of Papers}

\section{Included in the thesis}
\fullcite{stergiou2020right}
\\
\\
\fullcite{stergiou2021learn}
\\
\\
\fullcite{stergiou2020learning}
\\
\\
\fullcite{stergiou2019class}
\\
\\
\fullcite{stergiou2019saliency}
\\
\\
\fullcite{stergiou2019analyzing}

\section{Other works}

\fullcite{stergiou2021refining}
\\
\\
\fullcite{stergiou2021mind}
\\
\\
\fullcite{stergiou2019FAST}

    \cleartorecto
    \microtypesetup{protrusion = false}
    \tableofcontents*    % Or \tableofcontents*
    \cleartorecto
    %\listoffigures*      % Or \listoffigures*
    %\cleartorecto
    %\listoftables       % Or \listoftables*
    %\cleartorecto
    \chapter{Samenvatting}

\vspace{3em}

Dit proefschrift richt zich op het analyseren van menselijke handelingen en interacties in video’s. We beginnen met het identificeren van de belangrijkste uitdagingen met betrekking tot actieherkenning in video's en bekijken in hoeverre deze worden aangepakt met de huidige methoden.

Op basis van deze uitdagingen, en door ons te concentreren op het temporele aspect van acties, stellen we dat de huidige spatio-temporele kernels met een vaste grootte in 3D convolutionele neurale netwerken (CNNs) kunnen worden verbeterd om beter om te gaan met temporele variaties in de invoer. Onze bijdragen zijn gebaseerd op het vergroten van de convolutionele receptieve velden door het gebruik van ruimtelijk-temporele segmenten van video's die variëren in grootte, en daarnaast het bepalen van de lokale relevantie van de functie in relatie tot de gehele video. De zo geëxtraheerde functies bevatten informatie die het belang van lokale functies omvat over meerdere tijdsduren, waaronder de gehele video.

Vervolgens bestuderen we hoe we variaties tussen actieklassen beter kunnen aanpakken, door hun karakteristieke patronen over verschillende lagen van de architectuur te versterken. Met de hiërarchische extractie van kenmerken worden variaties van relatief vergelijkbare klassen op dezelfde manier gemodelleerd als zeer verschillende klassen. Daardoor is het minder waarschijnlijk dat onderscheid tussen vergelijkbare klassen effectief wordt gemodelleerd. De
voorgestelde aanpak regulariseert feature maps door features te versterken die overeenkomen met de klasse van de video die wordt bekeken. We stappen af van klasse-agnostische netwerken en doen vroege voorspellingen op basis van dit functieversterkingsmechanisme.

De voorgestelde vernieuwingen zijn geëvalueerd op verschillende benchmark-datasets voor actieherkenning en we laten daar competitieve resultaten op zien. Op het gebied van prestaties concurreren we met de nieuwste algoritmes terwijl onze methoden efficiënter zijn wat betreft het aantal GFLOPs.

Ten slotte presenteren we een voor mensen begrijpelijke benadering die is gericht op het bieden van visuele verklaringen voor functies die zijn geleerd via spatio-temporele netwerken. We isoleren spatio-temporele regio's in 3D-CNN's die informatief zijn voor een actieklasse. We breiden deze aanpak uit om de hele netwerkarchitectuur mogelijk te maken, incrementeel kernels met verschillende complexiteiten te ontdekken en lagen met betrekking tot een specifieke klasse te modelleren.

\clearpage
\newpage

\chapter{Abstract}

This thesis focuses on video understanding for human action and interaction recognition. We start by identifying the main challenges related to action recognition from videos and review how they have been addressed by current methods. 

Based on these challenges, and by focusing on the temporal aspect of actions, we argue that current fixed-sized spatio-temporal kernels in 3D convolutional neural networks (CNNs) can be improved to better deal with temporal variations in the input. Our contributions are based on the enlargement of the convolutional receptive fields through the introduction of spatio-temporal size-varying segments of videos, as well as the discovery of the local feature relevance over the entire video sequence. The resulting extracted features encapsulate information that includes the importance of local features across multiple temporal durations, as well as the entire video sequence. 

Subsequently, we study how we can better handle variations between classes of actions, by enhancing their feature differences over different layers of the architecture. The hierarchical extraction of features models variations of relatively similar classes the same as very dissimilar classes. Therefore, distinctions between similar classes are less likely to be modelled. The proposed approach regularises feature maps by amplifying features that correspond to the class of the video that is processed. We move away from class-agnostic networks and make early predictions based on feature amplification mechanism.

The proposed approaches are evaluated on several benchmark action recognition datasets and  show competitive results. In terms of performance, we compete with the state-of-the-art while being more efficient in terms of GFLOPs. 

Finally, we present a human-understandable approach aimed at providing visual explanations for features learned over spatio-temporal networks. We isolate spatio-temporal regions in 3D-CNNs that are informative for an action class. We extend this approach to allow for the traversal over the entire network architecture, incrementally discovering kernels at different complexities, and modelling layers related to a specific class.
    \cleartorecto
    \microtypesetup{protrusion = true}

    \mainmatter         % Folios in Arabic numerals, numbered chapters.

    \paper              % "Chapter" is renamed "Paper"
    %\paperpage          % Similar to \part*{Papers}, but appears in TOC
    \numberofpapers{8}  % Specify size of thumb indices
    
    \author
{}
\title{Introduction}

\maketitle
\label{ch1}

Many videos depict people. Their actions inform us of their activities, in relation to objects and other people, as well as the cultural and social setting. Research has aimed at automating the recognition of human actions in videos. In this chapter, we provide an overview of the challenges associated with the task of human action recognition from videos. We subsequently present the structure of this thesis by discussing the contents of each chapter.

\section{Challenges in video understanding}
\label{ch:1::sec:intro::sec:challenges}

% Challenges 
Based on how information is processed for images and for temporal sequences in videos, we identify two main groups of challenges.

% Visual variations
The first group of challenges is associated with visual changes. These changes can be based on differences in the environment, in which human actions or interactions are performed, as well as the recording settings for videos. Significant variations in terms of these conditions, and low correspondence to previously processed examples, can significantly increase the difficulty of recognising the actions and interactions performed. Most notably, the viewpoint of the camera has a large effect on how actions and interactions are perceived. The lack of stereo in common video data presents difficulties in the recognition of objects in the background and foreground. Difficulties may also arise because of occlusion of the actors or their body parts, especially in scenes that include physical interaction with objects or other actors. This presents obstacles in the recognition of human actions and interactions from a single viewpoint, as characteristic movements or poses of the key body parts may not be visible. Considering that movement is present in videos, camera motion and motion blur may introduce an additional degree of difficulty for the correct recognition of actions under different settings. Other visual-based challenges include variations such as those based on luminance. Examples include lighting from multiple directions, or the reflective capabilities of material that determine how much light is absorbed and how much is reflected under which direction. In addition, lighting conditions can impact the colour consistency in video frames.

The second set of variations in human actions in videos is based on differences in the performance of the actions themselves. This is strongly associated with the human factor. As actions can be identified in many different ways, the criteria used to describe actions remain unclear. Vallacher and Wegner \cite{vallacher2011action} have created a broad system to explain the formation of relationships between personality traits and how individuals perform everyday actions. The foundation of their proposed \textit{theory of action identification} is based on three principles. The first is that human actions are maintained with respect to the prepotent identity associated with the action. Prepotent identities can, for example, correspond to one person sustaining the act of passing the ball, by projecting the exact location that the receiver will catch it. This can also be sustained by a less experienced passer, as \enquote{throwing the ball within the general proximity of the receiver}. We can find that identities serve a guideline role for the action performed. The first point that we can draw from the passer example concerns the levels of complexity that the same action can be expressed as. The second principle states that the higher the level of identity, the more likely it is to become prepotent. Low identities include more general action identifications such as \enquote{moving hands} that is a component within a very large number of actions.

\begin{figure}[ht]
\includegraphics[width=\textwidth]{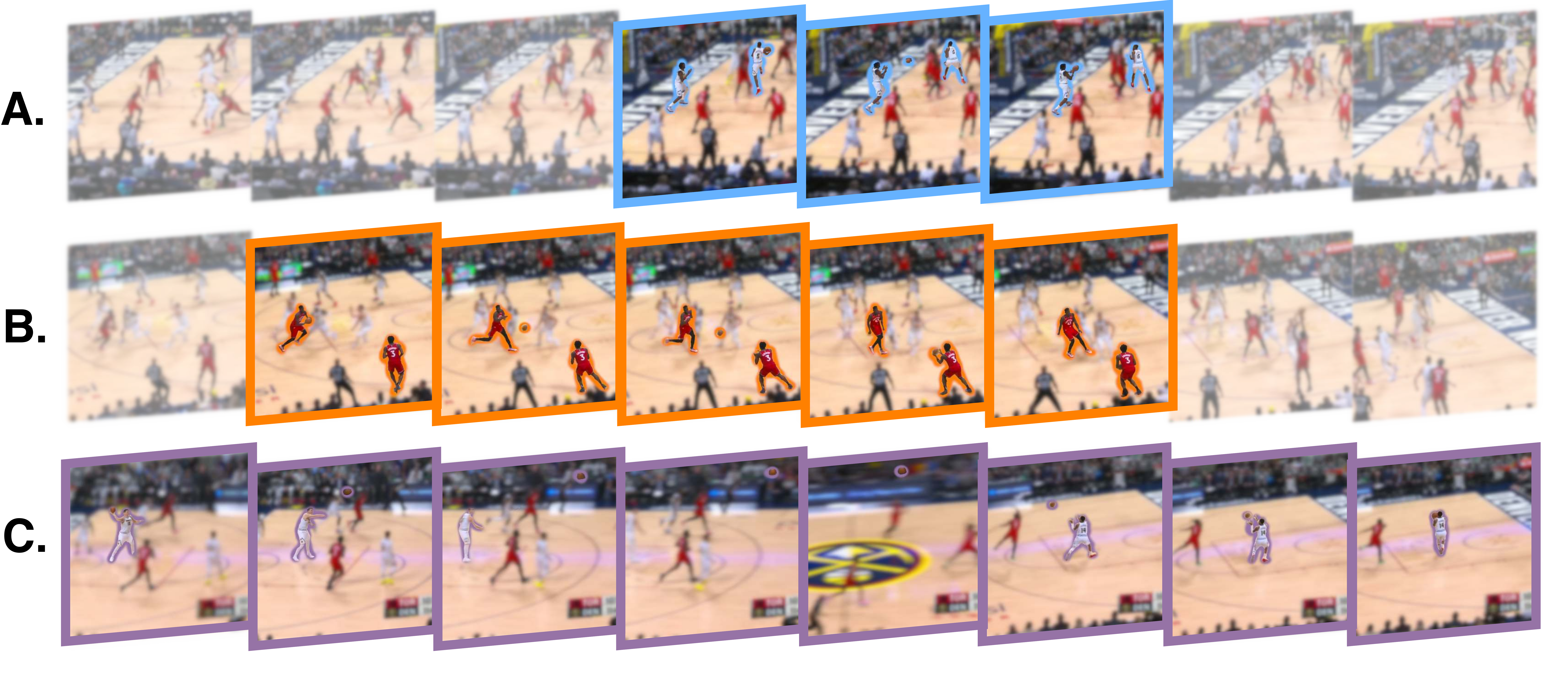}
\caption{\textbf{Examples of basketball passes from the same game}. (A) Brief hand-off pass, (B) Wing pass, (C) Full-court outlet pass.}
\label{fig:passes_examples}
\end{figure}

However, actions can be better understood with the use of higher-level identities such as \enquote{throwing ball to teammate},  \enquote{making a corner pass} or \enquote{pass over the defending team}. This however, leads to the second point that can be drawn from the example, which is that the actor's skill level can constrain his action identity creation ability. The third principle states that actions which cannot be performed by a high-level prepotent, tend to then be expressed by a lower-level identity. This is additionally connected to the individual's skills (or lack of thereof), fears or abstractions. For example, high-level identities such as \enquote{full court pass to teammate} are strongly associated with the above mentioned constraints. Instead, the person is more willing to adopt an identity such as \enquote{throw ball towards the baseline} in order to control the action.
%% How action identities are used in CV datasets
Considering how action and interaction identities are strongly connected with an individual's prepotent identity associated with that action, a great problem arises for the definition of action classes. We demonstrate in \Cref{fig:passes_examples} three examples that are under the same class \textit{\enquote{basketball pass}}.

\begin{figure}[ht]
\includegraphics[width=\textwidth]{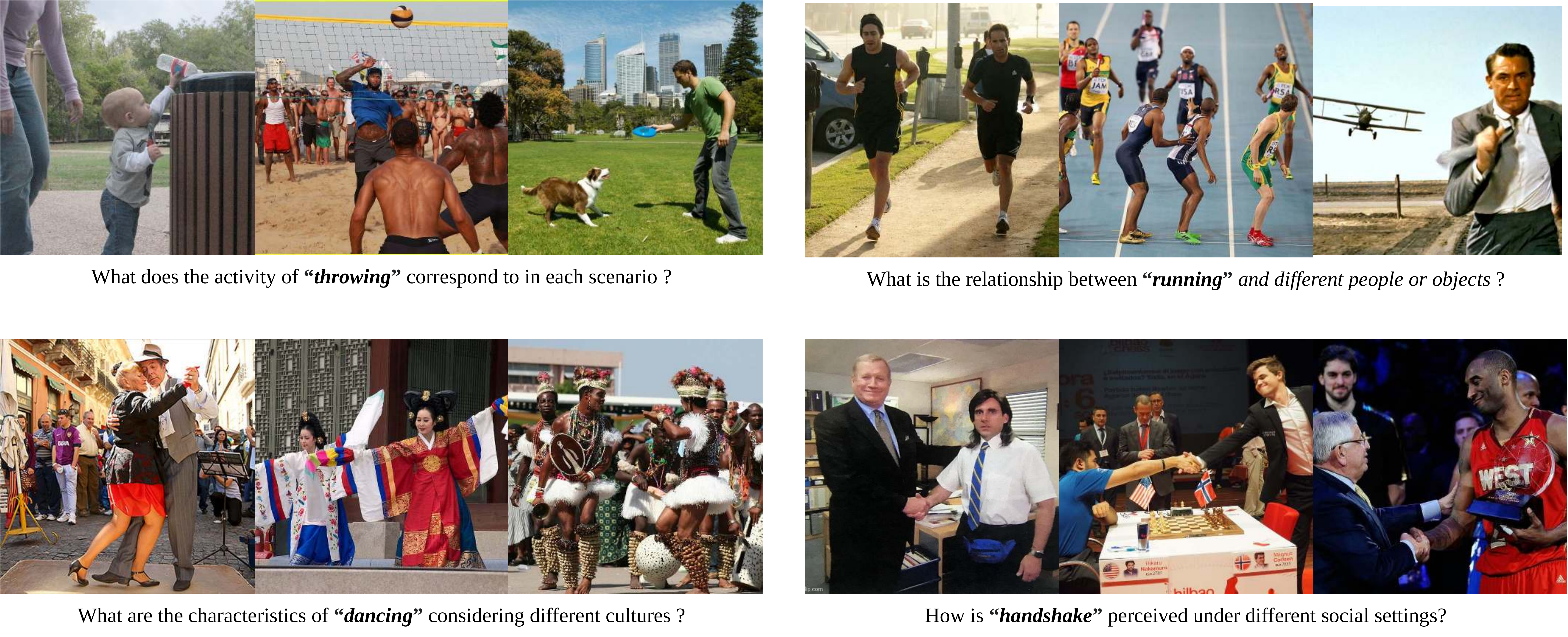}
\caption{\textbf{Instances of human actions}. Actions can be informative in terms of the activities that are presented (e.g. \textit{\enquote{throwing}} activity). Their relations to people and object may differ between examples (e.g. \textit{\enquote{running}} action). There is a cultural (e.g. \textit{\enquote{dancing}} action) or social (e.g. \textit{\enquote{handshake}} action) aspect in terms of their characteristics. }
\label{figure:actions_example}
\end{figure}

% criteria for understanding actions
Understanding actions is also based on their socially conveyed meanings and the identities that have been assigned to them. For example, considering the examples in \Cref{figure:actions_example}, actions can vary based on the scenarios that they are performed in. An activity such as \textit{throwing} an object relates differently in different action scenes. \textit{Throwing} an empty bottle serves the purpose of getting rid of that object, while \textit{throwing} a ball serves the purpose of scoring a point. In the volleyball example, other people are also important for understanding the context as throwing ball is perceived differently if the receiver is a teammate or a player of the opposite team. In the same sense, it can also be part of a playful interaction without a competitive nature (e.g. throwing a frisbee to a dog). The relations of actions to other people or objects is also important and should be considered. Given the action of \textit{running}, the correlation to other people and objects in the immediate environment can be telling for the context that the action is performed in. There is the possibility that the action is performed for an exercise or in a social manner, based on the accompanying people. Alternatively, the nature of the activity can be within the context of a race and thus, the participants competing against each other. The action could also be engaged in for completely different reasons as seen in the third example in \Cref{figure:actions_example}. Another important aspect of actions is their cultural meanings while being either general or specific. An example of this can see seen in the different forms of \textit{dancing}. \textit{Dancing} can be associated with a social occasion or being part of a tradition. In addition to culture, social settings can also change the interpretation of different actions. For example, a \textit{handshake} could be perceived differently. In relation to the settings, a \textit{handshake} could be a formal way of two people introducing each other. Within the context of a competition, a handshake could be a form of resignation as observed in chess or could be for congratulating the winner or receiver of an award.

% Summarise on challenges
The personal agency of each person represents significant challenges in the performance of actions. These factors and aspects describe the \textit{how} and the \textit{why} actions are performed in a specific manner. Through the inclusion of these factors a more distinct identity of each action can be perceived. All traits are important, as most actions are personally determined by individuals, there is a direct connection between these traits and the performance of actions.

\section{Thesis overview}
\label{ch:1::sec:overview}

In this section, we provide an outline of the structure of the thesis and the action recognition challenges that each chapter addresses. \Cref{ch2} gives an overview of how current action recognition methods in computer vision address human actions, while the corresponding datasets that have been collected for action recognition and video understanding tasks are summarised in \Cref{ch3}. Our main contributions are introduced in \Cref{ch4,ch5,ch6,ch7} while a discussion is provided in \Cref{ch8}.

% Chapter 2
\textbf{\Cref{ch2}, Related Work}. We detail a synopsis of current progress made in action recognition. We distinguish between approaches that have been based on hand-crafted features and methods that used learned features through optimisation. We then present the main motivation for the works that have been included in this thesis. We discuss how our approaches differe from previous efforts and the challenges that our methods address.  

% Chapter 3
\textbf{\Cref{ch3}, Datasets for Video Understanding}. We provide an overview of historic and current datasets used as action recognition benchmarks. We focus on datasets that exemplify milestones achieved in terms of data collection and increases in the data complexity. We then present the datasets that are used in this thesis.

% Chapter 4
\textbf{\Cref{ch4}, Improving Action Recognition through Time-Consistent Features}. We present a novel method to address variations in terms of the performances of actions. Specifically, we target variations in the durations of different motions of the action. We explore how the locality of spatio-temporal convolutional patterns can be extended in order to address the relevance of the local features within the context of the entire action sequence. The newly created feature volumes encapsulate these dynamics of the local features with respect to the entire video. Through the temporal attention of the most relevant features, we are able to strengthen informative signals. This is done through a recurrent sub-network with the \textit{Squeezed and Recursion} part of the method. The second part of the proposed method studies the cyclic consistency of the original and the global attention-aligned features. By measuring the distance in a common embedding space between the two feature volumes, a similarity measure between the two volumes can be established. Based on that similarity measure, we propose a \textit{Temporal Gating} function that can hold a closed or open state. In a closed state the similarity between the two feature volumes is not substantial and thus information is processed independently. Instead, in an open state, the gate fuses the two volumes to create a single coherent feature volume that highlights with temporal attention the features that are informative across the entire video. We show that this inclusion of temporal attention can improve the performance of spatio-temporal action recognition models with a minimal increase in the total number of computations required.

% Chapter 5
\textbf{\cref{ch5}, Time-Varying Convolutions for Video Understanding}. We further study the extraction of features that correspond to action patterns based on different temporal durations. This chapter focuses on the representation of temporal variations in the performance of actions that are not based in the order of magnitude. We propose the extraction of features over varied size spatio-temporal windows. Through this, short-term and long-term action patterns can be better modelled as convolutional receptive fields that are less strict to fixed-sized regions. The proposed convolutional blocks named \textit{Multi-Temporal Convolutions} utilise three branches. Each branch addresses different temporal durations over an interval. The first branch is the \textit{local} branch of small spatio-temporal regions. The \textit{prolonged} branch similarly models information within enlarged spatio-temporal windows while the information also relates to the features that have been extracted in the local branch. The final \textit{global aggregated feature importance} branch uses Squeeze and Recursion from \Cref{ch4} to align the features from the previous two branches and to discover their relevance within the entire video sequence. Experiments demonstrate competitive results over common benchmark action recognition datasets with an additional substantial reduction in terms of the number of computational operations used. We ablate over the methods to create feature volumes for the extraction of longer spatio-temporal sizes as well as the proportions of local and prolonged features that can be extracted.

% Chapter 6
\textbf{\Cref{ch6}, Class-Specific Regularisation Across Time}. We address the association of features to specific action classes. As features in current convolutional models are learned in a hierarchical fashion, only a small set of features targets the modelling of spatio-temporal patterns that are specific to a class or a smaller sub-set of action classes. By associating class and feature information, descriptors based on class-relative information can be created. The goal of the proposed approach, named \textit{Class Regularisation}, is to create descriptors that can amplify features specific to a class. Through this, subtle differences in the visual appearance and performance between similar classes can be intensified in order to create a distinction between them. 

% Chapter 7
\textbf{\Cref{ch7}, Spatio-Temporal Feature Interpretation}. We explore features that are learned by spatio-temporal convolutions. The main objective for this chapter is to study feature explanation as a qualitative measure for the representation of a model's performance. We first target the task of uncovering the spatio-temporal regions that are informative for an action that is performed and that 3D CNNs rely for their predictions. This is visualised as a saliency mask over time, creating a tube effect that corresponds to the space-time regional attention of the network. Our approach uses the class weights associated with the class and its features that are to be visualised alongside the respective feature activations produced over space-time locations. Based on the representation of class feature relevance over the video sequence, we also extend the approach to work hierarchically over features of multiple layers. Through a proposed cross-layer feature relevance exploration named \textit{back step}, we can construct an association between high-level and lower-level features over different network layers. This effectively deals with the curse of dimensionality problem of traversing over different layers inside the network while also presenting their causalities. The resulting few-to-many connections create a structure that resembles a pyramid. We demonstrate the merits of our method through providing visual insights into the spatio-temporal features of each layer that correspond to the action class. 

% Chapter 8
\textbf{\Cref{ch8}, Discussion and Future Research Directions}. We conclude with \Cref{ch8} in which we summarise the methods that have been proposed in this thesis. Additionally, we discuss prominent research directions and the challenges that such future works could address.

    \author
{}
\title{Related Work}
\maketitle
\label{ch2}

In this chapter we discuss the recent progress that has been made in the field of action recognition and video understanding. We start by reviewing methods that utilise hand-crafted features for the extraction of spatio-temporal information in combination with a classifier model. We then overview methods that use objective tasks to discover descriptive features. We refer to the product of these descriptors as learned-features. 

\section{Recognition from hand-crafted features}
\label{ch:2::sec:handcrafted}

Traditionally, the recognition of human actions and interactions from video starts with the extraction of image features to represent the scene. Subsequently, the extracted spatial-features are used for the classification of the video sequence with an action label. An important requirement in this process is that the extracted image features must be invariant to image conditions and the action performance. These conditions are in combination with maintaining a sufficient feature complexity to deal with subtle differences between classes.

% outline
We make a distinction between approaches that are based on local features utilising salient points in the video sequence, and approaches based on feature templates that take into account regions that roughly correspond to a person's body or body parts.

\subsection{Local features approach}
\label{ch:2::sec:handcrafted::sub:local}

In general, local feature methods use a bottom-up approach by first detecting points of interest in a video, and then aggregate the detected regions over time and space to represent the performed actions. This selection of points is performed locally, typically at edges or motion boundaries. Popular descriptors are based on Harris corners \cite{marin2013exploring,zhang2013recognition}, SIFT \cite{delaitre2010recognizing,lowe1999object} or optical flow \cite{yu2012propagative}. Typically, no direct correspondence between points and video actors or their body part exists. Consequently, factors such as camera motion, dynamic backgrounds and obstructions affect the presence of local features.

% 3D
When additional depth information is available, e.g. from RGB-D recordings, local features can take into account depth gradients \cite{li2010action}. By utilising multiple viewpoints for the efficient mapping of 3D points, Xia and Aggarwal \cite{xia2013spatio} considered the creation of a codebook based on depth sequences.

% bow and fisher vectors
To increase the robustness of local descriptors, the distribution of points can be described based on a bag-of-words (BoW) or Fisher Vectors (FV) \cite{gao2016constrained,oneata2013action}. The consideration based on the use of local descriptors is that instances of the same action class are to have similar descriptors. Improving on this and to allow a more complex feature distribution, Niebles \etal~\cite{niebles2008unsupervised} constructed a vocabulary using latent topics models. 

\begin{figure}[ht]
  \centering
  \includegraphics[width=.9\textwidth]{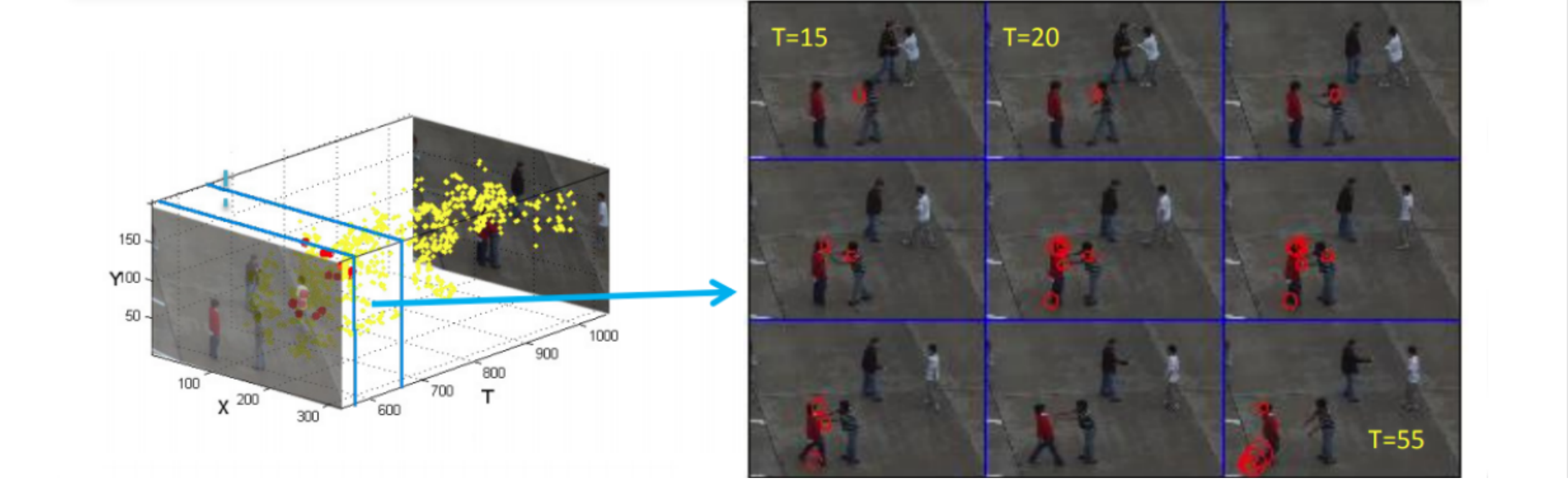}
  \includegraphics[width=.9\textwidth]{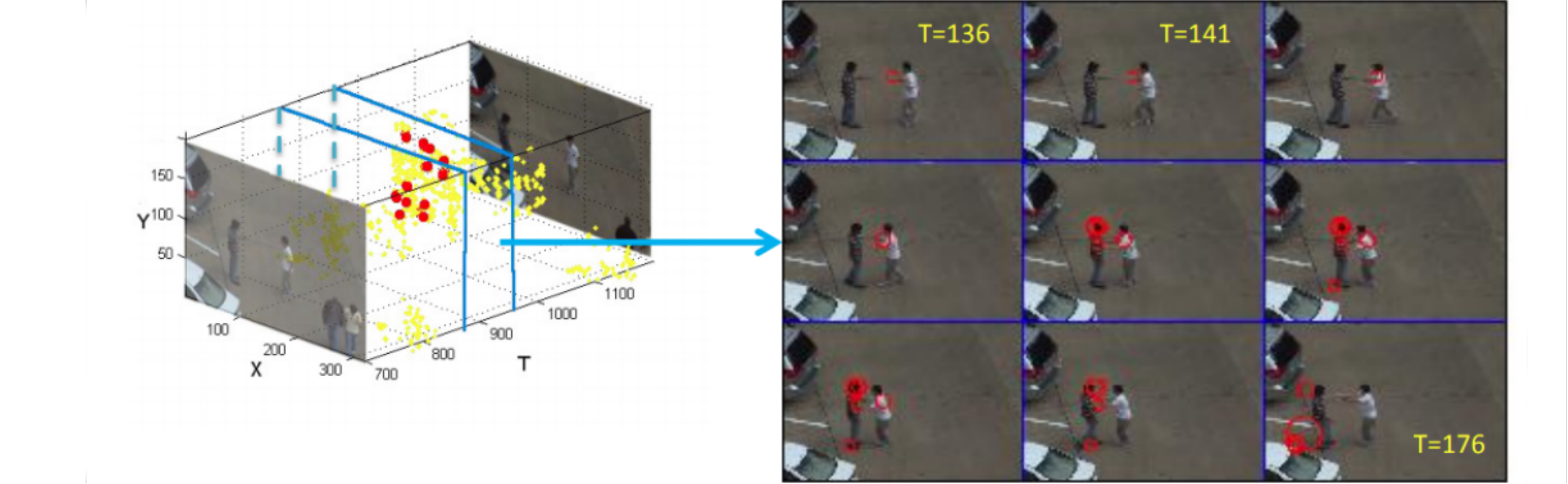}
  \caption{The red points are discovered co-occurring features from Zhang \etal~ The spatio-temporal phases are computed based on the causality between actor body parts. These causal relationships can represent relationships between actors over extended time durations. Figure sourced from \cite{zhang2012spatio}.}
\label{fig:phases}
\end{figure}

% spatio-temporal
As an alternative to modelling the trajectories of individual points in order to capture the motion of local points, other works focused on the temporally sequential nature of human actions and interactions through modelling the changes in the distribution of interest points over time. Zhang \etal~\cite{zhang2012spatio} used spatio-temporal phases to create a histogram of bag-of-phases. An example of this is visualised in \Cref{fig:phases}. Each phase is composed of local words with specific ordering and spatial position. Instead of jointly mapping both dimensions, others have addressed separation as well \cite{shariat2013new,tran2014activity}. The computed histograms represent similar features in single or multiple frames. Histograms of visual words have also been utilised by Kong \etal~\cite{kong2012learning} with the words derived from the quantisation of the spatial-temporal descriptors clustered. The produced clusters form high-level representations, termed interactive phases. These phases include motion relationships such as the shaking of two hands. This idea has been extended to localise interactions by spatially clustering phrases \cite{tran2013social}. Prabhakar and Rehg \cite{prabhakar2012categorizing} aimed at the inclusion of variation in the temporal domain by modelling the causality of the occurrence of visual words.

% bow for patches / trajectories
As not all motions and attributes are informative,  Kong \etal~\cite{kong2014discriminative} considered only body parts that are characteristic for a specific class. Their method pools BoW responses in a coarse grid, thus allowing them to identify specific motion patterns relative to a person’s location. A limit to the level of detail is the granularity of the patches and the accuracy of the person detector. Subsequent detections are linked to trajectories in order to additionally account for the temporal dimensionality of the video sequences. Mohammadi \etal~\cite{mohammadi2015violence} extend this grid-based approach by grouping the motion patterns as BoW vectors. Similarly, Turchini \etal~\cite{turchini2016understanding} used trajectories of multiple local feature types in order to localise where actions are performed. Wang and Cordelia \cite{wang2013action} have introduced Improved Dense Trajectories (DT), a widely adopted way of finding and describing trajectories of points. Dense Trajectories encode points as the temporal-based combination of Histograms of Oriented Gradients (HOG), Histograms of Oriented Flow (HOF) and Motion Boundary Histograms (MBH).

% person and context
Other approaches have been based on the use of local features to initially  isolate the actors in videos. The local features can be endings of HOG and HOF descriptors as persented by Caba \etal~\cite{caba2016fast}. Through the localisation of a person, the context of motions and actions of other people in relation to that actor can provide useful cues for the understanding of the scene. Reddy and Shah \cite{reddy2013recognizing} used information obtained through a scene context descriptor which combines the location and surroundings extracted with optical flow and 3D-SIFT, based on the moving and stationary pixels. Cho \etal~\cite{cho2017compositional} introduced a descriptor that takes into account the local, global and individual movement of actors in video sequences. By linking local features to persons, descriptions for the actor surroundings can also be made. Lan \etal~\cite{lan2012discriminative} presented an Action Context (AC) descriptor that is based on connected action probability vectors of several people. Similarly, Choi \etal~\cite{choi2014understanding} performed joint tracking, classification of the actions of an individual and the recognition of collective activities by considering bounding boxes of extracted local features.

\subsection{Template-based approaches}
\label{ch:2::sec:handcrafted::sub:template}

% general templates - hog and poselets
In a single frame, HOG descriptors can represent characteristic poses. For example, a high-five interaction can be described as two people facing each other with outstretched hands that meet above their heads. This notion was adopted by Bourdev \etal~\cite{bourdev2010detecting} to detect people engaged in specific actions, and was applied to human-human interactions by Raptis and Sigal \cite{raptis2013poselet}. Sefidgar \etal~\cite{sefidgar2015discriminative} formulated descriptors with discriminative key frames and their relative distance and timing within the interaction. Alternatively, Sener and  Ikizler-Cinbis \cite{sener2015two} formulated interaction detection as a multiple-instance learning problem to focus on relevant frames. This was done as not all frames in a video sequence are considered informative.

% hof
Based on HOG spatial feature descriptors, the motion around a characteristic pose can provide complementary information. Van Gemeren \etal~\cite{van2014dyadic} combined HOG and HOF descriptors to encode the characteristic frame of a two-person interaction. Yu and Yang \cite{yu2015fast} concatenated HOG and HOF descriptors and applied FV to make the detection linearly separable. This allowed the concurrent utilisation of spatial and temporal information by the model. Mousavi \etal~\cite{mousavi2015analyzing} introduced summarised tracked local key-point features through the Histogram of Oriented Tracklets (HOT). While optical flow can be seen as the representation of the motion between two subsequent frames, tracklets describe the path of local key-points over longer time intervals.

% combination with local
Temporal patches can be used in combination with local descriptors to take advantage of potentially conjoint salient pose or motion information \cite{ji2016multiple}. Yin \etal~\cite{yin2012small} employ 3D-SIFT to describe local motion events, but used a HOF to model the global motion of the video sequence. Similarly, Lathuiliere \etal~\cite{lathuiliere2017recognition} combined HOG descriptors and trajectory information from linked local features. Single-person and two-person interaction attributes, such as \enquote{two persons are standing side-by-side}, were calculated from these features.

% two-stage (face, body)
A different approach to interest points is to detect faces or bodies using a generic face or body detector \cite{patron2012structured,ryoo2011stochastic}. Given two close detections, actions and interactions can be classified based on extracted features within the detection region \cite{ryoo2011stochastic}. Various attributes, including gross body movement and proximity, have been employed to classify the interaction. Patron \etal~\cite{patron2012structured} also include the relative size and orientation of each person. Khodabandeh \etal~\cite{khodabandeh2015discovering} consider clusters of similar frames based on proximity and appearance of pairs of people. They included user feedback in order to improve the purity of the clusters and thus the classification of them. The drawback of this two-stage approach is that classification is sub-optimal when the person localisation fails, for example when people partly obstruct each other. This is a common situation, especially in cases where multiple actors are in close proximity.

% dpm
Such situations are mitigated with the use of Deformable Parts Models (DPMs) \cite{felzenszwalb2010object}. In DPMs, an articulated object such as a person or multiple people interacting are modelled as a set of parts with the inclusion of deformations between them. This provides a degree of flexibility in the spatial layout of the parts. As such, parts that are generally well detected, e.g. a person's head, can be coupled with parts that are traditionally more challenging to detect, such as a lower arm. Lu \etal~\cite{lu2015human} used DPMs to localise the rough outline of a person with optical flow being then employed to discover the motions performed in the subsequent frames. The resulting volume is then segmented into supervoxels to refine the person's outline in each frame, and classify the action. Instead of encoding the orientation of (pairs of) limbs as poselets, DPMs can also include a larger number of articulations by using a mixture of parts \cite{yang2011articulated}. This approach has been used to describe the joint poses of two interacting people \cite{yang2012recognizing}. Hoai and Zisserman \cite{hoai2014talking} extend the model to account for deformations in scale by using template examples of an action class and compare their similarity to the input video then used it to rank the detection scores. The main constraint of this method was the emphasis towards upper body movement based on the videos used.

% temporal
In order to include additional degrees of variations based on the temporal performance of interactions, works have introduced a variety of methods. Ji \etal~\cite{ji2017new} modelled the changes in HOG descriptors over time using a Hidden Markov Model (HMM) for human interactions. With the use of the distance between actors, they consider the frames in the start, middle or end stage of the interaction. The HMM scores for all stages are fused for the final classification. The same rationale of using phases has been adopted by Cao \etal~\cite{cao2013recognize} for human activities that addressed the task of classifying a video sequence with potentially missing frames. 

% temporal dpms
While DPMs only encode a particular pose or motion spatially, extensions for the inclusion of temporal information were proposed to deal with the time-varying nature of actions. Yao \etal~\cite{yao2014animated} focused on human-object interactions capturing the movement related to a key pose using a DPM and a linked set of motion templates that also correspond to different phases of the performance. Tian \etal~\cite{tian2013spatiotemporal} have extended DPMs for action detection to model changes in pose over time. These formulations work well for the representation of coarse movements, but finer-scale movements are difficult to model because the motion is not linked to specific parts of the body. Tran and Yuan \cite{tran2012max} also address a localisation task but consider linking regions over time based on HOG and HOF in a structured learning approach. A max-path algorithm is used to find the optimal volume that contains the action in space and time.

\section{Action recognition from learned features}
\label{ch:2::sec:learned}

The hand-coded feature descriptors described in \Cref{ch:2::sec:handcrafted} focus on local or global spatial or spatio-temporal information. The manual selection of descriptors leaves room for improvement because the process is agnostic to the specific classification task, application domain or class of behaviours.

% cnns
With the introduction of multiple convolutions \cite{lecun1998gradient}, Convolutional Neural Networks (CNNs or ConvNets) have been used for the classification across different modalities. CNNs allow for the discovery of informative patterns through updates based on the error of the task, e.g. predicting the action label. Consequently, they can overcome the issue of sub-optimal feature selection. While multiple convolution kernels allow for the selection of a wide range of image or video features, the stacking of consecutive convolution operations allows for a hierarchical extraction of complex features \cite{simonyan2014very}. Typically, the characteristics extracted in the first layers of the network correspond to features of limited complexity such as edges and simple textures. The complexity of the features increases in relation to the network depth.

% requirements
Methods based on neural networks have shown notable improvements for video understanding tasks as well. The benefit of improving accuracy based on large datasets, allows the architectures to generalise their feature assumptions over multiple instances, rather than being limited to the modelling capabilities of a predefined set of features, as in the hand-crafted methods. 

% aim of section
The purpose of this section is to present neural network architectures for human action recognition. We then show how temporal information is modelled and incorporated in the convolutions.

\subsection{Networks with individually-processed frames}
\label{ch:2::sec:learned::sub:single}

Initial works with CNNs for action recognition have been based on the use of single frames \cite{asadi2017survey,bilen2016dynamic,gkioxari2015contextual}. Similar to the use of hand-crafted features, several methods proposed possible extensions to additionally incorporate temporal information.

% Early, Late and Slow Fusion
Based on the classification of individual frames, Karpathy \etal~\cite{karpathy2014large} proposed three techniques to fuse the scores of multiple frames using different convolutional configurations. In the Early Fusion strategy, the input of the network is a stack of subsequent frames. Late Fusion combines the convolutional features of the first and last frames of a sequence in the final, fully connected layers. Slow Fusion is a combination of these two approaches, that empowers a progressive fusion over frames and activation maps, with the extension of convolutional layer connections through time. All three approaches are limited in their capability to deal with subtle temporal variations between classes, and large intra-class variations. It is a challenge to deal with these variations as they have to be modelled from a typically modest number of training videos.

% transfer learning
To partly mitigate this issue, authors have investigated the use of Transfer Learning \cite{bengio2011deep,bengio2012deep,caruana1998multitask,pan2010survey,yosinski2014transferable} from large image datasets. This is a process in which the network is first trained on a large dataset with general examples, and is subsequently re-purposed for another, more specific, classification task. In general, this means that the deeper layers are retrained for the specific domain. Consequently, fewer parameters need to be learned for the novel domain, which reduces the risk of overfitting.

\subsection{Motion-based and stream networks} 
\label{ch:2::sec:learned::sub:streams}

% two-Stream CNNS & multiple streams
Two-stream CNNs combine regular RGB frames and optical flow images as input \cite{simonyan2014two}, and are an alternative approach to model temporal information. The rationale is that through images, the spatial features of an action can be encoded, while the optical flow provides information about the motion. The network consists of two streams in the network structure. The spatial-based CNN is trained on individual video frames, and the temporal stream CNN which takes stacked optical flow frames as input. The results from the two networks are concatenated with late fusion. Different information fusion methods for each stream were explored by Park \etal~\cite{park2016combining}. Wang \etal~\cite{wang2016temporal} used sporadically sampled fragments from the video as inputs to two-stream CNN architectures. The resulting Temporal Segment Networks (TSN) share parameters across networks from different segments and make a prediction on each of the snippets independently. The predicted class is then the \enquote{point of agreement} between the video segments. This method capitalises on information from small temporal segments rather than using the video as a single input. Following the use of selected frames Diba \etal~\cite{diba2017deep} also propose a representation and encoding of the sequence features in a Temporal Linear Encoding (TLE) layer, after the convolution feature extraction is performed. It is based on the aggregation of appearance features from each of the individual temporal fragments. Works have also included the use of depth data as stream inputs \cite{garcia2018modality} in which features from the depth stream are distilled and simulated at test time as the test data does not include depth data.

% links between the streams
Typically, inputs in the two-stream CNNs are processed independently and only fused as a last step. This approach prevents the exchange of information between the streams. As such, it is not possible to develop attention mechanisms that focus on specific parts of the input in either stream. One way of establishing these links is by additional shortcut connections between convolutional layers of the motion stream to the spatial stream. This provides benefits in optimising the network architecture and increasing the network depth Feichtenhofer \etal~\cite{feichtenhofer2016spatiotemporal}. Residual learning \cite{he2016deep} enables the model to avoid degradation in deep structures, which relates to the saturation of accuracy as layers of the network are not able to effectively learn the identity map and instead \enquote{threshold} to zero mappings.

\begin{figure}[ht]
  \centering
  \includegraphics[width=.75\textwidth]{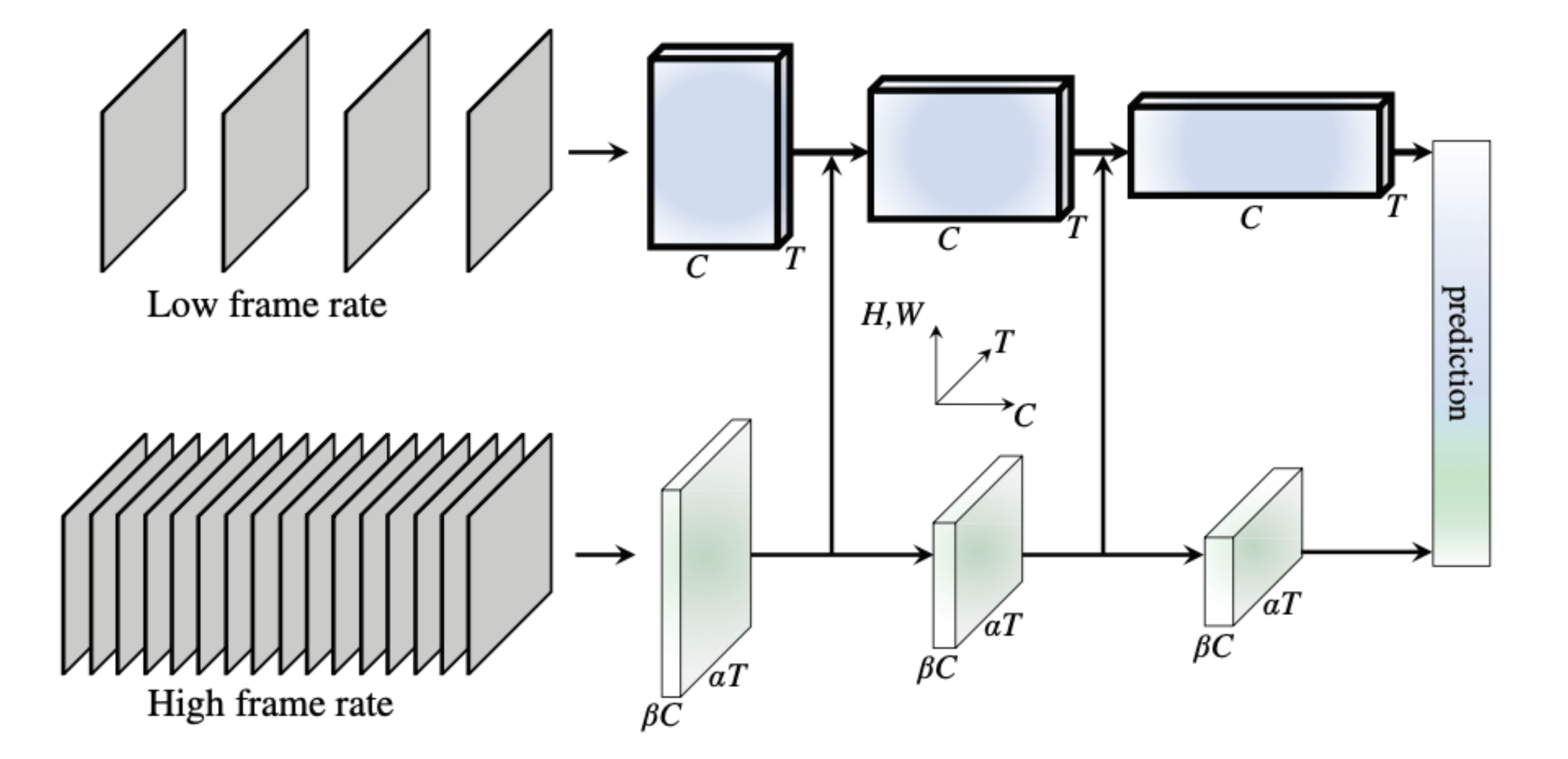}
  \caption{SlowFast network architecture. The slow (top) path uses lower frame rates to sample frames from the video sequence. The fast (bottom) path uses higher frame rates for sampling while include a fraction of the channels used by the slow path. Figure sourced from \cite{feichtenhofer2019slowfast}.}
\label{fig:slowfast}
\end{figure}

% R-CNN improvements
Works of Gkioxari \etal~\cite{gkioxari2015contextual} and Mettes \etal~\cite{mettes2017spatial} have been based on the adaptation of regional CNNs with the inclusion of multiple regions. Primary regions are considered the ones in which the main actor or actors are visible, while secondary regions contain contextual cues. The use of multiple independent or dependent regions as cues, and the separation of streams as a form to encode different features of an input, allows the focus towards discriminative regions \cite{miao2017multimodal,singh2016multi,tu2018multi,wu2016multi}

%3D-CNNs
Instead of treating the image and motion aspects of a video in separate streams, a video sequence can be represented as a 3D volume that is composed of stacked frames. Baccouche \etal~\cite{baccouche2011sequential} and Ji \etal~\cite{ji20133d} use 3D convolutions to simultaneously encode the spatial and temporal features of such a volume. This approach is essentially an extension of the standard 2D convolutions to 3D. The resulting feature maps encode informative spatio-temporal patterns in the video volume. Tran \etal~\cite{tran2015learning} presented the C3D architecture and demonstrated its superiority over 2D CNNs. Improvements based on the utilisation of spatio-temporal convolutions have been further studied by Hara \etal~\cite{hara2018can} and Kataoka \etal~\cite{kataoka2020would}. 3D convolutions can also be used concurrently with a two-stream network. Carreira and Zisserman \cite{carreira2017quo} have introduced a fusion of these two methodologies, two-stream inflated 3D-CNNs (I3D), that add a temporal dimension to the kernels of both convolutional and pooling layers. The work considers the creation of two I3D models that are applied to static image and optical flow inputs, and thus allows the 3D-CNNs to benefit from the additional information of motion patterns in optical flow streams. Spatio-temporal networks can be used as a base architecture to extend the type of information processed such as queries for people regions \cite{Girdhar_2019_CVPR}, position and motion \cite{choutas2018potion} and feature neighbourhood correspondence across time \cite{cao2019gcnet,wang2018non}.

% Pseudo, 2+1D and other spatio-temporal block variants
The larger number of parameters in 3D convolution blocks and, consequently, the demand of larger datasets for 3D-CNNs to train, have motivated the introduction of alternative convolution blocks. Notably, Qiu \etal~\cite{qiu2017learning} have proposed three supplementary blocks with different configurations of a single 2D convolutional kernel for the extraction of appearance information per frame and a temporal kernel responsible for the changes of pixel values over time loosely inspired by the separable convolutions of 2D-CNNs \cite{chollet2017xception,howard2017mobilenets}. This idea has also been used to separate spatio-temporal kernels into purely spatial and purely temporal ones by Tran \etal~\cite{tran2018closer} with the introduction of (2+1)D convolution blocks. Others have fused both solely-spatial and spatio-temporal convolutions in an effort to emphasise the spatial signal \cite{luo2019grouped,zhou2018mict}. Similarly, works have also focused on the use of pairs of temporal-vertical and temporal-horizontal movements \cite{li2019collaborative,stergiou2019FAST}. 3D grouped convolutions \cite{chen2018multifiber,tran2019video} have also shown benefits in performing convolutions in groups and decreasing the number of GFLOPs without a reduction in accuracy. Chen \etal~\cite{chen2019drop} have also studied how the creation of different modalities by sub-sampling part of the activations can improve computations and increase performance.

% Efficiency and performance improvement
Recent works have also aimed towards improving the efficiency of spatio-temporal convolutions architecturally. Feichtenhofer \etal~\cite{feichtenhofer2019slowfast}, motivated by the two-stream approach of 2D \cite{simonyan2014two} and 3D \cite{carreira2016human} CNNs, proposed a dual 3D-CNN architecture (pathway) with each of the sub-networks using RGB frames of different frame rates. Apart from the difference in frame rate sampling, with slow and fast rates, each pathway also used a different number of channels. Notably, the number of channels in the fast pathway is 1/8 of that of the slow pathway. This decision was made because the fast pathway is supplementary to the slow pathway as the representation spatio-temporal feature capabilities are weakened based on the number of frames that are omitted. An example of the network architecture is shown in \Cref{fig:slowfast}. Based on the recent success in terms of efficiency for image-based models \cite{howard2017mobilenets,pham2018efficient}, works on video-based models also focused on the creation of more efficient architectures while also improving performance. Models such as X3D \cite{feichtenhofer2020x3d} define a set of expansion parameters for the frame rate, temporal dimension input size, spatial dimensions input sizes, number of layers per block, number of channels and in-block channel numbers. Based on a backbone architecture, different models can be produced based on these parameters in order to address different computational complexities and memory usage. Other works \cite{kahatapitiya2021coarse} have further expanded the use of X3D with different temporal resolutions similar to SlowFast. Others \cite{yang2020temporal} have focused on the use of features across multiple feature complexity through a spatio-temporal fusion of information from different network layers. The created architecture was named Temporal Pyramid Network (TPN), as the information for the final predictions are used in a pyramid-like fashion. 

% 2D -> spatio-temporal
Other approaches towards spatio-temporal network efficiency have been towards the creation of modules that individually process the spatial and temporal modalities. Lin \etal~\cite{lin2019tsm} proposed a Temporal Shift Module (TSM) which uses 2D convolutions frame-wise but temporally shifts channels. This corresponds to frame-wise channel activations from preceding and succeeding frames, thus forming a temporal feature dependency across the frames. Works have further aimed at the encoding of the channel dependencies in order to improve this temporal shifting operation \cite{liu2020teinet}. Similar to this, Sudhakaran \etal~\cite{sudhakaran2020gate} used a learnable gate that splits information to be temporally processed by shifting channels as in TSM and spatially convolved activations as in a Grouped Spatio-temporal aggregation Module (GSM) \cite{luo2019grouped}. Jiang \etal~\cite{jiang2019stm} proposed a Spatio Temporal Motion (STM) network composed of two modules. The first is the Channel-wise Spatiotemporal Module (CSTM) which extracts spatio-temporal features similar to (2+1)D convolutions with kernels performed in groups. The second Channel-wise Motion Module (CMM) learns feature differences across frames in a channel-wise fashion. Further works \cite{li2020tea} have focused on the temporal excitation while spatially aggregating frame activations. AdaFused \cite{meng2021adafuse} has been build on top of TSM and uses a learnable policy for the frame channels to keep, reuse or skip during the adaptive temporal fusion of two neighbouring frame features. Other approaches \cite{liu2020tam} use a local feature branch and a global feature branch to address temporal cross-frame feature variations. Hussein \etal~\cite{hussein2019timeception} proposed a 2D module, named TimeCeption, utilising multiple frames as inputs and modelling the temporal feature dependencies through temporal depth-wise convolutions. Based on the Inception block, different kernel sizes can be used which enables the exploration of features across multiple time scales.

% specific architectures
Recent advances in reinforcement learning and evolutionary algorithms have contributed to a reduction in human supervision for creating robust network architectures with neural architecture search (NAS) \cite{zoph2017neural}. This trend has enabled the construction of architectures for specific tasks rather than general architectures \cite{Zoph_2018_CVPR}. In the video domain, NAS has been originally performed with the use of an acyclic graph with TSN being the backbone architecture \cite{peng2019video}. EvaNet \cite{piergiovanni2019evolving} was one of the first architectures to employ 3D convolutions within the search space. For the optimisation task, the connectivity between operations remains fixed, while the type of operations is optimised based on a search space of available operations. This effectively reduced the computational complexity of the architectural optimisation. Later works of Ryoo \etal~\cite{ryoo2019assemblenet} have used both RGB and optical flow inputs with each network architecture and their cross-layer connections discovered as part of the optimisation process. Recent works of Kondratyuk \etal~\cite{kondratyuk2021movinets} build on image-based MobileNet \cite{howard2017mobilenets} search spaces while also including the expansion parameters of X3D. However, as has been pointed out in the majority of the aforementioned works, the design of a 3D-CNN through a gradient-based method is especially challenging considering the extensive computational and time requirements.

\subsection{Recurrent networks}
\label{ch:2::sec:learned::sub:recurrent}

While CNNs can recognise image components and learn to combine them to classify different classes, they lack the ability to recognise patterns across time explicitly. Stream-based networks and 3D convolutions can take into account motion, but do not explicitly deal with variations in the temporal performance of an action or interaction. An alternative approach is to use recurrent neural networks (RNNs) that model temporal patterns using a state representation. The key idea is to use some form of temporal recursion in the network that allows the persistence of information through sequences of inputs. Thus the temporal variations in videos can be efficiently modelled alongside the spatial variations.

% use of rnns
Recurrent neural networks have been effectively used as a supplementary architecture to CNNs for extracting temporal features. In such architectures, spatial information is extracted though CNNs and is then passed to recurrent networks to learn the temporal characteristics of each interaction class \cite{bagautdinov2017social, deng2016structure}. Zhao \etal~\cite{zhao2017two} proposed an approach based on the normalisation of each layer of the network with batch normalisation \cite{ioffe2015batch}. The architecture is combined with a 3D-CNN using a two-stream fusion of the RNN and CNN. The use of multiple recurrent networks has also been extended to include tree structures (RNN-T) \cite{li2017adaptive}, to perform a hierarchical recognition process in which each RNN is responsible for learning an action instance based on an Action Category Hierarchy (ACH). This allows for the distinction between very dissimilar classes high in the hierarchy, while subtle differences between related classes such as a handshake and a fist bump are dealt with in the lower nodes.

% lstms
Recurrent Neural Networks suffer from vanishing gradients. This issue causes the updates in the network weights of the top layers to gradually diminish as the number of data-processing iterations increases. This hinders learning the temporal parameters effectively. To overcome this issue, Long Short-Term Memory (LSTM) RNNs \cite{hochreiter1997long} have been introduced that include additional \enquote{memory cell} modules that decide through gating whether to keep the processed information. As such, they are capable of maintaining information over longer periods, which allows them to learn long-term dependencies \cite{chung2014empirical}. This is essential for the modelling of interaction classes as the distinctive information is often present in different phases of the interaction.

% lstm temporal convolutions
Other works \cite{donahue2015long,li2018videolstm, varol17} have shown that the combination of convolutions and long-term recursions performs well for recognition tasks in videos. Donahue \etal~\cite{donahue2015long} were effective in both image and video description by directly connecting powerful feature extractors such as CNNs with recurrent models. Similarly, Baccouche \etal~\cite{baccouche2011sequential} extracted features from the 3D-CNN architecture and extended the work to a two-step recognition process with a LSTM. The first step was the use of 3D convolutions for the extraction of spatio-temporal features. The second step is based on these learned features that are passed to the LSTM so the model can make predictions on the entire video sequence. As such, the network can benefit from both short-term and long-term temporal information.

 \section{Addressing motion permutations and class-specific features}
\label{ch:2::sec:contrib}

% Differences to our contributions
Our work differentiates from these methods as our scope focuses on three different aspects of spatio-temporal information.

% Space time patterns of different size
We initially study patterns over different spatio-temporal sizes (in \Cref{ch4,ch5}). Through the extraction of features across different durations and spatial windows, a more robust and diverse representation of the video data can be created. We demonstrate that these patterns are not separate from each other. Instead, a strong dependence exists between features extracted from different spatio-temporal region sizes. Enforcing an alignment between these features to a common representation can create local features that also correspond to extended regions while also creating patterns of extended size that are respectively consistent along shorter sequences.

% Class-related features
We secondly study the correspondence between features and classes (in \Cref{ch6}). As certain features or combinations of features relate more to a specific class, we investigate how the regularisation of layer features based on their importance to a specific class can improve the accuracy of classifications. The direct link between classes and layer features can provide insights in terms of both features as well as the attention regions that are associated with a specific class across multiple layers within the network.

% Feature visualisation
Our final aim in this thesis is to provide visual explanations for the spatio-temporal features that are learned by 3D-CNNs (in \Cref{ch7}). We explore feature visualisations as a visually inoperable qualitative measure for spatio-temporal models. Methods that can demonstrate the inner workings of CNNs not only improve their overall transparency but can also demonstrate future directions towards improving spatio-temporal networks.

% Next chapter
The aforementioned aims of our work are evaluated based on commonly used action recognition datasets. We provide a detailed overview of the publicly available datasets in the following chapter alongside our chosen datasets for each task.
    
    \author
{}
\title{Datasets for Video Understanding}

\maketitle
\label{ch3}

In this chapter we overview available datasets for human action recognition. We explore the datasets both historically as well as based on their public availability. We also detail our choices for the datasets used in this thesis.

\startcontents[chapters]
%\printcontents[chapters]{}{1}{\section*{\contentsname}}

\section{Overview of video action datasets}
\label{ch:3::sec:intro}

% About
The past two decades have presented a large growth over both the number of action recognition datasets as well as their sizes. This relates to the inclusion of more complex spatio-temporal human actions. The introduction of CNNs to action recognition tasks has pushed the requirements for increased datasets in order to train accurate and robust models. Labelled datasets are suitable benchmarks because they allow for a direct comparison between methods. This generally leads to better understanding of the algorithmic advantages and limitations, and therefore leads to performance progression.

% Section outline
This chapter is structured as follows. We first list the available datasets for human action recognition. We also present each of the datasets that we believe have addressed a specific aspect of video understanding and explain their importance. The following sections \Cref{ch:3::sec:datasets::sub::hmdb_51,ch:3::sec:datasets::sub::ucf_101,ch:3::sec:datasets::sub::kinetics,ch:3::sec:datasets::sub::mit,ch:3::sec:datasets::sub::hacs} present the datasets that we used during our experiments and we detail our choices for selecting them.

\begin{table}[!ht]
\caption{\textbf{Video datasets for action recognition}. The presented datasets are arranged chronologically. Bold text denotes datasets used in our experiments.}
\centering
\resizebox{\textwidth}{!}{%
\begin{tabular}{|l|c|rr|cr|r|c|l|}
\hline
\multirow{2}{*}{Dataset} &
\multirow{2}{*}{Avail.} &
\multicolumn{2}{c|}{Action} &
\multicolumn{2}{c|}{Actors} &
\multirow{2}{*}{Duration} &
\multirow{2}{*}{Year} &
\multirow{2}{*}{Purpose}\\
& &
Classes &
Instances &
H./N.H. &
\# Act. &
&
&
\\[0.25em]
\hline
KTH \cite{schuldt2004recognizing} & \ding{51} &
6 &
2K &
\ding{51} / - &
25 &
$\sim$ 12s. &
2004 &
greyscaled videos\\[0.25em]
CAVIAR \cite{bins2004context} & \ding{51} &
9 &
28 &
\ding{51} / - &
<30 &
$\sim$ 8s. &
2004 &
Wide-angle view of actions\\[0.25em]
Weizmann \cite{blank2005actions} & \ding{51} &
10 &
90 &
\ding{51} / - &
8 &
$\sim$ 12s. &
2004 &
fixed camera axtions\\[0.25em]
ViSOR \cite{vezzani2008annotation,vezzani2008visor} &  &
N/A &
N/A &
\ding{51} / - &
$\sim$ 250 &
$\sim$ 12s. &
2005 &
Videos surveillance\\[0.25em]
IXMAS \cite{weinland2006free} & \ding{51} &
11 &
390 &
\ding{51} / - &
10 &
$\sim$ 5s. &
2006 &
RGB and motion caption data\\[0.25em]
Coffee \& Cigarettes \cite{laptev2007retrieving} & &
2 &
245 &
\ding{51} / - &
$\sim$ 5 &
$\sim$ 5s. &
2007 &
Smoking/drinking in movies and TV \\[0.25em]
CASIA Action \cite{casia} & \ding{51} &
15 &
1,446 &
\ding{51} / - &
24 &
N/A &
2007 &
Actions from video cameras \\[0.25em]
UCF Sports \cite{rodriguez2008action} & \ding{51} &
9 &
150 &
\ding{51} / -&
<100 &
$\sim$ 5s.&
2008 &
Actions from sports videos\\[0.25em]
Hollywood \cite{laptev2008learning} & \ding{51} &
8 &
475 &
\ding{51} / - &
<100 &
$\sim$ 16s.&
2008 &
Human actions in films\\[0.25em]
UIUC \cite{tran2008human} & \ding{51} &
14 &
532 &
\ding{51} / -&
<100 &
$\sim$ 6s.&
2008 &
Few examples dataset\\[0.25em]
UT-interaction \cite{ryoo2009spatio} & \ding{51} &
6 &
90 &
\ding{51} / - &
60 &
$\sim$ 17s. &
2009 &
Outside recordings \\[0.25em]
BEHAVE \cite{blunsden2010behave} & \ding{51} &
10 &
163 &
\ding{51} / - &
<50 &
40s.&
2009 &
Human interactions\\[0.25em]
UCF-11 \cite{liu2009recognizing} & \ding{51} &
11 &
1K &
\ding{51} / - &
100+ &
$\sim$ 5s.&
2009 &
YouTube videos of human actions\\[0.25em]
i3DPost MuHAVi \cite{gkalelis2009i3dpost} &  &
12 &
>1K &
\ding{51} / - &
100+ &
N/A &
2009 &
Multi-view human actions \\[0.25em]
Hollywood2 \cite{marszalek2009actions} & \ding{51} &
12 &
3K &
\ding{51} / - &
100+ &
$\sim$ 12s. &
2009 &
Human actions in films \\[0.25em]
TV-Human Interactions \cite{patron2010high} & \ding{51} &
4 &
300 &
\ding{51} / - &
100+ &
$\sim$ 3s. &
2010 &
Sourced from TV shows \\[0.25em]
UCF-50 \cite{reddy2013recognizing} & \ding{51} &
50 &
5K &
\ding{51} / - &
100+ &
$\sim$ 15s.&
2010 &
Web videos of human actions\\[0.25em]
Olympic Sports \cite{niebles2010modeling} & \ding{51} &
16 &
800&
\ding{51} / -&
100+ &
$\sim$ 3s. &
2010 &
Human actions in sports\\[0.25em]
\textbf{HMDB-51} \cite{kuehne2011hmdb} & \ding{51} &
51 &
7K &
\ding{51} / - &
100+ &
$\sim$ 3s. &
2011 &
Human motions from movies\\[0.25em]
CCV \cite{jiang2011consumer} & \ding{51} &
20 &
9K &
\ding{51} / - &
100+ &
$\sim$ 80s. &
2011 &
Videos sourced from the web\\[0.25em]
ASLAN \cite{kliper2011action} & \ding{51} &
432 &
4K &
\ding{51} / - &
100+ &
$\sim$ 5s. &
2011 &
Videos for action similarity\\[0.25em]
\textbf{UCF-101} \cite{soomro2012ucf101} & \ding{51} &
101 &
13K &
\ding{51} / - &
100+ &
$\sim$ 15s. &
2012 &
Web videos of human actions\\[0.25em]
CAD-60 \cite{sung2012unstructured} & \ding{51} &
12 &
60 &
\ding{51} / - &
<30 &
$\sim$ 45s. &
2012 &
Videos of human motions\\[0.25em]
THUMOS'13 \cite{idrees2017thumos,jian2013thumos} & \ding{51} &
101 &
13K &
\ding{51} / -&
100+ &
$\sim$ 15s. &
2013 &
Web-videos extending UCF-101\\[0.25em]
CAD-120 \cite{koppula2013learning} & \ding{51} &
12 &
120 &
\ding{51} / - &
<60 &
$\sim$ 45s. &
2013 &
Videos of human motions\\[0.25em]
Sports-1M \cite{karpathy2014large} & \ding{51} &
487 &
1M &
\ding{51} / -&
1,000+&
$\sim$ 9s&
2014 &
Multi-labelled spirts actions\\[0.25em]
THUMOS'14 \cite{idrees2017thumos,jian2014thumos} & \ding{51} &
101 &
16K &
\ding{51} / -&
100+ &
$\sim$ 15s. &
2014 &
extension of THUMOS'13 \\[0.25em]
ActivityNet-100 \cite{caba2015activitynet} & \ding{51} &
100 &
5K &
\ding{51} / - &
100+ &
$\sim$ 2m. &
2015 &
Untrimmed web videos\\[0.25em]
Watch-n-Patch \cite{wu2015watch} & \ding{51} &
21 &
2K &
\ding{51} / -&
7 &
$\sim$ 30s. &
2015 &
RGB-D data for daily activities\\[0.25em]
ActivityNet-200 \cite{caba2015activitynet} & \ding{51} &
200 &
15K &
\ding{51} / -&
100+&
$\sim$ 2m.&
2016 &
Untrimmed web videos \\[0.25em]
YouTube-8M \cite{abu2016youtube} &  &
N/A &
N/A &
\ding{51} / \ding{51}&
N/A &
N/A &
2016 &
Multi-labelled YouTube videos\\[0.25em]
Charades \cite{sigurdsson2016hollywood} & \ding{51} &
157 &
67K &
\ding{51} / - &
267 &
$\sim$ 30s.&
2016 &
Daily activities videos \\[0.25em]
ShakeFive2 \cite{van2016spatio} & \ding{51} &
5 &
153 &
\ding{51} / - &
33 &
$\sim$ 7s. &
2016 &
Human interactions w/ pose data\\[0.25em]
OA \cite{li2016recognition} & \ding{51} &
48 &
480 &
\ding{51} / \ding{51}&
<100 &
5s. &
2016 &
Ongoing actions in web videos\\[0.25em]
CONVERSE \cite{edwards2016pose} &  &
10 &
N/A &
\ding{51} / - &
N/A &
N/A &
2016 &
Human interactions\\[0.25em]
DALY \cite{weinzaepfel2016human} & \ding{51} &
10 &
4K &
\ding{51} / - &
100+ &
$\sim$ 8s. &
2016 &
Daily human activities\\[0.25em]
Okutama Action \cite{barekatain2017okutama} & \ding{51} &
12 &
4700 &
\ding{51} / -&
$\sim$ 400&
$\sim$ 60s.&
2017 &
Actions with aerial view\\[0.25em]
\textbf{Kinetics-400 (K-400)} \cite{kay2017kinetics} & \ding{51} &
400 &
306K &
\ding{51} / -&
1,000+&
$\sim$ 10s. &
2017 &
Large-scale human actions dataset\\[0.25em]
Someting-Someting v1 \cite{goyal2017something} & \ding{51} &
174 &
109K &
\ding{51} / -&
100+ &
$\sim$ 4.1s. &
2017 &
Human actions with objects\\[0.25em]
AVA \cite{gu2018ava} & \ding{51} &
80 &
392K &
\ding{51} / - &
100+ &
$\sim$ 2.7s. &
2017 &
Atomic human-object interactions\\[0.25em]
\textbf{Moments in Time (MiT)} \cite{monfort2018moments} & \ding{51} &
339 &
1M &
\ding{51} / \ding{51}&
1,000+ &
3s. &
2017 &
Event-based high variance data\\[0.25em]
MultiTHUMOS \cite{yeung2018every} & \ding{51} &
65 &
16K &
\ding{51} / - &
100+ &
$\sim$ 4.8s. &
2017 &
Densely-labelled action recognition\\[0.25em]
Diving-48 \cite{li2018resound} & \ding{51} &
48 &
18K &
\ding{51} / -&
N/A &
$\sim$ 3s.&
2018 &
Competitive diving videos dataset \\[0.25em]
EPIC-KITCHENS-55 \cite{damen2018scaling} & \ding{51} &
2,747 &
40K &
\ding{51} / - &
35 &
3.1s. &
2018 &
Ego-centric actions\\[0.25em]
Kinetics-600 (K-600) \cite{carreira2018short} & \ding{51} &
600 &
495K &
\ding{51} / - &
100+&
$\sim$ 10s. &
2018 &
Large-scale human actions dataset \\[0.25em]
VLOG \cite{Fouhey18} & \ding{51} &
30 &
122K &
\ding{51} / -&
10.7K&
$\sim$ 10s.&
2018 &
Lifestyle VLOGs dataset\\[0.25em]
Something-Something v2 \cite{goyal2017something} & \ding{51} &
174 &
221K &
\ding{51} / - &
100+ &
$\sim$ 3.8s. &
2018 &
Human actions with objects\\[0.25em]
\textbf{Kinetics-700} \cite{carreira2019short} & \ding{51} &
700 &
650K &
\ding{51} / - &
1,000+ &
$\sim$ 10s. &
2019 &
Large-scale human actions dataset\\[0.25em]
Jester \cite{materzynska2019jester} & \ding{51} &
27 &
148K &
\ding{51} / - &
1,376 &
3s. &
2019 &
Webcam hand gestures\\[0.25em]
\textbf{HACS-Clips} \cite{zhao2019hacs} & \ding{51} &
200 &
482K &
\ding{51} / - &
1,000+ &
2.0s.&
2019 &
Fixed-duration clips of human actions\\[0.25em]
HACS-Segments \cite{zhao2019hacs} & \ding{51} &
200 &
139K &
\ding{51} / - &
1,000+ &
2.0s. &
2019 &
Segments from YouTube videos\\[0.25em]
IG65M \cite{ghadiyaram2019large}  &  &
N/A &
65M &
N/A &
N/A &
N/A &
2019 &
Actions in Instagram videos\\[0.25em]
AViD \cite{piergiovanni2020avid} & \ding{51} &
887 &
450K &
\ding{51} / - &
1,000+ &
$\sim$ 9s.&
2020 &
Diverse with blurred faces\\[0.25em]
HVU \cite{diba2020large} & \ding{51} &
3K &
572K &
\ding{51} / - &
1,000+ &
$\sim$ 10s.&
2020 &
Multi-label/task video understanding\\[0.25em]
HAA500 \cite{chung2020haa500} & \ding{51} &
500 &
10K &
\ding{51} / - &
1,000+ &
$\sim$ 2.1s.&
2020 &
Diverse atomic actions \\[0.25em] 
\textbf{Kinetics-700 (2020)} \cite{smaira2020short} & \ding{51} &
700 &
647K &
\ding{51} / - &
1,000+ &
$\sim$ 10s.&
2020 &
Large-scale human actions dataset\\[0.25em]
FineGym \cite{shao2020finegym} & \ding{51} &
530 &
33K &
\ding{51} / - &
100+ &
$\sim$ 2s. &
2020 &
Actions from gymnastics\\[0.25em]
EPIC-KITCHENS-100 \cite{damen2020rescaling} & \ding{51} &
4,053 &
90K &
\ding{51} / - &
37 &
$\sim$ 3.1s. &
2020 &
Ego-centric actions\\ 
\hline
\end{tabular}%
}
\label{tab:Datasets_full}
\end{table}

In this section we provide a catalogue of the most widely used and popular datasets for action recognition chronologically. The majority of these datasets have also been a point of study in recent literature reviews \cite{chaquet2013survey,herath2017going,hutchinson2020video,koohzadi2017survey,poppe2010survey,singh2019video,stergiou2019analyzing,vrigkas2015review}.

% About the dataset
\Cref{tab:Datasets_full} demonstrates a comprehensive overview over action recognition datasets. It demonstrates the number of action classes that have been allocated alongside the total clips/videos that each dataset includes. Column H./N.H. indicates if the video actions are performed solely by human actors or could also include other actors (e.g. animals or animations). The duration label shows the average clip duration in either seconds (s.) or minutes (m.).

% KTH dataset
The KTH \cite{schuldt2004recognizing} was one of the first datasets that captured basic spatio-temporal actions. In total 2391 action sequences were used with greyscaled frames. The labels only included six actions: \textit{walk}, \textit{jog}, \textit{run}, \textit{box}, \textit{hand-wave} and \textit{hand-clap}. Variations between video sequences include indoor or outdoor settings and different clothes. The dataset has been widely used in earlier methods utilising spatio-temporal features such as spatio-temporal ROI extractors \cite{dollar2005behavior,oikonomopoulos2006spatiotemporal,willems2008efficient}, bags-of-features \cite{messing2009activity,niebles2007hierarchical,sadanand2012action,wang2009evaluation} and (Improved) Dense Trajectories \cite{wang2013dense}. KTH has been largely surpassed by modern datasets. 

% Hollywood 2
The Hollywood \cite{laptev2008learning} and Hollywood 2 \cite{marszalek2009actions} datasets were the first datasets to move from videos obtained in lab-environments to extraction of actions from media such as web videos, movies and TV series. Both datasets include action segments from movies with the initial version of the dataset including 475 videos and 8 action labels: \textit{answer phone}, \textit{get out car}, \textit{handshake}, \textit{hug person}, \textit{kiss}, \textit{sit down}, \textit{sit up} and \textit{stand up}. The second version increased the dataset size to 3669 videos with the addition of four additional action labels: \textit{drive car}, \textit{eat}, \textit{fight person} and \textit{run}. Although the datasets included more variations and added action label complexity, they were still with controlled video conditions. This included limited camera motion, motion blur and background clutter. Later datasets aimed towards the inclusion of such settings.

% UCF-50
At the time that UCF-50 \cite{reddy2013recognizing} was first introduced, it included the largest and most diverse collection of human actions in videos with 50 different action labels and over 5K videos. The videos had been sourced from YouTube and include both amateur and professional videos effectively addressing the requirements for diverse video conditions from previous datasets \cite{blank2005actions,ryoo2009spatio,laptev2008learning,marszalek2009actions}. The use of web-videos for the creation of datasets has been a basic property in later datasets for action recognition.

% Sports-1M
The first large-scale human actions video dataset was Sports-1M \cite{karpathy2014large}. It included at total of 487 labels for sport-related actions. The dataset was made specifically for CNN models in order to utilise the 1M clips sourced from YouTube as a common benchmark. The dataset was also intended for pre-training networks. Similarly to the UCF variants \cite{liu2009recognizing,reddy2013recognizing,soomro2012ucf101} the videos were obtained from YouTube. Because of the large size of the dataset, a taxonomy similar to ImageNet \cite{krizhevsky2012imagenet} was used to create six internal nodes for multiple actions: \textit{Aquatic Sports}, \textit{Team Sports}, \textit{Winter Sports}, \textit{Ball Sports}, \textit{Combat Sports} and \textit{Sports with Animals}. Each class has on average 2000 videos however the dataset is weakly annotated with approximately 5\% of the data including more than one label. However, as the labelling is noisy, motion cues are not clearly distinguishable and therefore certain actions are classified by specific objects or backgrounds that are associated with an action.     

% ActivityNet
Following Sports-1M as a dataset for deep learning models, ActivityNet \cite{caba2015activitynet} included general daily human actions that have been manually labelled. The initial version included 100 action categories and approximately 5K videos. The second version doubled the number of classes to 200 while also increasing the number of videos to 15K which was also used for the ActivityNet 2016 challenge \cite{heilbron2016activitynet_challenge}. Although the dataset was used as a common benchmark for models, it has been largely surpassed by larger and more diverse datasets such as Kinetics \cite{carreira2017quo}.

% Something Something
Something-something \cite{goyal2017something} dataset has been introduced for human-object interactions. The dataset consists of first person videos where the actions performed are based on every-day objects. The goal of the dataset is fine-grained video understanding with the first dataset being composed of 108,499 videos while the second version including 220,847 videos. The dataset is especially useful for training and benchmarking video models as the videos are of relatively high resolution. However, the dataset only addresses human-object interactions in first person and not general human actions and interactions as is done in other datasets (e.g. Kinetics, HACS and MiT).

% HVU
The Holistic Video Understanding (HVU) dataset \cite{diba2020large} contains approximately 572K clips organised hierarchically with different levels of taxonomies. The dataset is built over the AVA \cite{gu2018ava}, Kinetics-400 \cite{kay2017kinetics}, HMDB-51 \cite{kuehne2011hmdb}, UCF-101 \cite{soomro2012ucf101} and HACS \cite{zhao2019hacs} datasets creating a super-set of all of the aforementioned datasets.  It includes a total of over 3K labels based on the semantic aspects of video scenes, objects, actions and events. Apart from video classification, the dataset can also be used for the supplementary tasks of video captioning and video clustering. One limiting factor of the dataset as an action recognition benchmark is its recency. Although established datasets such as Kinetics and MiT still suffice in terms of the complexity and challenges, there is not a significant number of works based on the dataset. This means it is more difficult to understand the merits of a newly proposed method because it can only be compared to a limited number of other works.

% AViD
The Anonymised Videos from Diverse countries (AViD) \cite{piergiovanni2020avid} is a publicly available dataset from various countries. The aim of the dataset is to address data bias of datasets in which the majority of the videos are collected from specific countries. As culture is a factor that can effect the way an action is performed \cite{vallacher2011action}, the diversification of data through the inclusion of examples from different countries is important. In addition, the dataset is static instead of requiring a download script such as Kinetics, HACS or HVU and therefore the number of videos does not change based on their availability on online hosting websites (such as YouTube). Videos include blurred faces to anonymise the actor identities. The dataset includes a total of 887 classes with clips being sourced from Kinetics, Charades and Moments in Time. Actions that were dependant on facial expressions such as \enquote{smiling} or \enquote{applying eyeliner} were removed as the dataset is anonymised, resulting in 736 classes. The dataset introduces 157 additional action categories with a total number of 450K videos. Similar to HVU, due to its recency, the dataset has not yet been widely used for either benchmarking or pre-training.

Recent efforts have also been made towards the automation of the data collection process in order to enable the creation of datasets for action-related tasks with minimum human effort requirements. This includes the use of skeleton-based data to simulate the performance of an action based on frames \cite{khodabandeh2018diy} or through the use of virtual parametarised environments \cite{hwang2020eldersim}.

\section{Datasets used in this thesis}
\label{ch:3::sec:datasets}

% Our datasets
Based on the characteristics of the overviewed datasets, we select seven datasets for our experiments. Our choice has been motivated by either the requirements of a large pre-training dataset, datasets for benchmarking or transfer-learning. We present these datasets and explain our choices in the following subsections.

\subsection{Human Motion Database (HMDB-51)}
\label{ch:3::sec:datasets::sub::hmdb_51}

% About
One of the first datasets to address the limited number of action categories and classes of previous video datasets such as Hollywood \cite{laptev2005space} (with 8 action classes), UCF Sports \cite{rodriguez2008action} (with 9 actions), Hollywood2 \cite{marszalek2009actions} (with 12) and Olympic Sports \cite{niebles2010modeling} (with 16), was HMDB-51 \cite{kuehne2011hmdb}. Although the number of classes is close to UCF-50 \cite{reddy2013recognizing}, HMDB-51 is less associated with sports and instead includes various action classes. Some intra-class variations include visibility of body parts, camera motions and viewpoints and clip quality. The videos used have been sourced from movies segmented to smaller video segments corresponding to action categories such as \textit{hand-waving}, \textit{sword fighting} and \textit{running}. In total, it includes 6,766 clips with each action class having a minimum of 101 clips. The videos in the dataset have an average duration of approximately $3$ seconds with a frame rate of 30. However, there is a significant amount of variations in terms of the labelled completion times between actions. For example, the shortest example includes only 40 frames (for class \textit{kick}), while the longest clip from class \textit{brushing hair} consists of 649 frames. Although the dataset has been widely used for comparisons, the number of examples is now considered small and the number of classes limited. It is considered less effective for use by current models for benchmarking because of this.

% Our usage
Because of the small number of both classes and examples compared to current data requirements of models, we use HMDB-51 as a fine-tuning dataset to test the feature transferability of our models. 

\subsection{UCF-101}
\label{ch:3::sec:datasets::sub::ucf_101}

% About
Following the introduction of HMDB-51, UCF-101 \cite{soomro2012ucf101} is the enlargement of the prior UCF-50 dataset to 101 action classes. Most of the previous datasets have been based on the use of either actors with fixed backgrounds \cite{blank2005actions, schuldt2004recognizing, weinland2006free} or scripted actions and interactions \cite{kuehne2011hmdb,laptev2008learning,marszalek2009actions}. In contrast, UCF-101 was based on videos sourced from YouTube, which includes actions and interactions that are not pre-planned. The total number of clips is 13,320 with a fixed frame rate of 25. The average clip duration is approximately 15s. , with the amount of variation in terms of duration between classes being higher of that of HMDB-51. The class with the lower average number of frames is \textit{Jump Rope} with an average of 346 frames, while class \textit{Rock Climbing Indoor} reaches a maximum number of in-clip frames of 832. A subset with 24 classes has also been used in the THUMOS'13 challenge. Additionally, there are three different splits that can be used to create the training and validation sets.

% Our usage
In our experiments, we use UCF-101 similarly to HMDB-51 for testing the fine-tuning capabilities of our proposed models. All of our experiments use the first split (split1) for the creation of the training and validations sets.

\subsection{Kinetics Video Datasets}
\label{ch:3::sec:datasets::sub::kinetics}

% About
The previously described datasets only included a limited number of actions and a relatively small number of videos in total. Data in the range of tens of thousands does not suffice for accurate class predictions for human actions with large degrees of variations. The inclusion of larger and more diverse datasets not only benefits the feature variations that can be modelled, but also can effectively address data-related problems such as overfitting. The Kinetics dataset was originally introduced by Kay \textit{et al.} \cite{kay2017kinetics} with the original variant (K-400) including 400 action classes and a total of approximately 306K videos. The goal of the dataset is to provide a common benchmark similar to the image-based datasets of ImageNet \cite{deng2009imagenet} for image recognition or COCO \cite{lin2014microsoft} for object segmentation. The dataset is been widely popularised and used as a common benchmark for action recognition models. Later variants (K-600) \cite{carreira2018short} increased the dataset to 600 action classes with a new total of around 495K clips. The 600 expansion shares 368 classes with the original 400 variant with the 32 remaining classes being renamed or removed altogether. The later 700 (K-700) \cite{carreira2019short} variant included 100 additional classes in order to both enlarge the dataset to a greater number of classes as well as to deal with the approximate 5\% size reduction per year of the dataset based on YouTube video availability. It is also worth noting that this reduction does not account for region-restricted videos and thus reductions do vary across countries. The latest version \cite{smaira2020short} does not include additional classes from the later 700. It instead aims to maintain the number of total clips to around 650K as approximately 15K videos became unavailable. The average video duration is $\sim 10$ seconds for all variants.

% Our use
We use three of the Kinetics variants including K-400 and K-700 with both the 2019 and 2020 versions. Our main comparisons with models from the literature are done over the 400 variant as it contains the largest number of results to compare to. The additional experiments on the 700 (2019) variant in \Cref{ch4} and (2020) in \Cref{ch5} are performed in order to include experiments on datasets with increased number of classes and variations in videos. We do not distinguish results between the 2019 and 2020 versions as the difference only amounts to approximately 15K different videos between the two datasets, which is about 2\% of the dataset. In comparison to the 400 variant, K-700 is 2 times larger in terms of number of clips.

\subsection{Moments in Time (MiT)}
\label{ch:3::sec:datasets::sub::mit}

% About
The Moments in Time (MiT) \cite{monfort2018moments} is built on the same principles as Kinetics with the introduction of a collection of one million temporal action segments. The clips have durations of 3 seconds and can include both human and non-human actions. In comparison to Kinetics, the dataset is significantly more diverse in terms of both the 339 labels that are included, as well as the intra-class video variations that are observed. MiT also includes a significantly higher number of videos per action class in comparison to Kinetics with approximately 2.4K clips for each action class. Because of the diverse nature of MiT, in terms of both labelling and videos, class prediction is considered to be more difficult than other large-scale datasets that include more fine-grained action labels \cite{caba2015activitynet,goyal2017something,karpathy2014large,kay2017kinetics, sigurdsson2016hollywood}. 

% Our use
We use the MiT dataset in our experiments in order to include additional results on a large and highly variant dataset, apart from Kinetics. An example of how class labels of different complexity levels result in large intra-class video variations can be seen for class \textit{\enquote{opening}} which may include clips of opening doors, curtains, eyes or mouths. These differences in notions for class verbs can provide challenges in the visual recognition of a class.

\subsection{Human Action Clips and Segments (HACS)}
\label{ch:3::sec:datasets::sub::hacs}

% About
The Human Action Clips and Segments (HACS) \cite{zhao2019hacs} dataset includes two variants. The first is the \textit{Clips} variant that uses 200 action labels over 1.5M 2-second clips. These labels include clips from videos that contain both the underlying action classes (labelled as positive) and those that do not include the labelled action classes (labelled negative). All clips have been sourced from approximately 500K YouTube videos. There are about 482K positive clips while the rest are clips in which no actions are performed. Both the positive and negative segments are similar in terms of the actors, backgrounds and objects that may be used, but differ in terms of performing or not performing the corresponding action. The second annotation type, named \textit{Segments}, consists of temporal segments used for temporal action localisation. The HACS dataset is currently one of the largest datasets for human action recognition with approximately $2.4$ times the number of clips per class compared to K-700. This significantly helps accurate model benchmarking and makes HACS a suitable pre-training dataset.

% Our use
In our experiments in \Cref{ch4,ch5,ch6}, we use HACS as a pre-training dataset and as a dataset for benchmarking. Because of the large number of clips per class, the dataset is ideal for both tasks which allows us not only to pre-train the convolutional weights of our models but also produce benchmarks in comparison to models from the literature. 
    
    \author
{}
\title{Improving Action Recognition through Time-Consistent Features}

\maketitle
\label{ch4}

In this chapter we explore the challenges of modelling temporal performance variations across human actions and interactions in video sequences. Our focus for addressing the motion variations of actions is aimed at the use of a novel method named \textit{Squeeze and Recursion Temporal Gates}, that aligns the temporally short and spatially local feature patterns, and the motions that they represent within the context of the entire video sequence \footnote{The code corresponding to this chapter's method is available at: \url{https://git.io/JfuPi}} .

\startcontents[chapters]
%\printcontents[chapters]{}{1}{\section*{\contentsname}}

\section{Introduction}
\label{ch:4::sec:intro}

% Problem definition
Despite the significant progress in the state-of-the-art models for human action and interaction recognition, notable challenges in capturing temporal variations are still present. Problems such as inconsistencies in temporal durations of actions, differences in the performed sets of movements, as well as changes in appearance based on viewpoint, remain either partially, or not at all addressed. These factors can significantly impact performance of models with the accuracy fluctuating from class to class given intra-class dissimilarities.

% Principles that define action dissimilarities
Human actions and interactions can vary based on the actors or settings that they are preformed in. Even though the underlying action class in different action instances remains the same, it present notable degrees of variation, because the action identity \cite{vallacher1989levels} that is associated with the main actor of each instance is different. These variations are based on the interpretation that each actor has for an action or the different skill levels of actors in different instances. The varying degree of complexity of the labels assigned to an underlined action constitutes the creation of a common set of difficult to define movements.  The ambiguity of a common measure for identifying each action is reflected by the movements performed and their overall temporal length. Thus, although the action can be considered as variations of the same action class, their temporal durations that are relevant to the action are notably different. Specifically, addressing these temporal variations not only provides a broader set of features, that better correspond to a large number of intra-class instances, but also improves the overall prepotent identity \cite{vallacher1989levels} that is expected to be learned for that class.{\parfillskip0pt\par}

\begin{wrapfigure}{r}{5em}
%\begin{mdframed}
\vspace{-1em}
    \includegraphics[width=5em]{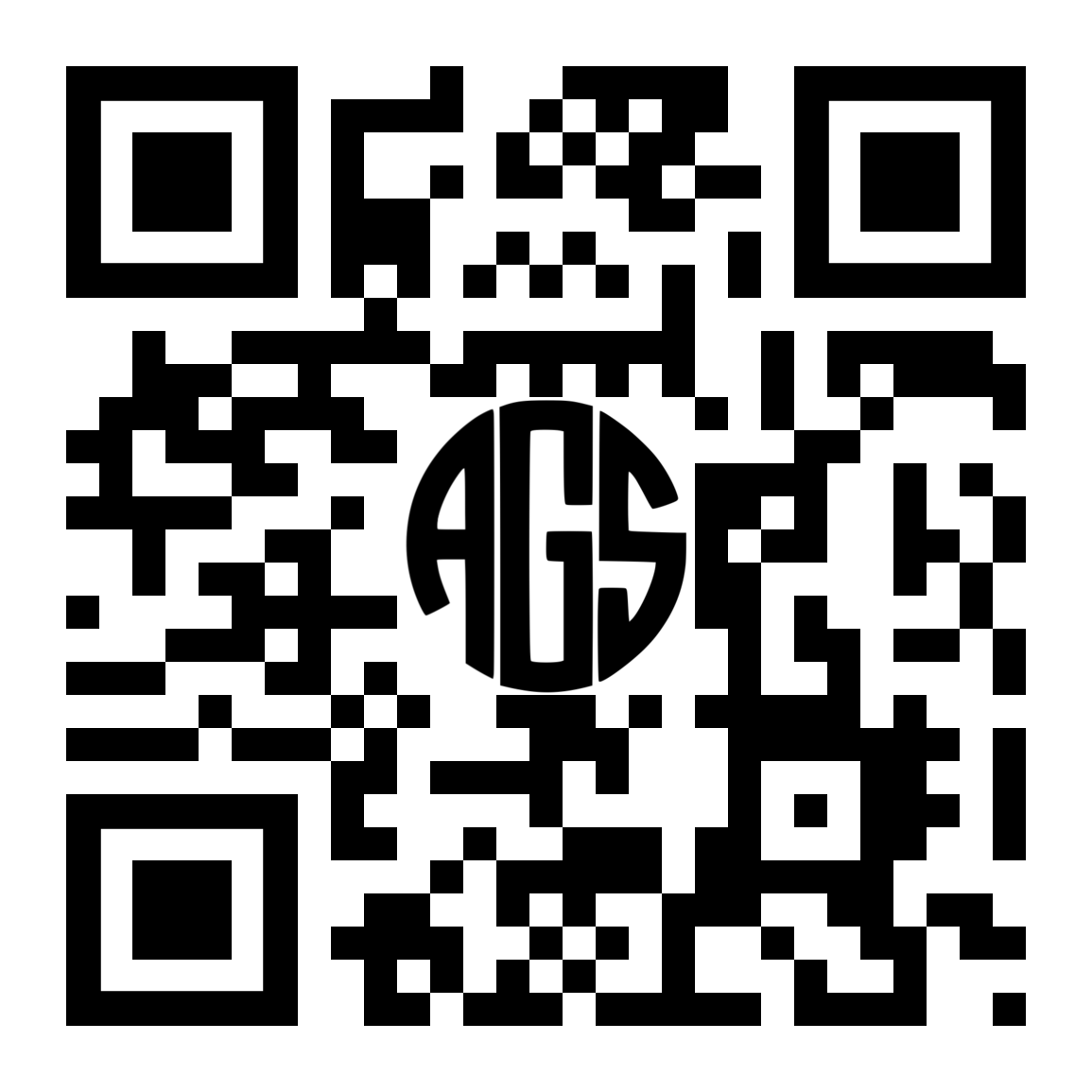}
%\end{mdframed}
\vspace{-4em}
\end{wrapfigure}
\noindent An approach based on which temporal information of each instance can be studied through attention. With the use of temporal attention, it is possible to study the relevance of temporal patterns over varying durations, with respect to the action classes that are performed. 

% Our focus for this chapter
Considering how differences in the temporal length and performance of an action are connected with the action's prepotent identity, we focus on creating a feature alignment between temporally short motion patterns and their extended counterparts across the entire video sequence. The proposed method named Squeeze and Recursion Temporal Gates (SRTG) uses the extracted convolutional features over constrained spatio-temporal locations and incorporates their general relevance across the entire temporal extent of the input. The created volume, that additionally encapsulates the dynamics of the local features with respect to the entire video sequence, is compared to the strictly local features. Through strengthening highly similar features across the two feature volumes, a generalised variant of the local patterns can be created. The resulting activation volume from feature fusion can increase the temporal locality and extend representations to incorporate feature changes across temporal length variations. 

% Chapter format
In \Cref{ch:4::sec:related} we review approaches that are based on the fusion of feature attention and local convolutional features. In \Cref{ch:4::sec:methodology} we provide a complete description of our proposed Squeeze and Recursion Temporal Gates. Our environment setting alongside our experiment results are presented in \Cref{ch:4::sec:results}. Fine-tuning experiments of our methods are evaluated in \Cref{ch:4::sec:TL}. We discuss our results and conclude in \Cref{ch:4::sec:discussion}.

\section{Attention fusion for convolutional features}
\label{ch:4::sec:related}
We identify two main groups of approaches that incorporate attention based on the extracted convolutional features. The first group of methods is based on the use of attention as a mask that can be applied over the input. The second category considers feature-based approaches that fuse attention between locally extracted patterns with globally informative features.

\subsection{Feature mask through attention}
\label{ch:4::sec:related::sub:mask}

% Mask-based image approaches
Initial attempts for masking convolutional features, in image tasks, were based on the creation of Soft-attention \cite{chen2016attention, jaderberg2015spatial} that weighs pixels across regions. The notion of attention was then later explored by Wang \etal~\cite{wang2017residual} where they proposed a separate \textit{Attention Module}. The Attention Module is based on an encoder-decoder structure that creates the corresponding feature mask which can relate to either low-level or high-level features, based on the complexity of the layer that the module is applied to. Similarly, Chen \etal~\cite{chen2017sca} used feature attention both spatially and channel-wise. This resulted in the creation of masks that not only corresponded to spatial locations of objects within the input images, but also solely addressing the features that relate to these objects. Later works have been explored that addressed both the use of recurrent connections across the encode-decoder attention structure \cite{zhang2018progressive}, spatial and global feature attention across residual encoder-decoder connections \cite{sindagi2019ha} and incorporation of localisation \cite{sun2019saliency}.

% Approaches for videos
In the video domain, works of Shikhar \etal~\cite{sharma2016action} create a (soft) attention mechanism by extracting spatial (2D) convolution features and use each frame instance as input to a recurrent network. Girdhar and Ramanan \cite{girdhar2017attentional} proposed a dual attention module to address class-agnostic and class-specific informative regions. Chen \etal~\cite{chen20182} also use a dual attention block inside a CNN where the first attention branch selects the features that best correspond to the video objects, while the second branch highlights their spatio-temporal locations across the video sequence. Other works that consider spatio-temporal attention have been based on the feature activation in a specific spatio-temporal location, in relation to the adjacent positions. This approach has been named \textit{Non-local neural networks} \cite{wang2018non} which captures dependencies across couples of adjacent space-time positions. 

% Our method - Short and Long kernels
Despite the great promise that these methods have shown for the extraction of robust spatio-temporal features, there is still a lack of explicitly addressing the locality of the extracted convolutional spatio-temporal features within the context of the entire video. Most of the works either focus on the creation of masks as soft-spatio-temporal-attention \cite{chen20182,girdhar2017attentional,sharma2016action} or the study of feature dependencies across pairs of space-time locations \cite{wang2018non}. There is room for improvement by exploring the relevance of local features within the context of the entire video sequence.

\subsection{Action recognition through attention-based calibration}
\label{ch:4::sec:related::sub:attention}

% Attention methods in image recognition
Hu \etal~\cite{hu2018squeeze} have investigated how cross-channel dependencies in convolutional layers can be used to emphasise specific image-based features. Their proposed \textit{Squeeze and Excitation} block uses a vectorised average version of convolutional activations, created through the squeeze function, which act as a global channel descriptor. The created down-sampled version of the activation map is then used by the excite method and utilised as a non-linear gating technique to produce a weighted activation map. The per-channel weighted map is then applied to the original features to create feature-calibrated activations. Based on the use of global information, additional works include \textit{Gather and Excite} \cite{hu2018gather}, which uses a regional-based version of Squeeze and Excitation, and \textit{Point-wise Self Attention} \cite{zhao2018psanet} which can connect feature map regions to create self-adaptive attention. Other works have been based on the use of residual connections with \textit{Bottleneck Attention Modules} (BAM) \cite{park2018bam} creating attention maps across the spatial and channel dimensions of the module inputs. BAM follows the notion of utilising global information \cite{hu2018squeeze,hu2018gather,zhao2018psanet} for discovering channel attention, while spatial attention is discovered by convolving the original input. Variations such as CBAM \cite{woo2018cbam} consider the creation of channel-based and spatial-based attention individually.

% Videos
In the field of video recognition, one of the first attempts for using attention as a method for calibration, was by Long \etal~\cite{long2018attention}. In their work, attention was expressed in the form of a per-channel condensed vector similar to that of \textit{Gather and Excite} and was used to cluster channels within the activation map. The clustering procedure was done for RGB frame features, convolutional features from optical flow and audio inputs. Clusters were then concatenated to create general descriptors for all the mentioned combinations of input types. Other approaches have drawn inspiration from non-local mean denoising operations \cite{buades2005non} using a non-local averaging over images and extend it to videos. Qiu \etal~\cite{qiu2019learning} have proposed the overall creation of an additional stream as a global attention path that can be updated by backpropagation.

% Our method -- distinction
Our proposed attention-calibration method (SRTG) differs from current video-based approaches as our aim is to directly address the divergence between short-term motions, through the exploration of coherence of the extracted convolutional patterns across the entire duration of the video segment. As CNN layer activations are constrained by the locality of their receptive fields, our method attempts to calibrate the locally-learned spatio-temporal patterns with the general motion patterns. This calibration is done by reconsidering the temporal attention of features across the entire video sequence and the correspondence to the relevance of the locally extracted patterns.

\section{Squeeze and Recursion Temporal Gates (SRTG)}
\label{ch:4::sec:methodology}

% Section outline
In this section, we provide a formal description of the proposed Squeeze and Recursion Temporal Gates (SRTG) blocks in \Cref{ch:4::sec:methodology::subsec:SR}. We additionally describe the criteria for calculating the feature alignment between the locally extracted convolutional features and the global information activations in \Cref{ch:4::sec:methodology::subsec:CC}. In \Cref{ch:4::sec:methodology::subsec:srtg_congig} we overview the possible configurations within Residual blocks. 

% Notations
Formally, layer activations are denoted as $\textbf{a}_{(C \! \times \! T \! \times \! H \! \times \! W)}$ with $C$ channels, $T$ frames, $H$ height and $W$ width, respectively. The backbone blocks, that SRTG is applied to, include residual connections, so we therefore formulate the final accumulated activations $a^{[l]}$ as the sum of the previous block activations $a^{[l-1]}$ and the current computed features $z^{[l]}$ ($a^{[l]} = z^{[l]} + a^{[l-1]}$). We use block indices based on $l$.

\subsection{Squeeze and Recursion}
\label{ch:4::sec:methodology::subsec:SR}

% Squeeze and Recursion formulation
Squeeze and Recursion (SR) can be used in conjunction with any produced spatio-temporal activation $\textbf{a}^{[l]} = g(\textbf{z}^{[l]})$ given a non-linear activation function $g()$ applied to a volume of convolutional features $\textbf{z}^{[l]}$. The process for global information creation is similar to that in Squeeze and Excitation \cite{hu2018squeeze} for images. For each SR block input, activation maps are averaged at their spatial dimension to create a vectorised temporal feature descriptor of the original volume. Therefore, the produced vector encapsulates average feature activation values of each frame squeezed, by their spatial size. This is used to find an average temporal attention for each feature.

% RNNs
\textbf{Recurrent cells}. We use a recurrent sub-network in order to determine the temporal feature importance of each channel of the vectorised input activation map $pool(\textbf{a}^{[l]})_{(t)}$. The sequential structure of recurrent cells allows for the discovery of features that are generally informative over temporal video sequences. We primarily consider LSTM cells \cite{hochreiter1997long} and provide a brief description of their inner workings as SR sub-network.

\begin{figure}[ht]
\centering
\includegraphics[width=0.85\linewidth]{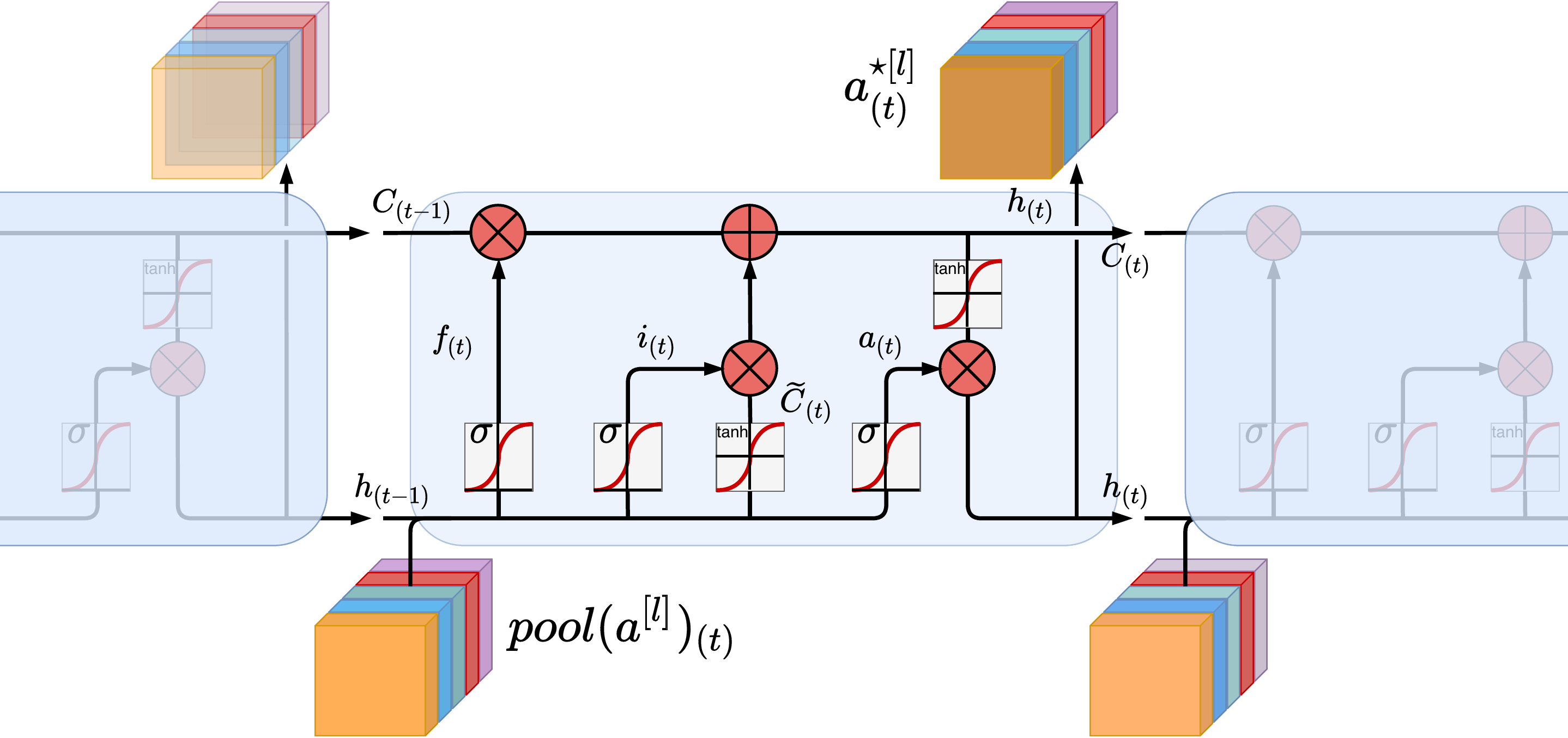}
\caption{\textbf{LSTMs cells}. Overview of a sequential chain of recurrent cells for the discovery of globally informative local features. Each input corresponds to a temporal activation map and produces a feature vector of the same size as the input.}
\label{fig:lstm}
\end{figure}

% Discarding features through  "forget gate layer" and storing features with "input gate layer".
\textbf{LSTM sub-net configuration}. A visual example of the sub-network is shown in \Cref{fig:lstm}. The effects of salient features are emphasised within the first operation of the recurrent cell with the \textit{forget gate layer} ($\textbf{f}_{(t)}$), where low intensity activations are discarded. Given the vectorised input $pool(\textbf{a}^{[l]})_{(t)}$, a decision is made with the inclusion of informative features from the previous frame $\textbf{h}_{(t-1)}$ with the use of channel weight $\textbf{w}_{f}$ and bias $\textbf{b}_{f}$. Features to be stored are discovered in two parts. First, the product of the sigmodial ($\sigma$) \textit{input gate layer} $\textbf{i}_{(t)}$, which determines the values that are to be updated. At the same time, the vector of candidate values $\widetilde{\textbf{C}}_{(t)}$ is created as:

\begin{align}
    \textbf{f}_{(t)} =\{ \sigma(\textbf{w}_{f} * [\textbf{h}_{(t-1)},pool(\textbf{a}^{[l]})_{(t)}] + \textbf{b}_{f}) \} \quad \label{eq:forget_gate} \quad\\
    \textbf{i}_{(t)} =\{ \sigma(\textbf{w}_{i} * [\textbf{h}_{(t-1)},pool(\textbf{a}^{[l]})_{(t)}] + \textbf{b}_{i}) \} \quad \label{eq:input_gate} \quad\\
    \widetilde{\textbf{C}}_{(t)} = \{ tanh(\textbf{w}_{C} * [\textbf{h}_{(t-1)},pool(\textbf{a}^{[l]})_{(t)}] + \textbf{b}_{C}) \} \label{eq:candidate_values} \quad
\end{align}

% Cell state updates
Updates to the previous cell state $\textbf{C}_{(t-1)}$ are done by initially deciding the features that are found inconsistent across time and that will be omitted ($\textbf{f}_{(t)} * \textbf{C}_{(t-1)}$). The second part of the update procedure consists of the forget and input gates which determine the new candidate values ($\textbf{i}_{(t)} * \widetilde{\textbf{C}}_{(t)}$), with the current cell state $\textbf{C}_{(t)}$ computed as:

\begin{equation}
\label{eq:cell_state}
    \textbf{C}_{(t)} = \textbf{f}_{(t)} * \textbf{C}_{(t-1)} + \textbf{i}_{(t)} * \widetilde{\textbf{C}}_{(t)}
\end{equation}

% Cell output
The final output of the recurrent cell $\textbf{h}_{(t)}$, for temporal state ($t$), is the combination of current cell state $\textbf{C}_{(t)}$, the previous hidden state $\textbf{h}_{(t-1)}$, and current input $pool(\textbf{a}^{[l]})_{(t)}$. The output is filtered by a sigmoid layer ($\textbf{o}_{(t)}$) which determines the part of the input to be updated:

\begin{equation}
\begin{split}
\label{eq:lstmout}
\textbf{h}_{(t)} = \textbf{o}_{(t)} * tanh(\textbf{C}_{(t)}), where \qquad \qquad \\
\textbf{o}_{(t)} =\{ \sigma(\textbf{w}_{o} * [\textbf{h}_{(t-1)},pool(\textbf{a}^{[l-1]})_{(t)}] + \textbf{b}_{o}) \}
\end{split}
\end{equation}

The produced hidden states ($\textbf{h}_{(t,\forall t \in T)}$) are concatenated in a coherent sequence of filtered spatio-temporal feature activations and used conjointly with the original input ($\textbf{a}^{[l]}$) through an element-wise multiplication operation. The produced activation map ($\textbf{a}^{\star[l]}$) effectively incorporates the global feature dynamics for the discovered features of different spatio-temporal region sizes.

\subsection{Cyclic Consistency}
\label{ch:4::sec:methodology::subsec:CC}

\begin{figure}[ht]
\centering
\includegraphics[width=0.9\linewidth]{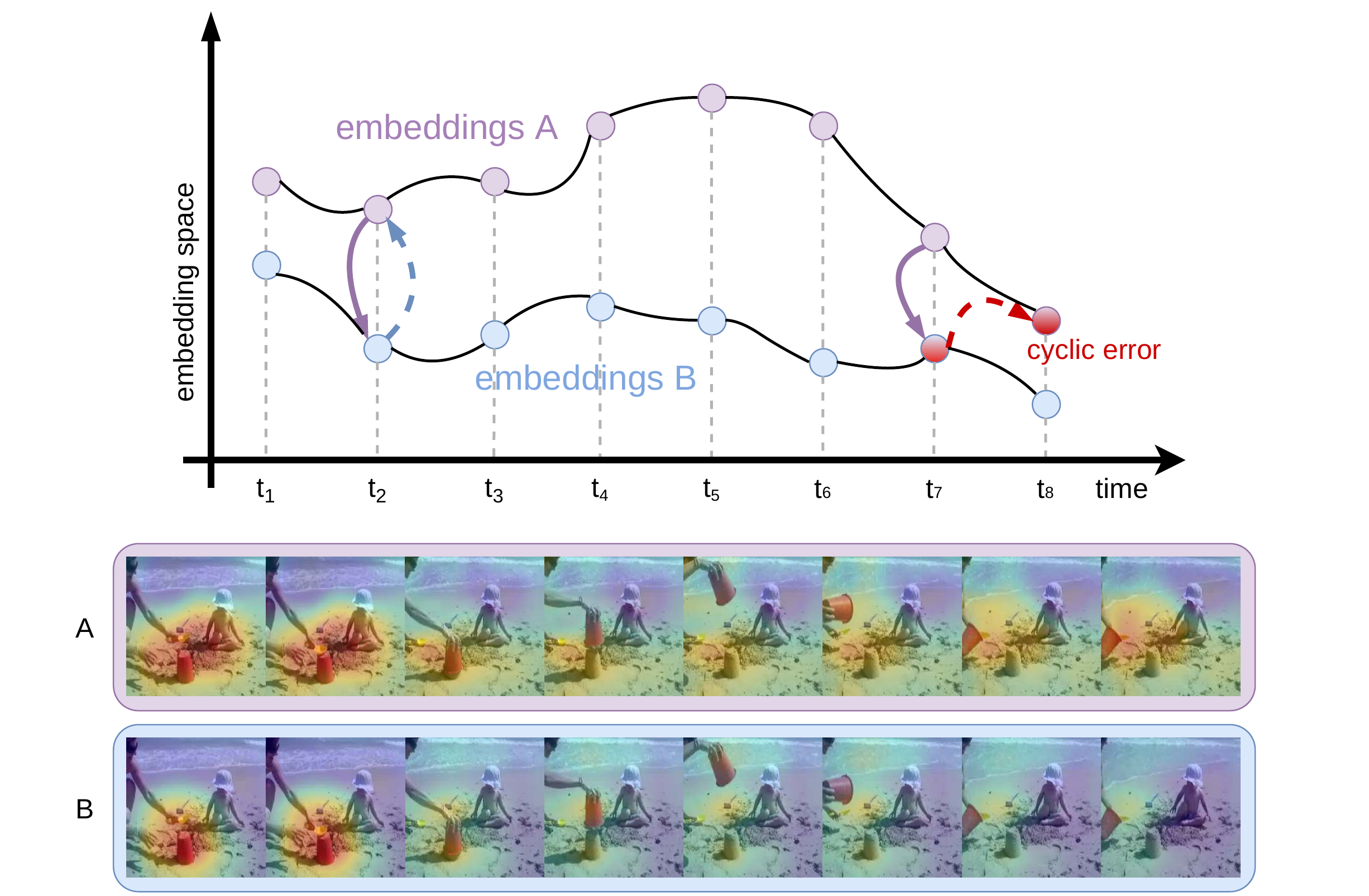}
\caption{\textbf{Temporal cyclic error.} Cycle-consistent points should cycle back to themselves (as with points at $t_{2}$). Points that do not present this trait, exhibit a temporal cyclic error (e.g. at $t_{7}$). Salient areas for each embedding are visualised with \textit{CFP} \cite{stergiou2019class}.}
\label{fig:Cyclic_consistency}
\end{figure}

% Cyclic Consistency definition.
 Supplementary to SR, cyclic mapping of each temporal instance is a widely used method \cite{dwibedi2019temporal,wang2019learning} to evaluate the similarity between pairs of temporal sequences. The basic premise considers the per-temporal location one-to-one mapping within two time sequences. We present a symbolic representation of the main idea in \autoref{fig:Cyclic_consistency}. We define each feature space of the two temporal sequences that are considered as an \textit{embedding space}. The defined embedding spaces are cycle-consistent if and only if each point at temporal instance $t$ in the embedding space \textbf{\textit{A}}, has a minimum distance point in embedding space \textbf{\textit{B}} at time $t$. Respectively, point $t$ in embedding space \textbf{\textit{B}} is required to also have a minimum distance point in embedding space \textbf{\textit{A}} at time $t$. We demonstrate how two points do not exhibit cyclic consistency in \Cref{fig:Cyclic_consistency}. In this case a temporal cyclic error occurs.

% The importance of cyclic consistency.
Through the use of cyclic-back consistency between points of two volumes, we can create a baseline in terms of their overall similarly. Considering the expected differences between individual features within the two feature spaces, they incorporate an overall similarity given a cyclic consistency alignment. Therefore, cyclic consistency is a suitable method to measure the (temporal) variations between embeddings.

% Cyclic soft-nearest neighbour
\textbf{Soft nearest neighbour distance.} Considering the vastness of embedding spaces in convolutional activations, the creation of a meaningful similarity measure is challenging. This challenge includes the selection of the nearest points in adjacent embedding spaces from points of a different space. The discovery of minimum distance spaces across the two representations is informative in terms of their feature correspondence. For this reason, \textit{soft matches} of points in projected embeddings \cite{goldberger2005neighbourhood} are preferred. The mindset behind soft matching is the selection of a point in an embedding space through the weighted sum of all possible matches with higher weights for points closer by. The closest actual observation is then selected based on its distance to the soft match point.

\begin{figure}[ht]
\centering
\includegraphics[width=\textwidth]{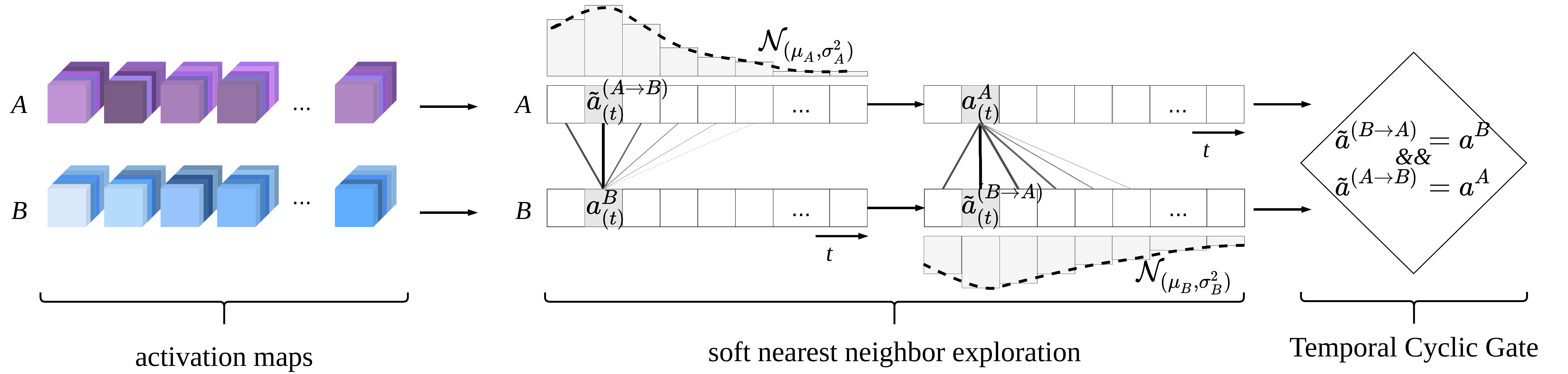}
\caption{\textbf{Fusion with Temporal Gating.} Activations ($\textbf{a}_{(t_i)}^{\textbf{\textit{B}}}$) from embedding space \textbf{\textit{B}} are compared to the activations ($\textbf{a}_{(t_j)}^{\textbf{\textit{A}}}$) in embedding space \textbf{\textit{A}}. Each frame-wise activation map ($\textbf{a}_{(t)}^{\textbf{\textit{B}}}$) is calculated based on its soft nearest neighbour ($\widetilde{\textbf{a}}^{\textbf{\textit{A}} \rightarrow \textbf{\textit{B}}}_{(t)}$) in space \textbf{\textit{A}}. Afterwards, $\widetilde{\textbf{a}}_{(t)}^{\textbf{\textit{B}} \rightarrow \textbf{\textit{A}}}$ in embedding space \textbf{\textit{B}} is also calculated. The gate is open when $\widetilde{\textbf{a}}^{\textbf{\textit{A}} \rightarrow \textbf{\textit{B}}}$ and $\widetilde{\textbf{a}}^{\textbf{\textit{B}} \rightarrow \textbf{\textit{A}}}$ are exactly and sequentially equal to $\textbf{a}^{\textbf{\textit{A}}}$ and $\textbf{a}^{\textbf{\textit{B}}}$.}
\label{fig:TCG_pipeline}
\end{figure}

% Main methodology
The soft nearest neighbour of an activation $\textbf{a}_{(t)}^{\textbf{\textit{A}}}$ in embedding space \textbf{\textit{B}} is discovered through the euclidean ($L_{2}$) distances between observation $\textbf{a}_{(t)}^{\textbf{\textit{B}}}$ and all points in \textbf{\textit{B}}. This process considers each frame as a separate instance for which we want to find the minimum distance point in the adjacent embedding space. We weigh the similarity to each point in adjacent embedding space \textbf{\textit{B}} to activation $\textbf{a}^{\textbf{\textit{A}}}_{(t)}$ using the softmax exponential difference between all activation pairs:

\begin{equation}
    \label{eq:soft_nn}
    \widetilde{\textbf{a}}^{(\textbf{\textit{B}} \rightarrow \textbf{\textit{A}})}_{(t)} = \sum_{i}^{T} \textbf{z}_{(i)}*\textbf{a}^{\textbf{\textit{B}}}_{(i)}, \; where: \quad \textbf{z}_{(i)} = \frac{e^{-||\textbf{a}^{\textbf{\textit{A}}}_{(t)} - \textbf{a}^{\textbf{\textit{B}}}_{(i)}||^{2}}}{\sum \limits_{i}^{T} e^{-||\textbf{a}^{\textbf{\textit{A}}}_{(t)} - \textbf{a}^{\textbf{\textit{B}}}_{(i)}||^{2}}}
\end{equation}

% From soft nearest neighbour to embedding 2 space frames
The softmax function creates a normal distribution of similarity weights $\mathcal{N}(\mu,\,\sigma^{2})\,$, centred on the adjacent space activation at the time instance with the minimum distance from activation $\textbf{a}^{\textbf{\textit{A}}}_{(t)}$. With the calculation of the nearest neighbour $\widetilde{\textbf{a}}^{(\textbf{\textit{B}} \rightarrow \textbf{\textit{A}})}_{(t)}$, the distance to the nearest frames in \textbf{\textit{B}} can be computed. Based on the initially considered frame $\textbf{a}^{\textbf{\textit{A}}}_{(t)}$, we obtain the closest time instance activations, through the selection of the minimum L2 distance from the found soft match: 

\begin{equation}
    \label{eq:soft_nn_to_emb2}
    \textbf{a}^{(\textbf{\textit{B}} \rightarrow \textbf{\textit{A}})}_{(t)} = \operatorname*{argmin}_{i}(||\widetilde{\textbf{a}}^{(\textbf{\textit{B}} \rightarrow \textbf{\textit{A}})}_{(t)} - \textbf{a}^{\textbf{\textit{B}}}_{(i)}||^{2}) 
\end{equation}

% Establishing cyclic consistency
Temporal embedding points are \textit{consistent} if and only if the initial temporal location $t$ matches precisely the temporal location of the discovered point in adjacent embedding space \textbf{\textit{B}}, $\textbf{a}^{(\textbf{\textit{B}} \rightarrow \textbf{\textit{A}})}_{(t)} = \textbf{a}^{\textbf{\textit{B}}}_{(t)} \: \forall t \in \{1,...,T\}$, as visualised in \Cref{fig:TCG_pipeline}. To establish a consistency check for temporal points of space \textbf{\textit{A}}, the same procedure is repeated with the consideration of every frame in embedding space \textbf{\textit{B}}, by calculating the soft nearest neighbour in \textbf{\textit{A}}. The embeddings are considered \textit{cycle consistent} if and only if all points on both embedding spaces map back to themselves: $\textbf{a}^{(\textbf{\textit{B}} \rightarrow \textbf{\textit{A}})}_{(t)} = \textbf{a}^{\textbf{\textit{B}}}_{(t)} \; and \, \textbf{a}^{(\textbf{\textit{A}} \rightarrow \textbf{\textit{B}})}_{(t)} = \textbf{a}^{\textbf{\textit{A}}}_{(t)} \: \forall t \in \{1,...,T\}$.

\begin{figure}[ht]
\centering
\includegraphics[width=.6\textwidth]{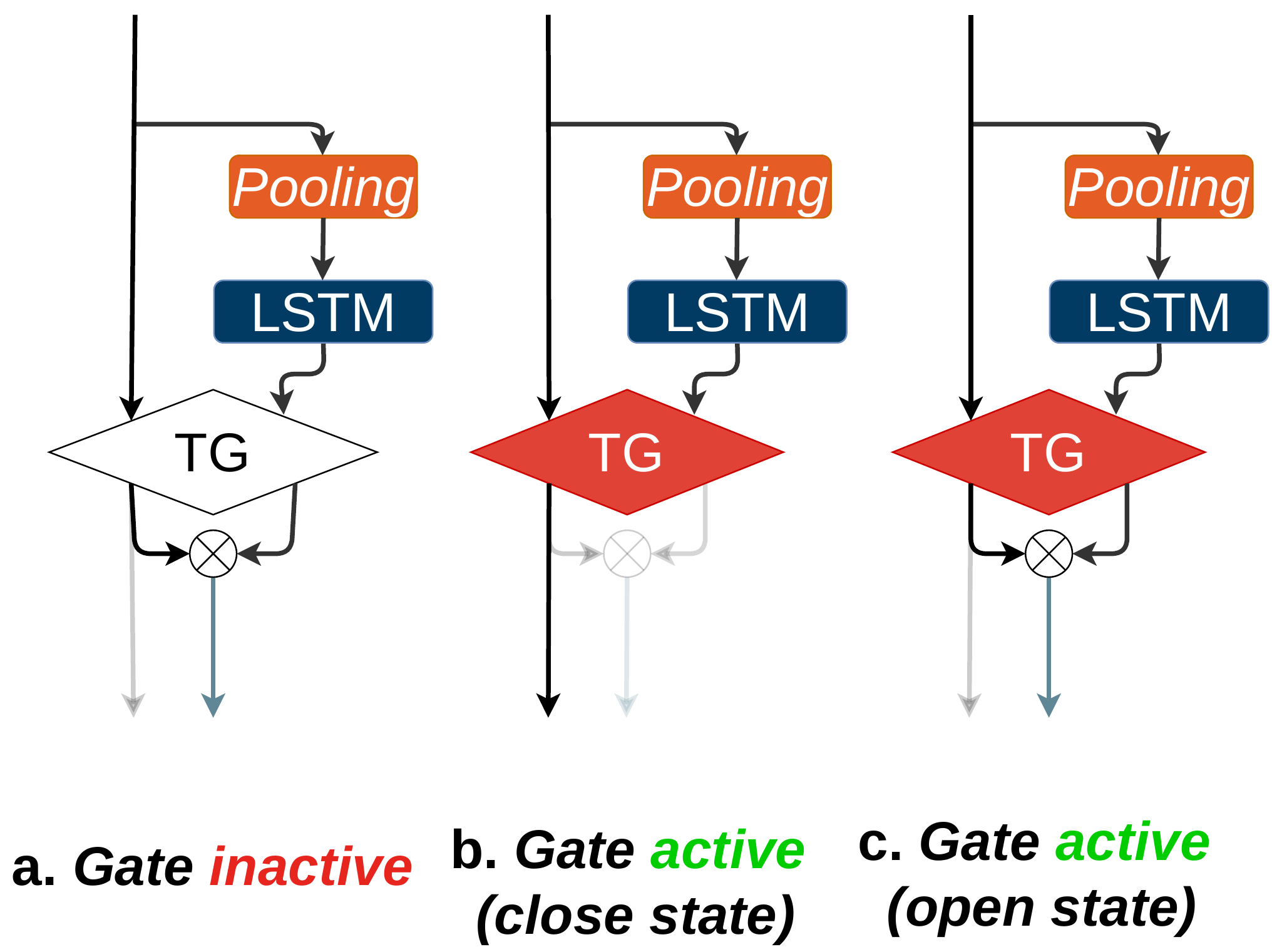}
\caption{\textbf{Temporal gate states}. (a) Inactive state of gate, where no cyclic consistency is calculated. (b) Gate in an active and closed state in which cyclic consistency is not established. (c) Active gate where the original activations and the calibrated ones are cyclic consistent and thus are fused together by element-wise multiplication ($\otimes$)\vspace{-2mm}}
\label{fig:TCG_states}
\end{figure}

% Globally and locally useful information through gating
\textbf{Temporal gates.} The averaged temporal feature attention is part of the produced activation vector. However, it does not necessarily correspond to a one-to-one similarity with the local spatio-temporal activations. The direct fusion of two activations without considering dissimilarities in terms of their representations can lead to unrepresentative volumes. We compute cyclic consistency between the pooled activations $pool(\textbf{a}^{[l]})$ and the outputted recurrent cells $\textbf{a}^{\star[l]}$ to evaluate the feature similarities between the two volumes. We utilise cyclic consistency as a gating mechanism to fuse the recurrent cell hidden states with unpooled versions of the activations when they are temporally cycle consistent. This ensures that only time-consistent information of the local patterns is added back to the network. An overview of the gate states is shown in \cref{fig:TCG_states}.

\subsection{SRTG configurations}
\label{ch:4::sec:methodology::subsec:srtg_congig}

\begin{figure}[t]
\includegraphics[width=\textwidth]{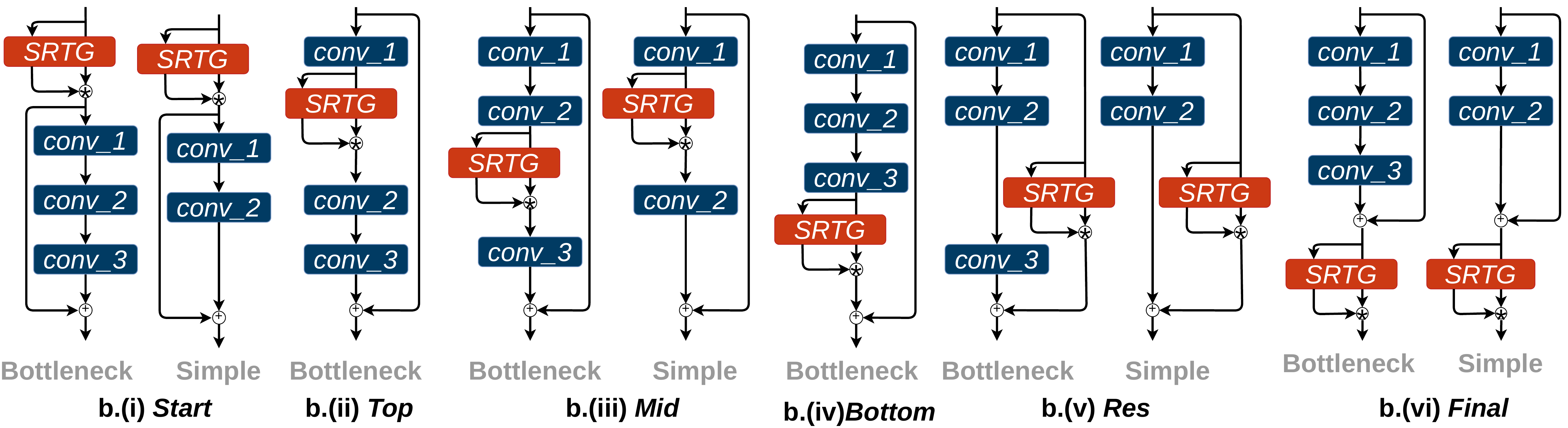}
\caption{\textbf{SRTG configurations} based on residual blocks. Residual Networks \cite{he2016deep} can include multiple SRTG configurations based on the use of either Simple blocks with two convolutional operations or Bottleneck blocks with three convolutional operations.}
\label{fig:srtg_pipeline}
\end{figure}

% Purpose of different configurations
Cycle-consistency calculations can be performed across multiple parts of a convolution block. Shown in \Cref{fig:srtg_pipeline}, we investigate six different approaches when testing SRTG as part of a convolution block. Each variant fuses global and local information in sections of the blocks. Configurations only differ in the relative locations of the SRTG and the LSTM input. Based on the residual version chosen, we consider Simple blocks with two conv operations and Bottleneck blocks with three conv operations. Not all SRTG configurations apply to the Simple blocks.

% Start configuration
\textbf{Start.} SRTG is performed on the block's input thus input information is aligned on both global and local information. This is used in both Simple and Bottleneck residual blocks.

% Top configuration
\textbf{Top.} Activations of the first convolution are used by the LSTM, with fused features being used by the final convolution. This is specific to Bottleneck blocks.

% Mid configuration
\textbf{Mid.} SRTG is added at the middle of Simple blocks and after the second convolution at Bottleneck blocks.

% End configuration
\textbf{End.} Local and global features are fused at the end of the final convolution, before the concatenation of the residual connection. This is only used in Bottleneck blocks.

% Residual configuration
\textbf{Res.} The SRTG module is applied to the residual connection. This transforms the residual connection to further include global spatio-temporal features combining them with convolutional activations from either Simple or Bottleneck blocks.

% Final configuration
\textbf{Final.} SRTG is added at the end of the residual block, which allows for the activations to be calculated jointly with their representations across time on the entire video. This can be used in both Simple and Bottleneck blocks.

\section{Main results}
\label{ch:4::sec:results}

For this section, we overview our experiment setting for training in \Cref{ch:4::sec:results::sub:settings}. The main results on the HACS dataset are presented in \Cref{ch:4::sec:results::sub:HACS}. Model comparisons on Kinetics-700 are shown in \Cref{ch:4::sec:results::sub:K700}. The final large-scale dataset on which we test the proposed method is Moments in Time and the results can be found in \Cref{ch:4::sec:results::sub:MiT}. Pairwise comparisons are summarised in \Cref{ch:4::sec:results::sub:ablation}.

\subsection{Experiment environment settings}
\label{ch:4::sec:results::sub:settings}

The proposed SRTG modules are evaluated on ResNet backbone architectures. The selection of the ResNet architecture was based on its wide application over multiple works in action recognition \cite{chen2019drop,chen2018multifiber,feichtenhofer2020x3d,feichtenhofer2019slowfast,tran2019video,tran2018closer}. To extend our comparisons, we consider both 3D and (2+1)D \cite{tran2018closer}, spatio-temporal convolutions over 3 architectures of different depths. For simplicity, we denominate ResNet architectures with 3D convolutions as r3d and with (2+1)D as r(2+1)d.

% Interval selection
\textbf{Interval selection.} The frame selection is performed by a uniform random sampling. Based on the average clip length, for each dataset, we use equivalently sized temporal strides. This is done in order to ensure that the frames across the majority of the video sequence will be used to create the input volume. The average clip length in HACS is 60 frames, and we therefore use a stride of 2. For Kinetics, the average clip length is 250 frames so we set the temporal stride to 5. Similarly, for MiT we use a temporal stride of 3 with average duration of 90 frames.

% GFLOPS x views
\textbf{Computational Inference.} We employ two different measures to report inference costs. We report computational costs (FLOPs) similar to works of \cite{fan2019more,feichtenhofer2020x3d,feichtenhofer2019slowfast,tran2019video} through sampling 10 clips from a single video and perform 3 crops along the spatial dimensions of size $256 \! \times \! 256$. The inference time is reported as the number of FLOPs used per spatio-temporal view (clips times crops). This provides a standardised measure of computing inference when comparing across models as shown for \Cref{table:HACS_accuracies_ch4,table:K700_accuracies_ch4,table:mit_accuracies_ch4}.

\begin{figure}[ht]
\centering
\includegraphics[width=\linewidth]{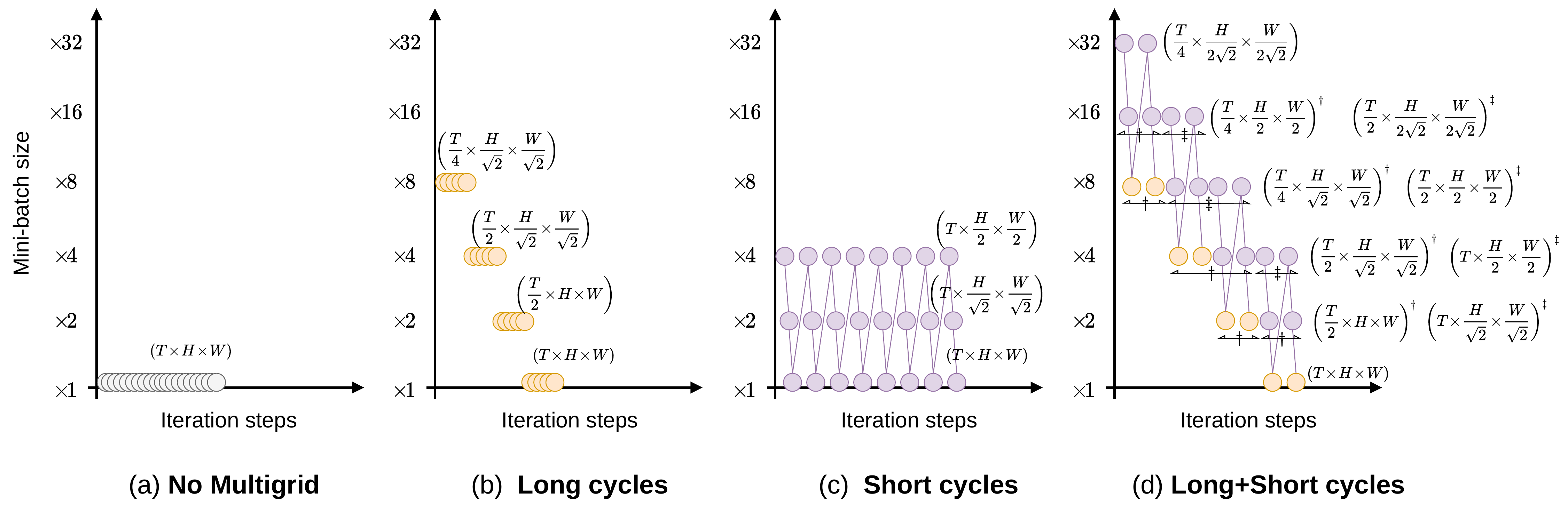}
\caption{\textbf{Multigrid training schedule.} (a) \textbf{Baseline} uses a fixed batch size. (b) \textbf{Multigrid long cycles} loop over 4 different batch sizes within an epoch. These changes are done in a sequential order. (c) \textbf{Multigrid short cycles} iterate over adjacent sizes with each iteration step re-adjusting the batch size. (d) \textbf{Multigrid long + short cycles} fuse both (b) and (c) together to a single schedule.}
\label{fig:multigrid}
\end{figure}

% Multigrid sampling
\textbf{Multigrid batch schedule}. All of our experiments utilise a multigrid scheme \cite{wu2020multigrid} for improvements in the training speeds. The method is based on variable-sized mini-batches created from a sampling grid of possible sizes. Scaling the mini-batch size is done with respect to the original batch and spatio-temporal dimensionality by satisfying $b \! \times \! t \! \times \! h \! \times \! w = B \! \times \! T \! \times \! H \! \times \! W$, in which ($b,t,h,w$) represent the scaled batch, time, height and width dimensions of the input data, while ($B,T,H,W$) are the original dimensions. This further ensures that computational costs (GFLOPs) remain similar between the scaled and the original batches. It is worth noting that multigrid considers a \textit{proportional} spatial scaling, thus increases and decreases in the height or width values are done in the same manner. We demonstrate the hierarchical schedule of multigrid in \Cref{fig:multigrid}.

% Long and Short cycles overview
Overall, the schedule alternates between two frequencies with \textit{long cycles} in which the batch size changes after a specified number of iterations and \textit{short cycles} that move to different mini-batch sizes at each iteration. In our experiments, long cycles iterate over 4 different batch sizes for a single epoch, with total input sizes of $\left( 8B \! \times \! \frac{T}{4} \! \times \! \frac{H}{\sqrt{2}} \! \times \! \frac{W}{\sqrt{2}} \right)$, $\left( 4B \! \times \! \frac{T}{2} \! \times \! \frac{H}{\sqrt{2}} \! \times \! \frac{W}{\sqrt{2}} \right)$, $\left( 2B \! \times \! \frac{T}{2} \! \times \! H \! \times \! W \right)$ and $\left( B \! \times \! T \! \times \! H \! \times \! W \right)$. Temporal reductions in frames are handled by uniform frame selection from the default sampled frames, while spatial reductions are done through bilinear interpolation. In contrast to long cycles, iterations in the short cycles take volume sizes of  $\left( 4B \! \times \! T \! \times \! \frac{H}{2} \! \times \! \frac{W}{2} \right)$, $\left( 2B \! \times \! T \! \times \! \frac{H}{\sqrt{2}} \! \times \! \frac{W}{\sqrt{2}} \right)$ and $\left( B \! \times \! T \! \times \! H \! \times \! W \right)$ which effectively only progressively reduce the spatial dimensions ($H,W$). Short cycles are combined with long cycles to form the batch size schedule that we used. This is based on a combination of sampling strategies as shown in \cref{fig:multigrid}. The learning rate follows the linear scaling rule \cite{goyal2017accurate} given the changes that are based on the long cycle. Reductions to batch sizes to half the size of the previous iteration will also result in a reduction by half for the learning rate. This is also done for increases in the batch size for which the learning rate is also increased by an equal amount.

% Params for SRTG nets
\textbf{Training}. Results on HACS are obtained with weights for all of our networks being randomly initialised based on the recipe described by He \etal~\cite{he2015delving}. Frames are resized to size $320 \! \times \! X$, where $X$ is the long side in the original frame, in order to maintain the aspect ratio. The frames are then cropped to $224 \! \times \! 224$ sizes. The base number of frames that we use is $16$. Both temporal and spatial dimensions are reduced/adjusted based on the defined multigrid mini-batch size schedule. The initial learning rate is set to $lr_{0}=0.1$ and reduced by 10 for steps [40, 70, 120] for a total of 150 epochs. We choose 150 total epochs as no further improvements were observed on higher number of iterations. We also use weight-decay which is set to $10^{-5}$. We use a standard SGD optimiser with momentum \cite{qian1999momentum} which we set to 0.9. The base mini-batch size is set to 32 and increased according to each cycle schedule. Structurally, we use a global spatio-temporal average pooling layer for feature vectorisation and a fully-connected layer for the class probabilities.

% Data augmentation
\textbf{Spatial data augmentation.} To improve the generalisation capabilities of our models, we further include spatial data augmentations, that are performed sequentially over frames in our data augmentation pipeline. We set a sequential probability of 80\% which translates to roughly 80\% of the data being further augmented after their spatial crops. A per-augmentation method probability of 40\%, which corresponds to the probability that each of the augmentation method has in order to be applied. Our spatial augmentations include Gaussian and mean blurring, RGB value changes ($\pm \{1,...,15\}$) and geometrical augmentations. Choices for augmentation methods was made based on \cite{wang2017effectiveness}.

% Table 1: Configuration comparison on HACS - res34
\begin{table}[!htb]
\caption{\textbf{Comparison of r3d-34 with SRTG configurations on HACS}.}
\centering
\resizebox{0.9\textwidth}{!}{%
\begin{tabular}{cccccc}
\hline
    \multirow{2}{*}{Config} &
    \multirow{2}{*}{Gates} &
      \multicolumn{2}{c}{top-1 (\%)} &
      \multicolumn{2}{c}{top-5 (\%)} \\
    & & 3D & (2+1)D & 3D & (2+1)D \\[.25em]
    \hline
    \hline
    No SRTG & \ding{55} & 74.82 & 75.70 & 92.84 & 93.57 \\[.1em]
    
    Start & \ding{51} & 75.70 \textcolor{applegreen}{($+0.88$)} & 76.44 \textcolor{applegreen}{($+0.74$)} & 93.23 \textcolor{applegreen}{($+0.39$)} & 93.78 \textcolor{applegreen}{($+0.21$)} \\[.1em]
    
    Mid & \ding{51} & 75.49 \textcolor{applegreen}{($+0.67$)} & 76.68 \textcolor{applegreen}{($+0.98$)} & 93.22 \textcolor{applegreen}{($+0.38$)} & 93.75 \textcolor{applegreen}{($+0.18$)} \\[.1em]
    
    Res & \ding{51} & 76.70 \textcolor{applegreen}{($+1.88$)} & 77.09 \textcolor{applegreen}{($+1.39$)} & 93.31 \textcolor{applegreen}{($+0.47$)} & 93.86 \textcolor{applegreen}{($+0.29$)} \\[.1em]

    Final & \ding{51} & \textbf{78.60} \textcolor{applegreen}{($+3.78$)} & \textbf{80.39} \textcolor{applegreen}{($+4.49$)} & \textbf{93.57} \textcolor{applegreen}{($+0.73$)} & \textbf{94.27} \textcolor{applegreen}{($+0.70$)} \\
\hline
\end{tabular}
}
\label{table:srtg_configurations}
\end{table}

\subsection{Results on HACS}
\label{ch:4::sec:results::sub:HACS}

% Configuration experiments
\textbf{SRTG module configurations}. Our initial comparisons are done over different SRTG module configurations with a 34-layer ResNet backbone. We use two networks with 3D convolutions (r3d-34) and (2+1)D convolutions (r(2+1)d-34). Architecturally, ResNet-34 contains Simple blocks with two conv layers instead of the Bottleneck blocks with three conv layers. We therefore only evaluate the Start, Mid, Res and Final configurations. Results for HACS are summarised in Table~\ref{table:srtg_configurations} and are obtained by training from scratch. We show that all of the tested SRTG module configurations perform better than the original network without SRTG. This demonstrates the merits of our more flexible treatment of the temporal dimension. This effect appears to be stronger when the filtering is applied later. Experiments with 3D convolutions show improvements in the range of 0.88--3.78\% for top-1 accuracy and 0.39--0.73\% for top-5 accuracy. The top performing configurations are the Final with +3.78\% top-1 and +0.73\% top-5 accuracies and Res with +1.88\% and 0.47\% top-1 and top-5 accuracies. Similar results are also obtained with (2+1)D convolutions, with accuracy increases in relation to the original network used as baseline, and range between 0.74--4.69\% for top-1 and 0.18--0.70\% for top-5 accuracies. Based on these results, following experiments use the final block configuration.

% Main results on HACS
\textbf{Main results}. In \Cref{table:HACS_accuracies_ch4} we present our results in comparison to both state-of-the-art models as well as against baseline ResNet models without SRTG modules. We also overview the number of parameters and the computational costs (GFLOPs) for each tested architecture.

% 3D convolutions
\textbf{3D convolutions}. For results obtained with 3D convolutions, we notice an average improvement of 2.3\% on the top-1 accuracy across all three configurations. For \textbf{SRTG r3d-34} this increase in accuracy is of 3.8\% for the top-1 and 0.8\% for the top-5. The achieved accuracies with SRTG are shown to be closer to larger corresponding networks without SRTG with \textbf{SRTG r3d-34} performing similar to r3d-50 and \textbf{SRTG r3d-50} accuracies being comparable to those of r3d-101. When considering state-of-the-art architectures, r3d-50 is close in performance to I3D \cite{carreira2017quo} while requiring less than half the number of floating-point operations (GFLOPs). \textbf{SRTG r3d-101} achieves comparable accuracies to TSM \cite{lin2019tsm} with margins of +0.2\% top-1 and 0.8\% for the top-1 and top-5.

% (2+1)D convolutions
\textbf{(2+1)D convolutions}. For SRTG networks that instead use (2+1)D convolutions, we notice a reasonable increase in accuracy within the ranges of 1.8--3.5\% for the top-1 and 0.5--1.1\% for the top-5 accuracies. However, this also translates to an increase in the number of GFLOPs, as (2+1)D convolutions use two operations with spatial-only kernels followed by temporal-only kernels. In terms of their accuracies compared to the baseline r(2+1)d networks without SRTG, increases are similar to those obtained with 3D convolutions. Improvement margins, across models, are between 1.3--4.7\% for top-1 and 0.5--1.1\% for top-5. The largest increase in performance is observed for \textbf{SRTG r(2+1)d-34} with +4.7\% top-1. The best performing model is \textbf{SRTG r(2+1)d-101} which however has significant computational and memory requirements. \textbf{SRTG r(2+1)d-50} shows similar performance to SlowFast-101 \cite{feichtenhofer2019slowfast} with only minimal increases in the number of GFLOPs and parameters. The smaller \textbf{SRTG r(2+1)d-34} can achieve accuracies comparable to the significantly larger r3d-101 and I3D while only using less than half of the number of GFLOPs of the other two models. This is also without being pre-trained to a different dataset.

% Accuracy w.r.t GFLOPs/params
\textbf{Computational complexity}. In \Cref{table:HACS_accuracies_ch4} the accuracies are presented with respect to the computational complexity of each model (GFLOPs) and the number of parameters. Compared to the original baseline models, SRTG does not significantly effect the number of floating-point operations of the entire network. The added number of GFLOPs amounts to $\ll 1\%$ of that of the original model making the our proposed block very lightweight to compute. The importance of this can be understood by the large number of GFLOPs that is already required for performing spatio-temporal 3D convolutions. We note that there is a significantly larger difference in the number of GFLOPs based on the space-time convolution used, with either 3D or (2+1)D than the inclusion of SRTG in the network.

\begin{table}[t]
\caption{\textbf{Action recognition model comparisons on HACS}. Weight initialisation sources are denoted by their respective indicators.}
\begin{threeparttable}[t]
\centering
\resizebox{.9\textwidth}{!}{%
\renewcommand{\arraystretch}{1.2} 
\begin{tabular}{c|c|c|c|c|c}
\hline
Model & 
Pre &
top-1 & top-5 & 
GFLOPs $\! \times \!$ views &
Params \\
\hline

MF-Net \cite{chen2018multifiber}$^{\dagger}$ &
\multirow{5}{*}{K-400} &
78.3 & 94.6 &
$11.1 \! \times \! 50$ &
8.0M \\

I3D \cite{carreira2017quo}$^{\dagger}$ &
 &
79.9 & 94.5 &
$108.0 \! \times \! 50$ &
12.0M \\

TSM \cite{lin2019tsm}$^{\dagger}$ &
 &
81.4 & 95.5 &
$65.0 \! \times \! 10$ &
24.3M \\

TAM \cite{fan2019more}$^{\dagger}$ &
 &
82.2 & 95.2 &
$86 \! \times \! 12$ &
25.6M\\

SF-101 \cite{feichtenhofer2019slowfast}$^{\dagger}$ &
 &
83.7 & \textbf{96.8} &
$213.0 \! \times \! 30$ &
53.7M \\
\hline

r3d-34 \cite{kataoka2020would}$^{*}$ &
\multirow{6}{*}{K-700} &
74.8 & 92.8 &
$26.6 \! \times \! 30$ &
63.7M\\

r3d-50 \cite{kataoka2020would}$^{*}$ &
 &
78.4 & 93.8 &
$52.6 \! \times \! 30$ &
36.7M\\

r3d-101 \cite{kataoka2020would}$^{*}$ &
 &
80.5 & 95.2 &
$78.0 \! \times \! 30$ &
69.1M\\

r(2+1)d-34 \cite{kataoka2020would}$^{*}$ &
 &
75.7 & 93.8 &
$37.8 \! \times \! 30$ &
61.8M\\

r(2+1)d-50 \cite{kataoka2020would}$^{*}$ &
 &
81.3 & 94.5 &
$83.3 \! \times \! 30$ &
34.8M\\

r(2+1)d-101 \cite{kataoka2020would}$^{*}$ &
 &
82.9 & 95.7 &
$163.0 \! \times \! 30$ &
67.2M\\
\hline

ir-CSN-101 \cite{tran2019video}$^{\dagger}$ &
\multirow{2}{*}{IG65} &
83.8 & 93.8 &
$63.6 \! \times \! 10$ &
22.1M \\

ip-CSN-101 \cite{tran2019video}$^{\dagger}$ &
 &
84.1 & 93.9 &
$63.6 \! \times \! 10$ &
24.5M \\
\hline

SRTG r3d-34 \cite{stergiou2021learn} \textbf{(ours)} &
- &
78.6 & 93.6 &
$26.6 \! \times \! 30$ &
83.8M \\

SRTG r3d-50 \cite{stergiou2021learn} \textbf{(ours)} &
- &
80.3 & 95.5 &
$52.7 \! \times \! 30$ &
56.9M \\

SRTG r3d-101 \cite{stergiou2021learn} \textbf{(ours)} &
- &
81.6 & 96.3 &
$78.1 \! \times \! 30$ &
107.1M \\

SRTG r(2+1)d-34 \cite{stergiou2021learn} \textbf{(ours)} &
- &
80.4 & 94.3 &
$37.8 \! \times \! 30$ &
82.1M \\

SRTG r(2+1)d-50 \cite{stergiou2021learn} \textbf{(ours)} &
- &
83.8 & 96.6 &
$83.4 \! \times \! 30$ &
55.0M \\

SRTG r(2+1)d-101 \cite{stergiou2021learn} \textbf{(ours)} &
- &
\textbf{84.3} & \textbf{96.8} &
$163.1 \! \times \! 30$ &
105.3M \\

\end{tabular}%
}
 \begin{tablenotes}
    \item[$\dagger$] models and weights from official repositories.  
    \item[$*$] re-implemented models and weights. 
   \end{tablenotes}
\end{threeparttable}%
\label{table:HACS_accuracies_ch4}
\end{table}

\subsection{Results on Kinetics-700}
\label{ch:4::sec:results::sub:K700}

\begin{table}[t]
\caption{\textbf{Comparison with K-700 state-of-the-art}. Flop calculation is similar to that in \Cref{table:HACS_accuracies_ch4}. $^{*}$ denotes our reproduced models.}
\centering
\resizebox{.85\textwidth}{!}{%
\renewcommand{\arraystretch}{1.5}
\begin{tabular}{c|c|c|cc}
\hline
Model & 
Pre-train &
GFLOPs $\! \times \!$ views &
top-1 & 
top-5 \\
\hline

I3D \cite{carreira2017quo} $^{*}$ &
K-600 &
$108 \! \times \! 50$ &
53.0 & 69.2 \\
\hline

TSM \cite{lin2019tsm} $^{*}$ &
\multirow{11}{*}{HACS} &
$65.0 \! \times \! 10$ &
54.0 & 72.2  \\

MF-Net \cite{chen2018multifiber} $^{*}$ &
 &
$11.1 \! \times \! 50$ &
54.3 & 73.4  \\

ir-CSN-101 \cite{tran2019video} $^{*}$ &
 &
$63.6 \! \times \! 10$ &
54.7 & 73.8  \\

SF-50 \cite{feichtenhofer2019slowfast} $^{*}$ &
 &
$65.7 \! \times \! 30$ &
56.2 & 75.7  \\

SF-101 \cite{feichtenhofer2019slowfast} $^{*}$ &
 &
$213.0 \! \times \! 30$ &
\textbf{57.3} & 77.2  \\\cline{1-1}\cline{3-5}

SRTG r3d-34 \cite{stergiou2021learn} \textbf{(ours)} &
 &
$26.6 \! \times \! 30$ &
49.1 & 72.7  \\

SRTG r3d-50 \cite{stergiou2021learn} \textbf{(ours)} &
&
$52.7 \! \times \! 30$ &
53.5 & 74.2  \\

SRTG r3d-101 \cite{stergiou2021learn} \textbf{(ours)} &
&
$78.1 \! \times \! 30$ &
56.5 & 76.8  \\

SRTG r(2+1)d-34 \cite{stergiou2021learn} \textbf{(ours)} &
&
$37.8 \! \times \! 30$ &
49.4 & 73.2 \\

SRTG r(2+1)d-50 \cite{stergiou2021learn} \textbf{(ours)} &
&
$83.4 \! \times \! 30$ &
54.2 & 74.6 \\

SRTG r(2+1)d-101 \cite{stergiou2021learn} \textbf{(ours)} &
&
$163.1 \! \times \! 30$ &
56.8 & \textbf{77.4} \\

\end{tabular}%
}
\label{table:K700_accuracies_ch4}
\end{table}

% K-700 intro
The selection of the Kinetics-700 dataset was done with the criteria of testing our proposed approach over an extended number of action classes to additionally demonstrate the generalisation capabilities of our approach on a larger scale. We present our findings in \Cref{table:K700_accuracies_ch4}. Due to the recency of the dataset, and its significant size increase over the previous 400/600 variants, there is sparsity in the reported results of \Cref{table:HACS_accuracies_ch4}.

% res_34
\textbf{ResNet 34}. The two tested 34-layer \textbf{SRTG r3d-34} and \textbf{SRTG r(2+1)d-34} networks achieve similar results with their top-1 and top-5 accuracy margins being $\leq 0.5\%$. Due to their limited computational complexity, their results are lower in comparison to models of higher complexity such as TSM \cite{lin2019tsm} and ir-CSN-101 \cite{tran2019video}.

% res_50
\textbf{ResNet 50}. Our \textbf{SRTG r3d-50} performs similar to I3D on the top-1 accuracy while only requiring half the number of GFLOPs. The top-5 accuracy is higher than that of the I3D and more competitive to the top-5 accuracy of MF-Net \cite{chen2018multifiber}. \textbf{SRTG r(2+1)d-50} with (2+1)D convolutions shows a small performance increase in comparison to the 3D variant. Accuracy rates achieved are similar to those of MF-Net and ir-CSN-101.

% res_101
\textbf{ResNet 101}. The two largest tested architectures that include SRTG are comparable to results obtained from state-of-the-art methods coming second to only SlowFast-101 \cite{feichtenhofer2019slowfast} in top-1 accuracy. They are also the top-performing models in terms of the top-5 accuracy. \textbf{SRTG r3d-101} performance can be compared to that of SlowFast-50 without however addressing temporal variations with two networks. This demonstrates that SRTG can be a viable approach for bridging the gap between image-based adapted models used for video data, and architectures created explicitly for spatio-temporal information. The (2+1)D \textbf{SRTG r(2+1)d-101} variant shows accuracy rates close to SlowFast-101 in both the top-1 and top-5 accuracies.

\subsection{Results on Moments in Time}
\label{ch:4::sec:results::sub:MiT}

\begin{table}[t]
\caption{\textbf{Comparison with MiT state-of-the-art}. Models denoted with ($^{\dagger}$) include additional information input sources. $^{*}$ denotes our own implementation.}
\centering
\resizebox{.7\linewidth}{!}{%
\renewcommand{\arraystretch}{1.5}
\begin{tabular}{c|c|cc}
\hline
Model &
Arch. size &
top-1 & top-5 \\
\hline

EvaNet \cite{piergiovanni2019evolving} &
\multirow{2}{*}{NAS\cite{zoph2017neural}} &
31.8 &	N/A \\

AssembleNet \cite{ryoo2019assemblenet} & &
\textbf{34.3} &	\textbf{62.7}\\

\hline
MF-Net \cite{chen2018multifiber} $^{*}$ &
\multirow{11}{*}{Fixed} &
27.3 &	48.2 \\

I3D \cite{carreira2017quo} &
&
29.5 &	56.1 \\

CoST \cite{li2019collaborative} & &
32.4 & 60.0\\

SoundNet \cite{monfort2018moments} $^{\dagger}$ & &
7.6 & 18.0 \\

TSN+Flow \cite{monfort2018moments} $^{\dagger}$ & &
15.7 & 34.7 \\\cline{1-1}\cline{3-4}

SRTG r3d-34 \cite{stergiou2021learn} \textbf{(ours)} & &
28.5 & 52.3 \\

SRTG r3d-50 \cite{stergiou2021learn} \textbf{(ours)} & &
30.7 & 55.6 \\

SRTG r3d-101 \cite{stergiou2021learn} \textbf{(ours)} & &
33.6 & 58.5 \\

SRTG r(2+1)d-34 \cite{stergiou2021learn} \textbf{(ours)} & &
29.0 & 54.2 \\

SRTG r(2+1)d-50 \cite{stergiou2021learn} \textbf{(ours)} & &
31.6 & 56.8 \\

SRTG r(2+1)d-101 \cite{stergiou2021learn} \textbf{(ours)} & &
33.7 & 59.1 \\

\end{tabular}%
}
\label{table:mit_accuracies_ch4}
\end{table}

% MiT results
\Cref{table:mit_accuracies_ch4} shows performance of the the top-1 and top-5 accuracies (\%) of current state-of-the-art models in Moments in Time (MiT). Comparisons made are based on both models on pre-defined architectures (denoted with \enquote{Fixed}) as well as models that employ different types of Neural Architecture Search (NAS) \cite{zoph2017neural} that effectively optimise the model structure to better address the data. Our SRTG models show promising results with  \textbf{SRTG r3d-101} and \textbf{SRTG r(2+1)d-101} having similar performance to AssembleNet and CoST, while outperforming the EvaNet \cite{piergiovanni2019evolving} learned architecture by +1.9\% in top-1 accuracy. The accuracies are also close to the top-performing AssembleNet with only a -0.6\% margin for the top-1 accuracy. In comparison to pre-defined architectures such as those of MF-Net, I3D and CoST, \textbf{SRTG r(2+1)d-101} outperforms them with a margin of 1.3--6.4\%. Additional results on 50-layered ResNets also show the merits of the proposed SRTG module with accuracies comparable to EvaNet. \textbf{SRTG r3d-50} largely outperforms similarly complex fixed architecture models such as I3D and MF-Net by margins in top-1 accuracy of +1.2\% and +3.4\% respectively. The \textbf{SRTG r(2+1)d-50} model shows almost identical performance to EvaNet without Neural Architecture Search. A marginal deficit in top-1 performance is observed in comparison to CoST (-0.8\%).  \textbf{SRTG r3d-34} and \textbf{SRTG r(2+1)d-34} also show promising results while having accuracy rates higher than models that use multiple data type inputs. In comparison to learned models \cite{piergiovanni2019evolving,ryoo2019assemblenet}, SRTG models come with added computational reductions, as there is no additional objective towards permuting the base model.

\subsection{Ablation studies}
\label{ch:4::sec:results::sub:ablation}

% Section focus
The scope of this section is the evaluation of models with and without SRTG. We demonstrate results in a pairwise fashion between the original baseline architectures and the proposed ones with SRTG. Experiments present a condensed view of these comparisons across the three large-scale datasets.

% 3D block comparison (3D and 2+1D) - with/without SRTG
\begin{table*}
\centering
           \captionsetup[subtable]{position = top}
           \captionsetup[table]{position=top}
           \caption{\textbf{Pairwise comparisons} for r3d networks with and without SRTG on HACS, K-700 and MiT.}
\begin{subtable}{\linewidth}
\caption{\textbf{Comparisons for r3d original architectures with/out SRTG enabled}}
\centering
\resizebox{\linewidth}{!}{%
\begin{tabular}{c|c|cc|cc|cc}
\hline
Dataset & 
accuracy &
\multicolumn{2}{c}{r3d-34} &
\multicolumn{2}{c}{r3d-50} &
\multicolumn{2}{c}{r3d-101} \\[0.25em]
& (\%) & None & \textbf{SRTG} & None & \textbf{SRTG} & None & \textbf{SRTG} \\[0.5em]\hline
\multirow{2}{*}{HACS} & top1 & 74.8 & \textbf{78.6} \textcolor{applegreen}{(+3.8)} & 78.4 & \textbf{80.4} \textcolor{applegreen}{(+2.0)} & 80.5 & \textbf{81.7} \textcolor{applegreen}{(+1.2)}\\[0.25em]
                      & top5 & 92.8 & \textbf{93.6} \textcolor{applegreen}{(+0.8)} & 93.8 & \textbf{95.4} \textcolor{applegreen}{(+1.7)} & 95.2 & \textbf{96.3} \textcolor{applegreen}{(+1.1)}\\[0.25em]\hline
\multirow{2}{*}{K-700} & top1 & 46.1 & \textbf{49.1} \textcolor{applegreen}{(+3.0)} & 49.1 & \textbf{53.5} \textcolor{applegreen}{(+4.4)} & 52.6 & \textbf{56.5} \textcolor{applegreen}{(+3.9)}\\[0.25em]
                      & top5 & 67.1 & \textbf{72.7} \textcolor{applegreen}{(+5.6)} & 72.5 & \textbf{74.2} \textcolor{applegreen}{(+1.7)} & 74.6 & \textbf{76.8} \textcolor{applegreen}{(+2.2)}\\[0.25em]\hline
\multirow{2}{*}{MiT} & top1 & 24.9 & \textbf{28.5} \textcolor{applegreen}{(+3.6)} & 28.2 & \textbf{30.7} \textcolor{applegreen}{(+2.5)} & 31.5 & \textbf{33.6} \textcolor{applegreen}{(+2.1)} \\[0.25em]
                      & top5 & 50.1 & \textbf{52.3} \textcolor{applegreen}{(+1.2)} & 53.5 & \textbf{55.6} \textcolor{applegreen}{(+2.1)} & 57.4 & \textbf{58.5} \textcolor{applegreen}{(+1.1)}\\[0.25em]
\end{tabular}
}
\label{table:srtg_blockcomparisons_r3d}
\end{subtable}%
\vspace{1em}
\begin{subtable}{\linewidth}
\caption{\textbf{Comparisons for r(2+1)d original architectures with/out SRTG enabled}}
\centering
\resizebox{\linewidth}{!}{%
\begin{tabular}{c|c|cc|cc|cc}
\hline
Dataset & 
accuracy &
\multicolumn{2}{c}{r(2+1)d-34} &
\multicolumn{2}{c}{r(2+1)d-50} &
\multicolumn{2}{c}{r(2+1)d-101} \\[0.25em]
& (\%) & None & \textbf{SRTG} & None & \textbf{SRTG} & None & \textbf{SRTG} \\[0.5em]\hline
\multirow{2}{*}{HACS} & top1 & 75.7 & \textbf{80.4} \textcolor{applegreen}{(+4.7)} & 81.3 & \textbf{83.8} \textcolor{applegreen}{(+2.5)} & 82.9 & \textbf{84.3} \textcolor{applegreen}{(+1.4)}\\[0.25em]
                      & top5 & 93.6 & \textbf{94.3} \textcolor{applegreen}{(+0.7)} & 94.5 & \textbf{96.6} \textcolor{applegreen}{(+2.1)} & 95.7 & \textbf{96.8} \textcolor{applegreen}{(+1.1)}\\[0.25em]\hline
\multirow{2}{*}{K-700} & top1 & 46.6 & \textbf{49.4} \textcolor{applegreen}{(+2.8)} & 49.9 & \textbf{54.2} \textcolor{applegreen}{(+4.3)} & 52.5 & \textbf{56.8} \textcolor{applegreen}{(+4.3)}\\[0.25em]
                      & top5 & 68.2 & \textbf{73.2} \textcolor{applegreen}{(+5.0)} & 73.3 & \textbf{74.6} \textcolor{applegreen}{(+1.3)} & 75.2 & \textbf{77.4} \textcolor{applegreen}{(+2.2)}\\[0.25em]\hline
\multirow{2}{*}{MiT} & top1 & 25.6 & \textbf{29.0} \textcolor{applegreen}{(+3.4)} & 29.3 & \textbf{31.6} \textcolor{applegreen}{(+2.3)} & 32.2 & \textbf{33.7} \textcolor{applegreen}{(+1.5)} \\[0.25em]
                      & top5 & 52.7 & \textbf{54.2} \textcolor{applegreen}{(+1.5)} & 55.2 & \textbf{56.8} \textcolor{applegreen}{(+1.6)} & 57.7 & \textbf{59.1} \textcolor{applegreen}{(+1.4)}\\[0.25em]
\end{tabular}
}
\label{table:srtg_blockcomparisons_r2plus1d}
\end{subtable}%
\label{table:srtg_blockcomparisons}
\end{table*}

% Comparisons
\textbf{SRTG pairwise comparisons}. In \Cref{table:srtg_blockcomparisons_r3d,table:srtg_blockcomparisons_r2plus1d} we detail full pairwise comparisons in terms of performance on HACS, K-700 and MiT for networks with and without SRTG. As presented in \Cref{table:srtg_blockcomparisons_r3d}, r3d configurations with SRTG can achieve an average +2.3\%, +3.7\% and +2.7\% top-1 accuracy improvements on HACS, K-700 and MiT datasets, respectively. Similarly, for top-5 this corresponds to +1.3\%, +2.8\% and +1.5\% increases in the accuracy rates for HACS, K-700 and MiT. For r(2+1)d architectures in \Cref{table:srtg_blockcomparisons_r2plus1d} top-1 and top-5 accuracies on HACS are improved on average by +2.9\% and +1.3\%, respectively. Similar increases are also observable for K-700 and MiT datasets with +3.8\% and +2.4\% improvement in their top-1 and +2.8\%, +1.5\% for their top-5 accuracies. The extended experimentation on both network architectures and datasets show that the inclusion of SRTG modules can significantly benefit the overall performance of spatio-temporal models without direct implications on the overall architecture.

% HACS SRTG with/out
\begin{figure}[t]
\centering
     \begin{subfigure}[b]{0.3\textwidth}
         \centering
         \includegraphics[width=\textwidth]{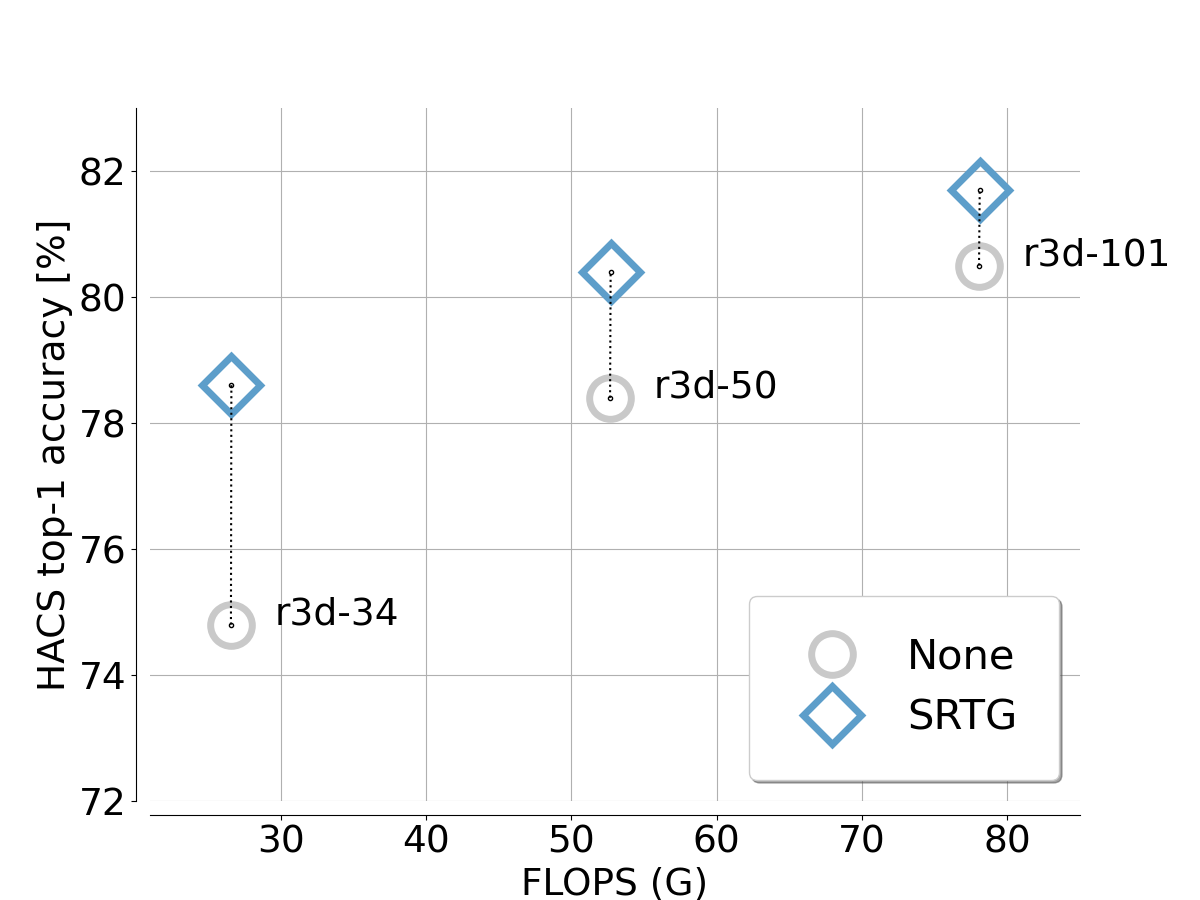}
         \caption{\textbf{HACS r3d top-1.}}
         \label{fig:srtg_r3d_HACS_top1}
     \end{subfigure}
     \hfill
     \begin{subfigure}[b]{0.3\textwidth}
         \centering
         \includegraphics[width=\textwidth]{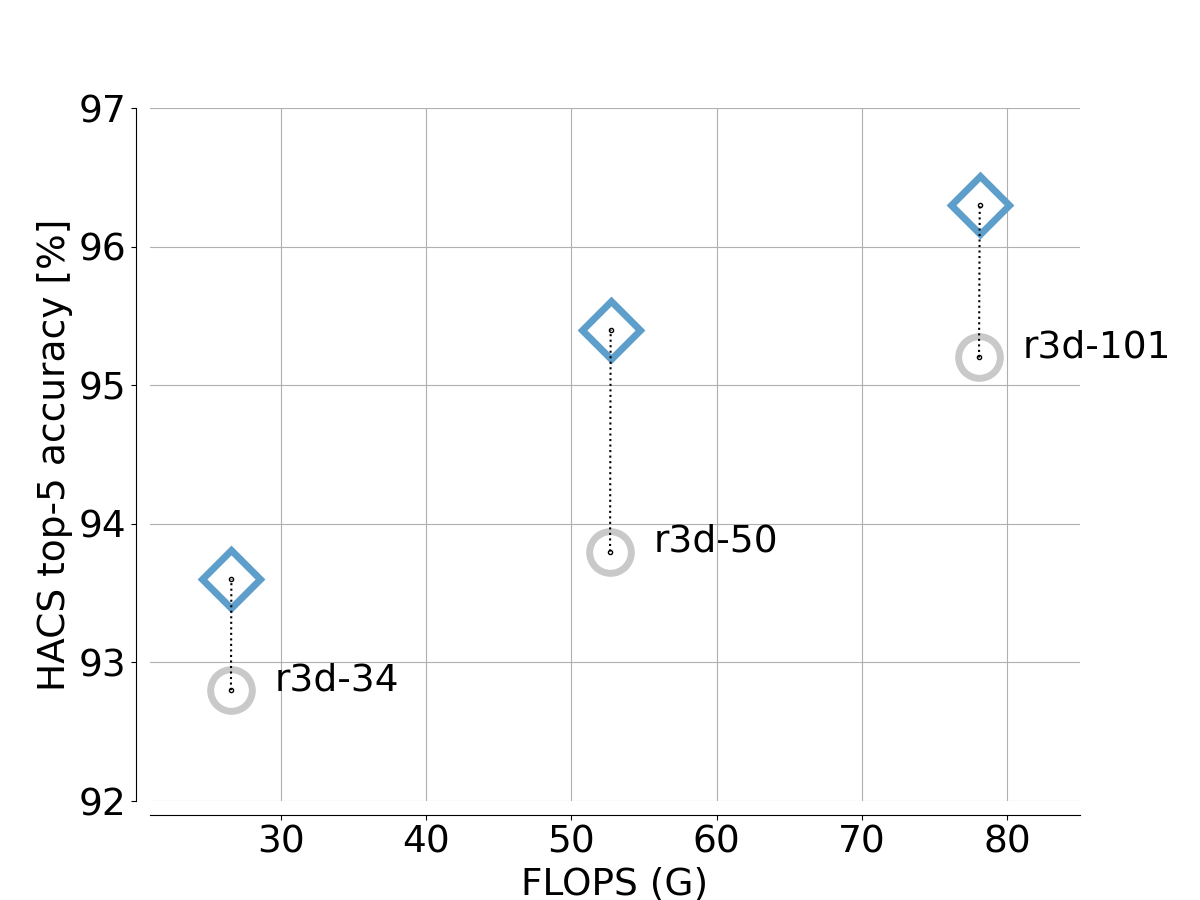}
         \caption{\textbf{HACS r3d top-5.}}
         \label{fig:srtg_r3d_HACS_top5}
     \end{subfigure}
     \hfill
     \begin{subfigure}[b]{0.3\textwidth}
         \centering
         \includegraphics[width=\textwidth]{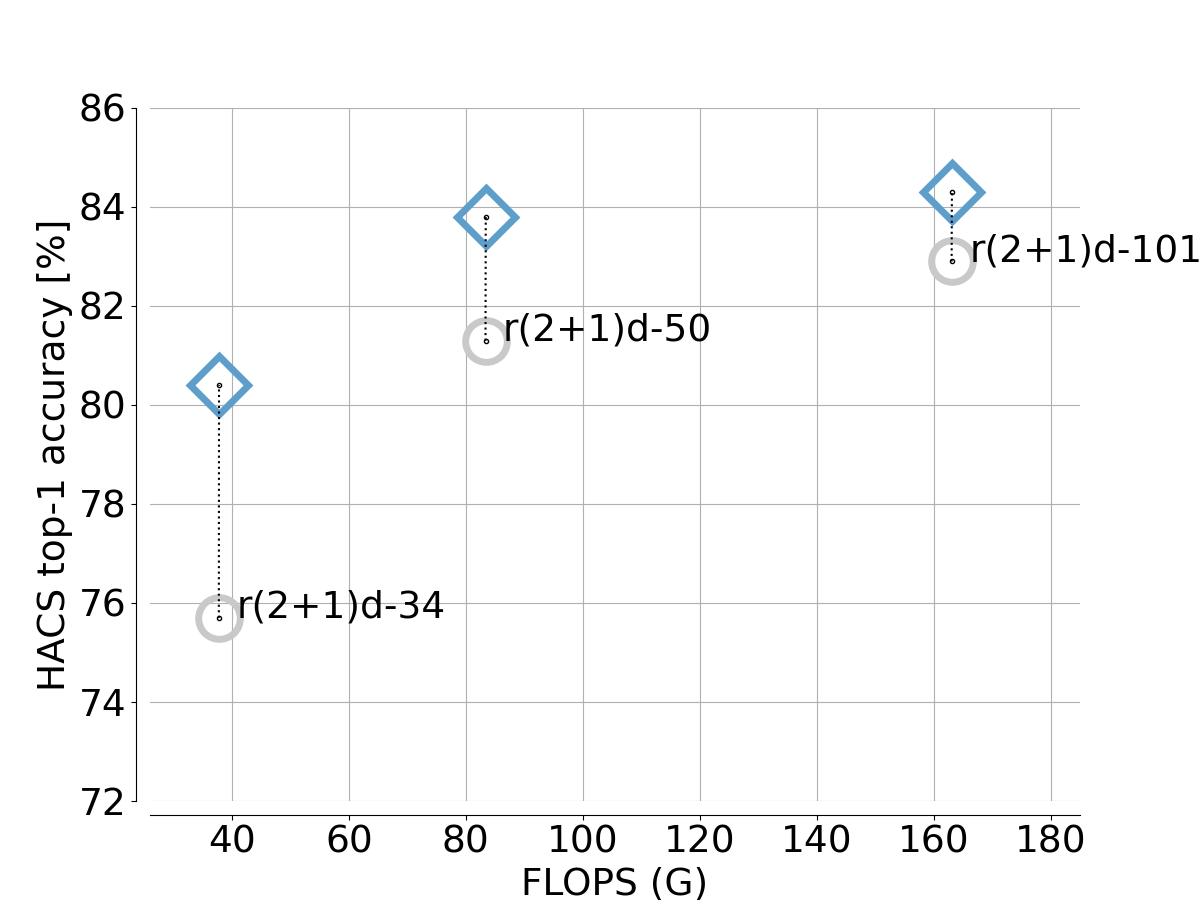}
         \caption{\textbf{HACS r(2+1)d top-1.}}
         \label{fig:srtg_r2plus1d_HACS_top1}
     \end{subfigure}\\
     \begin{subfigure}[b]{0.3\textwidth}
         \centering
         \includegraphics[width=\textwidth]{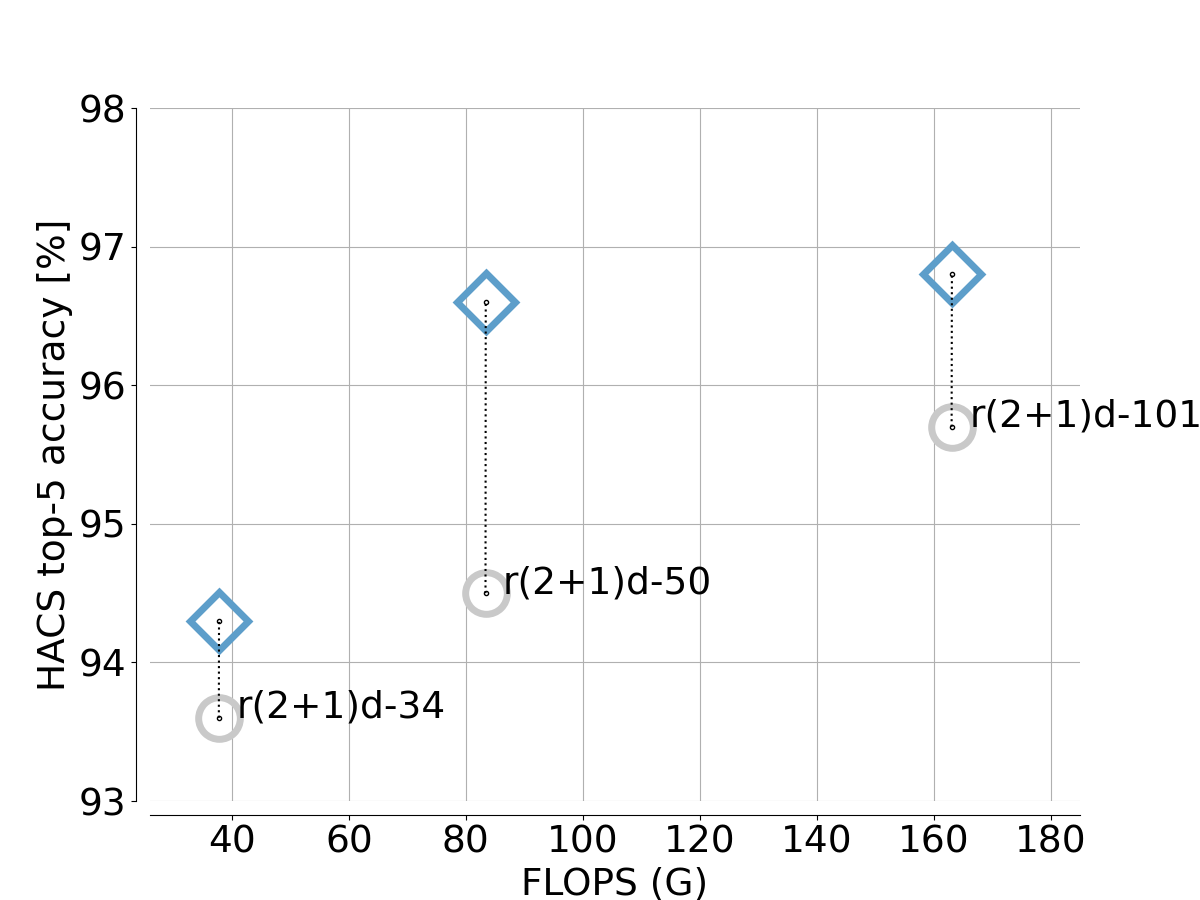}
         \caption{\textbf{HACS r(2+1)d top-5.}}
         \label{fig:srtg_r2plus1d_HACS_top5}
     \end{subfigure}
     \hfill
     \begin{subfigure}[b]{0.3\textwidth}
         \centering
         \includegraphics[width=\textwidth]{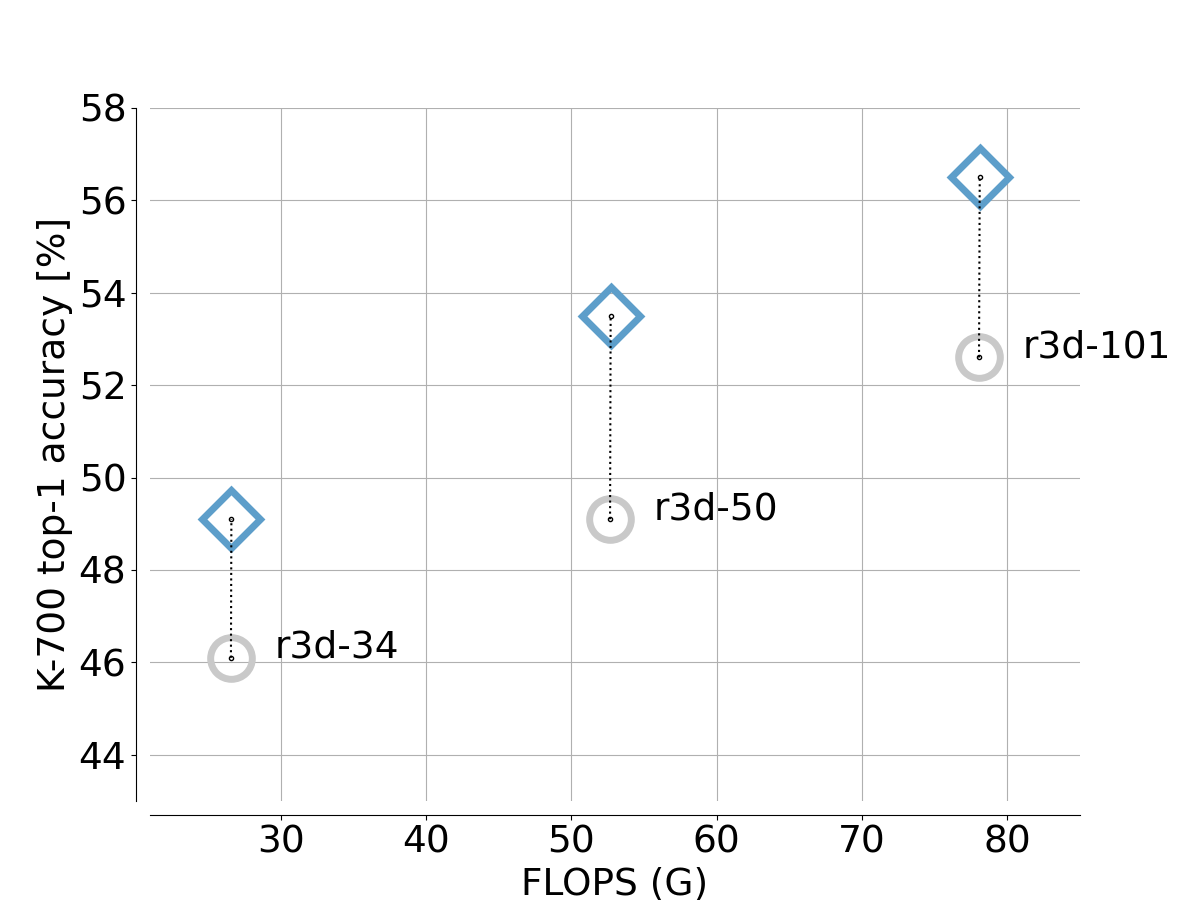}
         \caption{\textbf{K-700 r3d top-1.}}
         \label{fig:srtg_r3d_K700_top1}
     \end{subfigure}
     \hfill
     \begin{subfigure}[b]{0.3\textwidth}
         \centering
         \includegraphics[width=\textwidth]{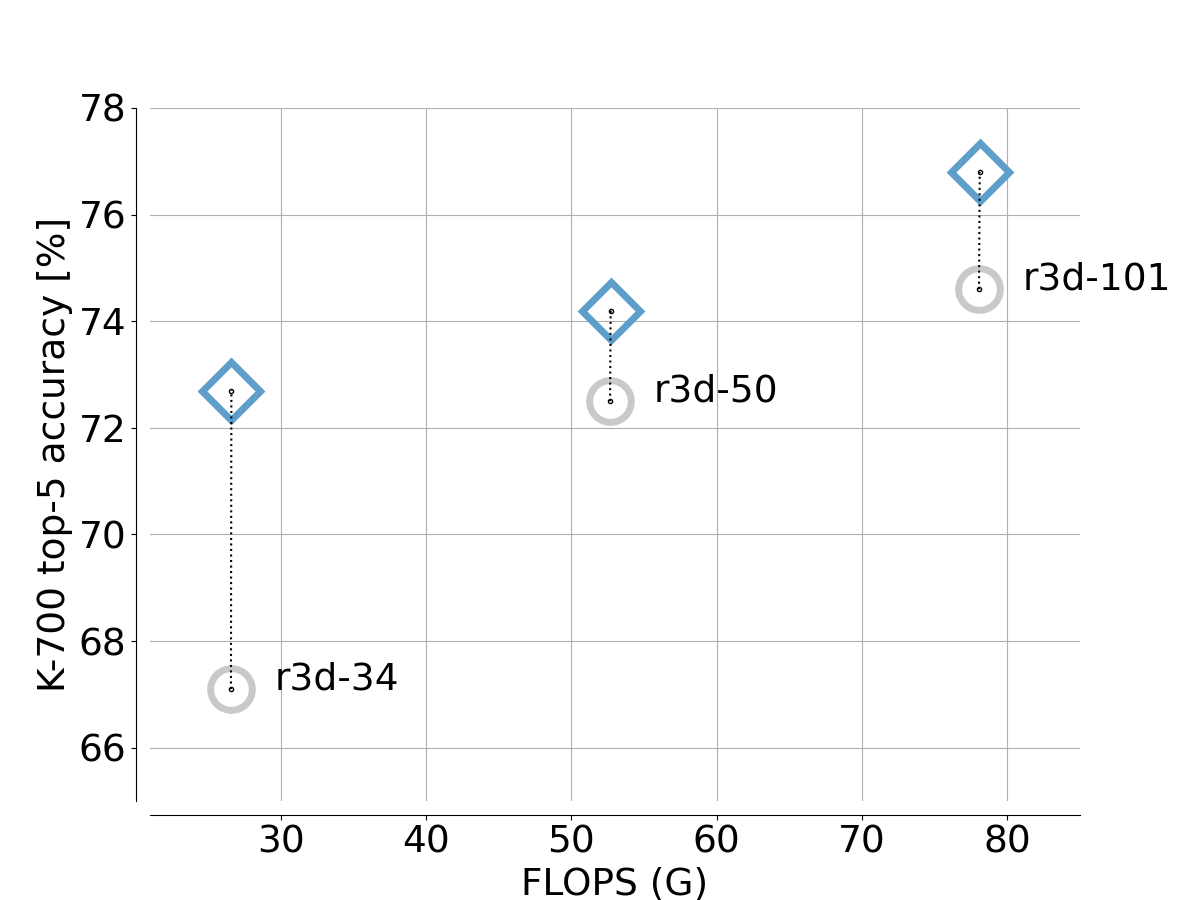}
         \caption{\textbf{K-700 r3d top-5.}}
         \label{fig:srtg_r3d_K700_top5}
     \end{subfigure}\\
     \begin{subfigure}[b]{0.3\textwidth}
         \centering
         \includegraphics[width=\textwidth]{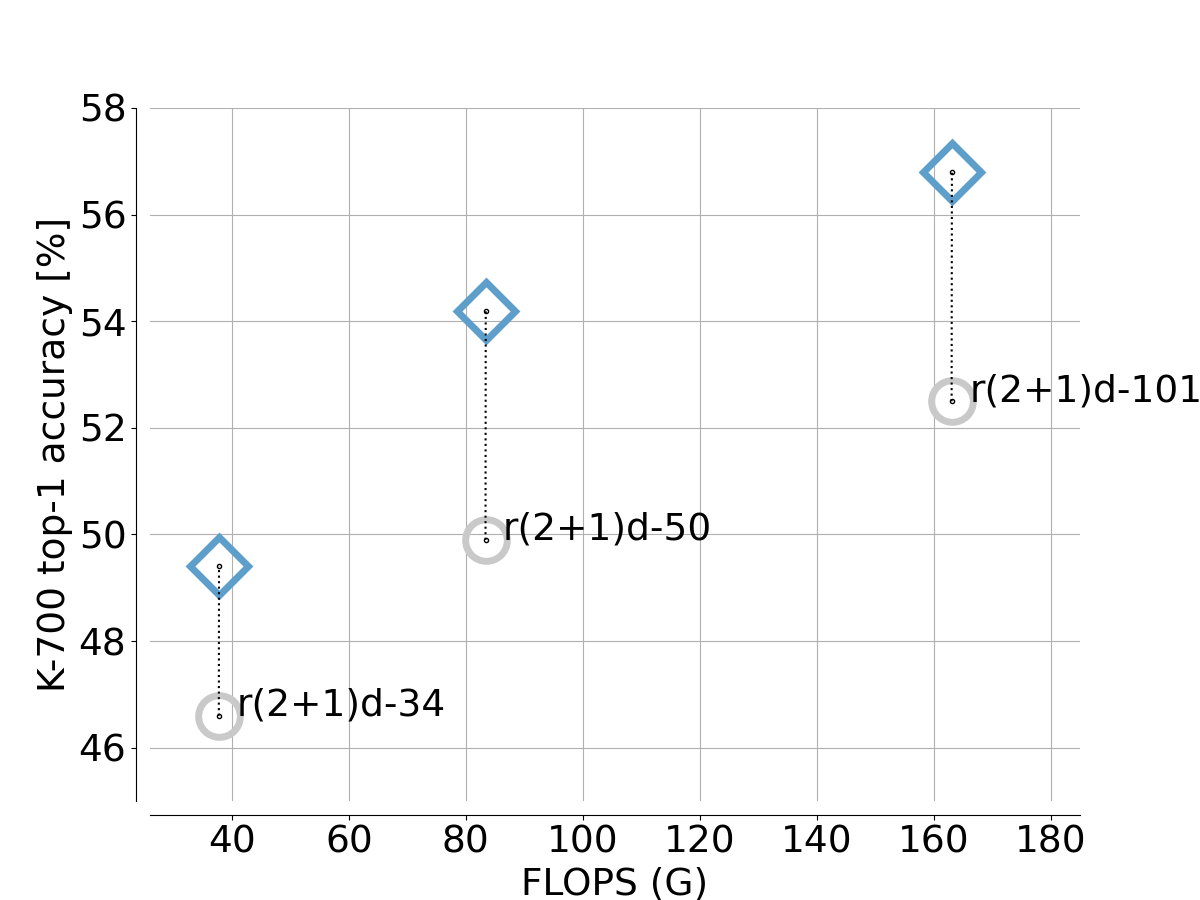}
         \caption{\textbf{K-700 r(2+1)d top-1.}}
         \label{fig:srtg_r2plus1d_K700_top1}
     \end{subfigure}
     \hfill
     \begin{subfigure}[b]{0.3\textwidth}
         \centering
         \includegraphics[width=\textwidth]{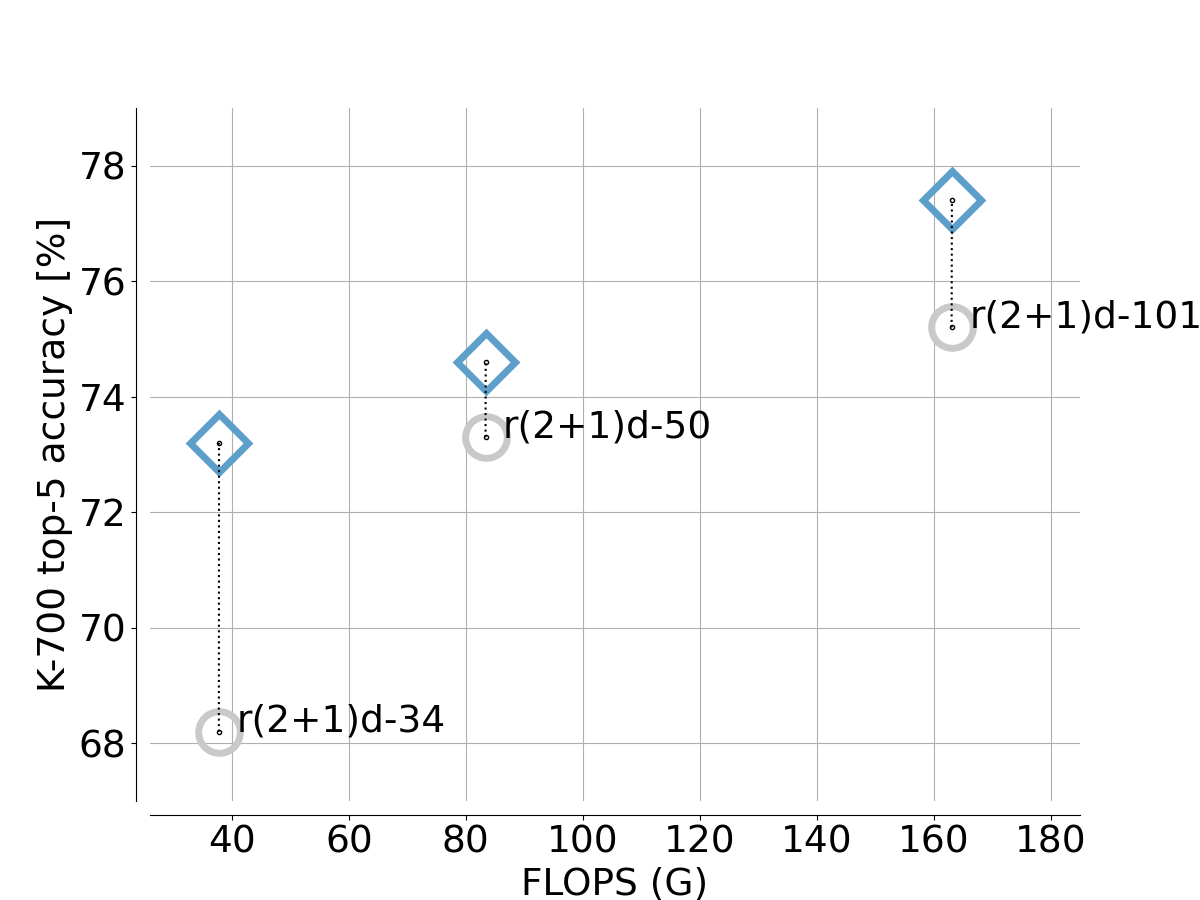}
         \caption{\textbf{K-700 r(2+1)d top-5.}}
         \label{fig:srtg_r2plus1d_K700_top5}
     \end{subfigure}
     \hfill
     \begin{subfigure}[b]{0.3\textwidth}
         \centering
         \includegraphics[width=\textwidth]{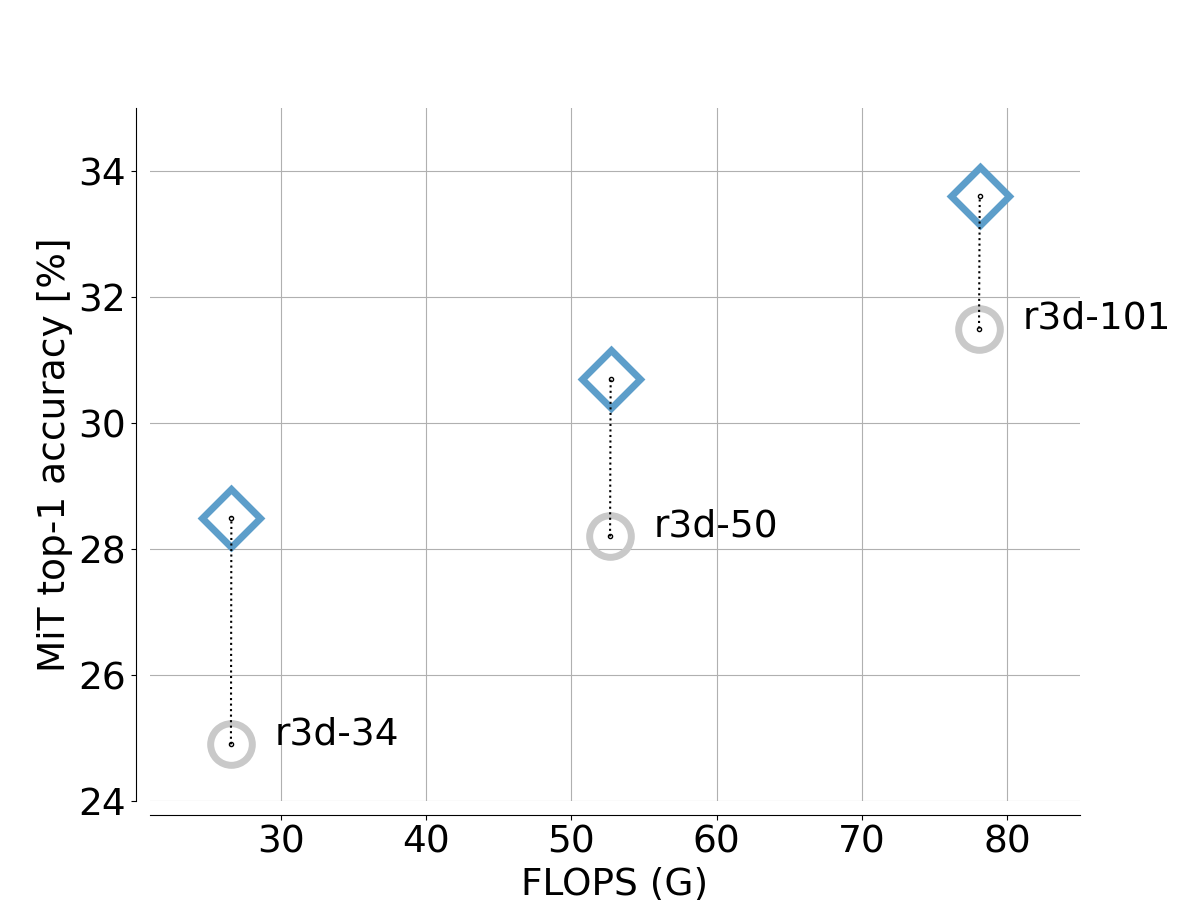}
         \caption{\textbf{MiT r3d top-1.}}
         \label{fig:srtg_r3d_MiT_top1}
     \end{subfigure}\\
     \begin{subfigure}[b]{0.3\textwidth}
         \centering
         \includegraphics[width=\textwidth]{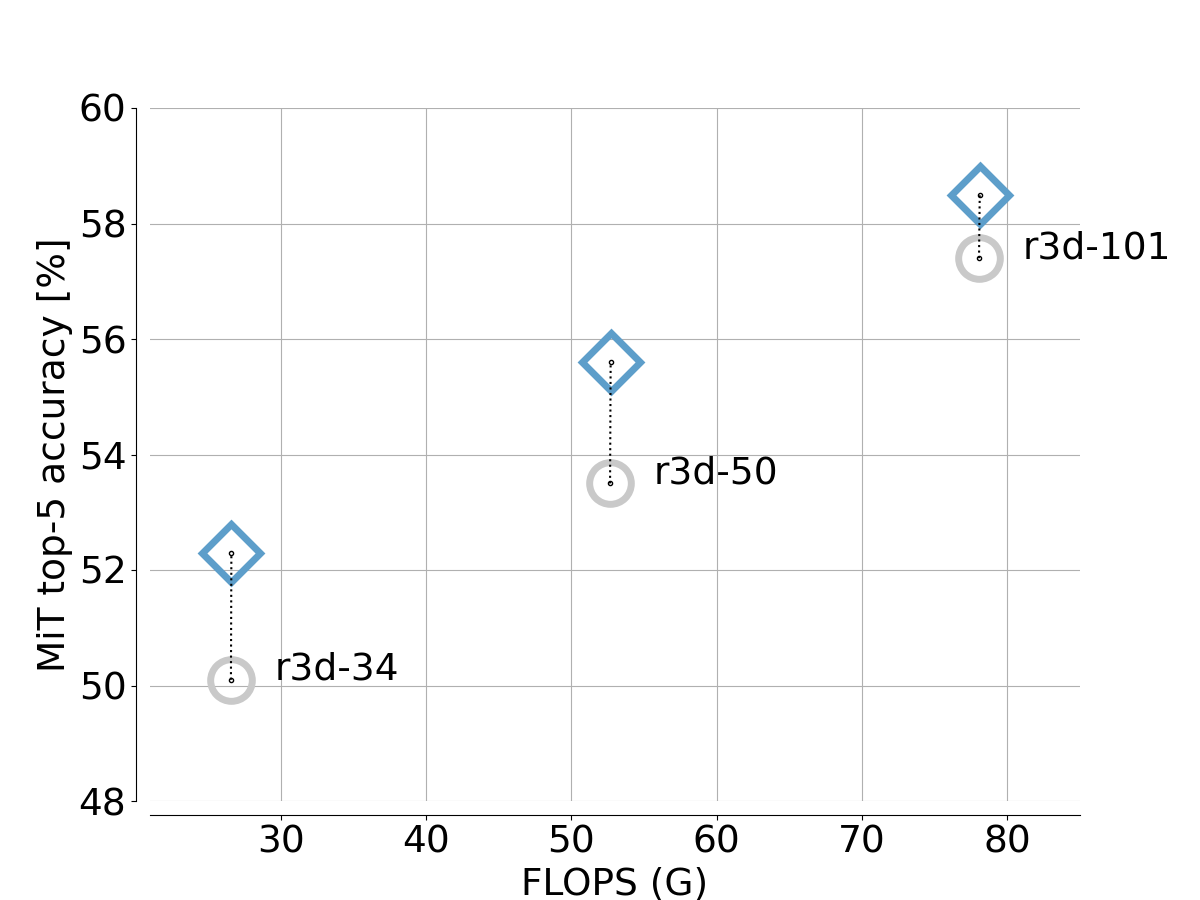}
         \caption{\textbf{MiT r3d top-5.}}
         \label{fig:srtg_r3d_MiT_top5}
     \end{subfigure}
     \hfill
     \begin{subfigure}[b]{0.3\textwidth}
         \centering
         \includegraphics[width=\textwidth]{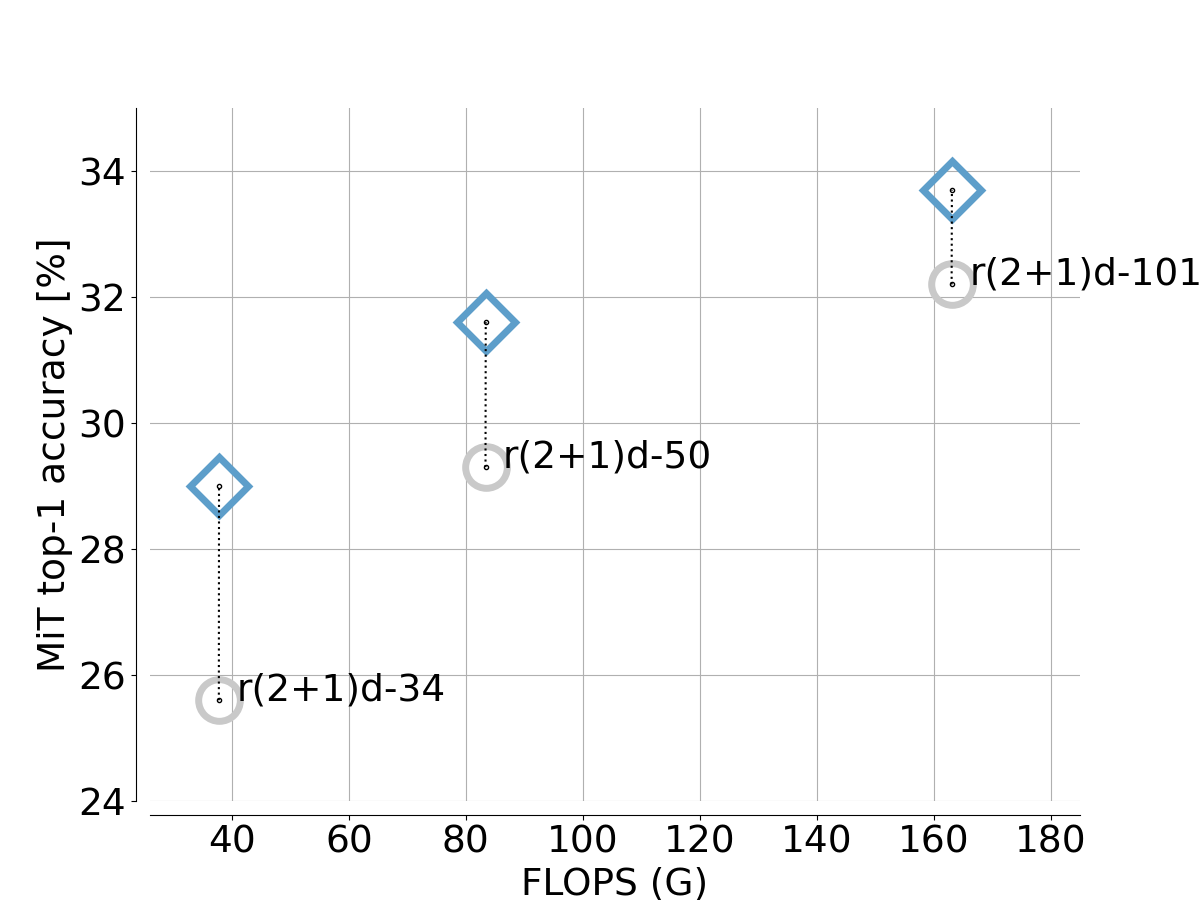}
         \caption{\textbf{MiT r(2+1)d top-1.}}
         \label{fig:srtg_r2plus1d_MiT_top1}
     \end{subfigure}
     \hfill
     \begin{subfigure}[b]{0.3\textwidth}
         \centering
         \includegraphics[width=\textwidth]{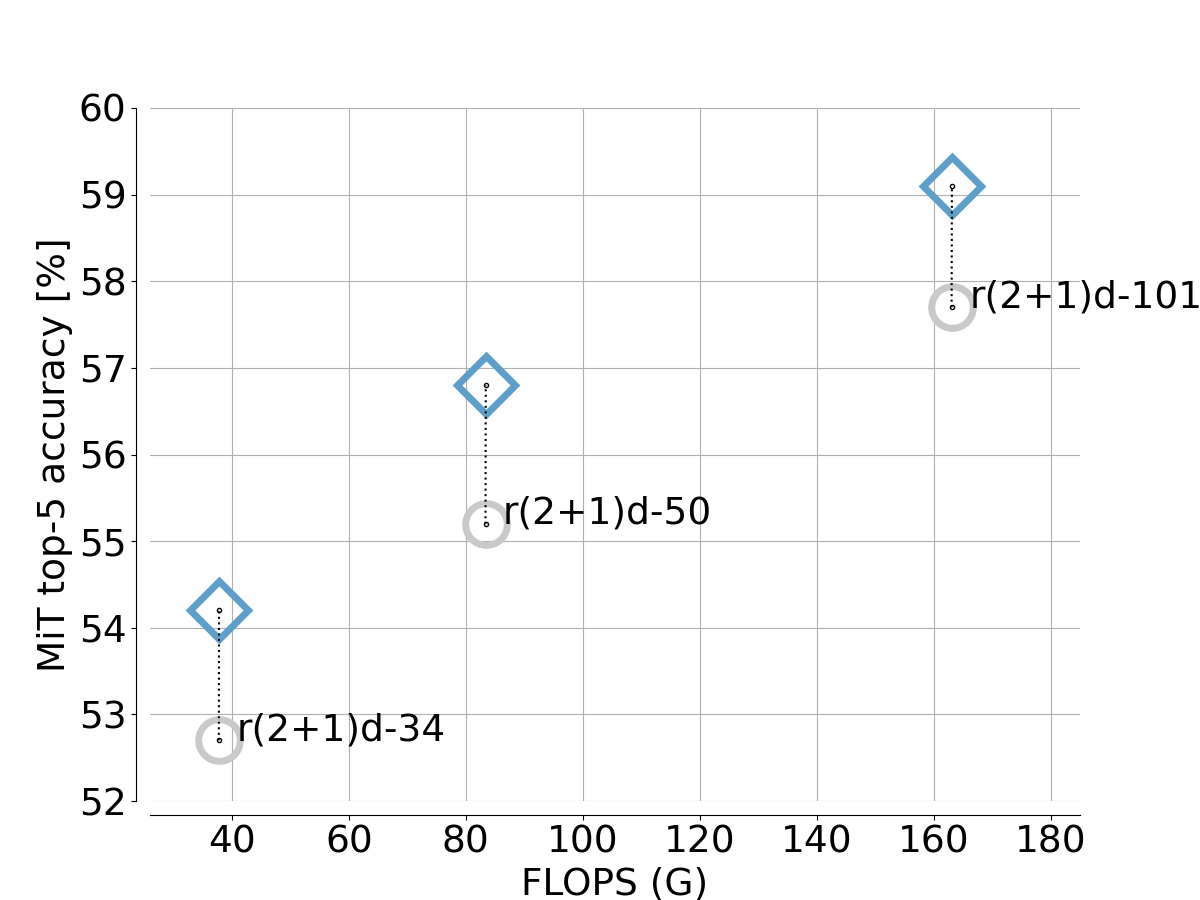}
         \caption{\textbf{MiT r(2+1)d top-5.}}
         \label{fig:srtg_r2plus1d_MiT_top5}
     \end{subfigure}
\caption{\textbf{Accuracy to computational complexity trade-offs} on HACS, K-700 and MiT for r3d and r(2+1)d architectures with and without SRTG.}
\label{fig:srtg_acc_2_flops}
\end{figure}

% Graphs
\textbf{Comparison visualisations}. The accuracy improvements are visualised in \Cref{fig:srtg_acc_2_flops} where the full pairwise comparisons are shown with respect to the number of FLOPs. Networks that include SRTG are denoted with blue while networks that do not are denoted with grey. Interestingly, the computational overhead of SRTG is $\ll 1\%$ of that of the entire architecture making the module computationally lightweight.

\section{Feature transferability evaluation}
\label{ch:4::sec:TL}

% About
One common aspect of CNNs is the utilisation of their weights trained on large-scale datasets and their transfer to smaller, typically more fine-grained,  datasets. Given that the learned features can be generalised, this weight transfer process can significantly benefit performance. To further evaluate the generalisation capabilities of SRTG-enabled networks, we present results with the corresponding models being pre-trained on large datasets and fine-tuned on the smaller UCF-101 \cite{soomro2012ucf101} and HMDB-51 \cite{kuehne2011hmdb}. Through this experimentation, we can eliminate biases that are based on the pre-training datasets used, while computing accuracy rates with respect to SRTG modules.
% Transfer learning set-up
For the feature transferability with SRTG ResNets, we only train the final class prediction layer while re-using the weights from the previously trained models in the convolutional layers. We additionally set the base learning rate to $0.01$ and reduce it by a factor of ten at epochs 40 and 60 for a total of 80 epochs. Our choice of 80 epochs was based on not observing further improvements. We include experiments for SRTG ResNets on both UCF-101 and HMDB-51 with their respected networks being pre-trained on multiple datasets.

% Table: UCF accuracies
\begin{table}[t]
\caption{\textbf{Transfer Learning on UCF-101:} Top-1 and top-5 accuracies after pre-training.}
\centering
\resizebox{.6\textwidth}{!}{%
\begin{tabular}{c|c|cc}
\hline
Model & 
Pre-training &
top-1 & top-5 \\[0.3em]
\hline
I3D & 
K-400 &
92.4 & 97.6 \\[0.3em]

TSM &
K-400 &
92.3 & 97.9 \\[0.3em]

ir-CSN-152 &
IG65M &
95.4 & 99.2 \\[0.3em]

MF-Net &
K-400 &
93.8 & 98.4 \\[0.3em]

SF r3d-50 &
ImageNet &
94.6 & 98.7 \\[0.3em]

SF r3d-101 &
ImageNet &
95.8 & 99.1 \\[0.3em]
\hline

r3d-101 &
\multirow{8}{*}{HACS+K-700} &
95.8 & 98.4 \\[0.3em]

r(2+1)d-101 &
 &
95.5 & 98.7 \\[0.3em]\cline{1-1}\cline{3-4}

SRTG r3d-34 \textbf{(ours)} &
 &
94.8 & 98.1 \\[0.3em]

SRTG r3d-50 \textbf{(ours)} &
 &
95.7 & 98.5 \\[0.3em]

SRTG r3d-101 \textbf{(ours)} &
 &
\textbf{97.3} & \textbf{99.5} \\[0.3em]

SRTG r(2+1)d-34 \textbf{(ours)} &
 &
94.1 & 97.8 \\[0.3em]

SRTG r(2+1)d-50 \textbf{(ours)} &
 &
95.7 & 98.8 \\[0.3em]

SRTG r(2+1)d-101 \textbf{(ours)} &
 &
\textbf{97.3} & 99.2 \\

\end{tabular}%
}
\label{table:accuracies_ucf_ch4}
\end{table}

% Table 4: Transfer learning (With/Without)
\begin{table}[ht]
\caption{\textbf{Pre-train dataset comparisons on UCF-101 and HMDB-51}.}
\centering
\resizebox{.75\linewidth}{!}{%
\begin{tabular}{c|c|c|cc}
\hline
Model & Pre-training & GFLOPs & UCF-101 & HMDB-51 \\[0.3em]
& & $\! \times \!$ views & top-1 (\%) & top-1 (\%)\\[0.2em]
\hline
\multirow{3}{*}{SRTG r3d-34} & HACS & \multirow{3}{*}{26.6$\! \times \!$30} & 94.8 & 74.3 \\[0.1em]
& HACS+K-700 & & 95.8 & 74.2 \\[0.1em]
& HACS+MiT & & 95.2 & 74.2 \\[0.1em]
\hline
\multirow{3}{*}{SRTG r(2+1)d-34} & HACS & \multirow{3}{*}{37.8$\! \times \!$30} & 94.1 & 72.9 \\[0.1em]
& HACS+K-700 &  & 94.6 & 73.2 \\[0.1em]
& HACS+MiT & & 95.6 & 74.5 \\[0.1em]
\hline
\multirow{3}{*}{SRTG r3d-50} & HACS & \multirow{3}{*}{52.7$\! \times \!$30} & 95.7 & 75.6 \\[0.1em]
& HACS+K-700 & & 96.8 & 76.0 \\[0.1em]
& HACS+MiT & & 96.5 & 76.0 \\[0.1em]
\hline
\multirow{3}{*}{SRTG r(2+1)d-50} & HACS & \multirow{3}{*}{83.4$\! \times \!$30} & 95.7 & 75.3 \\[0.1em]
& HACS+K-700 & & 96.0 & 75.7 \\[0.1em]
& HACS+MiT & & 96.3 & 76.0 \\[0.1em]
\hline
\multirow{3}{*}{SRTG r3d-101} & HACS & \multirow{3}{*}{163.1$\! \times \!$30} & 97.3 & 77.5 \\[0.1em]
& HACS+K-700 & & 97.4 & 78.0 \\[0.1em]
& HACS+MiT & & \textbf{97.6} & \textbf{78.4} \\[0.1em]

\end{tabular}
}
\vspace{-1mm}
\label{table:TL_UCF101_ch4}
\end{table}

% SOTA comparisons
We present our transfer-learning results in \Cref{table:accuracies_ucf_ch4}. The smaller \textbf{SRTG r3d-34} and \textbf{SRTG r(2+1)d-34} networks outperform more complex architectures such as I3D and TSM with margins of 1.7--2.4\% and 1.8--2.5\% respectively. This shows that SRTG can increase accuracy rates and also that the combination of the HACS and K-700 datasets has notable benefits when fine-tuning networks. Both 3D and (2+1)D variants perform similar to MF-Net in top-1 with only +1.0\% and +0.3\% top-1 accuracy increments. Both \textbf{SRTG r3d-50} and \textbf{SRTG r(2+1)d-50} demonstrate accuracy rates similar to the top-performing models with only SlowFast-101 achieving better results on top-1 with marginal difference of -0.1\%. The best performing model across our experiments was \textbf{SRTG r3d-101} with top-1 accuracy of 97.3\% and 99.5\% top-5 accuracy. The second highest performing model was the (2+1)D variant with a marginal decrease $\ll 0.1\%$ in top-1 accuracy compared to \textbf{SRTG r3d-101}. Compared to the non-SRTG counterpart, \textbf{SRTG r3d-101} demonstrates a +1.5\% top-1 accuracy increment and a +1.1\% for the top-5. Similarly, \textbf{SRTG r(2+1)d-101} improves by +1.8\% for the top-1 and +0.5\% for the top-5 accuracies compared to the r(2+1)d-101 baseline.

% Dataset comparisons
In our last set of comparisons in \Cref{table:TL_UCF101_ch4}, we present results from weight initialisation of models on different datasets. The accuracy rates remain consistent for the pre-training datasets with the margins between datasets being $< 1.5\%$. The consistency in accuracy rates is because of the large sizes of these datasets, thus including a large number of examples per class, and the overall robustness of our method. On average, the offset between pre-training models across datasets is 0.7\% for UCF-101 and 0.5\% for HMDB-51.

\section{Discussion and conclusions}
\label{ch:4::sec:discussion}

% Scope overview
We have focusded on the challenges of human action and interaction recognition from spatio-temporal data. Feature changes in performance are strongly associated to a person's prepotent identity for an action. We have addressed temporal feature imbalances through a temporal feature calibration module named Squeeze and Recursion Temporal Gates (SRTG).

% SRTG
The proposed SRTG module calibrates spatio-temporal convolutional features based on the importance of the extracted features across the video sequence. By processing the local spatio-temporal activations with the use of recurrent cells (LSTMs) we capture multi-frame feature dynamics. These feature dynamics capture the correspondence of the local features across the entirety of the clip. The created activations, with respect to information over the video sequence, are evaluated in terms of its cyclic consistency with the convolutional features. We introduce a gate function for fusing together cyclic consistent volumes that show relevance between their temporally local and extended features. Equivalently, volumes that present large dissimilarities and are not cyclic consistent are not fused together. Through this, we achieve a degree of relevance between local and extended features.

% Evaluation
We have evaluated our work on three large scale datasets: HACS, Kinetics-700 and Moments in Time, and over three ResNet architectures with either 3D or (2+1)D spatio-temporal convolutions. We have demonstrated competitive results to state-of-the-art architectures and in most cases outperform them. This also comes with negligible additional computations as our method is both memory and compute-efficient. Our ablation studies based on the original models without SRTG additionally validate our claims that the addition of SRTG to a 3D CNN architecture can yield further accuracy improvements. 

% Future direction
We believe that the study of feature relevance across time can further benefit current action recognition models by discovering temporal motions that are not constrained by the locality of kernels. Despite the alignment of the extracted spatio-temporal activations to information from the entire video sequence, limitations still exist. A dependency to the local spatio-temporal features is still present within the align features. A research direction in order to address this is the creation of kernels that can extract spatio-temporal features over different receptive field sizes. Such search can bring about a way to flexibly discover the relevance of features across multiple space-time sizes. This is explored in the following chapter.

%% Format bibliography like a section, not a chapter:
%\printbibliography[heading = subbibliography]
\stopcontents[chapters]

\author
{}
\title{Time-Varying Convolutions for Video Understanding}

\maketitle
\label{ch5}

This chapter addresses the locality of spatio-temporal convolutions by increasing their receptive fields. We focus on the creation of spatio-temporal activations that can capture features across different space-time modalities. The creation of these size-varying patterns is done through our novel \textit{Multi-Temporal Convolutions}, that model both local and prolonged features within a single activation volume. The proposed modules are integrated in 3D CNNs by replacing their original 3D convolutions. We explore the benefits of activation volumes that represent time-varying features through testing across multiple datasets \footnote{The code for the method is available at: \url{https://git.io/JfuPi} (${\dagger}$)\\
with the down-sampling operation available at: \url{https://git.io/JL5zL} (${\ddagger}$)} . 

\startcontents[chapters]
%\printcontents[chapters]{}{1}{\section*{\contentsname}}

\section{Introduction}
\label{ch:5::sec:intro}

% Problem definition
In the previous chapter we have focused on the relevance of local features in the context of the entire video. In our proposed method, we aligned features that can emphasise or suppress local features based on their relevance across the entire video segment used as input. By calibrating local features, their activations can reflect their overall significance as action descriptors. Although the inclusion of the feature dynamics across the video sequence within the local extracted patterns can be beneficial in terms of their representative capabilities, there is still a dependency on the fixed-size local patterns. Therefore, variations in the action and its feature durations that are not in the order of magnitude remain unaddressed. Based on this observation, we propose a method that can extract spatio-temporal patterns at different timescales. The recognition of temporal variations in this flexible manner can also maintain the spatial modelling power of the method.

\begin{figure}[t]
\centering
\includegraphics[width=\textwidth]{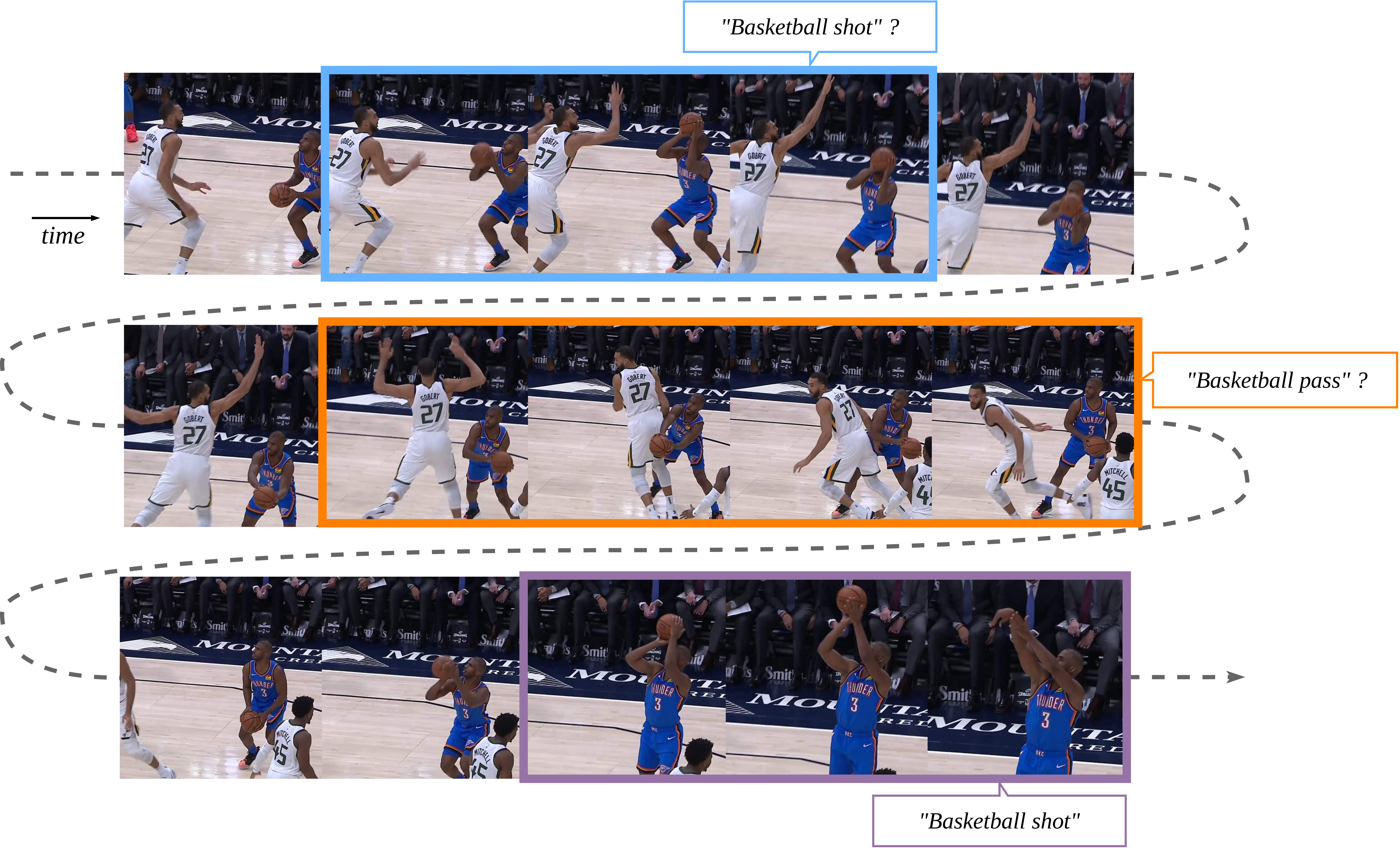}
\caption{\textbf{Temporal segments from class \enquote{basketball shot}}. Motions performed in different spatio-temporal local segments may relate to different action identities. The sole consideration of such features can lead to action misclassifications.}
\label{fig:fake_passes_examples}
\end{figure}

% Difference between features and action labels
The description of human actions based on their features strongly affects the temporal window in which they are performed \cite{vallacher2011action}. The action identity that is used to describe the sequence of movements a person performs is affected by their complexity. The general observation is that complexity also impacts duration with longer and more complex actions generally requiring larger durations. Vallacher \& Wagner \cite{vallacher1989levels} have described how action identities can include uncertainties or inadequacies based on different levels. For example, given the \enquote{basketball shot} example in \Cref{fig:fake_passes_examples}, if the sequence is broken into smaller parts, different action identities are discovered.{\parfillskip0pt\par}

\begin{wrapfigure}{r}{10em}
\vspace{-1em}
\centerline{%
    \includegraphics[width=5em]{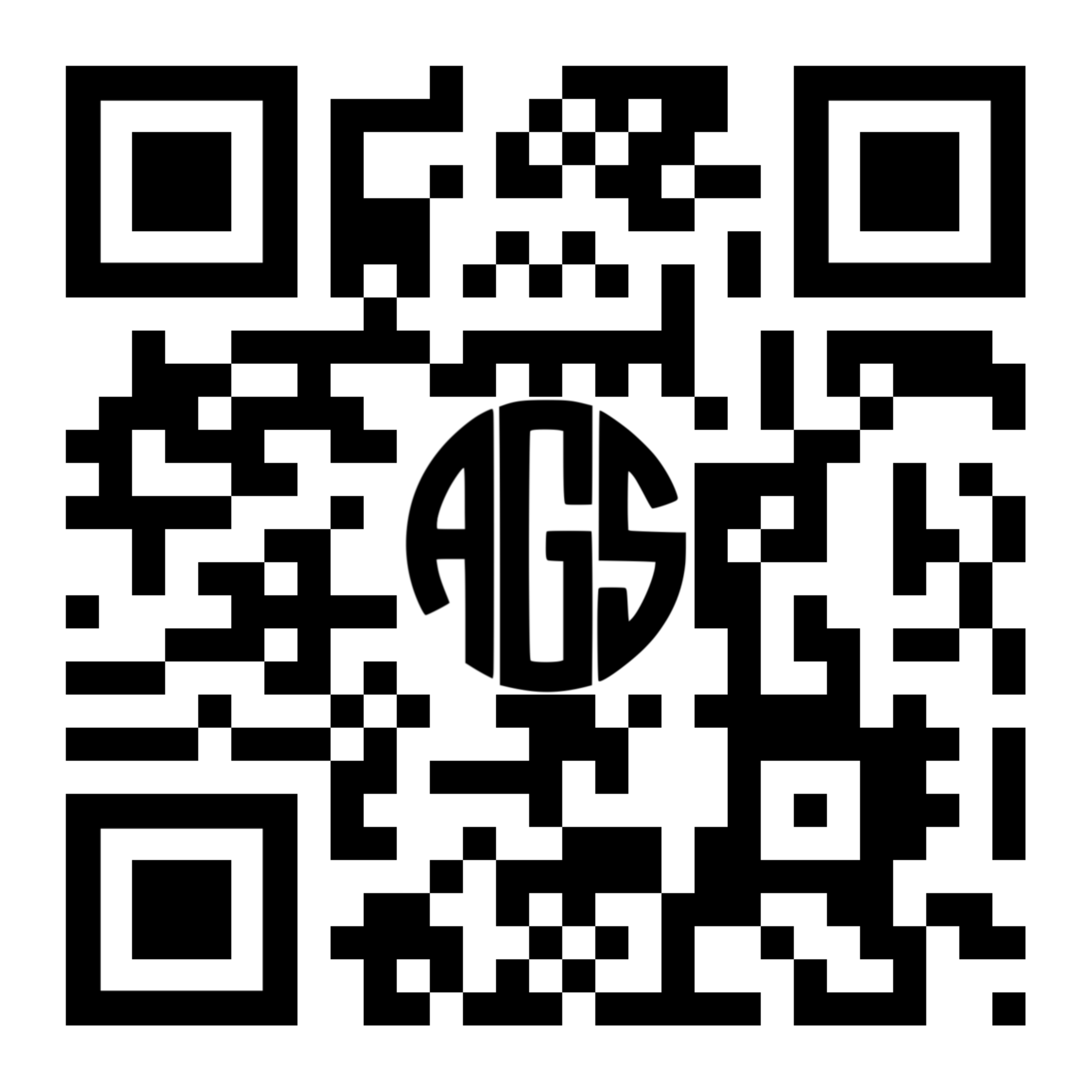}% 
    \includegraphics[width=5em]{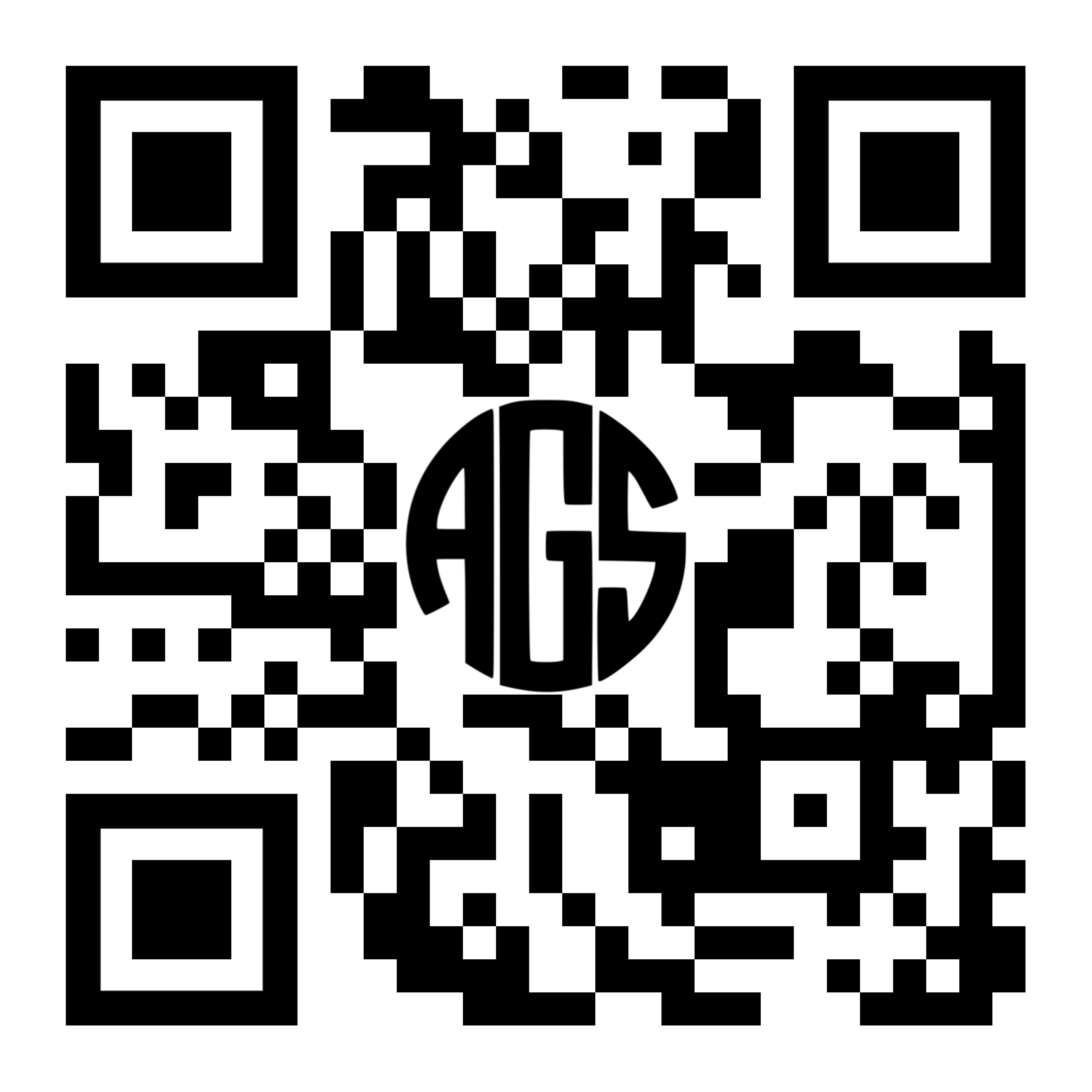}%
    }
\hspace{2em} \footnotesize{${\dagger}$} \hspace{5em} \footnotesize{${\ddagger}$}
\vspace{-3em}
\end{wrapfigure}
\noindent However, not all local sets of movements are sufficient to fully describe the action performed. In different local segments of the video, the sets of movements can lead to different individual identities. Based on the example, the second group of frames can include more fine-grained identities such as \enquote{looking for teammate}, \enquote{moving the ball to lower torso} and \enquote{extending the ball for a bounce pass}, which do not necessarily directly relate to \enquote{basketball shot} but could potentially relate to \enquote{basketball pass}. If however, a larger part of the clip is shown, we can understand that the underlying action identity that could better describe the example is indeed \enquote{basketball shot}. Based on this, we argue that actions have a degree of complexity which does not allow their description by solely considering small local patterns. Overall, action identities should be formed through the extraction of features across multiple temporal modalities. The inclusion of temporal segments of different durations can create features that are more descriptive of the action identity.

% Our focus for this chapter
Considering that the extraction of strictly local motion patterns may not directly correspond to the target action identity, we study how the feature extraction process can be improved with the inclusion of patterns under different lengths to better capture high-level short-term and long-term identities associated with actions. We investigate the effects and relations between short local patterns and features of prolonged durations through our proposed Multi-Temporal Convolutions (\textit{MTConv}) \cite{stergiou2020right} utilising three branches for the extraction of spatio-temporal features across varying durations. The branch structure includes a \textit{local branch} for spatio-temporal features within small spatio-temporal (local) segments, a \textit{prolonged branch} for patterns of extended durations that include an inherited correspondence to the local features, and finally a \textit{global aggregated feature importance branch} which further utilises SRTG to explore the feature dynamics of the aforementioned branches. Our proposed approach can address temporal patterns that are performed over different time-scales as well as align local and prolonged features in a shared feature space. Through the use of three branches for different temporal modalities, we focus on the discovery of multi-temporal patterns by extending the convolutional receptive field to spatio-temporal locations of varying sizes.

% Chapter format
In \Cref{ch:5::sec:related} we discuss related stream-based approaches for human action recognition in videos. In \Cref{ch:5::sec:methodology} we provide a complete description of the proposed convolution blocks and their layer architecture. Evaluations and tests based on our method with different settings are presented in \Cref{ch:5::sec:results}. A discussion alongside conclusions are found in \Cref{ch:5::sec:conclusions}.

\section{Temporal streams in video-based action recognition}
\label{ch:5::sec:related}
We discuss two different approaches that are based on the creation of streams for addressing temporal information. The first set of methods consists of approaches that are based on the use of additional hand-crafted temporal features, such as optical flow, to represent temporal variations. The second group utilises information from frame sequences in streams based on either frame rates, groups of features or by separating low and high frequencies of activations.

\subsection{Streams with hand-crafted temporal features}
\label{ch:5::sec:related::sub:modalities}

% Two-steams with optical flow
Optical flow is a straightforward approach for the representation of motion features in videos. Two-stream networks \cite{simonyan2014two} combine spatial features extracted from video frames in the first stream and the supplementary temporal patterns of the second stream that use optical flow sequences. Information from the two streams is fused at the end to produce class predictions. Additional works have studied approaches for information fusion across the two streams \cite{park2016combining}. Lateral connections between the two streams have also been proposed to include temporal information within the spatial pattern extraction process \cite{feichtenhofer2016spatiotemporal}. Other works have investigated the division of inputs into temporal segments \cite{wang2016temporal} and spatial-based and temporal-based encoding of segments \cite{diba2017deep}. Tu \etal~\cite{tu2018multi} proposed the utilisation of streams addressing different spatio-temporal regions concerning the entire regions for the actor in the video and the locations that movements are most prevalent. Each stream then uses both RGB frames and optical flow for the creation of a four-stream model. 

Although these works provide a straightforward approach to explicitly model temporal information and their subsequent variations, the strong dependence on hand-coded optical flow data is limiting. As features relating to the spatial or temporal extent are learned separately, it prevents learning complex spatio-temporal features jointly in an end-to-end approach.

\subsection{Temporal streams with 3D Convolutions}
\label{ch:5::sec:related::sub:3dconv}

% Spatio-temporal stream-based approaches
Considering the temporal modelling limitations of 2D two-stream CNNs, some works have adopted the stream-based approach with 3D convolutions \cite{carreira2017quo}. Through this adaptation, inputs to each stream include stacks of either RGB frames extracted from a video sequence or a stack of optical flow representations for the cross-frame movements. Although the use of 3D convolutions in the RGB stream can effectively also extract temporal features by convolving stacks of frames, there is still a dependency on the representation quality of optical flow features in the second stream. To mitigate this issue, later works of Feichtenhofer \etal~\cite{feichtenhofer2019slowfast} included a dual 3D model in which inputs were stacked RGB frames sampled over different temporal iteration steps within the same video sequence. The sampling-based slow and fast frame rates showed to be beneficial in the extraction of motion patterns of shorter and longer durations. This approach has further been employed for the creation of global feature paths using entire videos as inputs, and local feature paths with local spatio-temporal segments, using two separate network pathways \cite{qiu2019learning}. Block-based approaches with octave convolutions \cite{chen2019drop} have also been used to model temporal variation in the frequency domain.

% Our method - Short and Long kernels
Despite the great promise that these methods have shown for the extraction of robust spatio-temporal features, the extraction of local spatio-temporal patterns is not done in relation to how informative they are at a global scale. The aim of our tri-branch method is to address within convolutional blocks, temporal feature disparities through the extraction of periodically varying space-time features and utilise the correlations between these features through a global attention mechanism.

\section{Multi-Temporal convolutions}
\label{ch:5::sec:methodology}

\begin{figure}[ht]
\includegraphics[width=\textwidth]{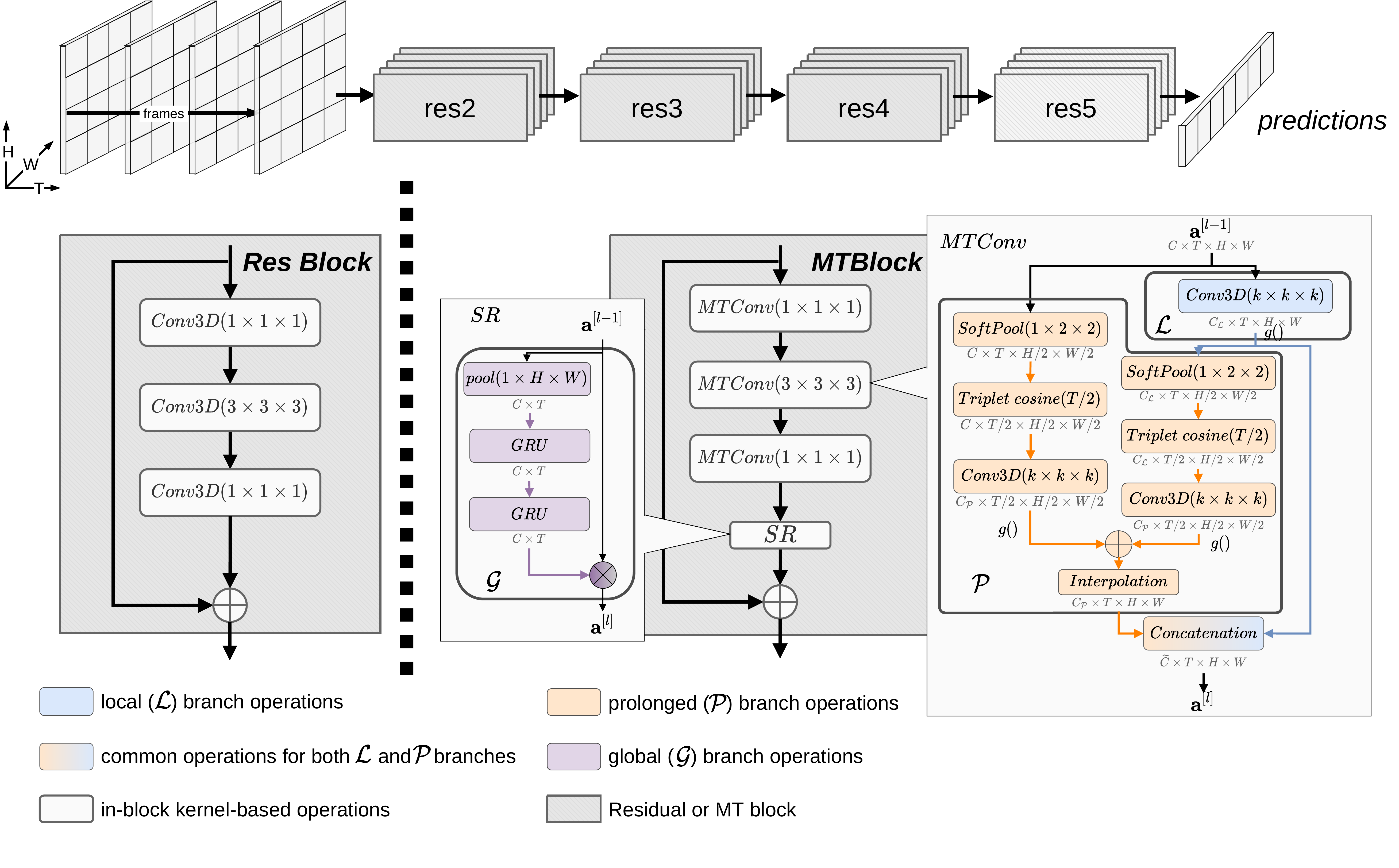}
\caption{\textbf{MTBlock structure}. Utilising X3D~\cite{feichtenhofer2020x3d} as backbone architectures, MTBlocks act as direct replacements for the Residual blocks (Res Block). In-block operations follow a sequence of Multi-Temporal Convolutions (MTConv) and a final Squeeze and Recursion (SR) feature alignment module as shown at the right side. We denote element-wise additions with $\oplus$ and element-wise multiplications with $\otimes$. }
\label{figure:mtconv}
\end{figure}

% Section outline
In this section, we describe multi-temporal convolutions (MTConvs) and their inner workings in terms of how information is processed. We then present the structure of the created blocks (MTBlocks, shown in \Cref{figure:mtconv}) that include the proposed modules.

% Notations
Formally and in line with the previous chapter, layer activations are denoted as $\textbf{a}_{(C \! \times \! T \! \times \! H \! \times \! W)}$ with $C$ channels, $T$ frames, $H$ height and $W$ width, respectively. Activations for each for the respective branches are denoted with $\textbf{a}^{\mathcal{L}}$ for the local branch ($\mathcal{L}$) and $\textbf{a}^{\mathcal{P}}$ for the prolonged branch ($\mathcal{P}$). Layers are indexed with $l$ and indicated as $\textbf{a}^{[l]}$ with $\textbf{a}^{[l],\mathcal{L}}$, $\textbf{a}^{[l],\mathcal{P}}$ in branch notation.

\subsection{Local and Prolonged branches}
\label{ch:5::sec:methodology::sub:lp}

% General 
A portion of layer $l$ channels ($\widetilde{C}$) is used by each of the local and prolonged branches. To determine the number of channels for each branch, a channel ratio parameter $\delta$ is used. The channel size of the input activations to the layer ($\textbf{a}^{[l-1]}$) is defined as $C$. Channels for the local branch ($C_{\mathcal{L}}$), based on ratio $\delta$, use the lowest integer value approximation (through the homonym function denoted with $\lfloor floor \rfloor$). Respectively, the prolonged branch channels ($C_{\mathcal{P}}$) use the maximum integer value approximation ($\lceil ceil \rceil$). We overview how the branch channel distribution is performed in \Cref{eq:rate}:

\begin{equation}
\label{eq:rate}
\begin{split}
    C_{\mathcal{L}} + C_{\mathcal{P}} = \widetilde{C} \; where \qquad \quad\\
    C_{\mathcal{L}} = \lfloor \delta * \widetilde{C}\rfloor \; and \; C_{\mathcal{P}} = \lceil(1-\delta) * \widetilde{C}\rceil \; 
\end{split}
\end{equation}

Inputs are first processed by the $\mathcal{L}$ and $\mathcal{P}$ branches. The local branch ($\mathcal{L}$) uses a single input with activation volumes $\textbf{a}^{[l-1]}$ of size $(C \! \times \! T \! \times \! H \! \times \! W)$. The prolonged branch ($\mathcal{P}$) uses a pair of inputs ($\textbf{a}^{[l],\mathcal{L}}, \textbf{a}^{[l-1]}$) with the first volume being the original layer input similar to the local branch. The second input is the output feature activations from the local branch $\mathcal{L}$ and of size $(C_{\mathcal{L}} \! \times \! T \! \times \! H \! \times \! W)$. Dual inputs are used in the prolonged branch ($\mathcal{P}$) as spatio-temporal patterns of elongated duration and spatial sizes are strongly correlated with local features, which are extracted by $\mathcal{L}$. With prolonged features incorporating the complexity of local short-term ones, the $\mathcal{P}$ branch effectively operates over $C_{\mathcal{L}}+C_{in}$ channels addressing the added complexity over $\mathcal{L}$. The $\mathcal{L}$ branch and $\mathcal{P}$ branch feature extraction process is summarised as seen in \Cref{eq:form1}:

\begin{equation}
\label{eq:form1}
    \textbf{a}^{[l]} = \mathcal{L}(\textbf{a}^{[l-1]}) \upSmallFrown (\mathcal{P}(\mathcal{L}(\textbf{a}^{[l-1]}), \textbf{a}^{[l-1]}))
\end{equation}

where $\upSmallFrown$ denotes the concatenation of the outputs from the two branches.

\textbf{Local branch in MTConv}. Short-term local motions in the input activations are extracted in the local branch. With layer input ($\textbf{a}^{[l-1]}$) we use a 3D convolution followed by batch normalisation (BN) \cite{ioffe2015batch} and compute feature volume ($\textbf{z}^{[l],\mathcal{L}}$) of $C_{\mathcal{L}}$ channels followed by non-linearity ReLU activation ($g()$). Unless otherwise stated, $g()$ refers to a ReLU activation. The final branch output takes the form of $\textbf{a}^{[l],\mathcal{L}}=g(\textbf{z}^{[l],\mathcal{L}})$.

\textbf{Prolonged branch in MTConv}. The extraction of patterns of extended duration is done over the prolonged branch. Information from the local branch ($\mathcal{L}$) and the layer input is used to extract features across larger spatio-temporal windows. The exploration of long-temporal features is done by reducing both inputs by a factor of two across their spatio-temporal dimensions. The size reduction is done by a factor of two as it provides a balanced trade-off between accuracy and computation. More aggressive strategies for spatio-temporal size reductions by larger factors might lead to significant information loss. Spatial downsampling over both inputs is done by their per-frame regional exponential maximum with \textit{SoftPool} \cite{stergiou2021refining} with the activations produced being of size $T \! \times \! H' \! \times \! W'$ (where $H'=H/2$ and $W'=W/2$). The activations are then downsampled temporally by a temporal triplet cosine frame selection to size $T' = T/2$. Detailed explanations for both methods are provided later in the section. The reduction in spatio-temporal size can increase the receptive fields of the prolonged kernels without requiring additional parameters while also reducing computational costs based on the size decrease of the activation volumes that are convolved. This inclusion of receptive fields, twice the duration of those in $\mathcal{L}$, allows for the exploration of temporal movements of larger spatio-temporal regions. Extended temporal patterns for inputs $\textbf{a}^{[l-1]}$ and $\textbf{a}^{[l],\mathcal{L}}$ are extracted by Conv3D operations followed by Batch Normalisation (BN). The complete process is formulated as:

\begin{equation}
\label{eq:prolonged}
    \textbf{a}^{[l],\mathcal{P}} = \mathcal{I}(g(\textbf{z}^{[l],\mathcal{L} \rightarrow \mathcal{P}}) \oplus g(\textbf{z}^{[l],\mathcal{P}}))
\end{equation}

in which element-wise addition is denoted by $\oplus$ and $\mathcal{I}()$ is the spatio-temporal tri-linear interpolation of the volume from size ($T' \! \times \! H' \! \times \! W'$) to original size ($T \! \times \! H \! \times \! W$). The feature volume $\textbf{z}^{[l],\mathcal{L} \rightarrow \mathcal{P}}$ corresponds to the extracted patterns from the reduced input $\textbf{a}^{[l],\mathcal{L}}$, while $\textbf{z}^{[l],\mathcal{P}}$ corresponds to features extracted from $\textbf{a}^{[l-1]}$:

\begin{equation}
\label{eq:prolonged_feats}
\centering
    \textbf{z}^{[l],\mathcal{L} \rightarrow \mathcal{P}} = \mathcal{T}(\overline{\textbf{a}}^{[l],\mathcal{L}}) * \textbf{w}^{\mathcal{L} \rightarrow \mathcal{P}} \; and \; \textbf{z}^{[l],\mathcal{P}} = \mathcal{T}(\overline{\textbf{a}}^{[l]}) * \textbf{w}^{\mathcal{P}}
\end{equation}

with $\mathcal{T}()$ the triplet cosine frame selection for a spatially pooled volume ($\overline{\textbf{a}}$). The convolutional weight vectors for the respective inputs are denoted as $\textbf{w}^{\mathcal{L} \rightarrow \mathcal{P}}$ and $\textbf{w}^{\mathcal{P}}$.

\begin{figure}[ht]
\centering
\includegraphics[width=0.85\linewidth]{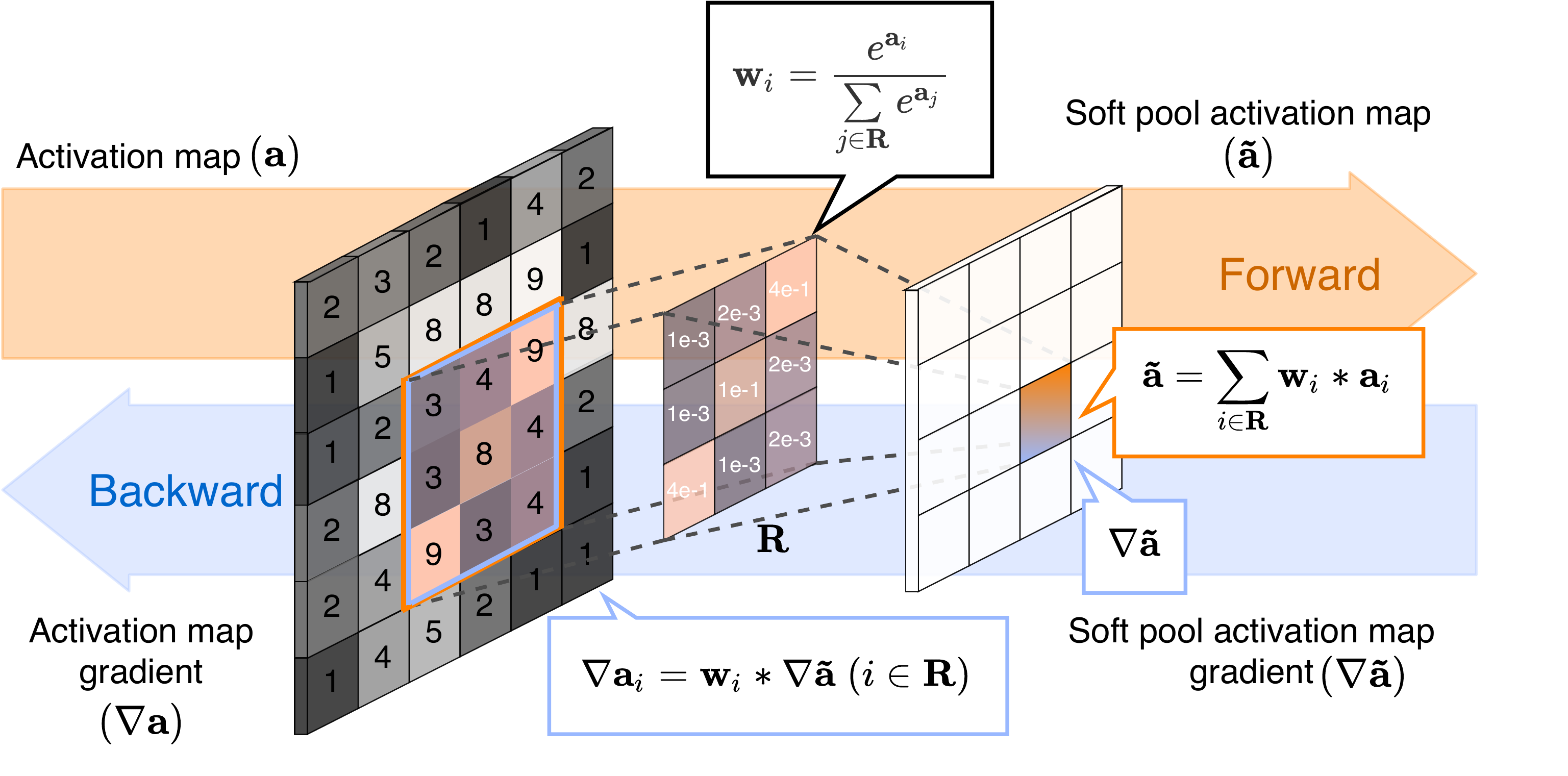}
\caption{\textbf{SoftPool spatial pooling}. Forward operations (in \textcolor{orange}{orange}) the kernel uses exponential maximum of each activation to produce a weighted sum of region \textbf{R}. The weights are also used during the backward gradient ($\nabla \textbf{a}$) calculations (in \textcolor{bblue}{blue}). }
\label{figure:softpool}
\end{figure}

% Downsampling methods - Spatial
\textbf{Prolonged branch spatial downsampling}. \textit{SoftPool} \cite{stergiou2021refining} uses soft-maximum approximation weighting to reduce the spatial dimensionality of the input. The method is based on the calculation of the softmax for each of the inputs within the kernel region. The softmax weights have a proportional effect to the output with higher-valued activations having a larger effect on the output than low-valued activations. A formulation based on input $\textbf{a}$ and frame $t$ with spatial region $\textbf{R}$ of size $H \! \times \! W$ is shown in \cref{eq:softpool}. 

\begin{equation}
\label{eq:softpool}
\centering
    \overline{\textbf{a}}_{t,r} = \sum_{r \in \textbf{R}}\frac{e^{\textbf{a}_{t,r}} * \textbf{a}_{t,r}}{\sum\limits_{k \in \textbf{R}} e^{\textbf{a}_{t,k}}}, \; \forall \; t \in |T|
\end{equation}

% Downsampling methods - Temporal
\textbf{Prolonged branch temporal downsampling}. Image-based pooling methods that have been extended for spatio-temporal data are based on the fusion of multiple frames. This fusion can degrade the quality of spatial details. The produced effects are similar to afterimages in which edges of objects within frames are less distinguishable as their cross-frame motions are joined together to a single frame. With our method being dependant on the preservation of such features, we instead use a temporal downsample method based on frame selection on the spatially-pooled activation volume ($\overline{\textbf{a}}$). 

\begin{figure}[ht]
\centering
\includegraphics[width=.75\linewidth]{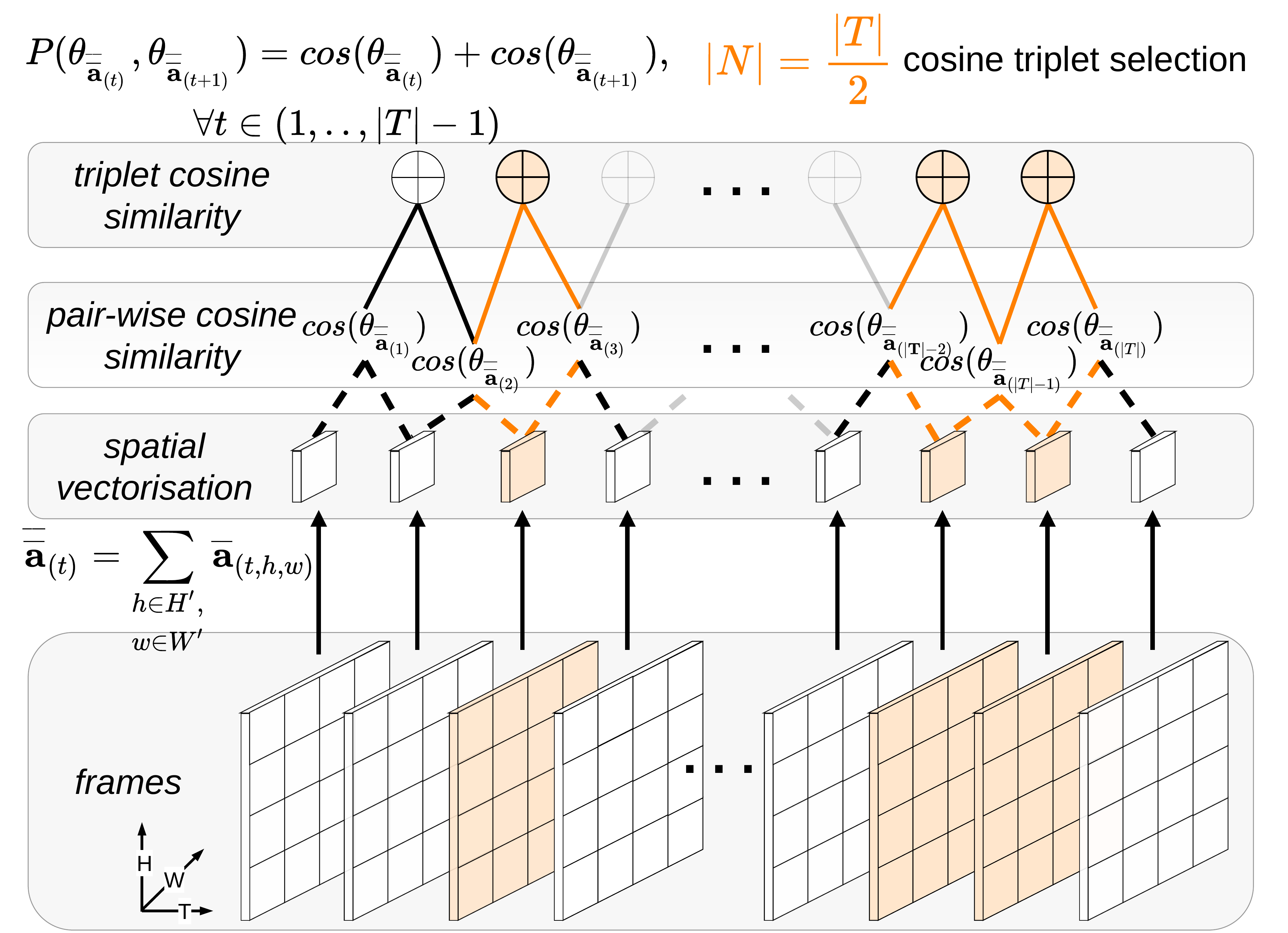}
\caption{\textbf{Temporal triplet cosine similarity pooling}. Frames are selected based on their channel-wise sum ($\oplus$) of their per-frame pair cosine similarity ($cos(\theta_{\overline{\overline{\textbf{a}}}_{(t)}})$) calculated from their spatially-summed volumes ($\overline{\overline{\textbf{a}}}$)}
\label{figure:triplet}
\end{figure}

The complete frame selection process is visualised in \Cref{figure:triplet}. For each frame $t$ the entire spatial regions is averaged to a single value for every channel. The activation volume produced ($\overline{\overline{\textbf{a}}}$) contains frame-wise vectors that can be used to measure their per-pair, feature-wise similarity. This is done based on their magnitude and sum of their dot products:

\begin{equation}
\label{eq:cosine}
\centering
    cos(\theta_{\overline{\overline{\textbf{a}}}_{(t)}}) = \frac{\sum\limits_{c \in C} \overline{\overline{\textbf{a}}}_{(t,c)} * \overline{\overline{\textbf{a}}}_{(t+1,c)}}{\sqrt{\sum\limits_{c \in C}\overline{\overline{\textbf{a}}}_{(t,c)}^{2}} * \sqrt{\sum\limits_{c \in C}\overline{\overline{\textbf{a}}}_{(t+1,c)}^{2}} }
\end{equation}

The cosine similarity pairs are then summed together for the creation of cosine similarity triplets ($P(\theta_{\overline{\overline{\textbf{a}}}_{(t)}},\theta_{\overline{\overline{\textbf{a}}}_{(t+1)}}) = cos(\theta_{\overline{\overline{\textbf{a}}}_{(t)}})+cos(\theta_{\overline{\overline{\textbf{a}}}_{(t+1)}})$). The produced triplet represents a concatenated view of how similar features in frame $t$ are in comparison to features from the preceding frame ($t-1$) and succeeding frame ($t+1$). The frame selection then takes the form of selecting the frame locations ($N$) with the lowest triplet cosine similarity value focusing on the most informative frames:

\begin{equation}
\label{eq:frame_selection}
\centering
\begin{split}
    \argmin_{\forall n \in N} P(\theta_{\overline{\overline{\textbf{a}}}_{(n)}},\theta_{\overline{\overline{\textbf{a}}}_{(n+1)}}) = cos(\theta_{\overline{\overline{\textbf{a}}}_{(n)}})+cos(\theta_{\overline{\overline{\textbf{a}}}_{(n+1)}}),\\
    \; where \; N \subset T, |N|=|T/2| \qquad \qquad
\end{split}
\end{equation}

The cross-frame similarity is used for frame selection, instead of frame regions temporally fused together, the per-frame activations remain consistent over the produced temporally sub-sampled volume.

\subsection{MTBlocks structure}
\label{ch:5::sec:methodology::sub::mtblocks}

% SR
\textbf{Global aggregated feature importance}. The concatenated activations of the local ($\mathcal{L}$) and prolonged ($\mathcal{P}$) branches are aligned based on the importance of each feature in the context of the entire video sequence. The role of the \textit{global aggregated feature importance} branch ($\mathcal{G}$) is the creation of coherent activations based on averaged feature attention through \textit{Squeeze and Recursion} \cite{stergiou2021learn} with GRU \cite{cho2014learning} recurrent cells. The branch operates over a vectorised version of the original volume pooled by its spatial dimensionality ($pool(\textbf{a}^{[l-1]})$). The pooled volume is processed through a dual-layer recurrent sub-network for the discovery and amplification of globally-informative features. Initial refinement of salient features is done by the update gate ($\textbf{z}_{(t)}$) that uses frame ($t$) input ($pool(\textbf{a}^{[l-1]})_{(t)}$) and the previous ($t-1$) hidden state ($\textbf{h}_{(t-1)}$) of the previous recurrent cell (for time $t-1$), through a sigmoid ($\sigma$) activation with weight $\textbf{W}_{z}$ and bias $\textbf{b}_{z}$:
 
\begin{equation}
\label{eq:update_gate}
    \textbf{z}_{(t)} =\{ \sigma(\textbf{W}_{z} * [\textbf{h}_{(t-1)},pool(\textbf{a}^{[l-1]})_{(t)}] + \textbf{b}_{z}) \} \quad \quad\\
\end{equation}

% Correspondence to LSTMs
The resulting update gate ($\textbf{z}_{(t)}$) can be seen as a single operation that corresponds to LSTM's forget ($\textbf{f}_{(t)}$) gate from \Cref{eq:forget_gate} and input ($\textbf{i}_{(t)}$) gate from \Cref{eq:input_gate}.

Cell input $pool(\textbf{a}^{[l-1]})_{(t)}$ and previous state outputs $\textbf{h}_{(t-1)}$ also pass through a reset gate ($\textbf{r}_{(t)}$), which uses weight ($\textbf{W}_{z}$) and bias ($\textbf{b}_{z}$), in order to ignore less time-consistent features: 

\begin{equation}
\label{eq:reset_gate}
    \textbf{r}_{(t)} =\{ \sigma(\textbf{W}_{r} * [\textbf{h}_{(t-1)},pool(\textbf{a}^{[l-1]})_{(t)}] + \textbf{b}_{z}) \} \quad \quad\\
\end{equation}

% LSTM equivalence
In LSTMs this is done by computing the candidate values ($\widetilde{\textbf{C}}_{(t)}$) in \cref{eq:candidate_values} and then the cell state ($\textbf{C}_{(t)}$) in \cref{eq:cell_state}. GRUs do not distinguish between cell states ($\textbf{C}_{(t)}$) and hidden states ($\textbf{h}_{(t)}$), as only their hidden states $h_{(t)}$ are passed across temporal cells and used as cell outputs.

Both update and reset gates act in a complementary manner with the same inputs. A candidate hidden state is computed ($\widetilde{\textbf{h}}_{(t)}$), based on the activations produced by the reset gate, through a $tanh$ activation, and includes reduced influence from the previous state ($\textbf{h}_{(t-1)}$) based on rate $\textbf{r}_{(t)}$. The produced cell state is the fusion of a portion of the previous state ($\textbf{z}_{(t)} * \textbf{h}_{(t-1)}$) and the supplementary portion of the candidate hidden state ($(1-\textbf{z}_{(t)}) * \widetilde{\textbf{h}}_{(t)}$):

\begin{align}
    \widetilde{\textbf{h}}_{(t)} = tanh(\textbf{W}_{h} * [\textbf{r}_{(t)}*\textbf{h}_{(t-1)},pool(\textbf{a}^{[l-1]})_{(t)}] + \textbf{b}_{h}) \label{eq:candidate_out}\\
    \textbf{h}_{(t)} = \textbf{z}_{(t)}*\textbf{h}_{(t-1)}+(1-\textbf{z}_{(t)}) * \widetilde{\textbf{h}}_{(t)} \qquad \quad \label{eq:gruout}
\end{align}

\begin{figure}[ht]
\centering
\includegraphics[width=0.85\linewidth]{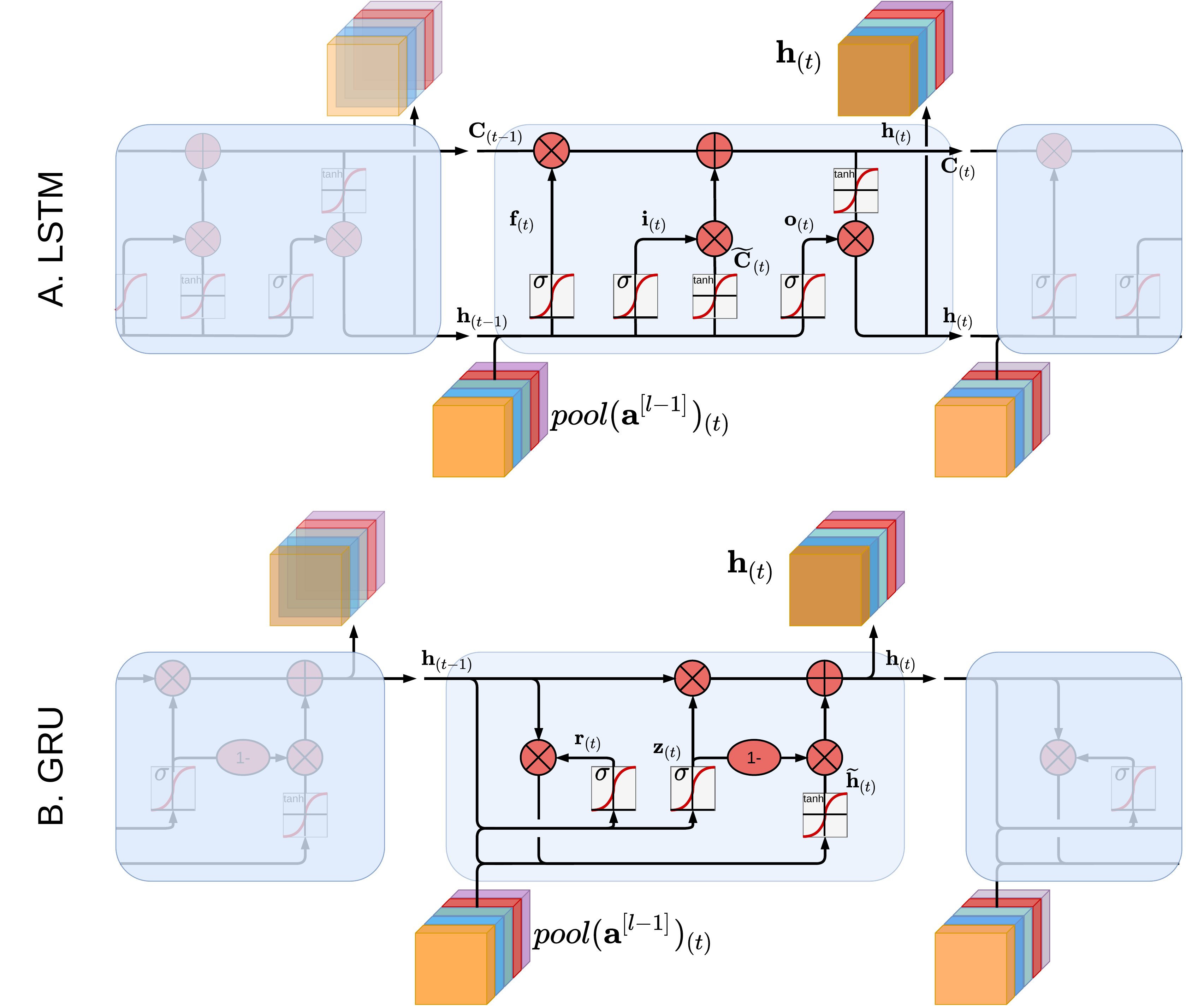}
\caption{\textbf{(A) LSTMs and (B) GRUs.} Overview of both architectures and their sequential chain of cells. GRUs only exchange their hidden states ($\textbf{h}_{(t)}$) across cells while LSTM compute both a cell state ($\textbf{C}_{(t)}$) and a hidden state ($\textbf{h}_{(t)}$).}
\label{figure:lstm2gru}
\end{figure}

% Reference to LSTMs
This approach significantly simplifies the feature selection process in comparison to LSTMs in \Cref{eq:candidate_values,eq:cell_state,eq:lstmout}. \Cref{figure:lstm2gru} demonstrates the structural differences between the two cell types. The inclusion of hidden states ($\textbf{h}$) in the activation maps ($\textbf{a}^{[l]}$) is performed the same as for LSTMs.

\subsection{Multi-Temporal Networks (MTNet)} 
\label{ch:5::sec:methodology::sub::mtnets}

% General architecture construction
We propose three MTNet architecture variants that use X3D \cite{feichtenhofer2020x3d} backbones, respectively X3D$_{S}$, X3D$_{M}$ and X3D$_{L}$. These models vary in size and GFLOP usage, and we replace their Residual blocks and 3D Convs with the proposed MTBlocks and MTConvs, as shown in \Cref{figure:mtconv}. We denote our models as MTNet$_{S}$, MTNet$_{M}$ and MTNet$_{L}$. The architectures follow a step-wise network and block expansion as recently proposed for video \cite{feichtenhofer2020x3d} and image-based models \cite{radosavovic2020designing}. Details of the three proposed models in terms of the number of parameters and GFLOPs appear in \Cref{tab:K400_accuracies_ch5}.

\section{Main results}
\label{ch:5::sec:results}

% About
The evaluation of our proposed method is done over the three MTNets of different sizes and on five action recognition benchmark datasets. We overview the experiment setting for our tests in \Cref{ch:5::sec:results::sub::settings}. We compare against state-of-the-art models on Kinetics-400 in \Cref{ch:5::sec:results::sub::kinetics-400}, on Moments in Time in \Cref{ch:5::sec:results::sub::mit}, on Kinetics-700 (2020) in \Cref{ch:5::sec:results::sub::kinetics-700} and on HACS in \Cref{ch:5::sec:results::sub::hacs}. In \Cref{ch:5::sec:results::sub::ablation} we perform additional ablation studies for different branch ratios ($\delta$), recurrent cells and pooling strategies for the prolonged branches. Experiments for transfer learning on UCF-101 are presented in \Cref{ch:5::sec:results::sub::TL}.

\subsection{Environment settings}
\label{ch:5::sec:results::sub::settings}

Our training environment is based on the settings that were also used by Feichtenhofer \cite{feichtenhofer2020x3d} for X3D networks. Some differences include the frame sampling process, the use of multigrid batch size schedule and spatial-data augmentations. We detail our choices below. 

\begin{figure}[ht]
\centering
\includegraphics[width=\linewidth]{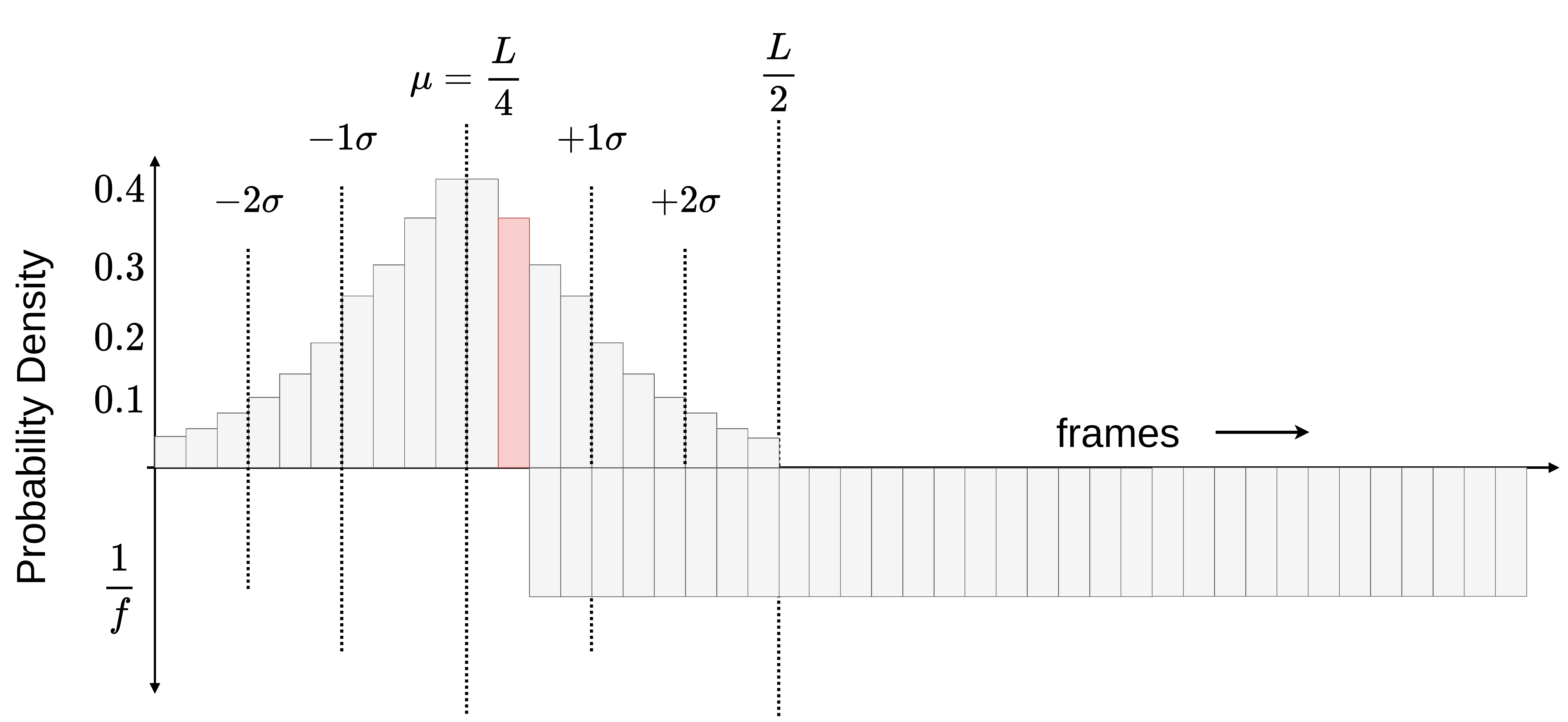}
\caption{\textbf{Frame sampling probability density.} Top normal distribution, with mean ($\mu = \frac{L}{4}$) and standard deviation ($\sigma = \frac{L}{6}$), is used to determine the start frame location ($s$). The probabilities for the rest of the frames are evenly distributed based on a uniform distribution of frame probabilities $p=\frac{1}{f}$, where $f$ is the remainder of frames $f=L-s$.}
\label{figure:frame_sampling}
\end{figure}

% Interval selection
\textbf{Interval selection.} The frame selection is done through uniform random sampling. The selection of the starting frame is done based on the assumption of a normal probabilistic distribution between (0 - $\frac{L}{2}$) where $L$ is the total length of the entire video sequence. This start frame initialisation is done to ensure that the majority of the middle frames will be selected, as in most datasets the primary actions are preformed in middle frames. The probability distribution of frames is shown in \cref{figure:frame_sampling}. Based on the average clip length, for each dataset, we use equivalently sized temporal strides. The average clip length in HACS is 60 frames, and we thus use a stride of 2. For the Kinetics variants, average clip length is 250 frames so we set the temporal stride to 5. Similarly for MiT we use a temporal stride of 3 with average duration of 90 frames.

% GFLOPS x views
\textbf{Computational inference.} We report computational cost over two different measures. We first report the computational inference as the measures used in \Cref{ch:4::sec:results::sub:settings}. This can express the computational costs (FLOPs) with respect to 10 clips with 3 spatial augmentations. The frame size used is $256 \! \times \! 256$. The inference time is then reported as the number of GFLOPs per spatio-temporal view (clips times crops). As our second measure, we report the inference cost in relation to latency (in msec).

% Multigrid sampling
\textbf{Multigrid batch schedule}. We use a multigrid scheme \cite{wu2020multigrid} for improvements in the training speeds. Similar to \Cref{ch:4::sec:results::sub:settings}, mini-batch size sampling is done proportionally to the original batch size through the condition that $b \! \times \! t \! \times \! h \! \times \! w = B \! \times \! T \! \times \! H \! \times \! W$, in which ($b,t,h,w$) represent the scaled batch size, number of frames, height and width dimensions. Equally, ($B,T,H,W$) are the original dimensions. This is done in order to maintain equivalence between the computational costs of the new and previous branches.

% Environment parameters MTNets
\textbf{Training parameters for MTConv.} We train our models on HACS for which we use \textit{random initialisation  without pre-training (\enquote{from scratch})}. We additionally set the initial mini-batch size to 16 clips per GPU (with the total mini-batch size of 64)\footnotemark \footnotetext{All experiments were performed with mixed half-precision (float16) for improved memory utilisation efficiency. Batch Normalisation is computed with single-precision (float32) for scaling stability. SoftPool is also computed with single precision to avoid gradient underflows.}. Frame selection is done by the aforementioned frame sampling procedure. For the spatial domain, we interpolate all frames so that the shorter side is of size 320. We then perform a random crop of $256 \! \times \! 256$. We note that the input frame size difference between MTNet$_S$ and MTNet$_M$ is part of the architectural parameters as with X3D$_S$ and X3D$_M$. In our case, inputs of $256 \! \times \! 256$ size in MTNet$_S$ are resized to $182 \! \times \! 182$ based on the $\sim \! 30\%$ decrease as proposed in \cite{feichtenhofer2020x3d}. All experiments were performed over 400 epochs with momentum of 0.9 and weight decay of $5 \! \times \! 10^{-5}$. Unless stated otherwise, all of our models use $\delta=0.875$. We explain this choice in \Cref{ch:5::sec:results::sub::ablation}.

% LR schedule X3D
\textbf{Cosine-based learning rate schedule on X3D}. Similarly to relevant works \cite{feichtenhofer2019slowfast}, as well as the suggested training recipe for X3D networks \cite{feichtenhofer2020x3d}, we use a cosine annealing learning rate decay schedule \cite{loshchilov2016sgdr}. This schedule is based on periodic decrease and increase of the learning rate through the use of a cosine function. The learning rate $lr_{n}$ for iteration $n$ is calculated as $lr_{n} = lr_{0} * 0.5[cos(\frac{n}{n_{max}}\pi)+1]$ in which $n_{max}$ is the total number of iterations and $lr_{0}$ is the starting learning rate. We use $lr_{0}=1.16$ and learning rate warmup for the first 8k iterations similar to \cite{feichtenhofer2020x3d,feichtenhofer2019slowfast}.

% Data augmentation
\textbf{Spatial data augmentation.} We use the same data augmentation strategies as for SRTG in \Cref{ch:4::sec:results::sub:settings}. We set the sequential data augmentation probabilities to 70\% to reduce data pre-processing times. The spatial augmentations include blurring, brightness increases, contrast increase and decrease as well as geometrical transformations.

\subsection{Results on Kinetics-400 (K-400)}
\label{ch:5::sec:results::sub::kinetics-400}

% Kinetics-400
% MTNet_L
We compare the proposed architectures (MTNets) in \cref{tab:K400_accuracies_ch5} with state-of-the-art results. We additionally present results for the two top-performing SRTG networks with 3D and (2+1)D convolutions. This allows for a direct evaluation in order to determine the improvements of MTConvs over the architectures introduced in \Cref{ch4}. Compared to the current top performing X3D-XL \cite{feichtenhofer2020x3d}, our largest \textbf{MTNet$_{L}$} architecture produces comparable performance (-1\% top-1 and -0.7\% top-5 lower accuracies), while considerably reducing the number of multiply-addition operations (measured in GFLOPs). $MTNet_{L}$ can reduce the computations by a factor of $\! \times  2.75$ in relation to X3D-XL. In general, the performance expected from \textbf{MTNet$_{L}$} is similar to that of SlowFast-101 (SF) \cite{feichtenhofer2019slowfast} which utilised more than 12 times the number of GFLOPs compared to our model. Considering the computational requirements, \textbf{MTNet$_{L}$} largely outperforms MFNet \cite{chen2018multifiber} with similar number of GFLOPs, by +5.3\% in top-1 and +2.8\% in top-5 accuracies. 

% MTNet_M
The smaller \textbf{MTNet$_{M}$} shows accuracy rates close to \textit{Channel-Separated Network} (ip-CSN-101) \cite{tran2019video}, \textit{Temporal Adaptive Module (TAM)} ResNet-50 and the ResNet 50 variant of SlowFast, while being significantly more efficient than any of the alternative architectures. Specifically, \textbf{MTNet$_{M}$} is $9.4$ times more efficient than ip-CSN-101, $9.7$ times more than TAM and $7.5$ times more efficient than SF-50. This reduction in multiply-add operations can significantly benefit the required training times in relation to the compared networks. Given its modest number of multiply-add operations (FLOPs), \textbf{MTNet$_{M}$} can still outperform R(2+1)D ResNet101 \cite{tran2018closer} and \textit{Temporal Shift Module} (TSM) \cite{lin2019tsm}.

% MTNet_S
Our smallest MTNet network (\textbf{MTNet$_{S}$}) performs on par with TSM while having the lowest number of GFLOPs from all tested networks. Despite the significant reduction of GFLOPs, \textbf{MTNet$_{S}$} can still perform better than both MFNet and I3D \cite{carreira2017quo} networks. We note that the proposed MTNet architectures are the only family of spatio-temporal architectures within the range of sub-20 GFLOPs that produce accuracy rates similar to that of the top-performing, and significantly more expensive, state-of-the-art models.

\begin{table}[t]
\caption{\textbf{Comparison with K-400 state-of-the-art}. For consistency with previous testing methods, we report the model complexity as the GFLOPs per single clip view $\! \times \!$ the number of clips with spatial cropping of size $256 \! \times \! 256$.}
\centering
\resizebox{\textwidth}{!}{%
\renewcommand{\arraystretch}{1.5}
\begin{tabular}{c|c|c|c|c|c|c}
\hline
Model & 
Input &
Backbone &
top-1 & top-5 & 
GFLOPs $\! \times \!$ views &
Params\\
\hline

R(2+1)D \cite{tran2018closer} &
$16 \! \times \! 224^{2}$ &
ResNet101 &
62.8 & 83.9 &
$152 \! \times \! 115$&
63.6M\\

I3D \cite{carreira2017quo}  &
$16 \! \times \! 224^{2}$ &
InceptionV1 &
71.6 & 90.0 &
$108 \! \times \! N/A$&
12M\\

MF-Net \cite{chen2018multifiber} &
$16 \! \times \! 224^{2}$ &
ResNet50 &
72.8 & 90.4 &
$11.1 \! \times \! 50$ &
\textbf{8.0}M\\

TSM \cite{lin2019tsm} &
$16 \! \times \! 224^{2}$ &
ResNet50 &
74.7 & 91.4 &
$65 \! \times \! 10$&
24.3M\\

ip-CSN-101 \cite{tran2019video} &
$8 \! \times \! 224^{2}$ &
ResNet101 &
76.7 & 92.3 &
$83.0 \! \times \! 30$ &
24.5M\\

TAM \cite{liu2021tam} &
$16 \! \times \! 256^{2}$ &
ResNet50 &
76.9 & 92.9 &
$86 \! \times \! 12$ &
25.6M\\

SF-50 \cite{feichtenhofer2019slowfast} &
$(32,4) \! \times \! 224^{2}$ &
ResNet50 &
77.0 & 92.6 &
$65.7 \! \times \! 30$ &
34.4M\\

ip-CSN-152 \cite{tran2019video} &
$8 \! \times \! 224^{2}$ &
ResNet152 &
77.8 & 92.8 &
$108.8 \! \times \! 30$ &
32.8M\\

SF-101  \cite{feichtenhofer2019slowfast} &
$(32,4) \! \times \! 224^{2}$ &
ResNet101 &
77.9 & 93.5 &
$213 \! \times \! 30$ &
53.7M\\

X3D-XL \cite{feichtenhofer2020x3d} &
$16 \! \times \! 224^{2}$ &
ResNet(X3D) &
\textbf{79.1} & \textbf{93.9} &
$48.4 \! \times \! 30$ &
11.0M\\
\hline

SRTG r3d-101 \cite{stergiou2021learn} \textbf{(ours)} &
$16 \! \times \! 224^{2}$ &
ResNet101 &
73.2 & 91.3 &
$78.1 \! \times \! 30$ &
107.1M\\

SRTG r(2+1)d-101 \cite{stergiou2021learn} \textbf{(ours)} &
$16 \! \times \! 224^{2}$ &
ResNet101 &
73.8 & 92.0 &
$163.1 \! \times \! 30$ &
105.3M\\
\hline

MTNet$_{S}$ \cite{stergiou2021refining} \textbf{(ours)} &
$16 \! \times \! 256^{2}$ &
ResNet(X3D) &
74.8 & 92.1 &
$\mathbf{5.8 \! \times \! 30}$ &
25.8M\\

MTNet$_{M}$ \cite{stergiou2021refining} \textbf{(ours)} &
$16 \! \times \! 256^{2}$ &
ResNet(X3D) &
76.6 & 92.5 &
$8.8 \! \times \! 30$ &
25.8M\\

MTNet$_{L}$ \cite{stergiou2021refining} \textbf{(ours)} &
$16 \! \times \! 256^{2}$ &
ResNet(X3D) &
78.1 & 93.2 &
$17.6 \! \times \! 30$ &
50.1M\\

\end{tabular}%
}
\label{tab:K400_accuracies_ch5}
\end{table}

\begin{figure}[ht]
\centering
\includegraphics[width=\linewidth]{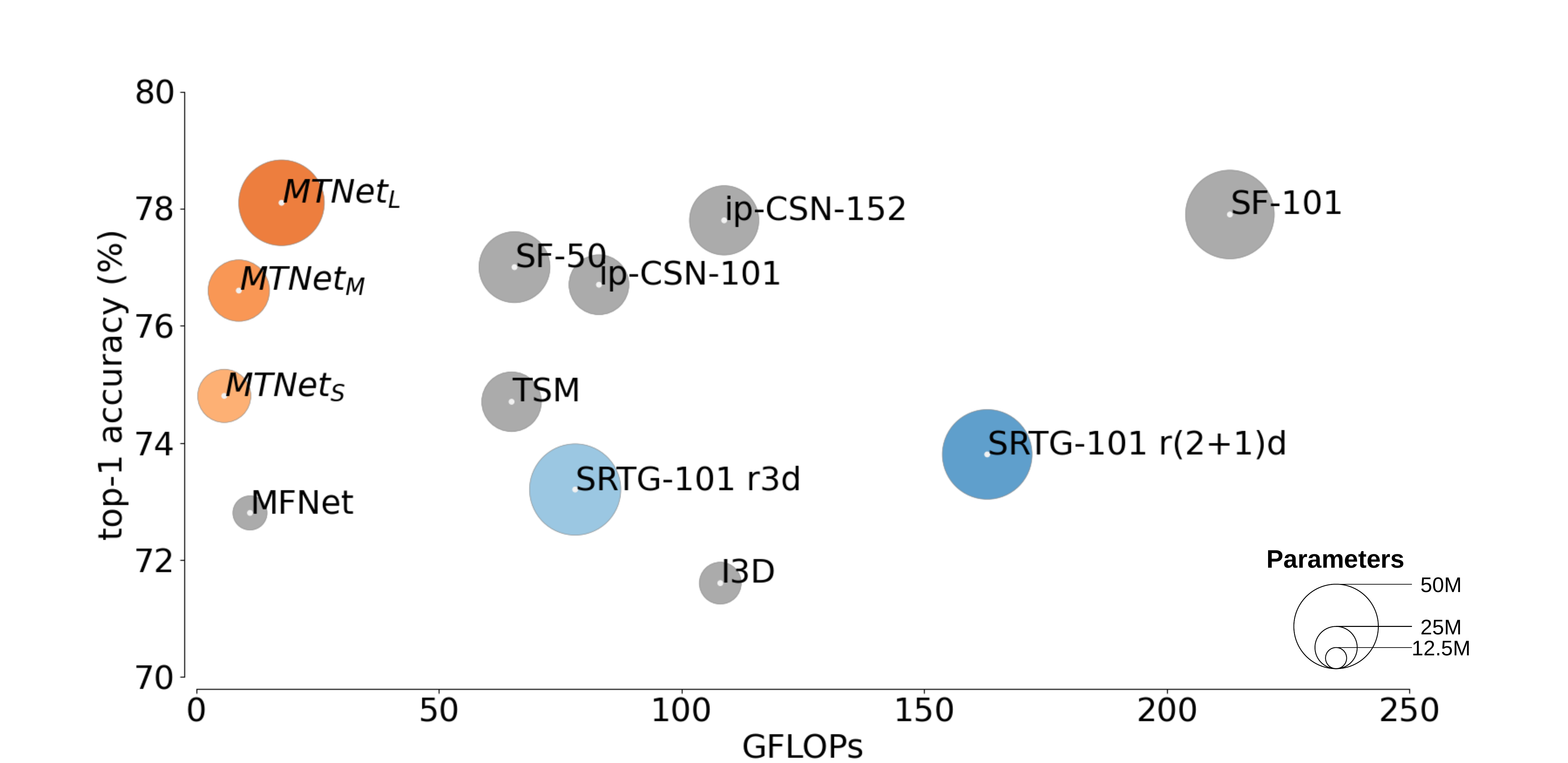}
\caption{\textbf{Accuracy to GFLOP/parameters comparison} on K-400. The size of the blob corresponds to the number of parameters that the network uses. As shown, MTNets are significantly more efficient in comparisons to state-of-the-art counterparts while maintaining competitive accuracy results.}
\label{figure:k400_acc_flop_params}
\end{figure}

% SRTG (2+1)D
The tested SRTG variant (\textbf{SRTG r(2+1)d-101}) was also trained with the parameters of \Cref{ch:4::sec:results::sub:settings}. The network demonstrates the ability to improve the accuracies in comparison to the originally proposed R(2+1)D \cite{tran2018closer}. The inclusion of SRTG modules for temporal channel dynamics calibration can produce accuracies similar to those of TSM and MF-Net. However, compared to the smaller MTNet$_{S}$ the accuracy benefits of extracting spatio-temporal patterns over varying sized space-time locations are visible.

% SRTG 3D
We additionally include in our experiments \textbf{SRTG r3d-101}. We show that the expected performance is similar to that of TSM and the previously mentioned \textbf{SRTG r(2+1)d-101}. Based on the overall simplicity of the architecture, the accuracy improvements with the inclusion of SRTG are notable. In comparison to the tested MTNet sizes, \textbf{SRTG r(2+1)d-101} shows both lower performance (1.6--4.9\%) as well as substantially larger computational requirements. The number of required GFLOPs is equivalent to 4.43--13.4 times that of the MTNets variants.

% Accuracy w.r.t GFLOPs/params
We present a visual representation of the accuracies in K-400 with respect to the computational complexity of each model (GFLOPs) and the number of parameters in \Cref{figure:k400_acc_flop_params}. MTNets can better balance the trade-off between computational complexity and accuracy.

\begin{table}[ht]
\caption{\textbf{Spatio-temporal method comparison on K-400}. Accuracy rates for spatio-temporal 3D conv methods on Kinetics with ResNet-50 as backbone. }
\centering
\resizebox{.9\textwidth}{!}{%
\renewcommand{\arraystretch}{1.5}
\begin{tabular}{c|ll|l|l}
\hline
Method & 
top-1 &
top-5 &
FLOPs (G) &
Params (M)\\
\hline

\textbf{(Baseline)} 3D \cite{hara2018can} &
61.3  &
83.1  &
53.2  &
36.7  \\[0.25em]
\hline

(2+1)D \cite{tran2018closer} &
61.8 \textcolor{applegreen}{($+0.5$)} &
83.5 \textcolor{applegreen}{($+0.4$)} &
56.0 \textcolor{cadmiumred}{($+2.8$)} &
38.8 \textcolor{cadmiumred}{($+2.1$)} \\[0.25em]

Multi-Fiber \cite{chen2018multifiber} &
72.8 \textcolor{applegreen}{($+11.5$)} &
90.4 \textcolor{applegreen}{($+7.3$)} &
\textbf{22.5} \textcolor{applegreen}{($-30.7$)} &
\textbf{8.0} \textcolor{applegreen}{($-28.7$)} \\[0.25em]

Slow-only \cite{feichtenhofer2019slowfast} &
72.6 \textcolor{applegreen}{($+11.5$)} &
90.3 \textcolor{applegreen}{($+7.2$)} &
27.3 \textcolor{applegreen}{($-25.9$)} &
26.6 \textcolor{applegreen}{($-10.1$)} \\[0.25em]

SlowFast \cite{feichtenhofer2019slowfast} &
74.3 \textcolor{applegreen}{($+13$)} &
91.0 \textcolor{applegreen}{($+7.9$)} &
39.8 \textcolor{applegreen}{($-13.4$)} &
34.4 \textcolor{applegreen}{($-2.3$)} \\[0.25em]

\hline

MTConv \textbf{(ours)} &
\textbf{74.8} \textcolor{applegreen}{($+13.5$)} &
\textbf{91.3} \textcolor{applegreen}{($+8.2$)} &
23.1 \textcolor{applegreen}{($-30.1$)} &
35.7 \textcolor{applegreen}{($-1.0$)} \\[0.25em]

\end{tabular}%
}
\label{tab:K400_accuracies_Res50_ch5}
\end{table}

\begin{figure}[ht]
\centering
     \begin{subfigure}[b]{0.48\textwidth}
         \centering
         \includegraphics[width=\textwidth]{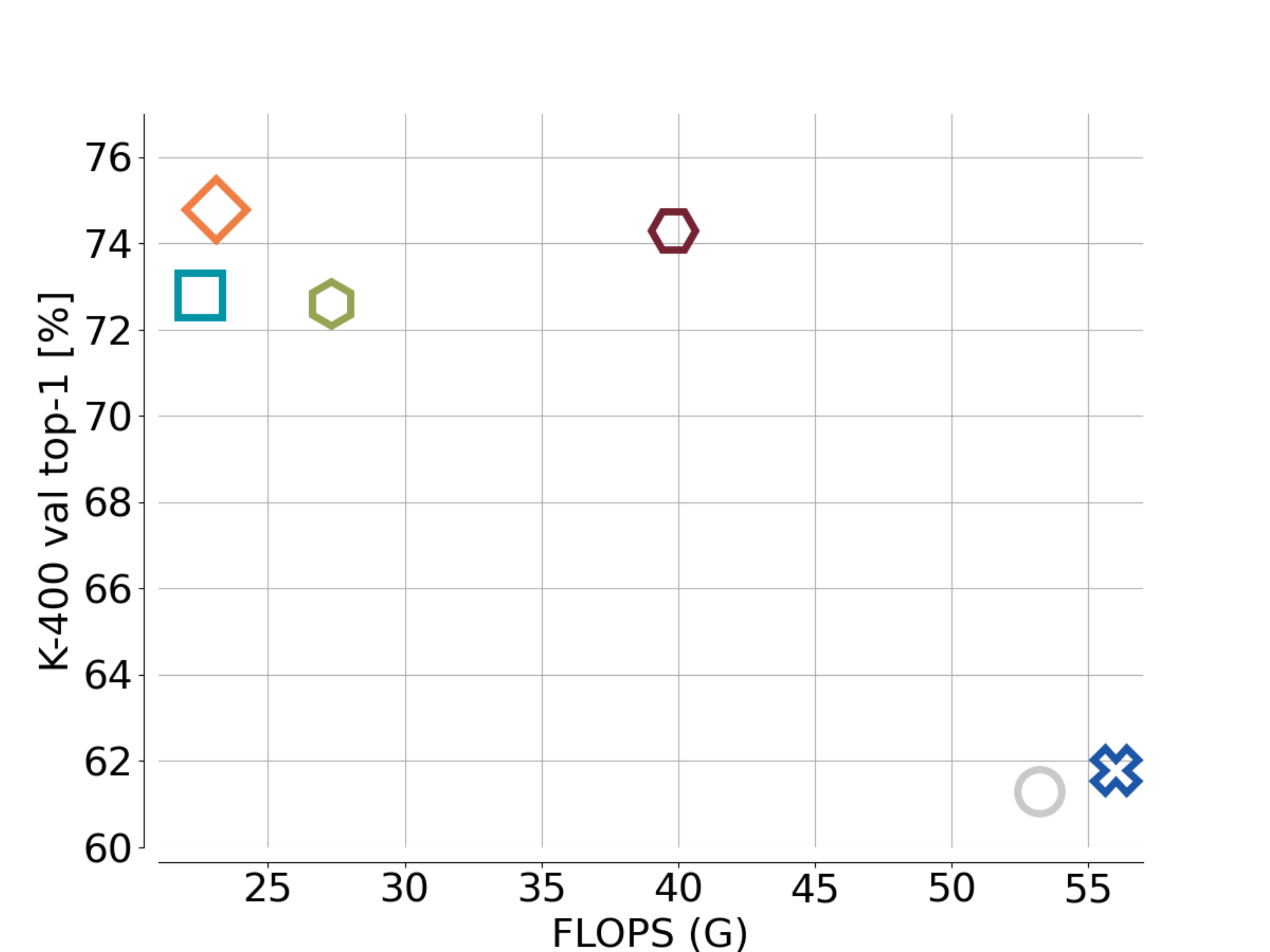}
         \caption{\textbf{Top-1 accuracy with respect to the number of GFLOPs}}
         \label{figure:k400_acc_ab_flops_par_top1_flops}
     \end{subfigure}
     \hfill
     \begin{subfigure}[b]{0.48\textwidth}
         \centering
         \includegraphics[width=\textwidth]{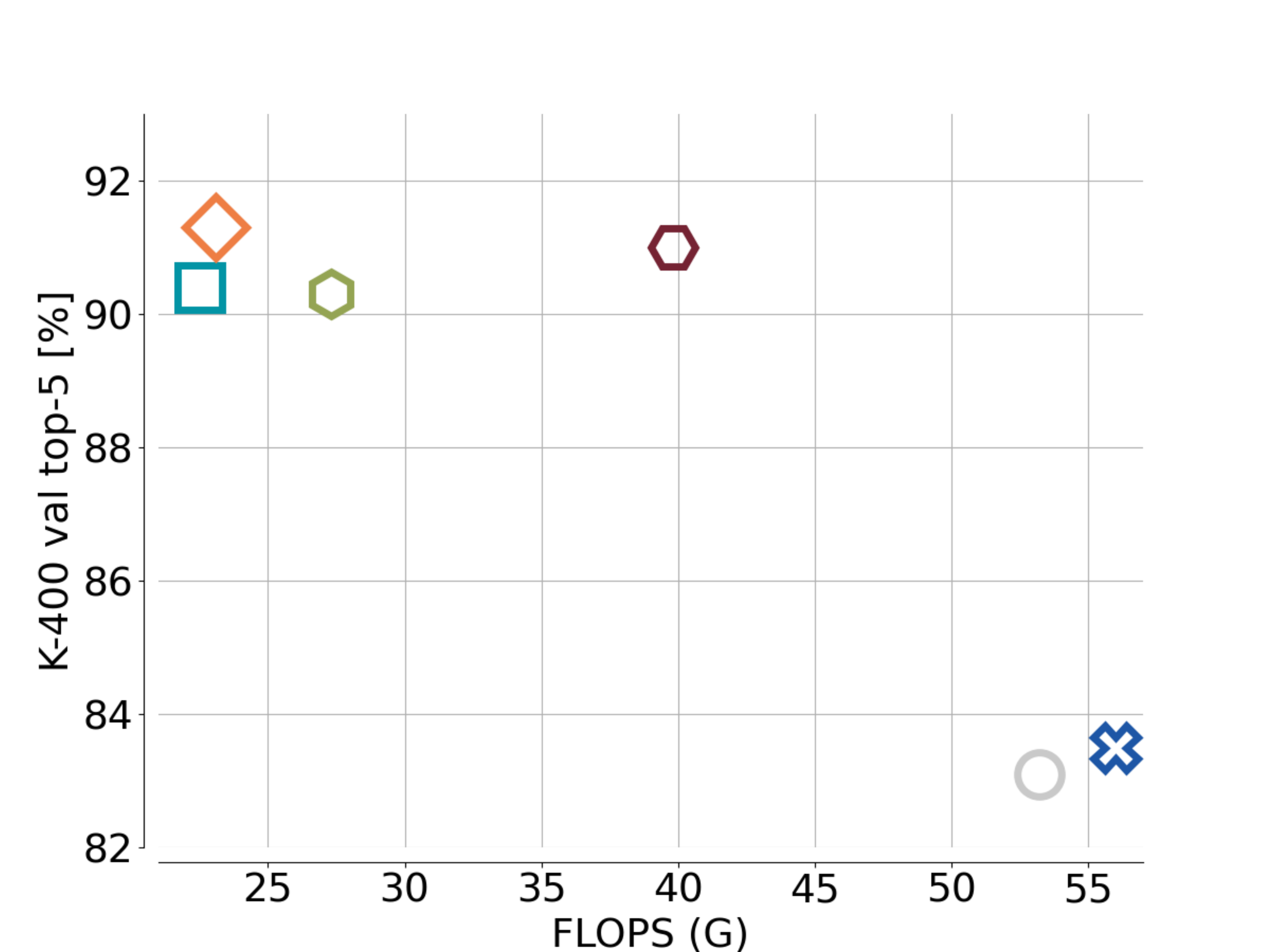}
         \caption{\textbf{Top-5 accuracy with respect to the number of GFLOPs}}
         \label{figure:k400_acc_ab_flops_par_top5_flops}
     \end{subfigure}\\\vspace{2em}
     \begin{subfigure}[b]{0.48\textwidth}
         \centering
         \includegraphics[width=\textwidth]{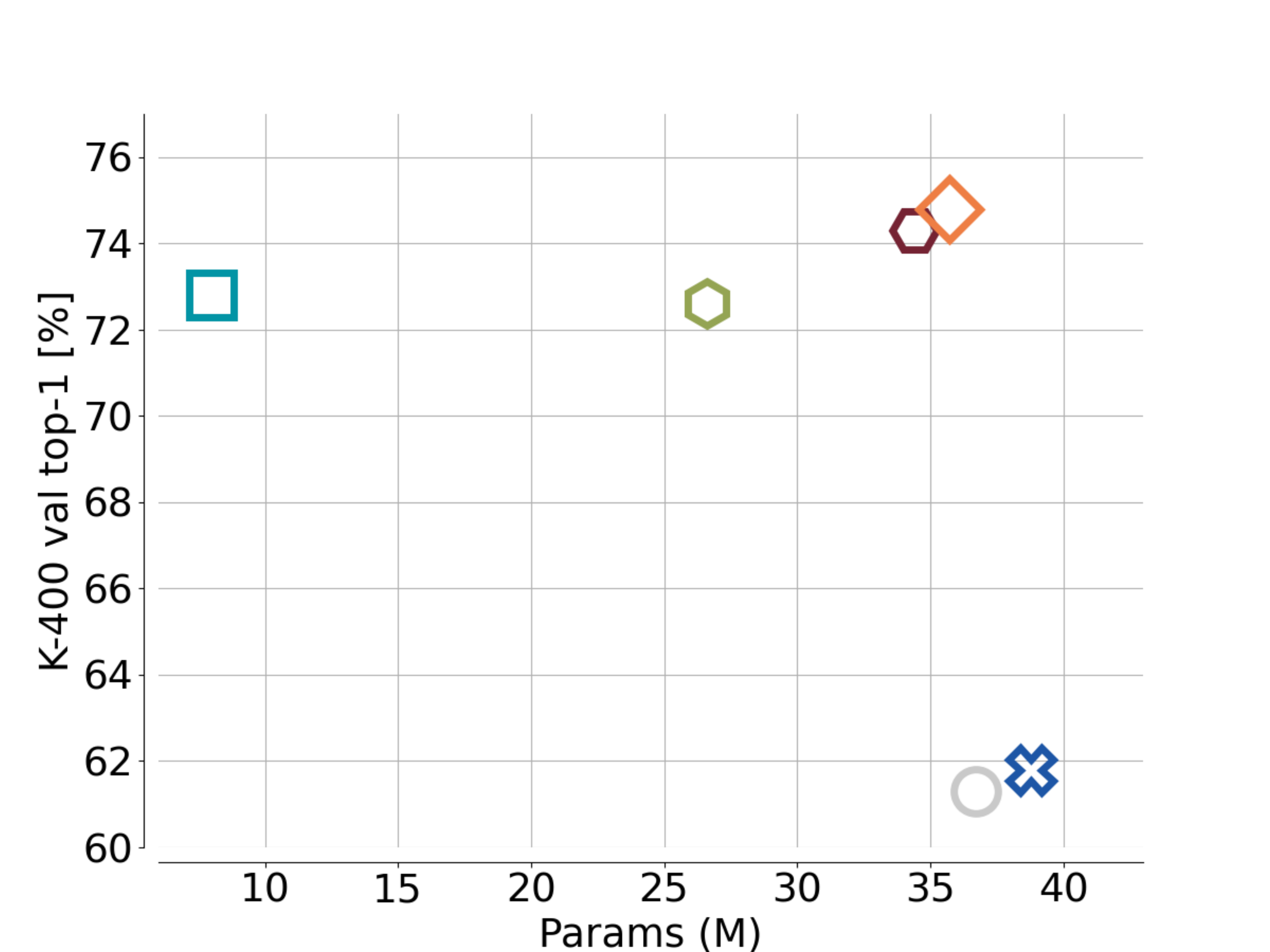}
         \caption{\textbf{Top-1 accuracy with respect to the number of parameters}}
         \label{figure:k400_acc_ab_flops_par_top1_params}
     \end{subfigure}
     \hfill
     \begin{subfigure}[b]{0.48\textwidth}
         \centering
         \includegraphics[width=\textwidth]{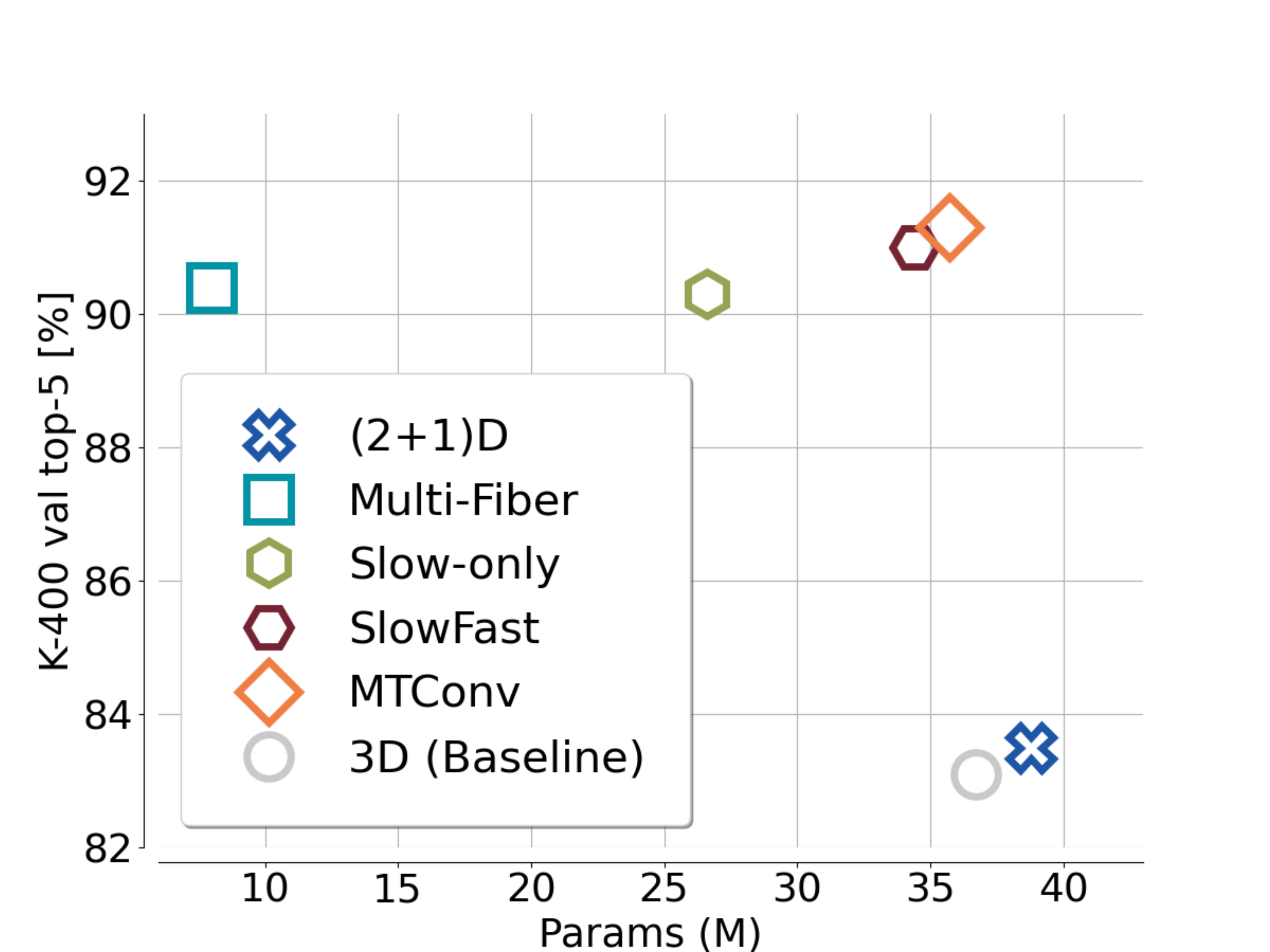}
         \caption{\textbf{Top-5 accuracy with respect to the number of parameters}}
         \label{figure:k400_acc_ab_flops_par_top5_params}
     \end{subfigure}\vspace{1em}
\caption{\textbf{Spatio-temporal data processing methods accuracy/complexity trade-offs} on K-400 based on \cref{tab:K400_accuracies_Res50_ch5}. Comparisons include the R(2+1)D model, MFNet, Slow-only path from SlowFast, SlowFast and the Proposed MTConvs. Baseline architecture is a 50-later ResNet.}
\label{figure:k400_acc_ab_flops_par}
\end{figure}

% Block changes on ResNet 50 on Kinetics
% Accuracies
We perform additional tests to better understand how each spatio-temporal strategy affects the accuracy. We compare all the aforementioned methods with a baseline ResNet-50 architecture in \Cref{tab:K400_accuracies_Res50_ch5}. The use of a common backbone architecture for all methods is done in order to provide a standardised comparison across the different spatio-temporal modelling schemes from the literature. The direct replacement of 3D convolutions with MTConvs demonstrate a significant performance improvement with +13.5\% in top-1 and +8.2\% in top-5 accuracies over 3D Convs. This also includes +(0.5--13)\% top-1 and +(0.3--7.8)\% top-5 accuracy performance increments over the compared spatio-temporal convolution schemes. MTConvs perform similar to SlowFast convolutions which follow the same trend as in \Cref{tab:K400_accuracies_ch5}, while performing better than the alternative Multi-Fibre or the SlowFast supplementary Slow-only approaches. For both results in the top-1 and top-5 accuracies we do not notice significant improvements between 3D and (2+1)D. 

% Computations
Computation-wise, MTConvs are also less complex with notable reductions in terms of the number of multiply-addition operations with a 56\% decrease in operation numbers in comparison to standard 3D convolutions. Although the performance of MTConvs and SlowFast is similar, their margin in number of GFLOPS is significant with MTConvs requiring approximately $1.7$ times less multiply-add operations. Additionally, MTConvs are within the same margin as other efficient methods such as Multi-Fibre which employ group-based convolutions while outperforming them in both top-1 and top-5. As shown in \Cref{figure:k400_acc_ab_flops_par}, the decrease in GFLOPs for MTNets is unrelated to the number of parameters.

\subsection{Results on Moments in Time (MiT)}
\label{ch:5::sec:results::sub::mit}

\begin{table}[t]
\caption{\textbf{Comparison with MiT state-of-the-art}. Models denoted with ($^{\dagger}$) include additional information input sources.}
\centering
\resizebox{.65\linewidth}{!}{%
\renewcommand{\arraystretch}{1.5}
\begin{tabular}{c|c|cc}
\hline
Model &
Arch. size &
top-1 & top-5 \\
\hline

EvaNet \cite{piergiovanni2019evolving} &
\multirow{2}{*}{NAS\cite{zoph2017neural}} &
31.8 &	N/A \\

AssembleNet \cite{ryoo2019assemblenet} & &
34.3 &	\textbf{62.7}\\

\hline

MF-Net \cite{chen2018multifiber} &
\multirow{13}{*}{Fixed} &
27.3 &	48.2 \\

I3D \cite{carreira2017quo} &
 &
29.5 &	56.1 \\

CoST \cite{li2019collaborative} & &
32.4 & 60.0\\

SoundNet \cite{monfort2018moments} $^{\dagger}$ & &
7.6 & 18.0 \\

TSN+Flow \cite{monfort2018moments} $^{\dagger}$ & &
15.7 & 34.7 \\\cline{1-1}\cline{3-4}

SRTG r3d-34 \cite{stergiou2021learn} \textbf{(ours)} & &
28.5 & 52.3 \\

SRTG r3d-50 \cite{stergiou2021learn} \textbf{(ours)} & &
30.7 & 55.6 \\

SRTG r3d-101 \cite{stergiou2021learn} \textbf{(ours)} & &
33.6 & 58.5 \\

SRTG r(2+1)d-34 \cite{stergiou2021learn} \textbf{(ours)} & &
29.0 & 54.2 \\

SRTG r(2+1)d-50 \cite{stergiou2021learn} \textbf{(ours)} & &
31.6 & 56.8 \\

SRTG r(2+1)d-101 \cite{stergiou2021learn} \textbf{(ours)} & &
33.7 & 59.1 \\\cline{1-1}\cline{3-4}

MTNet$_{M}$ \cite{stergiou2020right} \textbf{(ours)} &
 &
34.5 & 58.6 \\

MTNet$_{L}$ \cite{stergiou2020right} \textbf{(ours)} & &
\textbf{35.2} & 59.3 \\

\end{tabular}%
}
\label{tab:mit_accuracies_ch5}
\end{table}

% MiT results
We demonstrate the top-1 and top-5 accuracies achieved by state-of-the-art models on the Moments in Time (MiT) dataset in \Cref{tab:mit_accuracies_ch5}. We denote pre-defined architectures with \enquote{Fixed} and models created by Neural Architecture Search with \enquote{NAS} \cite{zoph2017neural}. NAS-based models have optimised structures in order to fit the data better. Our best performing architecture \textbf{MTNet$_{L}$} outperforms current state-of-the-art models in top-1 accuracy with 35.2\% while having a slightly lower top-5 rate (-3.4\%) than the AssembleNet \cite{ryoo2019assemblenet}. Notably, the comparisons also include models that exploit supplementary information. Specific architectures further utilise optical flow \cite{monfort2018moments} and sound-based \cite{monfort2018moments} inputs. In comparison, all of our models (MTNet and SRTG-enabled r3d and r(2+1)d) are trained solely on stacked RGB frames. Interestingly, the smaller \textbf{MTNet$_{M}$} performs similar to learned architectures such as AssembleNet \cite{piergiovanni2019evolving} while having a pre-defined architecture.

\subsection{Results on Kinetics-700 (K-700)}
\label{ch:5::sec:results::sub::kinetics-700}

\begin{table}[t]
\caption{\textbf{Comparison with K-700 state-of-the-art}. GFLOP calculation is similar to that in \Cref{tab:K400_accuracies_ch5}. (*) indicates reproduced models.}
\centering
\resizebox{.9\textwidth}{!}{%
\renewcommand{\arraystretch}{1.5}
\begin{tabular}{c|c|c|cc}
\hline
Model & 
Pre-train &
GFLOPs $\! \times \!$ views &
top-1 & 
top-5 \\
\hline

I3D \cite{carreira2017quo} &
K-600 &
$108 \! \times \! N/A$ &
53.0 & 69.2 \\\hline

TSM \cite{lin2019tsm} $^{*}$ &
\multirow{13}{*}{HACS} &
$65.0 \! \times \! 10$ &
54.0 & 72.2  \\

MF-Net \cite{chen2018multifiber} $^{*}$ &
 &
$11.1 \! \times \! 50$ &
54.3 & 73.4  \\

ir-CSN-101 \cite{tran2019video} $^{*}$ &
 &
$63.6 \! \times \! 10$ &
54.7 & 73.8  \\

SF-50 \cite{feichtenhofer2019slowfast} $^{*}$ &
 &
$65.7 \! \times \! 30$ &
56.2 & 75.7  \\

SF-101 \cite{feichtenhofer2019slowfast} $^{*}$ &
 &
$213.0 \! \times \! 30$ &
57.3 & 77.2  \\\cline{1-1}\cline{3-5}

SRTG r3d-34 \cite{stergiou2021learn} \textbf{(ours)} &
&
$26.6 \! \times \! 30$ &
49.1 & 72.7  \\

SRTG r3d-50 \cite{stergiou2021learn} \textbf{(ours)} &
&
$52.7 \! \times \! 30$ &
53.5 & 74.2  \\

SRTG r3d-101 \cite{stergiou2021learn} \textbf{(ours)} &
&
$78.1 \! \times \! 30$ &
56.5 & 76.8  \\

SRTG r(2+1)d-34 \cite{stergiou2021learn} \textbf{(ours)} &
&
$37.8 \! \times \! 30$ &
49.4 & 73.2 \\

SRTG r(2+1)d-50 \cite{stergiou2021learn} \textbf{(ours)} &
&
$83.4 \! \times \! 30$ &
54.2 & 74.6 \\

SRTG r(2+1)d-101 \cite{stergiou2021learn} \textbf{(ours)} &
&
$163.1 \! \times \! 30$ &
56.8 & 77.4 \\\cline{1-1}\cline{3-5}

MTNet$_{M}$ \cite{stergiou2020right} \textbf{(ours)} &
&
\textbf{$8.8 \! \times \! 30$} &
58.4 & 77.6 \\

MTNet$_{L}$ \cite{stergiou2020right} \textbf{(ours)} &
&
$17.6 \! \times \! 30$ &
\textbf{63.3} & \textbf{84.1} \\

\end{tabular}%
}
\label{table:K700_accuracies_ch5}
\end{table}

% Results on K-700
We additionally evaluate our models on the 700-variant of the Kinetics dataset. Our results achieved are presented in \Cref{table:K700_accuracies_ch5}. Our best performing architecture \textbf{MTNet}$_L$ outperforms the tested models from the literature by significant margins. In comparison to the previous top-performing SlowFast-101, \textbf{MTNet}$_L$ shows an accuracy improvement of 6.0\% for top-1 accuracy and 6.9\% for top-5 accuracies. In relation to the similarly complex MF-Net, we show +9.0\% top-1 and +10.7\% top-5 accuracies. In comparison to SRTG networks, \textbf{MTNet}$_{L}$ demonstrates accuracy increases in the range of 1.5--6.7\% for top-1 and 6.5--14.2\% for top-5. We also report the accuracy rates of \textbf{MTNet}$_M$ with performance comparable to that of SlowFast-101 while also demonstrating a +1.1\% improvement in the top-1 accuracy. Overall, \textbf{MTNet}$_M$ is the most lightweight architecture that we have tested while being the second best performing. In relation to SRTG networks, the accuracy margins range between 1.5--9.3\% depending on the model size. Based on this, it is evident that accuracy improvements that can be achieved through MTNet as well as the additional reductions in computational costs.

\subsection{Results on HACS}
\label{ch:5::sec:results::sub::hacs}

\begin{table}[t]
\caption{\textbf{Action recognition model comparisons on HACS}. Weight initialisation sources are denoted by their respective indicators.}
\begin{threeparttable}[t]
\centering
\resizebox{\textwidth}{!}{%
\renewcommand{\arraystretch}{1.2} 
\begin{tabular}{c|c|c|c|c|c|l}
\hline
\multirow{2}{*}{Model} & 
\multirow{2}{*}{Pre} &
\multicolumn{2}{c|}{Accuracy} & 
GFLOPs &
Params &
Latency\\

&
&
top-1 & top-5 & 
$\! \times \!$ views &
(M) &
($\downarrow$F / $\uparrow$B) \\
\hline

MF-Net \cite{chen2018multifiber}$^{\dagger}$ &
\multirow{4}{*}{K-400} &
78.3 & 94.6 &
$11.1 \! \times \! 50$ &
8.0 &
32.8/236.0 \footnotemark[3] \\

TAM \cite{liu2021tam}$^{\dagger}$ &
 &
82.2 & 95.2 &
$86 \! \times \! 12$ &
25.6 &
42.1/165.3\\

SF-101 \cite{feichtenhofer2019slowfast}$^{\dagger}$ &
 &
83.7 & 96.8 &
$65.7 \! \times \! 30$ &
53.7 &
39.3/125.1\\

X3D-L \cite{feichtenhofer2020x3d}$^{\dagger}$ &
&
85.8 & 96.1 &
$24.8 \! \times \! 30$ &
\textbf{6.1} &
73.6/ 457.4 \footnotemark[3] \\
\hline

r3d-34 \cite{kataoka2020would}$^{*}$ &
\multirow{6}{*}{K-700} &
74.8 & 92.8 &
$26.6 \! \times \! 30$ &
63.7 &
32.7/74.1\\

r3d-50 \cite{kataoka2020would}$^{*}$ &
 &
78.4 & 93.8 &
$52.6 \! \times \! 30$ &
36.7 &
28.2/87.7\\

r3d-101 \cite{kataoka2020would}$^{*}$ &
 &
80.5 & 95.2 &
$78.0 \! \times \! 30$ &
69.1 &
41.6/110.2\\

r(2+1)d-34 \cite{kataoka2020would}$^{*}$ &
 &
75.7 & 93.8 &
$37.8 \! \times \! 30$ &
61.8 &
40.8/152.0\\

r(2+1)d-50 \cite{kataoka2020would}$^{*}$ &
 &
81.3 & 94.5 &
$83.3 \! \times \! 30$ &
34.8 &
33.2/128.7\\

r(2+1)d-101 \cite{kataoka2020would}$^{*}$ &
 &
82.9 & 95.7 &
$163.0 \! \times \! 30$ &
67.2 &
49.9/163.6\\
\hline

ir-CSN-101 \cite{tran2019video}$^{\dagger}$ &
\multirow{2}{*}{IG65} &
83.8 & 93.8 &
$63.6 \! \times \! 10$ &
22.1 &
51.4/461.2 \footnotemark[3] \\

ip-CSN-101 \cite{tran2019video}$^{\dagger}$ &
 &
84.1 & 93.9 &
$63.6 \! \times \! 10$ &
24.5 &
64.3/512.6 \footnotemark[3] \\
\hline

SRTG r3d-34 \cite{stergiou2021learn} \textbf{(ours)} &
- &
78.6 & 93.6 &
$26.6 \! \times \! 30$ &
83.8 &
35.2/80.6\\

SRTG r3d-50 \cite{stergiou2021learn} \textbf{(ours)} &
- &
80.3 & 95.5 &
$52.7 \! \times \! 30$ &
56.9 &
31.8/96.9 \\

SRTG r3d-101 \cite{stergiou2021learn} \textbf{(ours)} &
- &
81.6 & 96.3 &
$78.1 \! \times \! 30$ &
107.1 &
49.2/131.6 \\

SRTG r(2+1)d-34 \cite{stergiou2021learn} \textbf{(ours)} &
- &
80.4 & 94.3 &
$37.8 \! \times \! 30$ &
82.1 &
46.3/157.0 \\

SRTG r(2+1)d-50 \cite{stergiou2021learn} \textbf{(ours)} &
- &
83.8 & 96.6 &
$83.4 \! \times \! 30$ &
55.0 &
37.6/141.5 \\

SRTG r(2+1)d-101 \cite{stergiou2021learn} \textbf{(ours)} &
- &
84.3 & 96.8 &
$163.1 \! \times \! 30$ &
105.3 &
58.9/172.2 \\
\hline

MTNet$_{S}$ \cite{stergiou2020right} \textbf{(ours)} &
- &
80.7 & 95.2 &
\textbf{$5.8 \! \times \! 30$}  &
25.8 &
50.8/199.6 \footnotemark[3] \\

MTNet$_{M}$ \cite{stergiou2020right} \textbf{(ours)} &
- &
83.4 & 95.9 &
$8.8 \! \times \! 30$  &
25.8 &
62.8/216.3 \footnotemark[3] \\

MTNet$_{L}$ \cite{stergiou2020right} \textbf{(ours)} &
- &
\textbf{86.6} & \textbf{96.7} &
$17.6 \! \times \! 30$  &
50.1 &
98.3/513.7 \footnotemark[3] \\

\end{tabular}%
}
 \begin{tablenotes}
    \item[$\dagger$] models and weights from official repositories.  
    \item[$*$] re-implemented models and weights. 
   \end{tablenotes}
\end{threeparttable}%
\label{tab:HACS_accuracies_ch5}
\end{table}

% Settings
We provide results over the Human Action Clips Segments (Clips) dataset (HACS). We present the performance achieved for different models in \cref{tab:HACS_accuracies_ch5}. Due to the lack of results reporting on HACS because of its recency, all results are re-implemented on our own machine and thus also provide a standard benchmark. Datasets that are utilised for weight initialisation are denoted by the \textit{Pre} column with the sources that models have been imported from described in the sub-caption. Latency times in milliseconds (msecs) are computed as the inference of a single batch during the forward and backward pass ($inf_{f}, inf_{b}$) divided by the number ($b$) of clips in the batch ($\downarrow \! F \; (\frac{inf_{F}}{b}), \uparrow \! B \; (\frac{inf_{B}}{b})$). The inference calculations are all performed with batch sizes of 32. For the MFNet \cite{chen2018multifiber}, TAM \cite{liu2021tam}, SF-101 \cite{feichtenhofer2019slowfast} and X3D-L \cite{feichtenhofer2020x3d} models, their weights are initialised based on the K-400 dataset. R3D and R(2+1)D models with their 34, 50 and 101 variants are pre-trained on K-700. Lastly, we include in our comparisons \textit{Channel-Separated Spatio-Temporal Convolutions} (CSN) that are pre-trained on the IG65M dataset that is not publicly available. The two tested CSN networks variants are an \textit{interaction-preserved} channel-separated network (ip-CSN) and \textit{interaction-reduced} Channel-Separated Network (ir-CSN). The ip-CSN network replaces $3 \! \times \! 3 \! \times \! 3$ convolutions with separable 3D convolutions, while ir-CSN replaces them with depth-wise 3D convolutions.

\footnotetext{Marked models are based on the use of Group/Channel-separated 3D Convolutions. X3D-L and MTNet models use the patched version of Channel-separated 3D convolutions in PyTorch \cite{paszke2019pytorch} 1.9.x while the other implementations are based on the channel-wise $O(n \! \times \! l)$ complex implementation. At the time of writing,  conversions are not possible due to weights being only available in specific versions.}

% Results MTNets
Our \textbf{MTNet$_{S}$} performs similarly to both r3d-101 \cite{kataoka2020would} and r(2+1)d-50 \cite{kataoka2020would}, while not being pre-trained on another dataset nor including their large number of parameters. Additionally, \textbf{MTNet$_{S}$} shows to perform better than MF-Net which requires roughly $1.9$ times more GFLOPs. The latency times of \textbf{MTNet}$_S$ are similar to those of SRTG r3d-101 and SRTG r(2+1)d-101. This is owing to the use of channel-separated convolutions which require additional $O(n \! \times \! l)$ time compared to 3D convolutions, where $n$ is the number of groups of channels and $l$ is the number of channel-separated convolution layers. The larger \textbf{MTNet$_{M}$} shows an overall increase in performance compared to \textbf{MTNet}$_S$ with +2.7\% and +0.7\% top-1 and top-5 accuracies. The achieved accuracies are similar to those of the SlowFast-101 and ir-CSN-101 models with only a fraction of the parameters and compute and without pre-training. Latency times follow an increasing trend compared to \textbf{MTNet}$_S$. Compared to X3D-L, \textbf{MTNet$_{L}$} shows a +0.8\% for the top-1 and +0.6\% top-5 accuracy improvements while having $\sim\!$ 29\% fewer GFLOPs. In our tests, \textbf{MTNet$_{L}$} achieved the best accuracy rates out of the tested models significantly outperforming SF-101 and ir/ip-CSN-101 without any previous kernel initialisation. We also note that partial latency overheads are also due to our proposed frame downsampling method based on cosine similarity triplets. Although the number of computations is not significantly affected, tensor slicing and indexing is computationally slow during parallelisation.

\subsection{Ablation Studies}
\label{ch:5::sec:results::sub::ablation}

% Section focus
The scope of this section is to perform ablation studies on HACS for our proposed models. For MTConvs, we compare results from different ratios ($\delta$) for the local ($\mathcal{L}$) and prolonged ($\mathcal{P}$) branches. We additionally evaluate the effects of different recurrent cell types on the global aggregated feature importance branch ($\mathcal{G}$) of MTBlocks. Finally, we demonstrate the resulting accuracies by employing different spatio-temporal pooling methods applied to inputs of $\mathcal{P}$.

% MTNet channel ratio
\begin{table}[ht]
\caption{\textbf{Branch channel ratio}: Varying channel ratio ($\delta$) across MTNet$_{M}$ and MTNet$_{L}$ architectures.}
\centering
\resizebox{.77\linewidth}{!}{%
\begin{tabular}{p{0.5cm}| l |cc|c|c}
\hline
Net. &
$\delta$ setting &
top-1 &
top-5 &
FLOPs (G) &
Params (M)\\ [0.15ex]
\hline
\multirow{7}{*}{\rotatebox[origin=c]{90}{MTNet$_{M}$}} &  $\delta=1.0$ (No $\mathcal{P}$) & 82.2 & 93.6 & 10.8 & 29.7\\
[0.5ex] 
 & $\qquad 7/8$ & \textbf{83.4} & \textbf{95.9} & 8.8 & 25.8\\ 
[0.5ex]
 & $\qquad 3/4$ & 83.1 & 95.6 & 6.7 & 21.8\\ 
[0.5ex]
 & $\qquad 5/8$ & 81.6 & 93.2 & 4.8 & 19.3\\ 
[0.5ex]
 & $\qquad 1/2$ & 79.7 & 91.8 & 3.6 & \textbf{18.6}\\ 
[0.5ex]
 & $\qquad 3/8$ & 78.6 & 89.4 & 2.6 & 19.2\\ 
[0.5ex]
 & $\qquad 1/4$ & 77.1 & 88.6 & \textbf{2.1} & 21.0\\ 
[0.5ex]
\hline
\multirow{6}{*}{\rotatebox[origin=c]{90}{MTNet$_{L}$}} &  $\delta=1.0$ (No $\mathcal{P}$) & 84.9 & 95.7 & 20.6 & 53.5\\
[0.5ex] 
 & $\qquad 7/8$ & \textbf{86.6} & \textbf{96.7} & 17.6 & 50.1\\ 
[0.5ex]
 & $\qquad 3/4$ & 86.1 & 96.2 & 12.5 & 45.3\\ 
[0.5ex]
 & $\qquad 1/2$ & 83.2 & 95.3  & 7.09 & \textbf{42.7}\\ 
[0.5ex]
 & $\qquad 3/8$ & 82.1 & 93.9 & 5.2 & 45.3\\ 
[0.5ex]
 & $\qquad 1/4$ & 80.3 & 92.4 & \textbf{4.1} & 47.8\\ 
[0.5ex]
\end{tabular}%
}
\label{tab:HACS_MTNet_delta}
\end{table}

% experiments with ratio
\textbf{Branch channel ratio}. As one primary feature of MTConvs is the bifurcation of channels assigned to local ($\mathcal{L}$) and prolonged ($\mathcal{P}$) branches, we further explore how different ratio values ($\delta$) affect performance accuracy-wise, and in terms of computations and parameters. Evident from \cref{tab:HACS_MTNet_delta}, the best ratios ($\delta$) in terms of accuracies, are in range of 0.875 $\! \sim \!$ 0.75 with changes in performance being marginal ($\pm 0.3 \! \sim \! 0.5\%$) for both the top-1 and top-5 accuracies. Compared to the use of standard 3D Convs, these ratios lead to further reductions in both multiply-add operations as well as the number of parameters. The improvements in the number of GFLOPs given these ratios are further demonstrated by a significant reduction of 25 $\! \sim \!$ 37 \% with the use of both $\mathcal{L}$ and $\mathcal{P}$ branches in comparison to using solely the local branch ($\mathcal{L}$). The sole use $\mathcal{L}$ branch is equivalent to a single standard 3D Conv. Computation-wise, the best ratio is $\delta=1/2$ as it demonstrates the largest balanced combined reduction in terms of GFLOPs (-66\%) and number of parameters (-37\%). The attributing factor for the loss in performance when using small ratios is the strong dependency of branch $\mathcal{P}$ to branch $\mathcal{L}$. Interestingly, the decrease in local feature dimensionality corresponds to the inability of the prolonged features to encapsulate substantial video action details by themselves. We note that $\delta$ decreases do not relate directly to reductions in parameters as seen in \Cref{tab:HACS_MTNet_delta}, since further reductions with $\delta < 1/2$ show an increasing trend, with branch $\mathcal{P}$ employing a larger number of parameters. Based on this, the smallest number of parameters is observed when the ratio is split equally (balanced) between the two channels ($\delta=1/2$). Lastly, we note that zero-valued ratios $\delta=0$ are not architecturally feasible as branch $\mathcal{P}$ includes the outputs of branch $\mathcal{L}$.

\begin{table}[ht]
\caption{\textbf{Recurrent cell configurations:} Alternative recurrent cells for the \textit{global aggregated feature importance branch} ($\mathcal{G}$). Branch ratio of $\delta=7/8$.}
\centering
\resizebox{.8\linewidth}{!}{%
\begin{tabular}{c|c|cc|cc|cc}
\hline
\multirow{2}{*}{Net} &
\multirow{2}{*}{Cell type} & 
Params &
FLOPS &
\multicolumn{2}{c}{Latency (msec)} &
\multirow{2}{*}{top-1} & 
\multirow{2}{*}{top-5}\\
&
& 
(M) &
(G) &
$\downarrow$F & $\uparrow$B &
&
\\[0.3em]
\hline
\parbox[t]{2mm}{\multirow{5}{*}{\rotatebox[origin=c]{90}{MTNet$_S$}}} &
RNN \cite{rumelhart1985learning} &
24.3 &
5.8 &
58 & 78 &
78.8 &
93.7\\ [0.25em]
&
LSTM \cite{hochreiter1997long} &
26.5 &
5.8 &
61 & 79 &
79.9 &
94.3\\[0.25em]
&
LSTM (peephole) \cite{gers2000recurrent} &
26.5 &
5.8 &
68 & 85 &
80.1 &
94.5\\[0.25em]
&
GRU \cite{cho2014learning} &
25.8 &
5.8 &
65 & 80 &
\textbf{80.7} &
\textbf{95.2}\\[0.25em]
\hline
\parbox[t]{2mm}{\multirow{5}{*}{\rotatebox[origin=c]{90}{MTNet$_M$}}} &
RNN \cite{rumelhart1985learning} &
24.3 &
8.8 &
84 & 113 &
82.5 &
94.8\\ [0.25em]
&
LSTM \cite{hochreiter1997long} &
26.5 &
8.8 &
86 & 109 &
83.1 &
95.4\\[0.25em]
&
LSTM (peephole) \cite{gers2000recurrent} &
26.5 &
8.8 &
94 & 120 &
83.2 &
95.6\\[0.25em]
&
GRU \cite{cho2014learning} &
25.8 &
8.8 &
90 & 111 &
\textbf{83.4} & 
\textbf{95.9} \\[0.25em]
\end{tabular}
}
\label{tab:HACS_MTNet_recurrent_cells}
\end{table}

% recurrent cells
\textbf{Recurrent cell configuration}. Evaluations are done in terms of the accuracy effects based on changes of the recurrent methodology. \Cref{tab:HACS_MTNet_recurrent_cells} demonstrates recurrent layer method changes for \textbf{MTNet$_{S}$} and \textbf{MTNet$_{M}$}. These changes only affect the global aggregated feature branch ($\mathcal{G}$). Latency times are reported based on the time (in msecs) required for a full forward ($\downarrow$F) and backward ($\uparrow$B) pass of a single $16 \! \times \! 256 \! \times \! 256$ clip to the entire network. We distinguish this from the reported latencies in \Cref{tab:HACS_accuracies_ch5} which are the per clip averages over entire batches of 32 clips. The use of GRUs \cite{cho2014learning} has been motivated by the (small) improvements compared to alternative recurrent cell structures of LSTMs and RNNs. Considering \textbf{MTNet$_{S}$}, GRUs seem to perform slightly better than regular RNN cells \cite{rumelhart1985learning} with +1.9\% top-1 and +1.5\% top-5 accuracies. However, the overall simplicity of RNNs proves to be more efficient in terms of parameters, with a -8\% overall network parameter reduction as well as marginally faster forward and backward latency times. This observation is also present for \textbf{MTNet$_{M}$} as GRU's top-1 and top-5 accuracies improve the RNN baseline by +0.9\% and +1.1\% respectively. In comparison to LSTMs \cite{hochreiter1997long} and LSTMs with peepholes variants \cite{gers2000recurrent}, GRUs also show marginally improved accuracy rates. However, GRUs merge LSTM's \textit{forget} and \textit{input} states as well as their \textit{cell} and \textit{hidden} states, thus simplifying the overall recurrent cell structure by requiring a smaller number of operations and weights. This is demonstrated by the reduction in number of parameters of GRUs in comparison to both LSTM variants.

\begin{table}[t]
\caption{\textbf{Spatio-temporal pooling methodology:} Top-1 accuracy for different pooling methods used on inputs for the \textit{prolonged branch} ($\mathcal{P}$). Ratio is set to $\delta=7/8$.}
\centering
\resizebox{\linewidth}{!}{%
\begin{tabular}{c|cccccc}
\hline
\multirow{2}{*}{Net} &
\multicolumn{6}{c}{Pooling} \\[0.2em]\cline{2-7}
&
Avg &
Max &
Stochastic \cite{zeiler2013stochastic} &
SoftPool \cite{stergiou2021refining} &
Avg + $cos$ &
SoftPool \cite{stergiou2021refining} + $cos$ \\
\hline
MTNet$_{S}$ &
77.8 &
75.9 &
76.8 &
77.8 &
80.5 &
\textbf{80.7} \\[0.25em]
MTNet$_{M}$ &
79.8 &
77.6 &
78.2 &
80.7 &
82.6 &
\textbf{83.4} \\[0.25em]
MTNet$_{L}$ &
83.8 &
82.1 &
82.9 &
84.2 &
85.9 &
\textbf{86.6} \\[0.25em]
\end{tabular}%
}
\label{tab:HACS_MTNet_poling}
\end{table}

% Pooling methods
\textbf{Spatio-temporal pooling methodology}. We include in our ablation studies accuracy changes attributed to different pooling methods for $\mathcal{P}$ branch's inputs. The methods tested can be both temporally and spatially symmetric (with the downsampling operations being performed similarly across all three dimensions) and asymmetric (with spatial pooling done separately from temporal). Our findings are reported in \Cref{tab:HACS_MTNet_poling}, in which we demonstrate the top-1 accuracies for different pooling configurations. Our frame selection method with the proposed temporal triplet cosine ($cos$) similarity produces improvements over all of the tested symmetric methods. Average pooling with triplet $cos$, shows accuracy rate improvements of +2.7\% for \textbf{$MTNet_{S}$}, +2.8\% on \textbf{$MTNet_{M}$} and +2.1\% for \textbf{$MTNet_{L}$}. Following the same trend, SoftPool \cite{stergiou2021refining} in combination with a triplet $cos$ frame selection increases top-1 accuracy by +2.9\%, +2.7\% and +2.4\% for each of the aforementioned models, respectively, in comparison to symmetric baseline SoftPool. Overall, the gap in accuracy rates between both asymmetric methods utilising average pooling or SoftPool is only considered marginal with $\pm 0.57\%$. The improvement is observed with the use of SoftPool instead of average pooling. With this considered, changes in performance are primarily associated with the use of triplet $cos$ in comparison to temporally-extended spatial methods. Therefore, we base our choice of temporal pooling on the significant gains in performance through the use of triplet $cos$ similarity pooling in comparison to spatio-temporally symmetric pooling methods, that apply pooling operations similarly across all dimensions. 

% Table: UCF accuracies
\begin{table}[t]
\caption{\textbf{Transfer Learning on UCF-101:} Top-1 and top-5 accuracies after pre-training.}
\centering
\resizebox{.6\textwidth}{!}{%
\begin{tabular}{c|c|cc}
\hline
Model & 
Pre-training &
top-1 & top-5 \\[0.3em]
\hline
I3D & 
K-400 &
92.4 & 97.6 \\[0.3em]

TSM &
K-400 &
92.3 & 97.9 \\[0.3em]

ir-CSN-152 &
IG65M &
95.4 & 99.2 \\[0.3em]

MF-Net &
K-400 &
93.8 & 98.4 \\[0.3em]

SF r3d-50 &
ImageNet &
94.6 & 98.7 \\[0.3em]

SF r3d-101 &
ImageNet &
95.8 & 99.1 \\[0.3em]
\hline

SRTG-101 (3D) &
HACS+K-700 &
97.3 & 99.6 \\[0.3em]

SRTG-101 (2+1)D &
HACS+K-700 &
97.2 & 99.1 \\[0.3em]

MTNet$_{S}$ \textbf{(ours)} &
HACS &
94.2 & 98.0 \\[0.3em]

MTNet$_{M}$ \textbf{(ours)} &
HACS &
95.4 & 98.1 \\[0.3em]

MTNet$_{L}$ \textbf{(ours)} &
HACS &
\textbf{97.4} & \textbf{99.2} \\
\end{tabular}%
}
\label{tab:accuracies_ucf_ch5}
\end{table}

\subsection{Feature Transferability evaluation}
\label{ch:5::sec:results::sub::TL}

% Transfer learning set-up
We compare transfer learning capabilities of MTNets with state-of-the-art video models on UCF-101 for fine-tuning. In order to compare with other methods, all tested models use the same weight initialisation as in \Cref{tab:HACS_accuracies_ch5}. We note that \textbf{$MTNet_{L}$} pre-trained on HACS, achieves similar performance to that of \textbf{SRTG r(2+1)d-101} and \textbf{SRTG r3d-101} which have been pre-trained on both HACS and K-700 datasets. Both of the models perform better than the rest of the tested architectures within our settings. \textbf{$MTNet_{M}$} demonstrates accuracy rates close to those of ir-CSN which uses the IG65M dataset sourced from Instagram \cite{ghadiyaram2019large} as well as the 101 variant of SlowFast. The smallest of the MTNet architectures, \textbf{$MTNet_{S}$}, shows accuracies above those of TSM and I3D while the overall performance is comparable to that of SlowFast-50. With this, we further demonstrate the generalisation capabilities of our varying spatio-temporal feature extraction approach as well as the dynamic temporal feature calibration module.

\section{Discussion and conclusions}
\label{ch:5::sec:conclusions}

% Scope overview
We have presented in this section the challenges that are faced when using local patterns as descriptors of actions of variant complexity and duration. Action identities of local video segment may not directly relate to the underlined action. Our aim is the extension of the temporal receptive fields of convolutions by encoding features across different temporal modalities.

% MTConvs
Our proposed Multi-Temporal Convolutions (MTConvs) are based on the extraction of features across variant spatio-temporal windows. Our multi-temporal blocks are built on three branches. The local branch extracts motion characteristics within a short temporal location, the prolonged branch is used for motion features of extended durations that include relation to the local branch. The global aggregated feature importance branch calibrates branch information based on the feature dynamics.

% Evaluation
We have evaluated our work on four large scale datasets Kinetics-400, Kinetics-700, Moments in Time and HACS, demonstrating competitive results to state-of-the-art architectures and in most cases outperforming them. This also comes with a significant reduction in the computational complexity signifying the overall efficiency of our method. Our ablation studies further validate our claims and motivate our design choices. Based on the obtained results, we believe that modelling variable-duration spatio-temporal patterns is a viable research direction to inspire future works in the field of video action recognition.

%% Format bibliography like a section, not a chapter:
%\printbibliography[heading = subbibliography]
\stopcontents[chapters]
    
    \author
{}
\title{Class-Specific Regularisation Across Time}

\maketitle
\label{ch6}

\startcontents[chapters]
%\printcontents[chapters]{}{1}{\section*{\contentsname}}

% Scope of the chapter
In this chapter we consider the inclusion of class-related information during the feature extraction process. The proposed method fuses class-based information to the extracted features by amplifying activations that better correspond to the specific class. We evaluate how the proposed feature regularisation method, through activation amplification, can improve classification performance based on classes in which their overall features demonstrate relative similarities or overlap.

\section{Introduction}
\label{ch:6::sec:intro}

% Summation of previous chapter
In the previous chapters, we explored the modelling difficulties of motion variations and proposed an approach to directly address these through the inclusion of temporally local patterns and prolonged temporal features. The resulting multi-temporal convolutions were based on the use of a triple-branch approach and the subsequent fusion of their produced features. We additionally proposed an attention-based, feature alignment method to integrate the relationships between short temporal motion characteristics with respect to their importance in the context of the entire video sequence. Through this, temporally important features are highlighted by recurrent cells, enabling the creation of coherent activations based on the discovered averaged feature attention. Even though the proposed feature extraction process has shown significant gains over fixed-sized spatio-temporal convolution alternatives, we believe that there is still room for improvement. Specifically, we consider the uniform nature of the feature extraction process even for classes that exhibit a high degree of similarity in terms of their temporal features and movements. Therefore, in this chapter we examine the regularisation of feature activations through features of the corresponding class that they represent.

% Feature inclusion for specific predictions
The main architectural principle of CNNs is that they include a large stack of multiple subsequent layers within a single hierarchical architecture. Based on this, the feature extraction process takes the form of applying successive convolutions alongside non-linearities over an input. This adds an additional layer of complexity through every successive operation. Kernels at early layers of the architecture are capable of extracting simple textures and patterns, while deeper layers target features with higher degrees of complexity. Through this, the feature dependency to the preceding layer's weighted neural connections becomes more apparent in later layers. Only a portion of these features and their inherited connections is specific to an individual class \cite{bau2019visualizing,gilpin2018explaining}. As all of the kernels in CNNs are learned in a class-agnostic way, the main features that are descriptive of each class, and discriminative between instances, are predominantly computed at the very last layer. This not only limits the capabilities of the network, by optimising towards features corresponding to multiple classes, but also their interpretability. This is because there is no direct association between features and a specific class. This link additionally cannot be easily discovered.

% Batch class regularisation
Our aim in this chapter is to study the effects of using class-specific feature activations to propagate information through the network. The proposed method termed \textit{Class Regularisation} (CR) \cite{stergiou2020learning} utilises class information within the feature extraction process in convolution blocks. This class-relative information is added back to the network through the amplification and suppression of activations, with respect to the predicted classes in the batch. An additional benefit of regularising activations based on corresponding class features, is that the effects of activations are further modulated, which is a significant benefit on the application of non-linearities. Through this class-relative information fusion, the CNN can differentiate between features in terms of their relevance to a specific class. This leads to a reduction of uncorrelated features that include noise in the final fully-connected layers that are responsible for the class predictions.

% Chapter structure
The chapter is structured as follows. We first overview approaches that are based on normalisation and regularisation of convolutional features in \Cref{ch:6::sec:related}. We then provide a formal description of our method in \Cref{ch:6::sec:regularisation}. Our experiments on the task of action recognition are shown in \Cref{ch:6::sec:experiments}. We conclude and discuss possible future research pathways in \Cref{ch:6::sec:discussion}.

\section{Normalisation and regularisation of features}
\label{ch:6::sec:related}

The inclusion of regularised information in CNNs is challenging given the increased level of ambiguity in terms of the model's inner workings with respect to the network depth. The two main methods used to change the distributions of activation maps are based on either activation normalisation or regularisation. 

% Normalisation methods
\textbf{Normalisation}. One of the first normalisation-based methods has been \textit{Batch Normalisation} (BN) \cite{ioffe2015batch}. BN aims at addressing activation value distribution changes during updates, referred to as \textit{Internal Convergence Shift}, by using the mean and standard deviation for each of the input activations. However, normalisation is not specifically bound to feature activations. \textit{Weight Normalisation} (WN) \cite{salimans2016weight} introduces a parameterisation of a layer's weight vector which utilises the unit's length, effectively decoupling the parameters from their directions within the weight space. Santurkar \etal~\cite{santurkar2018does} have argued that instead of reducing \textit{Internal Convergence Shift}, BN instead produces a smoother loss landscape. This new landscape improves the overall stability of models by allowing larger learning rates and faster overall convergence. Van Laarhoven \cite{van2017l2} has further demonstrated how both BN and WN significantly reduce the effects of \textit{weight decay} ($L_{2}$ regularisation). This is due to the fact that BN and WN create scaling-invariant weights. This has further been explored by Ba \etal~\cite{ba2016layer}, who demonstrated that BN and WN back-propagated gradients are scale-invariant. Other normalisation methods also include the use of \textit{Local Response Normalisation} (LRN) \cite{jarrett2009best,krizhevsky2012imagenet,lyu2008nonlinear,sermanet2014overfeat,zeiler2014visualizing} which computes the statistics within neighbourhoods of pixels rather than creating a uniform normalisation criterion for the image in its entirety, as BN does. Works that have refrained from utilising batch-based information include \textit{Layer Normalisation} (LN) \cite{ba2016layer}, which performs mean and variance computations channel-wise for layer input activations and per example. This allows for the process to be executed similarly during both training and testing. Supplementary to this, \textit{Instance Normalisation} (IN) \cite{ulyanov2016instance} performs LN individually for each channel which reduces the cross-channel dependencies. Lastly, \textit{Group Normalisation} (GN) \cite{wu2018group} is a combination of both LN and IN, and performs normalisation over a group of input channels. If the group number is equal to one then GN takes the form of LN, while if the number of groups is equal to the number of input channels, it is similar to IN.

% Regularisation methods
\textbf{Regularisation}. A smaller number of works have also considered BN as a method that can regularise information \cite{gitman2017comparison,szegedy2016rethinking,zhang2017understanding}. However, the two most popular regularisation techniques in the literature have been the inclusion of random noise in the data augmentation process \cite{krogh1992generalization,rifai2011adding} and the use of Dropout \cite{srivastava2014dropout} with the utilisation of a generalised linear model \cite{wager2013dropout}. 

% Spatio-temporal data + our contribution
As described, works in terms of regularisation of activation data have been scarce. In addition, current normalisation and regularisation techniques take the form of general solutions in CNNs, and a specific method for spatio-temporal data does not exist. In addition, to our knowledge no regularisation-based technique has been proposed that can bridge the general nature of convolutional feature extracted and the corresponding descriptive features for classes specifically. Our proposed method named Class Regularisation fills this void as a general solution that can be used in any architecture with minimum computational cost and regularises the feature distribution by including each feature's proportional importance to the target class.

\begin{figure}[ht]
\centering
    \includegraphics[width=0.8\linewidth]{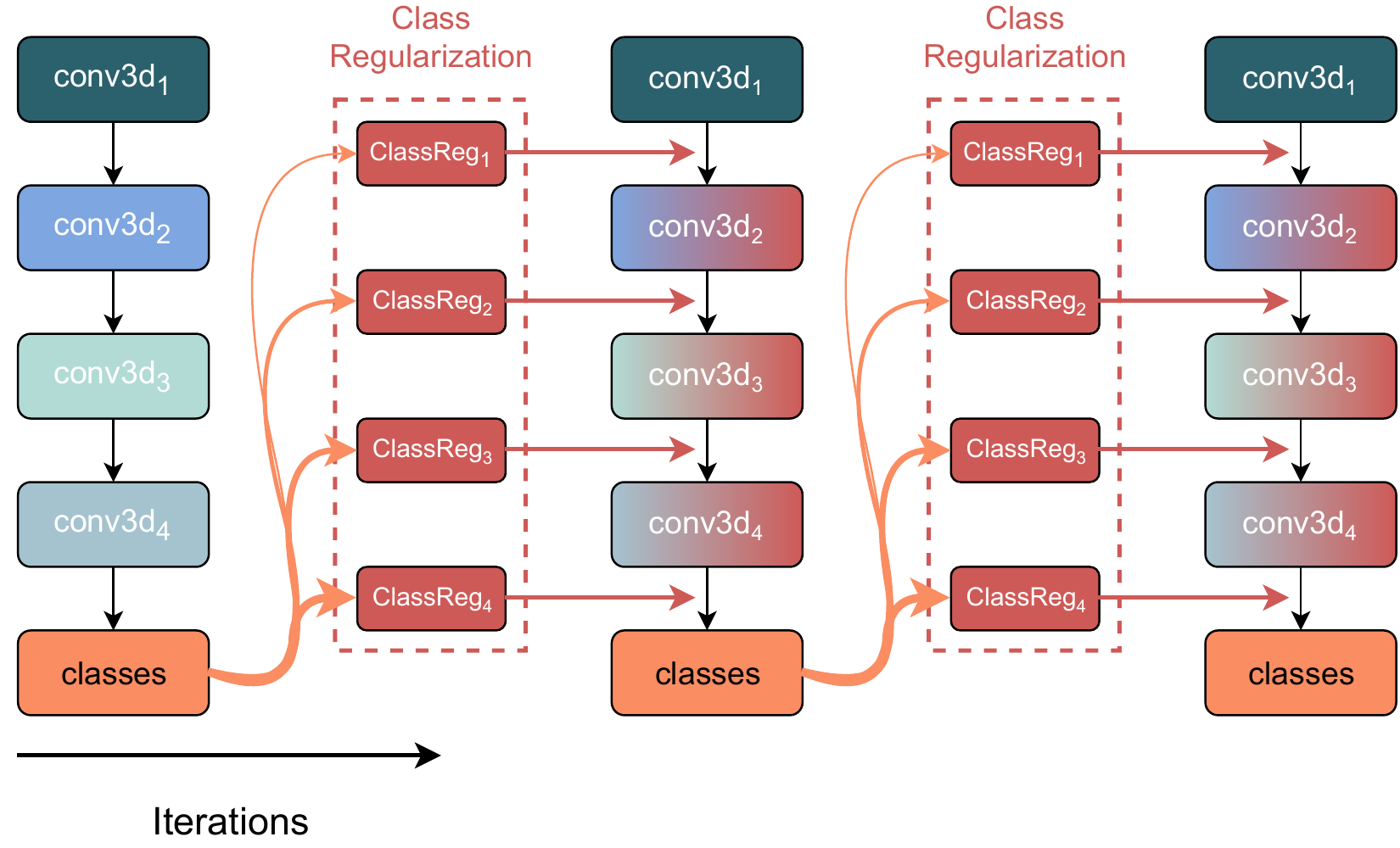}
   \caption{\textbf{Iterative Class Regularisation}. Pathway creation between class weights and intermediate layer features. An overview of the in-block operations is presented in \Cref{fig:classreg_block}.}
\label{fig:classreg_pipeline}
\end{figure}

\section{Regularisation over convolution blocks}
\label{ch:6::sec:regularisation}

In this section we first formulate Class Regularisation in \Cref{ch:6::sec:regularisation::fusion} and explain how class features are used alongside extracted convolution features. We then provide a description of the parameter update process in \Cref{ch:6::sec:regularisation::updates}.

\subsection{Layer and class feature fusion}
\label{ch:6::sec:regularisation::fusion}

% Notation used
Class Regularisation is built on the main notion that different kernels focus on spatio-temporal patterns that appear in different classes. 
We use a standard notation based on which activation maps of layer $l$ are denoted by $\textbf{a}^{[l]}$ and have a size of $C$ channels, $T$ temporal extent and $H$ and $W$ spatial dimensionality. We further denote the number of classes as $N$ with the weight vector ($\textbf{W}$) in the final fully-connected layer for predictions of size $N \! \times \! C'$, where $C'$ is the number of channels.  

\begin{figure}[t]
\centering
    \includegraphics[width=\linewidth]{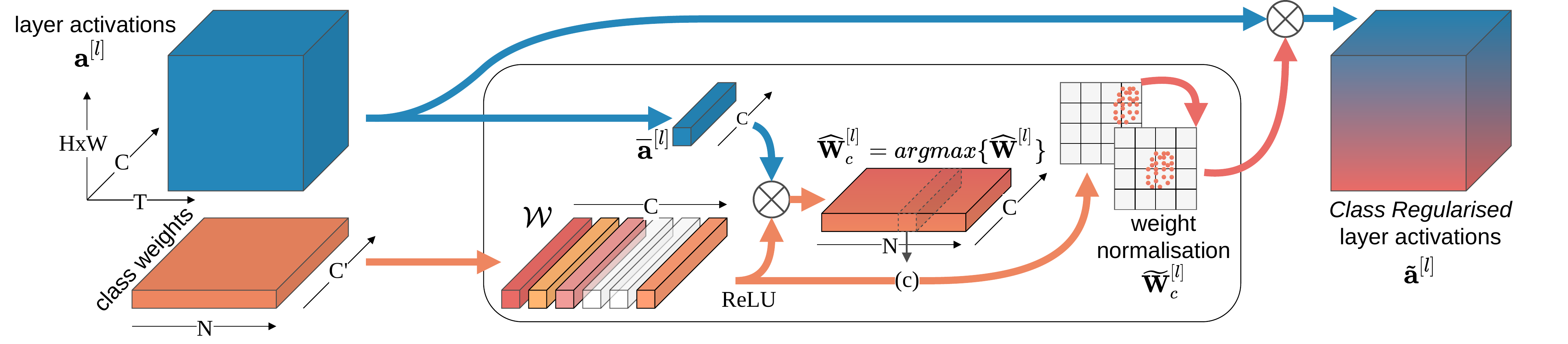}
   \caption{\textbf{Class Regularisation operations}. Class to feature correlations are encoded in weight vector $\widetilde{\textbf{W}}^{[l]}$ with the class ($c$) vector with the maximum correspondence being selected ($\widehat{\textbf{W}}^{[l]}_{c}$) and scaled based on the layer used to $\widetilde{\textbf{W}}^{[l]}_{c}$. The weights are then applied to the input activation maps creating Class Regularised activations ($\widetilde{\textbf{a}}^{[l]}$). ($\otimes$) denotes channel-wise multiplications.}
\label{fig:classreg_block}
\end{figure}

% Class weight convolution
 A visualisation of our approach can be found in \Cref{fig:classreg_block}. When considering the inclusion of class-related features to a layer $[l]$, one of the main problems is the representation of weights $\textbf{W}$. Class weights discover the  feature relevance to classes, in feature space $C'$. However, layer $[l]$ features are based on feature space $C$. This task comes with information loss due to the \textit{curse of dimensionality} when traversing from one feature space to another. In order to create a single differentiable mapping of the class weights over feature space $C'$, we use a single point-wise convolution function that acts as a transformation \cite{jimenez2014convolution}, remapping activations in space $C$ to corresponding values in $C'$. The weight vector used is denoted as $\mathcal{W}$ and is learned over training with the parameters being added to the optimiser. We use a $ReLU$ non-linear function to discover the final weight representations and to perform updates based on the calculated gradients during back-propagation. The new weight is formulated as $\textbf{W}^{C' \rightarrow C}$.

% Class weight and layer activation fusion
Although the size of the created embedding space of $\textbf{W}^{C' \rightarrow C}$ is equal to that of $\textbf{a}^{[l]}$ ($|W^{C' \rightarrow C}|=|a^{[l]}|$), their embeddings do not necessarily conform to the same representations. Therefore, in order to make the new weights $\textbf{W}^{C' \rightarrow C}$ compatible to the representation space of activation maps, we first create a sub-sampled volume of $a^{[l]}$ through pooling over its spatio-temporal extent. The new volume $\overline{\textbf{a}}^{[l]}$ encapsulates a spatio-temporal average feature activation of the original volume and provides a concatenated view of the activations. Based on the averaged feature map, we perform a point-wise multiplication of  $\overline{\textbf{a}}^{[l]}$ over each class ($n \in N$) location in weight vector $\textbf{W}^{C' \rightarrow C}$. The resulting class weights $\widehat{\textbf{W}}^{[l]}$ contain weights based on classes with respect to the current layer activations. This is similar to the application of class weights in the final layer with the exception of the transformation between feature spaces $C' \rightarrow C$.

% Class selection
In order to associate a class with the related features of the layer, we select the vectors with the highest feature activations. Therefore, the selection of the corresponding class that is best described by the averaged global features of vector $\overline{\textbf{a}}^{[l]}$ takes the form of discovering the maximum activation instance within vector $\widehat{\textbf{W}}^{[l]}$. This is again similar to the top-1 selection in the final class prediction layer in the network. The selection is done to discover the index ($c$) of the maximum activation instance $\widehat{\textbf{W}}^{[l]}_{c}$. We refrain from using the volume $\widehat{\textbf{W}}^{[l]}_{c}$ directly, as it represents the maximum activations based on classes rather than the feature weights that correspond to the respective classes. Instead, the index $c$ is used to select the class weights from weight vector $\textbf{W}^{C' \rightarrow C}$.

% Regularisation
Given that each layer $l$ is at a different location inside the network, the descriptive capabilities of their features, in terms of the classes that they best correspond to, are significantly impacted. As the feature complexity increases in the later network layers, this connection of features to classes becomes more apparent. Therefore, as the feature quality with respect to the target classes can vary across the architecture, we introduce an \textit{affection rate} ($\lambda$) term. The main aim of the term is to scale weights, and correspondingly their effect over the activation maps, based on the depth of the regularised layer. Through this, weights are then normalised given their maximum and minimum values and they are shifted accordingly.

\begin{algorithm}[t]
\SetAlgoLined
\KwData{Layer $[l]$ activation map ($\textbf{a}^{[l]}$).\\
        \qquad \: \: Class weights (\textbf{W}) for $N$ classes.}
\KwResult{Class-regularised activations $(\widetilde{\textbf{a}}^{[l]})$.}
\: \\
$\overline{\textbf{a}}^{[l]} \gets pool(\textbf{a}^{[l]})$\\
\For{n in N}{
    $\textbf{W}^{C' \rightarrow C}_{n} \gets max\{0,\mathcal{W} * \textbf{W}_{n}\}$\\
    $\widehat{\textbf{W}}^{[l]}_{n} \gets \textbf{W}^{C' \rightarrow C}_{n} \otimes \overline{\textbf{a}}^{[l]}$
}
$c \gets 0$\\
\For{n in |N|}{
    \If {$\widehat{\textbf{W}}^{[l]}_{n} > \widehat{\textbf{W}}^{[l]}_{c}$}{
        $c \gets n$
        }
}
$\widetilde{\textbf{W}}_{c} \gets \lambda \; \frac{(\textbf{W}^{[C' \rightarrow C]}_{c} - min\{\textbf{W}^{[C' \rightarrow C]}_{c}\}) * (1-\lambda)}{max\{\textbf{W}^{[C' \rightarrow C]}_{c}\} - min\{\textbf{W}^{[C' \rightarrow C]}_{c}\}}$ \label{alg:norm_cr}\\

$\widetilde{\textbf{a}}^{[l]} \gets \textbf{a}^{[l]} \otimes \widetilde{\textbf{W}}_{c}$

\caption{Class Regularisation computational overview}
\label{alg:class_reg}
\end{algorithm}

% Application over input and algorithm
The final normalised class weights ($\widetilde{\textbf{W}}$) are applied over the spatio-temporal layer activations ($\textbf{a}^{[l]}$), through a point-wise multiplication similar to the use of weights. Based on the selected class ($c$) specific features in the activation volume are amplified or reduced. A full overview of the execution sequence of the processes performed by Class Regularisation are shown in \Cref{alg:class_reg}. We use (*) to simplify convolution denomination from Boyce and DiPrima \cite{boyce1986elementary}. We further denote ReLU activations as the maximum of a given operation ($f(x)$) over a given volume (x) as the maximum between the produced function value and zero ($max\{0,f(x)\}$).

% Differences between \widehat{W}^{[l]}_{c} and \widetilde{a}^{[l]}
A possible question that may arise during the formulation is based on the calculation of a layer-feature inclusive weight vector $\widehat{\textbf{W}}^{[l]}$ and the re-calculation of class-based activations with $\widetilde{\textbf{a}}^{[l]} \gets\textbf{a}^{[l]} \otimes \widetilde{\textbf{W}}_{c}$. We make this distinction in our method as $\widehat{\textbf{W}}^{[l]}$ utilises a spatio-temporally pooled vector of feature activations. This disregards the feature localities that are represented in $\textbf{a}^{[l]}$ and thus does not allow for the corresponding class and feature inclusive volume to be directly used. Instead, the normalised weights $\widetilde{\textbf{W}}_{c}$ are applied over the original input and thus can discover class-relative features over regions in the activation map. It is also possible to apply the weights $\textbf{W}^{C' \rightarrow C}$ directly to $\textbf{a}^{[l]}$ and decrease the overall complexity. However, this will result in a significant increase in both computations and memory requirements as it will require the discovery of the maximum class feature activations from the created spatio-temporal activations.

\begin{figure}[t]
\centering
    \includegraphics[width=\linewidth]{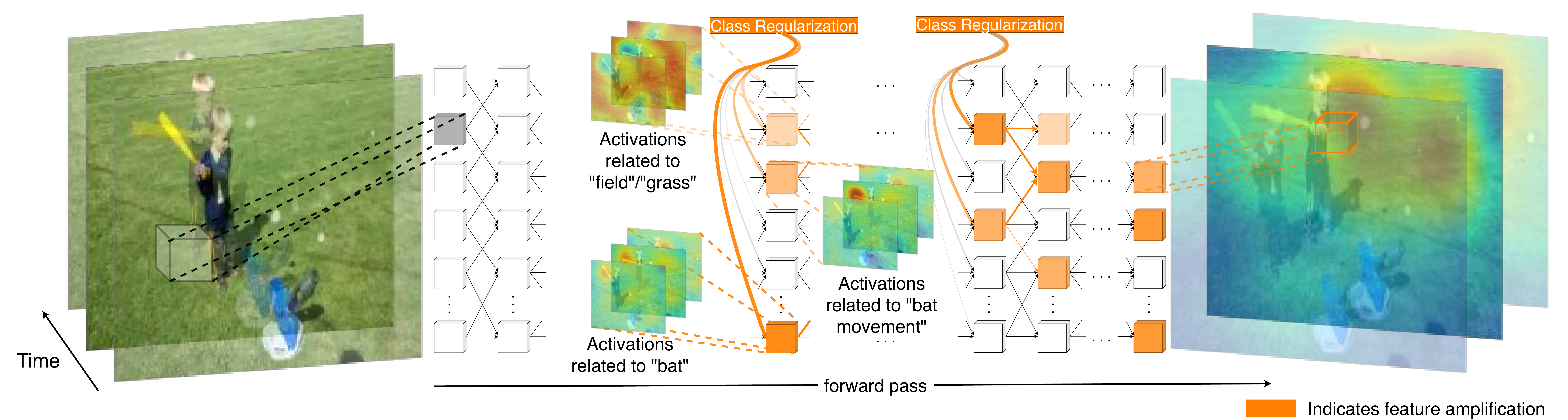}
   \caption{\textbf{Visualisation of feature amplification}. Class informative spatio-temporal features are intensified during iterations. The effect of amplification is propagated to later layers through the layer connections in which Class Regularisation is applied.}
\label{fig:CR_tubes}
\end{figure}

\subsection{Updating class regularisation weights}
\label{ch:6::sec:regularisation::updates}

% Motivation
As our proposed method utilises learnable parameters, there is a requirement of integrating the used weights ($\mathcal{W}$) as part of the learning procedure. We distinguish between the weights ($\mathcal{W}$) in our method that are used for dimensionality matching and the prediction layer's weights ($\textbf{W}$) they are applied to. The weights in the prediction layer are updated as normal based on the gradients calculated given the probabilistic class distribution. In order to re-use their values for Class Regularisation, we decouple them from the original computation graph and include them as a supplementary input as shown in \Cref{fig:classreg_block}. As the class prediction weights are used from the previous iteration, this decoupling of the weights has no effect on the class prediction weights in the current iteration.  

% Performing updates
The Class Regularisation parameters are updated based on the probabilistic loss calculated from the final prediction layer and thus are not required for the creation of an individual criterion. Dissimilarities in the feature representation space between layer activations $a^{[l]}$ and the transformed class weights $\textbf{W}^{C' \rightarrow C}$ are discovered based on the error corresponding to the class predictions. The alignment of the two feature spaces can also be achieved through a distance minimisation cost function (similar to an autoencoder style loss \cite{kramer1991nonlinear,roweis2000nonlinear,tenenbaum2000global}). However, we refrain from using a distance minimisation criterion for each Class Regularisation, or a concatenated version of it, at all locations that Class Regularisation is applied to, in order to limit the backwards traversal of information and the subsequent number of operations and FLOPs that are required. In addition, specific loss functions would mean that multiple gradients would be calculated for the remainder of the network parameters. Bifurcating the gradients and performing more than one step for each parameter during each iteration step has the potential of leading to sub-optimal parameter space directions and locations. Instead, by integrating the Class Regularisation parameters in the prediction task, the task of matching between the two created spaces is accumulated to the main classification task.

\begin{figure}[h]
\centering
    \includegraphics[width=\linewidth]{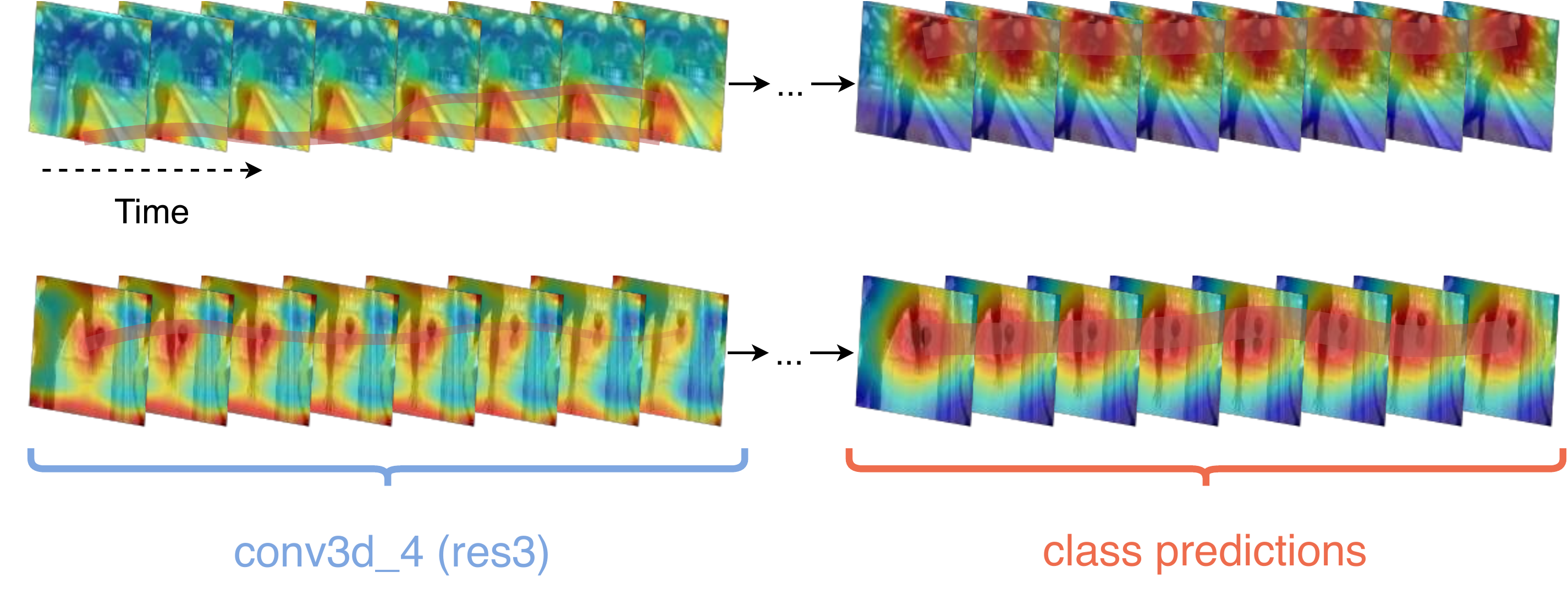}
   \caption{\textbf{Layer-wise class to feature correlation}. We use the features for (\textit{res3}) in wide-ResNet-50 \cite{zagoruyko2016wide} with 3D convolutions. Examples are drawn randomly from Kinetics-400 \cite{kay2017kinetics} and are both from class \textit{\enquote{bowling}}.}
\label{fig:CR_tubes_kinetics}
\end{figure}

\subsection{Class Regularisation for visual explanations}
\label{ch:6::sec:visual}

% Visualising individual block predictions with Class Reg.
Class Regularisation focuses on the inclusion of class-relative information inside the extracted features. This results in the creation of a direct correspondence between classes and features for each convolutional block that the method is applied to. An additional attribute, of this representation of class features in a feature space relative to the layer activations, is the improvement in terms of the overall explainability capabilities of the model. The curse of dimensionality problem has led to the dependence of back-propagating from the predictions to a particular layer. By actively alleviating it for the visualisation of relevant features, direct feature to class visual explanations can be created. Since the classes are represented in the same feature space as the activation maps, spatio-temporal regions that are informative over multiple network layers, can be discovered. To the best of our knowledge, this is the first method that enables such a direct feature correspondence.

% Use of Saliency tubes.
We extend the proposed regularisation method to enable visualisation with the inclusion of \textit{Saliency Tubes} \cite{stergiou2019saliency} at each block that it is applied to. Based on this, we create representations of features with the highest activations with respect to the selected class. We show two visual examples of the class \textit{\enquote{bowling}} in Kinetics-400 to demonstrate class activations in different layers of the network (\Cref{fig:CR_tubes_kinetics}). As observed, in both cases features in the early layers lack in terms of their deterministic capabilities for their feature to class correspondence. Most of the features are targeted at the distinction between foreground and background. In later layers however, this focus shifts from the actor to the background in the predictions layer in the top clip. As the bowling ball and bowling pins are part of a wallpaper, the overall focus shifts which demonstrates that a strong spatial-based visual signal is favoured over certain cases during the feature extraction process. In the bottom clip, the main field of focus of the network in the last layers is more limited to the area following the ball's trail. We note that the experimented architectures fuse spatio-temporal features together in the last convolution layers of the network, thus only the spatial extent can be visualised for the final class predictions.

% Visualisations for feature amplifications
The amplification of layer features based on corresponding classes is additionally presented in \Cref{fig:CR_tubes}. In \Cref{fig:CR_tubes}, the top-3 kernels to be amplified for an example of class \textit{\enquote{baseball hit}} can have a correspondence to the spatio-temporal features, such as the appearance of the bat, the field and the movement of the bat during a swing. In addition, these amplifications are also propagated to deeper layers in the network through the connections of the most informative kernels.

\section{Experiments}
\label{ch:6::sec:experiments}

% Network selection
We present the merits of our proposed Class Regularisation for the task of action recognition on ResNet-based architectures \cite{he2016deep,hara2018can,tran2018closer,zagoruyko2016wide} as well as on MTNets \cite{stergiou2020right}. The ResNet architectures chosen include both 3D and (2+1)D convolution variants as well as differences in their channel dimension size with wide-ResNets \cite{zagoruyko2016wide}. We include wide Residual Networks in our tests in order to have a greater understanding of the effects of larger feature dimension sizes. We do not explicitly include experiments with SR \cite{stergiou2021learn} networks from \Cref{ch4}, as MTNets include SR as part of their architectures.

\subsection{Architectural overview based on Class Regularisation}
\label{ch:6::sec:experiments::arch}

% Architectural changes based on Class Regularisation
An overview of the architectures alongside how Class Regularisation is applied over the networks can be viewed in \Cref{tab:CR_architectures}. We first present the ResNet architectures in \Cref{tab:CR_architectures_ResNet} alongside the progressive change in $\lambda$. Our choice of values is based on the number of layers with $\lambda=1$ denoting that Class Regularisation will have no effect over the volume that it is applied to based on the regularisation shift. Instead, when $\lambda \rightarrow 0$ the effect of Class Regularisation increases over the volume that it is applied to. Based on this, we progressively decrease $\lambda$ as the feature complexity increases. In \Cref{tab:CR_architectures_MTNet} we show how the MTNets that use a X3D backbone are structured. For simplicity, we do not elaborate on the processes performed in a MTConv. Therefore, for example, a $3 \! \times \! 3 \! \times \! 3$ MTConv with $54$ channels will correspond to a  $3 \! \times \! 3 \! \times \! 3$ 3D Conv for the local ($\mathcal{L}$) branch with $47$ channels alongside a $3 \! \times \! 3 \! \times \! 3$ 3D Conv for the prolonged ($\mathcal{P}$) branch with $7$ channels and a single point-wise convolution for transforming the local branch feature space to that of the prolonged $47 \rightarrow 7$. The primary change between MTNet$_S$ and MTNet$_M$ is in their input size similarly to the X3D backbones that they employ \cite{feichtenhofer2020x3d}. MTNet$_S$ uses inputs of size $13 \! \times \! 160^{2}$, while MTNet$_M$ uses inputs of size $16 \! \times \! 224^{2}$. The affection rate $\lambda$ follows a scaling similar to the ResNets.

\begin{table}[ht]
    \caption{\textbf{Architectures of 3D convolutions with and without Class Regularisation}. The convolution kernels are denoted first followed by the number of channels and the number of blocks. Throughout all the models, Class Regularisation accounts for a small number of parameters.}
    
    \begin{subtable}{.9\linewidth}
        \caption{\textbf{ResNet-based models with and without Class Regularisation}}
        \resizebox{\textwidth}{!}{
        \begin{tabular}[t]{c|c|c|c|c}
            \hline
        
            % Header
            layer name & 3D ResNet50 (w/o CR) & (2+1)D ResNet50 (w/o CR) & 3D Wide ResNet 50 (w/o CR) & 3D ResNet101 (w/o CR)  \\[0.3em] \hline
        
            %First convolution layer
            $con3d_{1}$ & \multicolumn{4}{c}{$7 \times 7 \times 7,64$ conv} \\[0.25em] \hline
        
            % First identity block
            $con3d_{2}$ 
            & $\begin{bmatrix}1 \times \! 1 \! \times 1 \\3 \! \times \! 3 \! \times \! 3 \\1 \! \times \! 1 \! \times \! 1  \end{bmatrix} (\times 64) \times 3$
            
            & $\begin{bmatrix}1 \! \times \! 1 \! \times \! 1  \\1 \! \times \! 3 \! \times \! 3 \\3 \! \times \! 1 \! \times \! 1 \\1 \! \times \! 1 \! \times \! 1  \end{bmatrix} (\times 64) \times 3$
            
            & $\begin{bmatrix}1 \! \times \! 1 \! \times \! 1  \\3 \! \times \! 3 \! \times \! 3  \\1 \! \times \! 1 \! \times \! 1  \end{bmatrix} (\times 128) \times 3$
            
            & $\begin{bmatrix}1 \! \times \! 1 \! \times \! 1  \\3 \! \times \! 3 \! \times \! 3  \\1 \! \times \! 1 \! \times \! 1  \end{bmatrix} (\times 64) \times 3$ \\[0.2em] \hline
            
            % Class Regularisation
            $ClassReg_{1}$ & - / $\lambda=0.9$ & - / $\lambda=0.9$ & - / $\lambda=0.9$ & - / $\lambda=0.9$\\[0.2em] \hline
            
            % Second identity block
            $con3d_{3}$ 
            & $\begin{bmatrix}1 \! \times \! 1 \! \times \! 1 \\3 \! \times \! 3 \! \times \! 3 \\1 \! \times \! 1 \! \times \! 1  \end{bmatrix} (\times 128) \times 4$
            
            & $\begin{bmatrix}1 \! \times \! 1 \! \times \! 1  \\1 \! \times \! 3 \! \times \! 3 \\3 \! \times \! 1 \! \times \! 1 \\1 \! \times \! 1 \! \times \! 1  \end{bmatrix} (\times 128) \times 4$
            
            & $\begin{bmatrix}1 \! \times \! 1 \! \times \! 1  \\3 \! \times \! 3 \! \times \! 3  \\1 \! \times \! 1 \! \times \! 1  \end{bmatrix} (\times 256) \times 4$
            
            & $\begin{bmatrix}1 \! \times \! 1 \! \times \! 1  \\3 \! \times \! 3 \! \times \! 3  \\1 \! \times \! 1 \! \times \! 1  \end{bmatrix} (\times 128) \times 4$ \\[0.2em] \hline
            
            % Class Regularisation
            $ClassReg_{2}$ & - / $\lambda=0.8$ & - / $\lambda=0.8$ & - / $\lambda=0.8$ & - / $\lambda=0.8$\\[0.2em] \hline
            
            % Third identity block
            $con3d_{4}$ 
            & $\begin{bmatrix}1 \! \times \! 1 \! \times\! 1 \\3 \! \times \! 3 \! \times \! 3 \\1 \! \times \! 1 \! \times \! 1  \end{bmatrix} (\times 256) \times 6$
            
            & $\begin{bmatrix}1 \! \times \! 1 \! \times \! 1  \\1 \! \times \! 3 \! \times \! 3 \\3 \! \times \! 1 \! \times \! 1 \\1 \! \times \! 1 \! \times \! 1  \end{bmatrix} (\times 256) \times 6$
            
            & $\begin{bmatrix}1 \! \times \! 1 \! \times \! 1  \\3 \! \times \! 3 \! \times \! 3  \\1 \! \times \! 1 \! \times \! 1  \end{bmatrix} (\times 512) \times 6$
            
            & $\begin{bmatrix}1 \! \times \! 1 \! \times \! 1  \\3 \! \times \! 3 \! \times \! 3  \\1 \! \times \! 1 \! \times \! 1  \end{bmatrix} (\times 256) \times 23$ \\[0.2em] \hline
            
            %Class Regularisation
            $ClassReg_{3}$ & - / $\lambda=0.7$ & - / $\lambda=0.7$ & - / $\lambda=0.7$ & - / $\lambda=0.7$\\[0.2em] \hline
            
            %Fourth identity block
            $con3d_{5}$ 
            & $\begin{bmatrix}1 \! \times \! 1 \! \times \! 1 \\3 \! \times \! 3 \! \times \! 3 \\1 \! \times \! 1 \! \times \! 1  \end{bmatrix} (\times 512) \times 3$
            
            & $\begin{bmatrix}1 \! \times \! 1 \! \times \! 1  \\1 \! \times \! 3 \! \times \! 3 \\3 \! \times \! 1 \! \times \! 1 \\1 \! \times \! 1 \! \times \! 1  \end{bmatrix} (\times 512) \times 3$
            
            & $\begin{bmatrix}1 \! \times \! 1 \! \times \! 1  \\3 \! \times \! 3 \! \times \! 3  \\1 \! \times \! 1 \! \times \! 1  \end{bmatrix} (\times 1024) \times 3$
            
            & $\begin{bmatrix}1 \! \times \! 1 \! \times \! 1  \\3 \! \times \! 3 \! \times \! 3  \\1 \! \times \! 1 \! \times \! 1  \end{bmatrix} (\times 512) \times 3$ \\[0.2em] \hline
            
            %Class Regularisation
            $ClassReg_{4}$ & - / $\lambda=0.6$ & - / $\lambda=0.6$ & - / $\lambda=0.6$ & - / $\lambda=0.6$\\[0.2em] \hline
            
             %Global Average Pooling and SoftMax
            predictions & \multicolumn{4}{c}{global average pool, softmax unit group} \\[0.2em] \hline
        
        \end{tabular}
        }
\label{tab:CR_architectures_ResNet}
\end{subtable}
\vspace{1.2em}
\\

\begin{subtable}{.9\textwidth}
        \caption{\textbf{MTNets with and without Class Regularisation}.}
        \resizebox{\textwidth}{!}{
        \begin{tabular}[t]{c|c|c|c}
            \hline
        
            % Header
            layer name & MTNet$_S$ (w/o CR) & (2+1)D MTNet$_M$ (w/o CR) & MTNet$_L$ (w/o CR) \\[0.3em] \hline
        
            %First convolution layer
            $con3d_{1}$ & \multicolumn{3}{c}{$1 \times 3 \! \times \! 3,24$ conv} \\[0.25em] \hline
        
            % First identity block
            $con3d_{2}$ 
            & $\begin{bmatrix}1 \! \times \! 1 \! \times \! 1 \\3 \! \times \! 3 \! \times \! 3 \\1 \! \times \! 1 \! \times \! 1  \end{bmatrix} \times 
            \begin{bmatrix} 54 \\54 \\24  \end{bmatrix} 
            \times 3$
            
            & $\begin{bmatrix}1 \! \times \! 1 \! \times \! 1 \\3 \! \times \! 3 \! \times \! 3 \\1 \! \times \! 1 \! \times \! 1  \end{bmatrix} \times 
            \begin{bmatrix} 54 \\54 \\24  \end{bmatrix} 
            \times 3$
            
            & $\begin{bmatrix}1 \! \times \! 1 \! \times \! 1 \\3 \! \times \! 3 \! \times \! 3 \\1 \! \times \! 1 \! \times \! 1  \end{bmatrix} \times 
            \begin{bmatrix} 54 \\54 \\24  \end{bmatrix} 
            \times 5$ \\[0.2em] \hline
            
            % Class Regularisation
            $ClassReg_{1}$ & - / $\lambda=0.9$ & - / $\lambda=0.9$ & - / $\lambda=0.9$ \\[0.2em] \hline
            
            % Second identity block
            $con3d_{3}$ 
            & $\begin{bmatrix}1 \! \times \! 1 \! \times \! 1 \\3 \! \times \! 3 \! \times \! 3 \\1 \! \times \! 1 \! \times \! 1  \end{bmatrix} \times 
            \begin{bmatrix} 108 \\108 \\48  \end{bmatrix} 
            \times 5$
            
            & $\begin{bmatrix}1 \! \times \! 1 \! \times \! 1 \\3 \! \times \! 3 \! \times \! 3 \\1 \! \times \! 1 \! \times \! 1  \end{bmatrix} \times 
            \begin{bmatrix} 108 \\108 \\48  \end{bmatrix} 
            \times 5$
            
            & $\begin{bmatrix}1 \! \times \! 1 \! \times \! 1 \\3 \! \times \! 3 \! \times \! 3 \\1 \! \times \! 1 \! \times \! 1  \end{bmatrix} \times 
            \begin{bmatrix} 108 \\108 \\48  \end{bmatrix} 
            \times 10$ \\[0.2em] \hline
            
            % Class Regularisation
            $ClassReg_{2}$ & - / $\lambda=0.8$ & - / $\lambda=0.8$ & - / $\lambda=0.8$ \\[0.2em] \hline
            
            % Third identity block
            $con3d_{4}$ 
            & $\begin{bmatrix}1 \! \times \! 1 \! \times \! 1 \\3 \! \times \! 3 \! \times \! 3 \\1 \! \times \! 1 \! \times \! 1  \end{bmatrix} \times 
            \begin{bmatrix} 216 \\216 \\96  \end{bmatrix} 
            \times 11$
            
            & $\begin{bmatrix}1 \! \times \! 1 \! \times \! 1 \\3 \! \times \! 3 \! \times \! 3 \\1 \! \times \! 1 \! \times \! 1  \end{bmatrix} \times 
            \begin{bmatrix} 216 \\216 \\96  \end{bmatrix} 
            \times 11$
            
            & $\begin{bmatrix}1 \! \times \! 1 \! \times \! 1 \\3 \! \times \! 3 \! \times \! 3 \\1 \! \times \! 1 \! \times \! 1  \end{bmatrix} \times 
            \begin{bmatrix} 216 \\216 \\96  \end{bmatrix} 
            \times 25$ \\[0.2em] \hline
            
            %Class Regularisation
            $ClassReg_{3}$ & - / $\lambda=0.7$ & - / $\lambda=0.7$ & - / $\lambda=0.7$ \\[0.2em] \hline
            
            %Fourth identity block
            $con3d_{5}$ 
            & $\begin{bmatrix}1 \! \times \! 1 \! \times \! 1 \\3 \! \times \! 3 \! \times \! 3 \\1 \! \times \! 1 \! \times \! 1  \end{bmatrix} \times 
            \begin{bmatrix} 432 \\432 \\192  \end{bmatrix} 
            \times 7$
            
            & $\begin{bmatrix}1 \! \times \! 1 \! \times \! 1 \\3 \! \times \! 3 \! \times \! 3 \\1 \! \times \! 1 \! \times \! 1  \end{bmatrix} \times 
            \begin{bmatrix} 432 \\432 \\192  \end{bmatrix} 
            \times 7$
            
            & $\begin{bmatrix}1 \! \times \! 1 \! \times \! 1 \\3 \! \times \! 3 \! \times \! 3 \\1 \! \times \! 1 \! \times \! 1  \end{bmatrix} \times 
            \begin{bmatrix} 432 \\432 \\192  \end{bmatrix} 
            \times 15$ \\[0.2em] \hline
            
            %Class Regularisation
            $ClassReg_{4}$ & - / $\lambda=0.6$ & - / $\lambda=0.6$ & - / $\lambda=0.6$ \\[0.2em] \hline
            
             %Global Average Pooling and SoftMax
            predictions & \multicolumn{3}{c}{global average pool, softmax unit group} \\[0.2em] \hline
        
        \end{tabular}
        }
\label{tab:CR_architectures_MTNet}
\end{subtable}
\label{tab:CR_architectures}
\end{table}

\subsection{Experiment setup}
\label{ch:6::sec:experiments::setup}

% Datasets
We demonstrate our proposed method on two benchmark action recognition datasets. For our experiments we use the widely popular Kinetics-400 and HACS for baseline comparisons.   

% Training configuration
Training is only performed for the weights associated with Class Regularisation. The rest of the network weights are from pre-trained models and no updates are performed over them. On Kinetics-400 (K-400) and HACS we train the Class Regularisation weights from scratch while using a standard Kaiming initialisation \cite{he2015delving}. The initial mini-batch is set to 16 with 4 clips per GPU. For all of our experiments we use a SGD optimiser with momentum \cite{sutskever2013importance} which we set to 0.9. We use weight-decay of value 1e-5. We use spatio-temporal data augmentations and image crops similar to the previous chapter. All models were trained over a total of 120 epochs as no further improvements were observed afterwards. Based on this, we further apply a step-wise learning rate reduction every 50 epochs to one tenth of the previous learning rate value. The initial learning rate is set to 0.1. For fine-tuning we decrease the learning rate to 1e-3 and set the step-wise learning rate reduction to every 70 steps.

\subsection{Results on HACS}
\label{ch:6::sec:experiments::hacs}

% HACS table overview
In our comparisons on the HACS \cite{zhao2019hacs} dataset we include Class Regularisation in the current top-performing MTNet \cite{stergiou2020right} models and compare with other state of the art architectures in \Cref{table:HACS_accuracies_cr}. In order to maintain comparison standards with the rest of the works, we use the same cropping and sampling size as in \cite{feichtenhofer2020x3d,stergiou2020right}, which results in 30-view spatio-temporal input clips. For each architecture we report, from left to right, the pre-training datasets (Pre) that the models used to initialise their weights following the average achieved top-1 and top-5 accuracy rates on the validation set. Models that do not include a pre-trained dataset are trained from scratch. For the last two metrics, we report the inference cost expressed as the required GFLOPs per single clip times the number of clips and their spatio augmentations (denoted as \enquote{views}). We finally report the overall number of parameters for each network. 

\begin{table}[t]
\caption{\textbf{Action recognition model comparisons on HACS}. Weight initialisation sources are denoted by their respective indicators.}
\begin{threeparttable}[t]
\centering
\resizebox{.9\textwidth}{!}{%
\renewcommand{\arraystretch}{1.2} 
\begin{tabular}{c|c|c|c|c|c}
\hline
Model & 
Pre &
top-1 & top-5 & 
GFLOPs $\! \times \!$ views &
Params \\
\hline

MF-Net \cite{chen2018multifiber}$^{\dagger}$ &
\multirow{4}{*}{K-400 \cite{kay2017kinetics}} &
78.3 & 94.6 &
$11.1 \! \times \! 50$ &
8.0M \\

TAM \cite{fan2019more}$^{\dagger}$ &
 &
82.2 & 95.2 &
$86 \! \times \! 12$ &
25.6M\\

SF-101 \cite{feichtenhofer2019slowfast}$^{\dagger}$ &
 &
83.7 & 96.8 &
$65.7 \! \times \! 30$ &
53.7M \\

X3D-L \cite{feichtenhofer2020x3d}$^{\dagger}$ &
&
85.8 & 96.1 &
$24.8 \! \times \! 30$ &
\textbf{6.1}M\\
\hline

ir-CSN-101 \cite{tran2019video}$^{\dagger}$ &
\multirow{2}{*}{IG65 \cite{ghadiyaram2019large}} &
83.8 & 93.8 &
$63.6 \! \times \! 10$ &
22.1M \\

ip-CSN-101 \cite{tran2019video}$^{\dagger}$ &
 &
84.1 & 93.9 &
$63.6 \! \times \! 10$ &
24.5M \\
\hline

r3d-34 \cite{kataoka2020would}$^{*}$ &
\multirow{6}{*}{-} &
74.8 & 92.8 &
$26.6 \! \times \! 30$ &
63.7M\\

r3d-50 \cite{kataoka2020would}$^{*}$ &
 &
78.4 & 93.8 &
$52.6 \! \times \! 30$ &
36.7M\\

r3d-101 \cite{kataoka2020would}$^{*}$ &
 &
80.5 & 95.2 &
$78.0 \! \times \! 30$ &
69.1M\\

r(2+1)d-34 \cite{kataoka2020would}$^{*}$ &
 &
75.7 & 93.8 &
$37.8 \! \times \! 30$ &
61.8M\\

r(2+1)d-50 \cite{kataoka2020would}$^{*}$ &
 &
81.3 & 94.5 &
$83.3 \! \times \! 30$ &
34.8M\\

r(2+1)d-101 \cite{kataoka2020would}$^{*}$ &
 &
82.9 & 95.7 &
$163.0 \! \times \! 30$ &
67.2M\\
\hline

SRTG r3d-34 \cite{stergiou2021learn} &
\multirow{6}{*}{-} &
78.6 & 93.6 &
$26.6 \! \times \! 30$ &
83.8M \\

SRTG r3d-50 \cite{stergiou2021learn} &
 &
80.3 & 95.5 &
$52.7 \! \times \! 30$ &
56.9M \\

SRTG r3d-101 \cite{stergiou2021learn} &
 &
81.6 & 96.3 &
$78.1 \! \times \! 30$ &
107.1M \\

SRTG r(2+1)d-34 \cite{stergiou2021learn} &
 &
80.4 & 94.3 &
$37.8 \! \times \! 30$ &
82.1M \\

SRTG r(2+1)d-50 \cite{stergiou2021learn}  &
 &
83.8 & 96.6 &
$83.4 \! \times \! 30$ &
55.0M \\

SRTG r(2+1)d-101 \cite{stergiou2021learn}  &
 &
84.3 & 96.8 &
$163.1 \! \times \! 30$ &
105.3M \\
\hline

MTNet$_{S}$ \cite{stergiou2020right} &
\multirow{3}{*}{-} &
80.7 & 95.2 &
\textbf{$5.8 \! \times \! 30$}  &
25.8M\\

MTNet$_{M}$ \cite{stergiou2020right} &
 &
83.4 & 95.9 &
$8.8 \! \times \! 30$  &
25.8M\\

MTNet$_{L}$ \cite{stergiou2020right} &
 &
86.6 & 96.7 &
$17.6 \! \times \! 30$  &
50.1M\\
\hline

MTNet$_{S}$ (CR) &
\multirow{3}{*}{-} &
81.8 & 96.1 &
\textbf{$5.8 \! \times \! 30$}  &
26.5M\\

MTNet$_{M}$ (CR) &
 &
84.7 & 96.7 &
$8.9 \! \times \! 30$  &
26.5M\\

MTNet$_{L}$ (CR) &
 &
\textbf{87.5} & \textbf{97.4} &
$17.7 \! \times \! 30$  &
51.2M\\

\end{tabular}%
}
 \begin{tablenotes}
    \item[$\dagger$] models and weights from official repositories.  
    \item[$*$] re-implemented models.
   \end{tablenotes}
\end{threeparttable}%
\label{table:HACS_accuracies_cr}
\end{table}

% Results comparison
In comparison to the top-preforming models on HACS, enabling Class Regularisation can increase the overall performance. Notably, it also comes at a negligible computational overhead when added over MTNet ($<+1\%$). More specifically, considering the overall efficiency of the MTNets in addition to the proposed methods, our MTNet$_S$ with CR can achieve accuracy rates similar to those of ResNet-101 with 3D or ResNet-50 with (2+1)D convolutions \cite{tran2018closer}, as well as TAM \cite{fan2019more} and SRTG r3d-50 \cite{stergiou2021learn}. However, the GFLOP requirements are very low, similar to the original MTNet$_S$ with only an additional $>0.1$ GFLOPs that account for less than $10\%$ of the total $5.8$ GFLOPs. We additionally do not notice a significant increase in terms of the computational overhead for MTNet$_M$ and MTNet$_L$ with the inclusion of Class Regularisation. For MTNet$_M$ (CR), we note a performance increase of +1.3\% in the top-1 and +0.8\% in the top-5 compared to the original network without CR. These accuracy rates are similar to both of the larger SRTG r3d-101 \cite{stergiou2021learn} and Channel-Separated Networks (ip-CSN-101) \cite{tran2019video}. This increase in performance does not come at a cost in terms of the computational inference or the number of parameters with the additional number of parameters accounting for $\sim 10\%$ of the total number of network parameters. The larger MTNet$_L$ (CR) outperforms all other tested models by a margin of at least $+0.9\%$ in the top-1 and $+0.7\%$ (from MTNet$_L$) while maintaining almost identical computational complexity as MTNet$_L$. Considering the fact that the tested models were not fine-tuned, with their weights previously initialised from a significantly larger dataset, the performance benefits of Class Regularisation are evident.

\subsection{Results on Kinetics-400}
\label{ch:6::sec:experiments::kinetics}

The results presented on the K-400 dataset include experiments on pre-trained state-of-the art architectures alongside pair-wise comparisons for networks with weights that have been randomly initialised similarly to the experiments on HACS. The distinction between the two is made in order to demonstrate the direct impact of Class Regularisation across different training environments. The first one explores the capabilities of the proposed method on pre-trained models while the other is done \textit{from scratch}, similar to HACS.

% Main results on KInetics w/ MTNet
\textbf{Main accuracy results}. We demonstrate in \Cref{tab:K400_accuracies_cr} the main accuracy rates achieved through the inclusion of Class Regularisation in the top-performing MTNet architectures. The accuracy rates follow a similar trend as those that have been presented on HACS, with MTNet$_L$ with Class Regularisation (CR) being the top-performing architecture. This further decreases the gap between the significantly larger X3D-XL and MTNet$_L$ to only 0.2\% for the top-1 accuracy. However, the computational complexity of the MTNet$_L$ architecture still remains small, corresponding to approximately 36\% of that of X3D-XL. Through this we show that the inclusion of CR comes with no additional computational costs, thus being a straightforward approach for lightweight architectures to further improve their classification accuracies. The smaller MTNet$_M$ also shows competitive results to those of the top performing models when also including CR. The top-1 accuracy is drawn closer to networks such as SlowFast-101 and MTNet$_L$ while only including a fraction of the computational requirements. In terms of efficiency, it still retains its overall efficiency with computations being reducing further from the lightweight MFNet and TSM. The only two more efficient models are the MTNet$_S$ variants with and without Class Regularisation. The smallest tested architecture, MTNet$_S$ (CR), shows a +1.3\% improvement on the top-1 accuracy in comparison to the baseline model MTNet$_S$ without CR. This comes with less than 0.1 GFLOPs of additional computations. In terms of additional parameters, across all of the tested architectures the parameter increments with CR remain constant at approximately +0.76M parameters. This increase is more noticeable to small architectures, such as MTNet$_S$ in which the number of parameters is already limited. In larger architectures, such as MTNet$_L$ the additional parameters are proportionally less significant.

\begin{table}[t]
\caption{\textbf{Comparison with K-400 state-of-the-art}. For consistency with previous testing methods, we report the model complexity as the GFLOPs per single clip view $\! \times \!$ the number of clips with spatial cropping of size $256 \! \times \! 256$.}
\centering
\resizebox{\textwidth}{!}{%
\renewcommand{\arraystretch}{1.5}
\begin{tabular}{c|c|c|c|c|c|c}
\hline
Model & 
Input &
Backbone &
top-1 & top-5 & 
GFLOPs $\! \times \!$ views &
Params\\
\hline
I3D \cite{carreira2017quo}  &
$16 \! \times \! 224^{2}$ &
InceptionV1 &
71.6 & 90.0 &
$108 \! \times \! N/A$&
12M\\

TSM \cite{lin2019tsm} &
$16 \! \times \! 224^{2}$ &
ResNet50 &
74.7 & 91.4 &
$65 \! \times \! 10$&
24.3M\\

R(2+1)D \cite{tran2018closer} &
$16 \! \times \! 224^{2}$ &
ResNet101 &
62.8 & 83.9 &
$152 \! \times \! 115$&
63.6M\\

ip-CSN-101 \cite{tran2019video} &
$8 \! \times \! 224^{2}$ &
ResNet101 &
76.7 & 92.3 &
$83.0 \! \times \! 30$ &
24.5M\\

ip-CSN-152 \cite{tran2019video} &
$8 \! \times \! 224^{2}$ &
ResNet152 &
77.8 & 92.8 &
$108.8 \! \times \! 30$ &
32.8M\\

MF-Net \cite{chen2018multifiber} &
$16 \! \times \! 224^{2}$ &
ResNet50 &
72.8 & 90.4 &
$11.1 \! \times \! 50$ &
8.0M\\

SF-50 \cite{feichtenhofer2019slowfast} &
$(32,4) \! \times \! 224^{2}$ &
ResNet50 &
77.0 & 92.6 &
$65.7 \! \times \! 30$ &
34.4M\\

SF-101  \cite{feichtenhofer2019slowfast} &
$(32,4) \! \times \! 224^{2}$ &
ResNet101 &
77.9 & 93.5 &
$213 \! \times \! 30$ &
53.7M\\

X3D-XL \cite{feichtenhofer2020x3d} &
$16 \! \times \! 224^{2}$ &
ResNet(X3D) &
\textbf{79.1} & \textbf{93.9} &
$48.4 \! \times \! 30$ &
11.0M\\

TAM \cite{fan2019more} &
$16 \! \times \! 256^{2}$ &
ResNet50 &
76.9 & 92.9 &
$86 \! \times \! 12$ &
25.6M\\

SRTG r3d-101 \cite{stergiou2021learn} &
$16 \! \times \! 224^{2}$ &
ResNet101 &
73.2 & 91.3 &
$78.1 \! \times \! 30$ &
107.1M\\

SRTG r(2+1)d-101 \cite{stergiou2021learn} &
$16 \! \times \! 224^{2}$ &
ResNet101 &
73.8 & 92.0 &
$163.1 \! \times \! 30$ &
105.3M\\

MTNet$_{S}$ \cite{stergiou2020right} &
$13 \! \times \! 182^{2}$ &
ResNet(X3D) &
74.8 & 92.1 &
$\mathbf{5.8 \! \times \! 30}$ &
25.8M\\

MTNet$_{M}$ \cite{stergiou2020right} &
$16 \! \times \! 256^{2}$ &
ResNet(X3D) &
76.6 & 92.5 &
$8.8 \! \times \! 30$ &
25.8M\\

MTNet$_{L}$ \cite{stergiou2020right} &
$16 \! \times \! 256^{2}$ &
ResNet(X3D) &
78.1 & 93.2 &
$17.6 \! \times \! 30$ &
50.1M\\
\hline

MTNet$_{S}$ (CR) &
$13 \! \times \! 182^{2}$ &
ResNet(X3D) &
76.1 & 92.4 &
$5.8 \! \times \! 30$ &
26.5M\\

MTNet$_{M}$ (CR) &
$16 \! \times \! 256^{2}$ &
ResNet(X3D) &
77.7 & 92.9 &
$8.9 \! \times \! 30$ &
26.5M\\

MTNet$_{L}$ (CR) &
$16 \! \times \! 256^{2}$ &
ResNet(X3D) &
78.9 & 93.6 &
$17.7 \! \times \! 30$ &
51.0M\\

\end{tabular}%
}
\label{tab:K400_accuracies_cr}
\end{table}

% Comparisons in pairs
\textbf{Pair-wise comparisons}. In \Cref{tab:k400_nopre} we present pair-wise accuracies over models that are not pre-trained on larger datasets and instead are trained with their weights being randomly initialised. The accuracies of the models that do not include CR are re-calculated to ensure the same training setting. Over all of the four tested models, architectures that include CR outperform their counterparts without. The largest margins are observed for wide-r3d-50 with a 1.3\% increase in the top-1 accuracy and r3d-101 with +1.3\% increase for the top-5 accuracy. In contrast, for smaller models such as r3d-50 the improvements show to be less with +0.6\% for the top-1 and top-5 accuracies. This seems to suggest that larger models such as r3d-101 or models with an increased number of channels per layer, such as wide-r3d-50, benefit more from the proposed regularisation method. The added floating-point operations (FLOPs) remain constant as the method is not affected by the increase in the number of convolutional layers per block. However, an increase is observed for wide-r3d-50, in which the number of channels per block is expanded. This corresponds to the latent space of the feature representations per convolution being enlarged which in turn also affects the representation of class-relative features in that space as well. Similar tendencies are also observed for the number of parameters, with the only exception being wide-r3d-50 and with the increased convolutional feature space.

\begin{figure*}[ht]
\centering
\includegraphics[width=1\textwidth]{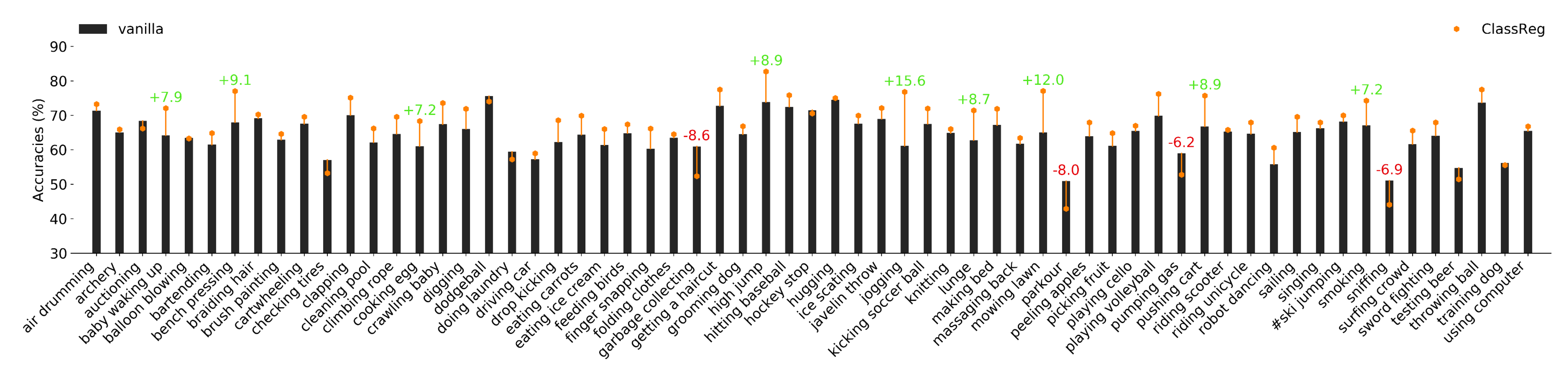}

\begin{minipage}[b]{0.12\linewidth}
    \centering
    \includegraphics[width=.9\linewidth]{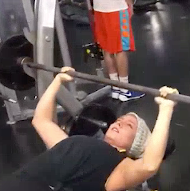} 
    \vspace{2ex}
\end{minipage}%%
\begin{minipage}[b]{0.12\linewidth}
    \centering
    \includegraphics[width=.9\linewidth]{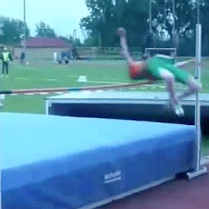} 
    \vspace{2ex}
\end{minipage}%%
\begin{minipage}[b]{0.12\linewidth}
    \centering
    \includegraphics[width=.9\linewidth]{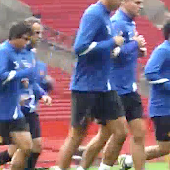} 
    \vspace{2ex}
\end{minipage}%%
\begin{minipage}[b]{0.12\linewidth}
    \centering
    \includegraphics[width=.9\linewidth]{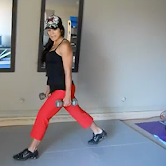} 
    \vspace{2ex}
\end{minipage}%%
\begin{minipage}[b]{0.12\linewidth}
    \centering
    \includegraphics[width=.9\linewidth]{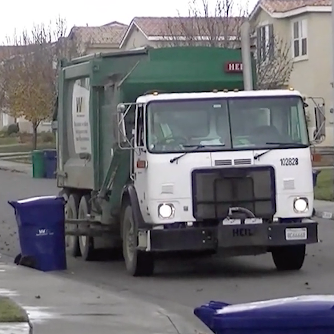} 
    \vspace{2ex}
\end{minipage}%%
\begin{minipage}[b]{0.12\linewidth}
    \centering
    \includegraphics[width=.9\linewidth]{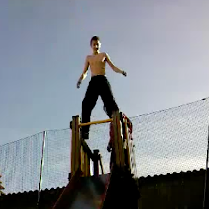} 
    \vspace{2ex}
\end{minipage}%%
\begin{minipage}[b]{0.12\linewidth}
    \centering
    \includegraphics[width=.9\linewidth]{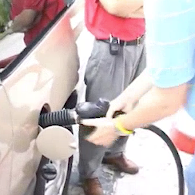} 
    \vspace{2ex}
\end{minipage}%%
\begin{minipage}[b]{0.12\linewidth}
    \centering
    \includegraphics[width=.9\linewidth]{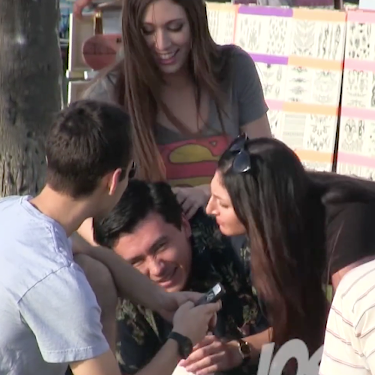} 
    \vspace{2ex}
\end{minipage}%%

\vspace{-2ex}

\begin{minipage}[b]{0.12\linewidth}
    \centering
    \includegraphics[width=.9\linewidth]{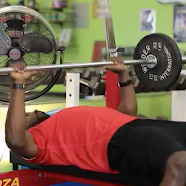} 
    \vspace{2ex}
\end{minipage}%%
\begin{minipage}[b]{0.12\linewidth}
    \centering
    \includegraphics[width=.9\linewidth]{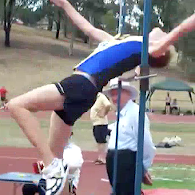} 
    \vspace{2ex}
\end{minipage}%%
\begin{minipage}[b]{0.12\linewidth}
    \centering
    \includegraphics[width=.9\linewidth]{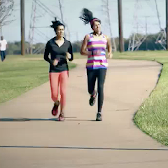} 
    \vspace{2ex}
\end{minipage}%%
\begin{minipage}[b]{0.12\linewidth}
    \centering
    \includegraphics[width=.9\linewidth]{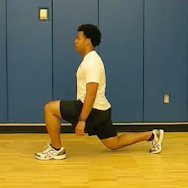} 
    \vspace{2ex}
\end{minipage}%%
\begin{minipage}[b]{0.12\linewidth}
    \centering
    \includegraphics[width=.9\linewidth]{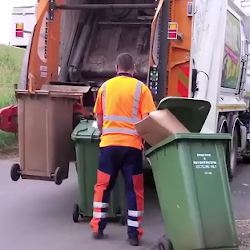} 
    \vspace{2ex}
\end{minipage}%%
\begin{minipage}[b]{0.12\linewidth}
    \centering
    \includegraphics[width=.9\linewidth]{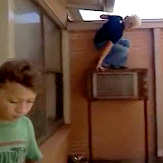} 
    \vspace{2ex}
\end{minipage}%%
\begin{minipage}[b]{0.12\linewidth}
    \centering
    \includegraphics[width=.9\linewidth]{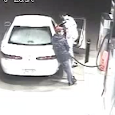} 
    \vspace{2ex}
\end{minipage}%%
\begin{minipage}[b]{0.12\linewidth}
    \centering
    \includegraphics[width=.9\linewidth]{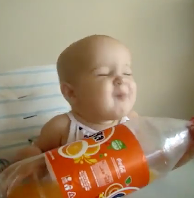} 
    \vspace{2ex}
\end{minipage}%%

\vspace{-2ex}

\begin{minipage}[b]{0.12\linewidth}
    \centering
    \includegraphics[width=.9\linewidth]{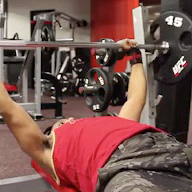}
    \vspace{2ex}
\end{minipage}%%
\begin{minipage}[b]{0.12\linewidth}
    \centering
    \includegraphics[width=.9\linewidth]{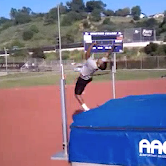}
    \vspace{2ex}
\end{minipage}%%
\begin{minipage}[b]{0.12\linewidth}
    \centering
    \includegraphics[width=.9\linewidth]{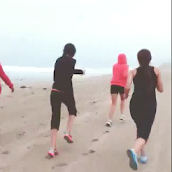}
    \vspace{2ex}
\end{minipage}%%
\begin{minipage}[b]{0.12\linewidth}
    \centering
    \includegraphics[width=.9\linewidth]{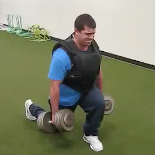}
    \vspace{2ex}
\end{minipage}%%
\begin{minipage}[b]{0.12\linewidth}
    \centering
    \includegraphics[width=.9\linewidth]{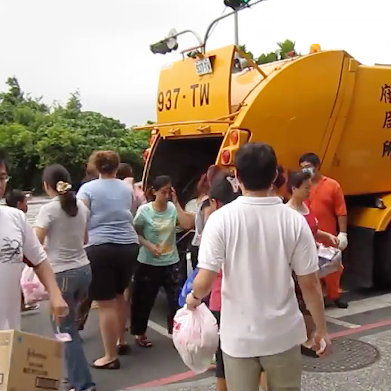}
    \vspace{2ex}
\end{minipage}%%
\begin{minipage}[b]{0.12\linewidth}
    \centering
    \includegraphics[width=.9\linewidth]{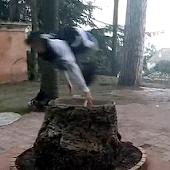}
    \vspace{2ex}
\end{minipage}%%
\begin{minipage}[b]{0.12\linewidth}
    \centering
    \includegraphics[width=.9\linewidth]{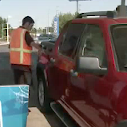}
    \vspace{2ex}
\end{minipage}%%
\begin{minipage}[b]{0.12\linewidth}
    \centering
    \includegraphics[width=.9\linewidth]{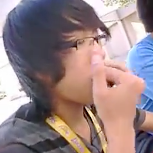}
    \vspace{2ex}
\end{minipage}%%

\vspace{-1ex}

\begin{minipage}[b]{0.12\linewidth}
    \centering
    bench pressing
    \vspace{2ex}
\end{minipage}%%
\begin{minipage}[b]{0.12\linewidth}
    \centering
    high jump
    \vspace{3ex}
\end{minipage}%%
\begin{minipage}[b]{0.12\linewidth}
    \centering
    jogging
    \vspace{3ex}
\end{minipage}%%
\begin{minipage}[b]{0.12\linewidth}
    \centering
    lunge
    \vspace{3ex}
\end{minipage}%%
\begin{minipage}[b]{0.12\linewidth}
    \centering
    garbage collecting
    \vspace{2ex}
\end{minipage}%%
\begin{minipage}[b]{0.12\linewidth}
    \centering
    parkour
    \vspace{3ex}
\end{minipage}%%
\begin{minipage}[b]{0.12\linewidth}
    \centering
    pumping gas
    \vspace{2ex}
\end{minipage}%%
\begin{minipage}[b]{0.12\linewidth}
    \centering
    sniffing
    \vspace{3ex}
\end{minipage}%%
\vspace{-1em}

\caption{\textbf{Per-class performance for ResNet-101 with and without Class Regularisation} with illustrative examples of classes with large performance gains or losses in K-400.}
\label{fig:class_acc}
%\vspace{-5mm}
\end{figure*}

As shown in \Cref{fig:class_acc}, with the inclusion of Class Regularisation, overall accuracy improvements across the majority of the K-400 classes can be observed. The largest margins are shown for instances and classes in which the execution can be descriptive. Examples of such cases include the \textit{\enquote{bench pressing}} , \textit{\enquote{high jump}} and \textit{\enquote{jogging}} classes with increases in their accuracies in the range of  8.9\% to 15.6\%. In contrast, classes that are more likely to contain significant feature variations are more likely to be wrongly classified. For example, considering the \textit{\enquote{parkour}} class, no standard features can be established as there are significant variations across examples. This also includes classes such as \textit{\enquote{garbage collection}}, which is performed either mechanically (top), by a single person (mid) or by multiple people (bottom). Factors such as the oscillations in \textit{\enquote{pumping gas}} or contextual information in \textit{\enquote{sniffing}} also effect the descriptive capabilities of features.

\begin{table}[t]
\centering
\caption{\textbf{Pair-wise comparisons for K-400}. We compare with popular residual architectures trained from scratch with and without Class Regularisation on Kinetics.}
\resizebox{.8\textwidth}{!}{%
\begin{tabular}{c|c|c|ll|l|l}
\hline
\multirow{2}{*}{Model} &
\multirow{2}{*}{CR} &
\multirow{2}{*}{Input} &
\multicolumn{2}{c|}{Accuracy (\%)} &
FLOPs &
Params\\[0.2em]
&
&
&
top-1 & top-5 &
(G) &
(M) \\[0.3em]
\hline

\multirow{2}{*}{r3d-50 \cite{stergiou2020learning}} &
\ding{55} &
\multirow{8}{*}{$16 \! \times \! 112^{2}$} &
63.6 & 84.5 &
$52.6$ &
36.7\\[0.3em]

&
\ding{51} &
&
\textbf{64.2} \textcolor{applegreen}{(+0.6)} & \textbf{85.1} \textcolor{applegreen}{(+0.6)} &
$53.8$ \textcolor{red}{(+1.2)} &
37.6 \textcolor{red}{(+0.9)}\\[0.3em]
\cline{1-2}\cline{4-7}

\multirow{2}{*}{r(2+1)d-50 \cite{stergiou2020learning}} &
\ding{55} &
&
64.5 & 85.2 &
$83.3$ &
34.8\\[0.3em]

&
\ding{51} &
&
\textbf{65.2} \textcolor{applegreen}{(+0.7)} & \textbf{86.4} \textcolor{applegreen}{(+1.2)} &
$84.5$ \textcolor{red}{(+1.2)} &
35.7 \textcolor{red}{(+0.9)}\\[0.3em]
\cline{1-2}\cline{4-7}

\multirow{2}{*}{wide-r3d-50 \cite{stergiou2020learning}} &
\ding{55} &
&
64.0 & 85.4 &
$168.6 $ &
140.9\\[0.3em]

&
\ding{51} &
&
\textbf{65.3} \textcolor{applegreen}{(+1.3)} & \textbf{86.1} \textcolor{applegreen}{(+0.7)} &
$171.3$ \textcolor{red}{(+2.7)} &
143.4 \textcolor{red}{(+2.5)}\\[0.3em]
\cline{1-2}\cline{4-7}

\multirow{2}{*}{r3d-101 \cite{stergiou2020learning}} &
\ding{55} &
&
65.2 & 86.3 &
$78.0$ &
69.1\\[0.3em]

&
\ding{51} &
&
\textbf{67.7} \textcolor{applegreen}{(+2.5)} & \textbf{87.6} \textcolor{applegreen}{(+1.3)} &
$79.2$ \textcolor{red}{(+1.2)} &
70.0 \textcolor{red}{(+0.9)}\\[0.3em]
\end{tabular}
}
\label{tab:k400_nopre}
\vspace{-.2em}
\end{table}

\subsection{Statistical significance}
\label{ch:6::sec:experiments::mcnemar}

% Overview of previous section results
In the previous sections we have overviewed the accuracy rates achieved over large-scale datasets. Initial experiments included top-performing models with the inclusion of Class Regularisation in comparison to state-of-the-art architectures. In addition, we have demonstrated pair-wise results, within the same training setting, for architectures with and without Class Regularisation. In both cases, the effect of Class Regularisation to the accuracy rates has been visible with a uniform increase in classification performance across both tasks and tested architectures. However, this only provides a partial understanding of the overall improvements that the proposed method can provide as it remains unclear how statistically different the predictions made were to those of the original models. 

% Requirements for statistical significance
As the proposed method is an additional module that can be added to existing architectures, and does not perform any architectural changes at the network level, a reflection on the rates achieved is required. Therefore, there is an additional need for a testing measure in order to conclude if the resulting difference in performance does not indeed fall within the expected error margins. This can be expressed through a null hypothesis ($h_{0}$) such that the resulting accuracies are indeed similar. The statistical significance level ($\rho$) can be then defined as the probabilistic confidence that the null hypothesis ($h_{0}$) does hold true. Alternatively, based on a threshold value ($a$), probabilities smaller than the threshold value signify that the null hypothesis should be rejected as the homogeneity between results is small. This threshold value is commonly set to 0.05, as motivated by Fisher \cite{fisher1992statistical}, and Caparo \cite{craparo2007significance}. In our experiments we set the null hypothesis ($h_{0}$) as the achieved top-1 performance of a base model without Class Regularisation and the same model with Class Regularisation being statistically equivalent and within the margin of error.

% Introduction of McNemars test
As the hypothesis is based on pairs of model accuracies, we use a standard McNemar's significance test \cite{mcnemar1947note} to study the predictive accuracies of the models. The test highlights the proportions of examples that the two models disagree in terms of the predictions.

\begin{table}[ht]
\caption{\textbf{Contingency table} for the calculation of the $\chi^{2}$ statistic.}
    \centering
    \begin{tabular}{l l p{2.8em} | p{3em} p{2em}}
        & & \multicolumn{2}{c}{Model 1} & \multirow{2}{*}{Total}\\[.15em]
        & & Correct & Incorrect & \\[.15em] \cline{3-5}
        \multirow{2}{*}{\rotatebox{90}{Model 2}} & \multicolumn{1}{c|}{Correct} & \multicolumn{1}{c|}{a} & \multicolumn{1}{c|}{b} & \multicolumn{1}{c|}{a+b}\\ [.15em] \cline{2-5}
         & \multicolumn{1}{c|}{Incorrect} & \multicolumn{1}{c|}{c} & \multicolumn{1}{c|}{d} & \multicolumn{1}{c|}{c+d}\\ [.15em] \cline{3-5}
        \multicolumn{2}{c|}{Total} & \multicolumn{1}{c|}{a+c} & \multicolumn{1}{c|}{b+d} & \multicolumn{1}{c}{n} \\ [.15em] \cline{3-4}
       \end{tabular}
    \label{tab:mcnemar_example}
\end{table}

\begin{table}[ht!]
\caption{\textbf{McNemar's statistical significance test on K-400}.}
\label{table:mcnemar_cr_k400}
\begin{subtable}[h]{0.45\linewidth}
        \centering
        \begin{tabular}{l l p{2.8em} | p{3em} p{2em}}
        & & \multicolumn{2}{c}{Original} & \multirow{2}{*}{Total}\\[.15em]
        & & Correct & Incorrect & \\[.15em] \cline{3-5}
        \multirow{2}{*}{\rotatebox{90}{CR}} & \multicolumn{1}{c|}{Correct} & \multicolumn{1}{c|}{8112} & \multicolumn{1}{c|}{659} & \multicolumn{1}{c|}{8771}\\ [.15em] \cline{2-5}
         & \multicolumn{1}{c|}{Incorrect} & \multicolumn{1}{c|}{576} & \multicolumn{1}{c|}{4314} & \multicolumn{1}{c|}{4890}\\ [.15em] \cline{3-5}
        \multicolumn{2}{c|}{Total} & \multicolumn{1}{c|}{8688} & \multicolumn{1}{c|}{4973} & \multicolumn{1}{c}{13661} \\ [.15em] \cline{3-4}
       \end{tabular}
       \begin{flushleft}
            statistic ($\chi^{2}$) = 5.445\\
            p-Val ($\rho$) = 1.9e$^{-2}$
       \end{flushleft}
       \caption{ResNet50 3D}
       \label{tab:mcnemar_resnet50}
    \end{subtable}
    \hspace{1em}
    \begin{subtable}[h]{0.45\linewidth}
        \centering
        \begin{tabular}{l l p{2.8em} | p{3em} p{2em}}
        & & \multicolumn{2}{c}{Original} & \multirow{2}{*}{Total}\\[.15em]
        & & Correct & Incorrect & \\[.15em] \cline{3-5}
        \multirow{2}{*}{\rotatebox{90}{CR}} & \multicolumn{1}{c|}{Correct} & \multicolumn{1}{c|}{8253} & \multicolumn{1}{c|}{654} & \multicolumn{1}{c|}{8904}\\ [.15em] \cline{2-5}
         & \multicolumn{1}{c|}{Incorrect} & \multicolumn{1}{c|}{557} & \multicolumn{1}{c|}{4197} & \multicolumn{1}{c|}{4754}\\ [.15em] \cline{3-5}
        \multicolumn{2}{c|}{Total} & \multicolumn{1}{c|}{8810} & \multicolumn{1}{c|}{4851} & \multicolumn{1}{c}{13661} \\ [.15em] \cline{3-4}
       \end{tabular}
       \begin{flushleft}
            statistic ($\chi^{2}$) = 7.610\\
            p-Val ($\rho$) = 5.8e$^{-3}$
       \end{flushleft}
       \caption{ResNet50 (2+1)D}
       \label{tab:mcnemar_r2plus1d_50}
    \end{subtable}
    \\
    \begin{subtable}[h]{0.45\linewidth}
        \centering
        \begin{tabular}{l l p{2.8em} | p{3em} p{2em}}
        & & \multicolumn{2}{c}{Original} & \multirow{2}{*}{Total}\\[.15em]
        & & Correct & Incorrect & \\[.15em] \cline{3-5}
        \multirow{2}{*}{\rotatebox{90}{CR}} & \multicolumn{1}{c|}{Correct} & \multicolumn{1}{c|}{8216} & \multicolumn{1}{c|}{703} & \multicolumn{1}{c|}{8919}\\ [.15em] \cline{2-5}
         & \multicolumn{1}{c|}{Incorrect} & \multicolumn{1}{c|}{527} & \multicolumn{1}{c|}{4215} & \multicolumn{1}{c|}{4742}\\ [.15em] \cline{3-5}
        \multicolumn{2}{c|}{Total} & \multicolumn{1}{c|}{8743} & \multicolumn{1}{c|}{4918} & \multicolumn{1}{c}{13661} \\ [.15em] \cline{3-4}
       \end{tabular}
       \begin{flushleft}
            statistic ($\chi^{2}$) = 24.898\\
            p-Val ($\rho$) = 6.0e$^{-7}$
       \end{flushleft}
       \caption{wide-ResNet50 3D}
       \label{tab:mcnemar_wide_r3d_50}
    \end{subtable}
    \hspace{1em}
    \begin{subtable}[h]{0.45\linewidth}
        \centering
        \begin{tabular}{l l p{2.8em} | p{3em} p{2em}}
        & & \multicolumn{2}{c}{Original} & \multirow{2}{*}{Total}\\[.15em]
        & & Correct & Incorrect & \\[.15em] \cline{3-5}
        \multirow{2}{*}{\rotatebox{90}{CR}} & \multicolumn{1}{c|}{Correct} & \multicolumn{1}{c|}{8775} & \multicolumn{1}{c|}{473} & \multicolumn{1}{c|}{9248}\\ [.15em] \cline{2-5}
         & \multicolumn{1}{c|}{Incorrect} & \multicolumn{1}{c|}{134} & \multicolumn{1}{c|}{4279} & \multicolumn{1}{c|}{4413}\\ [.15em] \cline{3-5}
        \multicolumn{2}{c|}{Total} & \multicolumn{1}{c|}{8909} & \multicolumn{1}{c|}{4752} & \multicolumn{1}{c}{13661} \\ [.15em] \cline{3-4}
       \end{tabular}
       \begin{flushleft}
            statistic ($\chi^{2}$) = 188.211\\
            p-Val ($\rho$) = 7.8e$^{-43}$
       \end{flushleft}
       \caption{ResNet-101 3D}
       \label{tab:mcnemar_r3d101}
    \end{subtable}
    \\
    \begin{subtable}[h]{0.45\linewidth}
        \centering
        \begin{tabular}{l l p{2.8em} | p{3em} p{2em}}
        & & \multicolumn{2}{c}{Original} & \multirow{2}{*}{Total}\\[.15em]
        & & Correct & Incorrect & \\[.15em] \cline{3-5}
        \multirow{2}{*}{\rotatebox{90}{CR}} & \multicolumn{1}{c|}{Correct} & \multicolumn{1}{c|}{9516} & \multicolumn{1}{c|}{878} & \multicolumn{1}{c|}{10394}\\ [.15em] \cline{2-5}
         & \multicolumn{1}{c|}{Incorrect} & \multicolumn{1}{c|}{707} & \multicolumn{1}{c|}{2560} & \multicolumn{1}{c|}{3267}\\ [.15em] \cline{3-5}
        \multicolumn{2}{c|}{Total} & \multicolumn{1}{c|}{10223} & \multicolumn{1}{c|}{3438} & \multicolumn{1}{c}{13661} \\ [.15em] \cline{3-4}
       \end{tabular}
       \begin{flushleft}
            statistic ($\chi^{2}$) = 18.233\\
            p-Val ($\rho$) = 1.9e$^{-5}$
       \end{flushleft}
       \caption{MTNet$_S$}
       \label{tab:mcnemar_mtnets}
    \end{subtable}
    \hspace{1em}
    \begin{subtable}[h]{0.45\linewidth}
        \centering
        \begin{tabular}{l l p{2.8em} | p{3em} p{2em}}
        & & \multicolumn{2}{c}{Original} & \multirow{2}{*}{Total}\\[.15em]
        & & Correct & Incorrect & \\[.15em] \cline{3-5}
        \multirow{2}{*}{\rotatebox{90}{CR}} & \multicolumn{1}{c|}{Correct} & \multicolumn{1}{c|}{9931} & \multicolumn{1}{c|}{678} & \multicolumn{1}{c|}{10609}\\ [.15em] \cline{2-5}
         & \multicolumn{1}{c|}{Incorrect} & \multicolumn{1}{c|}{538} & \multicolumn{1}{c|}{2514} & \multicolumn{1}{c|}{3052}\\ [.15em] \cline{3-5}
        \multicolumn{2}{c|}{Total} & \multicolumn{1}{c|}{10469} & \multicolumn{1}{c|}{3192} & \multicolumn{1}{c}{13661} \\ [.15em] \cline{3-4}
       \end{tabular}
       \begin{flushleft}
            statistic ($\chi^{2}$) = 15.889\\
            p-Val ($\rho$) = 6.7e$^{-5}$
       \end{flushleft}
       \caption{MTNet$_M$}
       \label{tab:mcnemar_mtnetm}
    \end{subtable}
    \\
    \begin{subtable}[h]{\linewidth}
        \centering
        \begin{tabular}{l l p{2.8em} | p{3em} p{2em}}
        & & \multicolumn{2}{c}{Original} & \multirow{2}{*}{Total}\\[.15em]
        & & Correct & Incorrect & \\[.15em] \cline{3-5}
        \multirow{2}{*}{\rotatebox{90}{CR}} & \multicolumn{1}{c|}{Correct} & \multicolumn{1}{c|}{10176} & \multicolumn{1}{c|}{614} & \multicolumn{1}{c|}{10790}\\ [.15em] \cline{2-5}
         & \multicolumn{1}{c|}{Incorrect} & \multicolumn{1}{c|}{496} & \multicolumn{1}{c|}{2375} & \multicolumn{1}{c|}{2871}\\ [.15em] \cline{3-5}
        \multicolumn{2}{c|}{Total} & \multicolumn{1}{c|}{10672} & \multicolumn{1}{c|}{2989} & \multicolumn{1}{c}{13661} \\ [.15em] \cline{3-4}
       \end{tabular}
       \begin{flushleft}
           \hspace{11em} statistic ($\chi^{2}$) = 12.332\\
           \hspace{11em} p-Val ($\rho$) = 4.4e$^{-4}$
       \end{flushleft}
       \caption{MTNet$_L$}
       \label{tab:mcnemar_mtnetl}
    \end{subtable}
\vspace{-1em}
\end{table}

More specifically, considering \cref{tab:mcnemar_example}, the null hypothesis holds true if the marginal probabilities of $p_{a} + p_{b} \approx p_{a} + p_{c}$ and $p_{c} + p_{d} \approx p_{b} + p_{d}$. This corresponds to $h_{0} : p_{b} \approx p_{c}$.
The test statistic is then calculated as:

\begin{equation}
    \chi^{2} = \frac{(|b-c|-1)^{2}}{(b+c)}
\end{equation}

which uses the Edwards continuity correction \cite{edwards1948note} and has a Chi-squared distribution with a single degree of freedom. Based on the threshold value ($a$), we can compute the probability ($\rho$) that the resulting differences in accuracies are indeed due to $h_{0}$. Probabilities lower than the defined threshold ($a=0.05$) reject the null hypothesis of statistical similarity in the top-1 accuracy rates of the two models. 

% Kinetics 400 val
Our statistical significance experiments were preformed over the K-400 validation set and were based on the provided models from \Cref{tab:K400_accuracies_cr,tab:k400_nopre}. This demonstrates both settings with models trained from scratch shown in \Cref{tab:mcnemar_resnet50,tab:mcnemar_r2plus1d_50,tab:mcnemar_wide_r3d_50,tab:mcnemar_r3d101} and state-of-the-art models that have been pre-trained on HACS shown in \Cref{tab:mcnemar_mtnets,tab:mcnemar_mtnetm,tab:mcnemar_mtnetl}.

\textbf{From scratch models}. The architectures presented in \Cref{tab:k400_nopre} are further explored based on their statistical significance. The four ResNet variants show that the improvements achieved with the inclusion of Class Regularisation is not within the margin of statistical error. All of the probabilities of statistical homogeneity between the model accuracies are significantly below the Fisher threshold of $a=0.05$. We observe that based on the contingency \cref{tab:mcnemar_resnet50,tab:mcnemar_r2plus1d_50,tab:mcnemar_wide_r3d_50,tab:mcnemar_r3d101}, $\rho_{i} < 0.98, \forall i \in M$, where $M$ is the set of the four tested models trained from scratch. The largest probability margins from $a$ are also shown to correlate with either networks with larger number of channels per block, such as wide-ResNet-50 or an increased number of layers (ResNet-101). This follows the notion that there is dependency on both the feature complexity as well as the feature size as more complex or a larger number of features can be more discriminative for the recognition of a class. This therefore also impacts the application of Class Regularisation and meaningful feature and class feature correlations can be better explored in more complex spaces.

\textbf{Pre-trained models}. Statistical significance results, on models that have their weights initialised from pre-training on HACS, show similar probabilities to the ones trained with randomly initialised weights. Across the three MTNet variants, the improvements by including Class Regularisation are determined to not correspond to statistical error with a confidence $\gg 99\%$. This enforces the notion that increases in accuracy rates, in both training from scratch and pre-trained settings, are indeed because of CR.

\section{Discussion and conclusions}
\label{ch:6::sec:discussion}

% Problem overview
In CNNs, feature extraction takes the form of a sequential application of kernels over an input volume. Features are learned in a hierarchical order. Based on this hierarchy, only a small fraction of these features and cross-layer connections would correspond to a specific class. This leaves room for improvement as all of the extracted features are considered uniformly during the final class probability distribution. In this section we have introduced a method named Class Regularisation which can strengthen or weaken layer activations based on the batch of videos that are processed.

Although the regularisation of activations has been explored as a general calibration method for activation distributions in CNNs layers \cite{ioffe2015batch, wu2018group}, there is a lack of techniques that can utilise class-based information as part of the regularisation process. This relates to features being calibrated individually of its overall descriptive capabilities, in terms of a class. Our proposed Class Regularisation can discover the features that best address a specific class and incorporate this discovered association as part of the training process. Its general application is not bound by the architecture type and utilises information from previous training iterations. We use as inputs both the class weights extracted from the previous iterations, and the activation feature maps of the layer that is to be regularised. The method can discover the class to feature correspondence and apply it over the input activation map creating \textit{Class Regularised} activations.

We evaluate the proposed method over two large-scale datasets: HACS and Kinetics-400, and over two additional training environments. We show that, with the inclusion of Class Regularisation, further improvements on top-performing models can be achieved with minimal additional computations or parameters. These observations hold true for both models that have been trained with their weights being initialised with random values (from scratch), as well as for models previously trained on a different dataset and their weights being initialised from them. We additionally demonstrate that performance difference to the baseline models due to CR is statistically significant.

Additionally, Class Regularisation aids in improving the explainability of 3D-CNNs. Through the direct connections between classes and features, we can visualise the spatio-temporal features that correlate to specific classes and thus provide further insights in terms of the specific patterns that classes best relate to.

We believe that the direct application of Class Regularisation, regardless of the model used, alongside its minimum additional computations can provide significant benefits, both the accuracies and well as the understanding of spatio-temporal CNNs.

\clearpage
\newpage
\stopcontents[chapters]
    
    \author
{}
\title{Spatio-Temporal Feature Interpretation}

\maketitle
\label{ch7}

% Scope of the chapter
In this chapter we present explainable representations of spatio-temporal features extracted from 3-dimensional convolutions. We argue that the creation of a methodology to provide visual interpretations for spatio-temporal features can aid in the explanation of failure cases given an architecture. For example, visualisations can be used to understand cases of confusion between classes through the analysis of misclassified outputs. \footnote{The code for \textit{Saliency Tubes} $({\dagger})$ is available at \url{https://git.io/JmXIH}\\
and for \textit{Class Feature Pyramids} $(\ddagger)$ is available at  \url{https://git.io/fjDCW}}

\startcontents[chapters]
%\printcontents[chapters]{}{1}{\section*{\contentsname}}

\section{Introduction}
\label{ch:7::sec:intro}

% Previous chapters
% MTConv and SRTG
In the previous chapters, we have shown the creation of robust spatio-temporal descriptors through convolutional receptive fields of flexible temporal size. The proposed method achieved this with the introduction of a triple branch approach, addressing both local spatio-temporal features and prolonged patterns of extended durations. The computed features are then aligned based on their overall temporal importance of features in the third branch. We have detailed the principals of our feature alignment method utilising attention discovered with recurrent-based sub-networks over time. We have provided an overview of our gating mechanism based on soft-nearest neighbour and cyclic consistency distance to achieve information fusion between the temporally aligned and the originally discovered patterns. Through this gate, feature fusion is either enabled or disabled based on the features exhibiting cyclic consistency. Our extensive experimentation on both large-scale and smaller datasets has showed significant improvements over fixed-sized alternatives while also further reducing computation costs.

% Class Regularisation
Subsequently, we have presented a direction for improvement as the uniform extraction of features can be limiting for action and human interaction classes that are similar in terms of their overall motion features. The regularisation method that we have presented uses feature amplification over activations in order to incorporate class-based information within the extracted features. Relationships that form between specific features and classes during training can be further exploited in order to strengthen their association and increase intra class variance.

% What to address in this chapter
Throughout the experiments in previous chapters,  we have shown score-based quantitative improvements across different models. Although quantitative measures such as{\parfillskip0pt\par}

\begin{wrapfigure}{r}{10em}
%\begin{mdframed}
\vspace{-1em}
\centerline{%
    \includegraphics[width=5em]{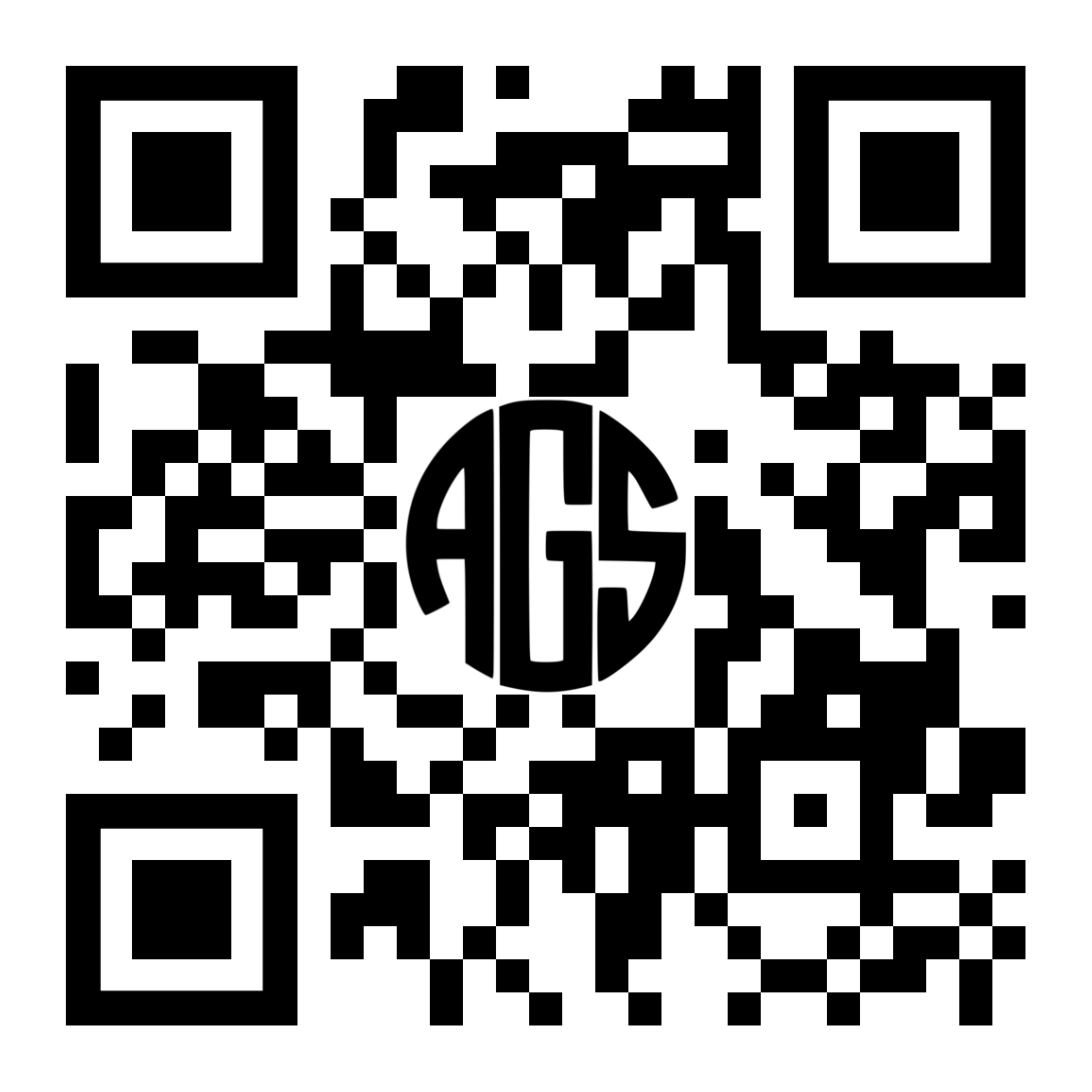}% 
    \includegraphics[width=5em]{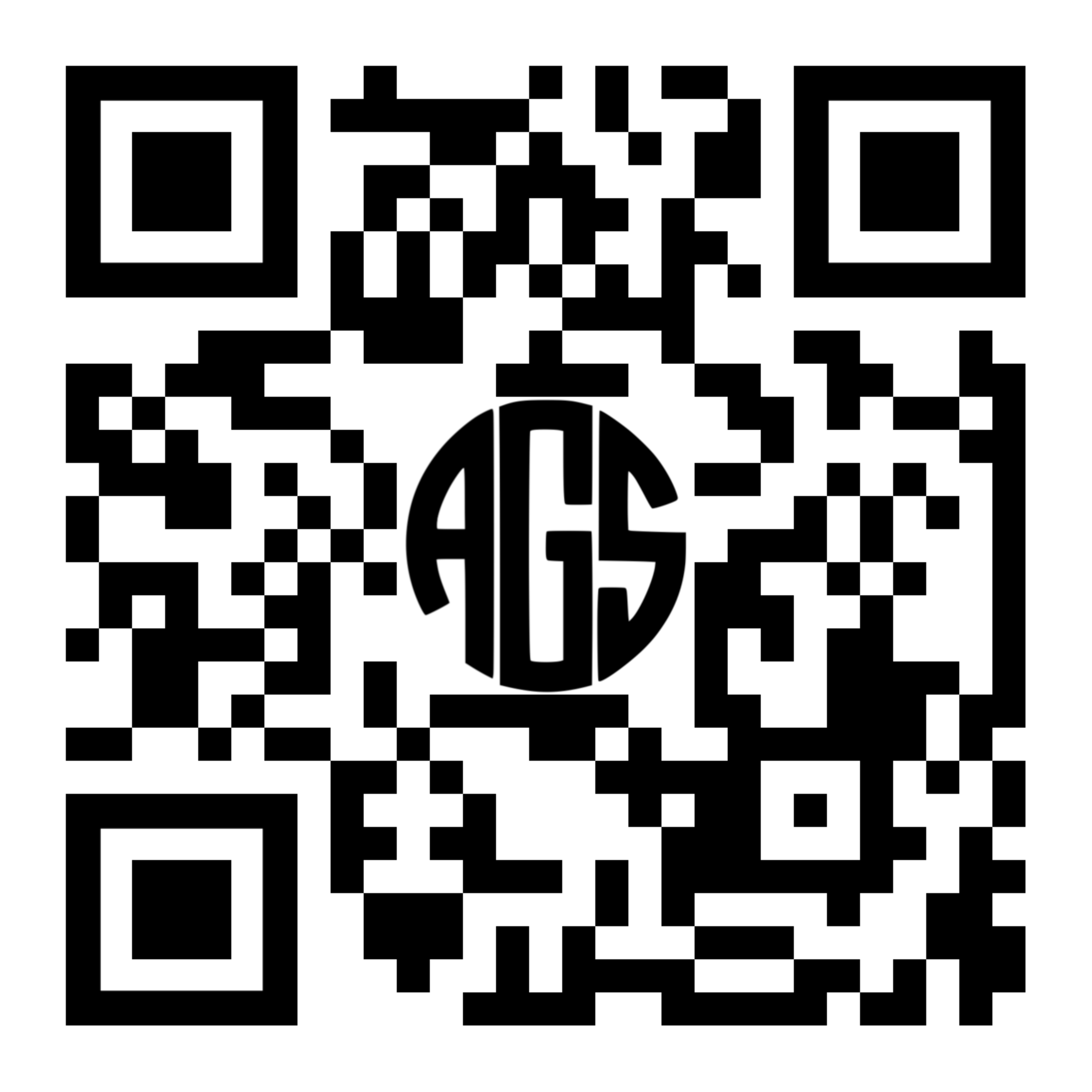}%
    }
\hspace{2em} \footnotesize{${\dagger}$} \hspace{5em} \footnotesize{${\ddagger}$}
\vspace{-5em}
%\end{mdframed}
\end{wrapfigure}
\noindent validation set accuracies can be indicative of the overall performance, they do not provide a qualitative measure to present the effects of each network's selected architectural aspects. The need for such a measure is further evident by the overall limited transparency of CNNs as models are typically regarded as \textit{black boxes}. 

\begin{figure}[ht]
\includegraphics[width=\textwidth]{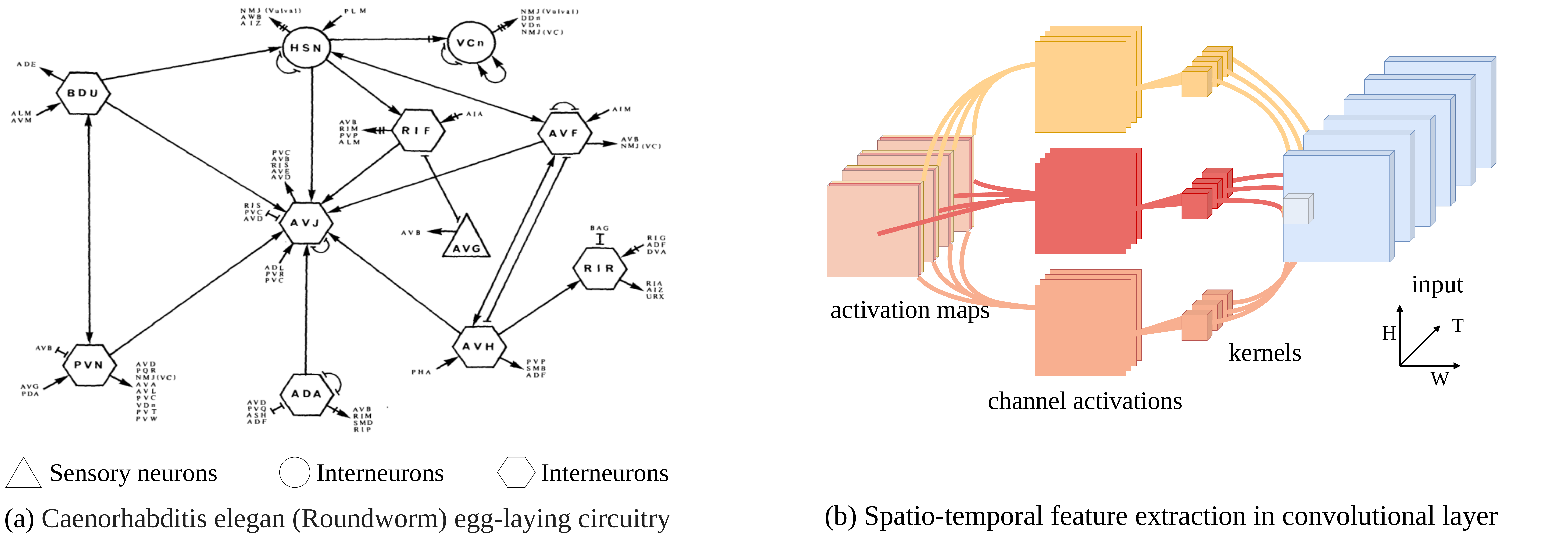}
\caption{\textbf{Examples of Biological neural circuits and convolutional feature extraction}. The biological neural circuit (a) represents round-worm's egg-laying process. The spatio-temporal convolution kernels (b) are responsible for the extraction of space-time features. Image (a) is sourced from \cite{white1986structure}.}
\label{fig:neural_circuits}
\end{figure}

% Interpretability biological and artificial neurons
Even though there is only a weak connection between current artificial neurons and biological equivalents, there are commonalities in terms of the methods that aim to understand them. A significant portion of current works in modern neuroscience studies biological neural connectomes mappings \cite{white1986structure}. A Connectome mapping can present information paths of the synaptic connections between different neurons and neural types in a visually comprehensible format. However, these representations are limited in a sense that they can only provide neuroscientists with a visual description of the synaptic connections between neurons. They do not provide any information about the magnitude of a synaptic connection nor if the connection excites or inhibits. This is in direct contrast to artificial neural networks where not only it is possible to discover their non-zero weights but also the produced feature activation values. We compare the information mapping in two examples of biological and artificial neural processes in \Cref{fig:neural_circuits}, where the neuronal circuit of a roundworm's egg-laying process and forward information pass in a spatio-temporal (3D) convolution are shown. Despite their obvious structural differences, a key aspect of artificial neurons in comparison to biological neurons, that benefits their overall explainability is their sequential forward-to-backward information flow. Inversely, gradient computations are calculated in a backward-to-forward manner. The chain rule of backpropagation provides a direct connection between classes and their associated features across different layers. 

% Saliency Tubes
Given the hierarchical connection between classes and their layer features, we propose two novel spatio-temporal visual attention methods. We first explore the visual interpretations in the form of the spatio-temporal clip regions that are found to be feature rich for a specific class. Our approach, named Saliency Tubes, can uncover the regions and frames that 3D CNNs rely on, class predictions by utilising the weights of a selected class and their produced feature mask that corresponds to the regional attention of the network. 

% Class Feature Pyramids
Based on this method of representing spatio-temporal attention, we further study hierarchically the dependencies of network features. In the extended method named Class Feature Pyramids, we use a cross-layer exploration method, termed \textit{back step}, which constructs an association between high-level features of layers in greater architectural depths and low-level features in earlier layers. Through back step the entire network can be traversed while capturing the causalities of feature activations between the discovered most informative features of the preceding and succeeding layers. The resulting few-to-many connections across adjacent network layers can be best described as a pyramid-like structure. 

% Chapter structure
In \Cref{ch:7::sec:spatial} we overview approaches in the literature for the visualisation of spatial patterns. Advances in spatio-temporal network interpretation methods are discussed in \Cref{ch:7::sec:spatio_temporal}. We overview the main methodology of Saliency Tubes in \Cref{ch:7::sec:saliencyTubes}. The extension to Class Feature Pyramids is then presented in \Cref{ch:7::sec:cfp}. We conclude in \Cref{ch:7::sec:discussion}.

\section{Spatial convolutional feature interpretations}
\label{ch:7::sec:spatial}

%In this section, we present spatial visualisation methods considering feature activations maps in \Cref{ch:7::sec:spatial::sub:activations}. In \Cref{ch:7::sec:spatial::sub:weights} we overview weight-optimising methods, while in \Cref{ch:7::sec:spatial::sub:attributions} visualisations that study the cross layer influence of neurons are discussed.

\begin{figure}[t]
\centering
     \begin{subfigure}[b]{0.23\textwidth}
         \centering
         \includegraphics[width=\textwidth]{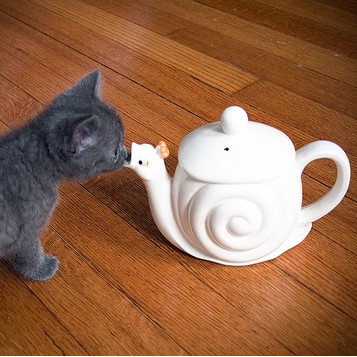}
         \caption{\textbf{Original}\\ \:}
         \label{fig:image_vis::original}
     \end{subfigure}
     \hfill
     \begin{subfigure}[b]{0.23\textwidth}
         \centering
         \includegraphics[width=\textwidth]{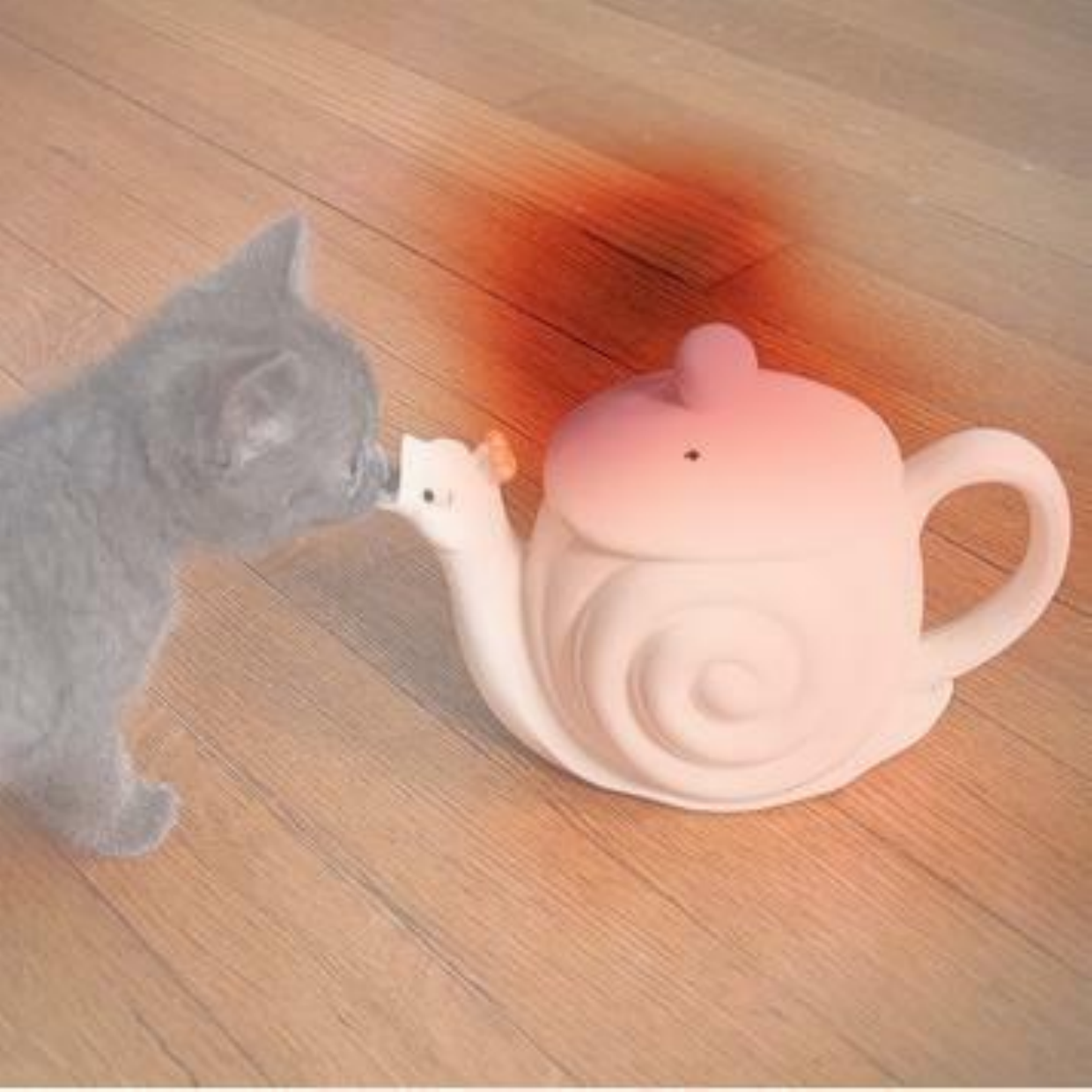}
         \caption{\textbf{CAM}\\ \cite{zhou2016learning}}
         \label{fig:image_vis::CAM}
     \end{subfigure}
     \hfill
     \begin{subfigure}[b]{0.23\textwidth}
         \centering
         \includegraphics[width=\textwidth]{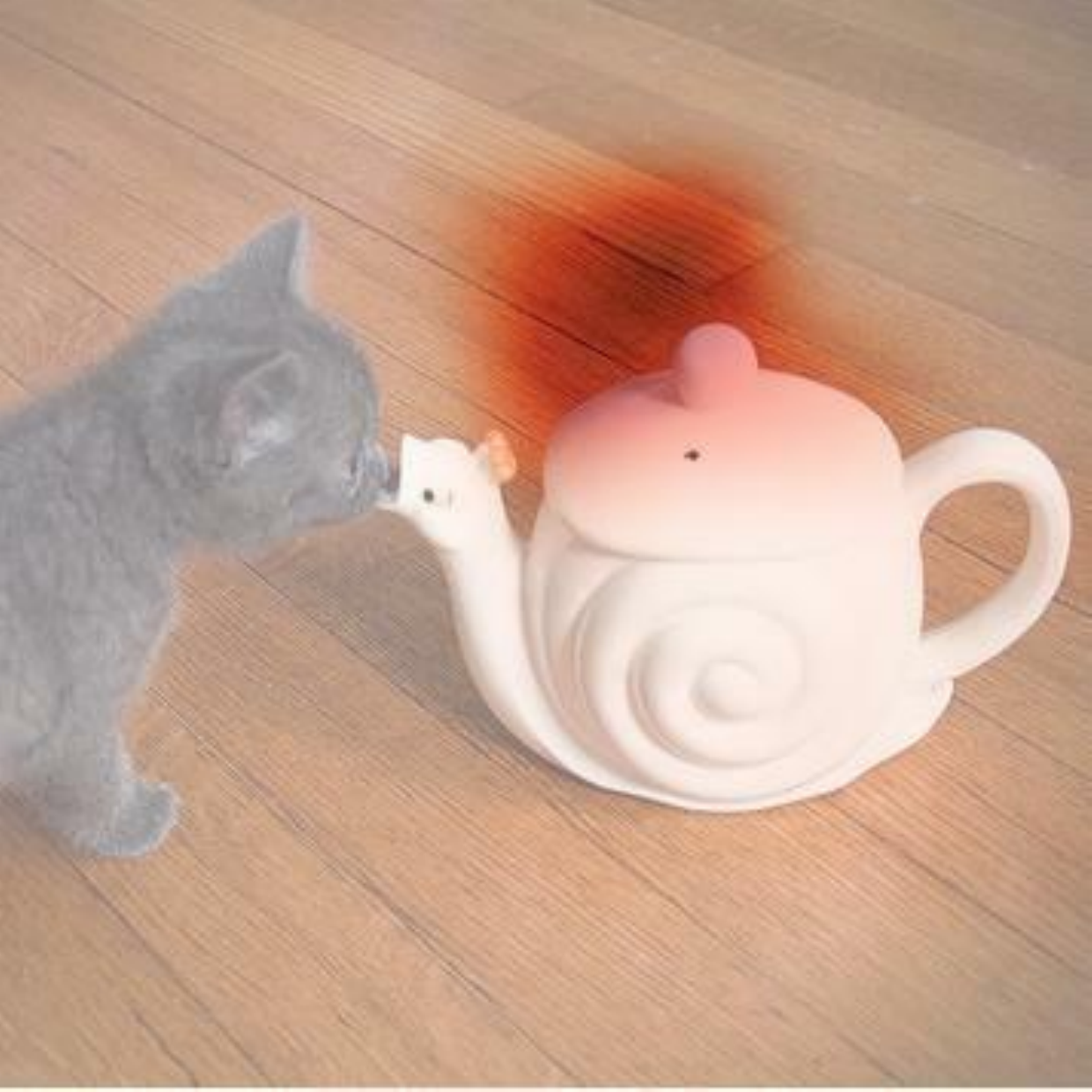}
         \caption{\textbf{Grad-CAM}\\ \cite{selvaraju2017grad}}
         \label{fig:image_vis::GradCAM}
     \end{subfigure}
     \hfill
     \begin{subfigure}[b]{0.23\textwidth}
         \centering
         \includegraphics[width=\textwidth]{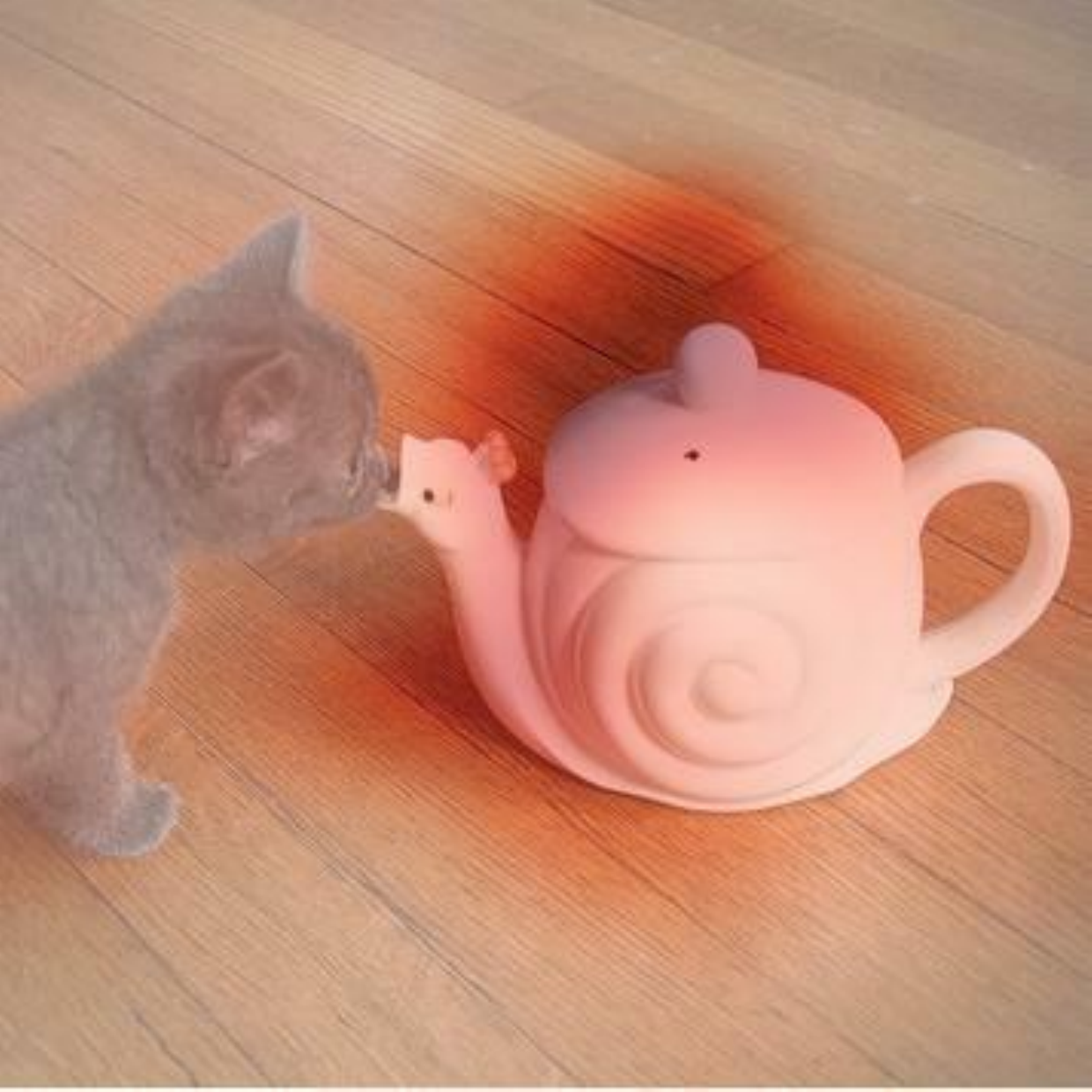}
         \caption{\textbf{Grad-CAM++}\\ \cite{chattopadhay2018grad}}
         \label{fig:image_vis::GradCAMPP}
     \end{subfigure}
     \\
     \begin{subfigure}[b]{0.23\textwidth}
         \centering
         \includegraphics[width=\textwidth]{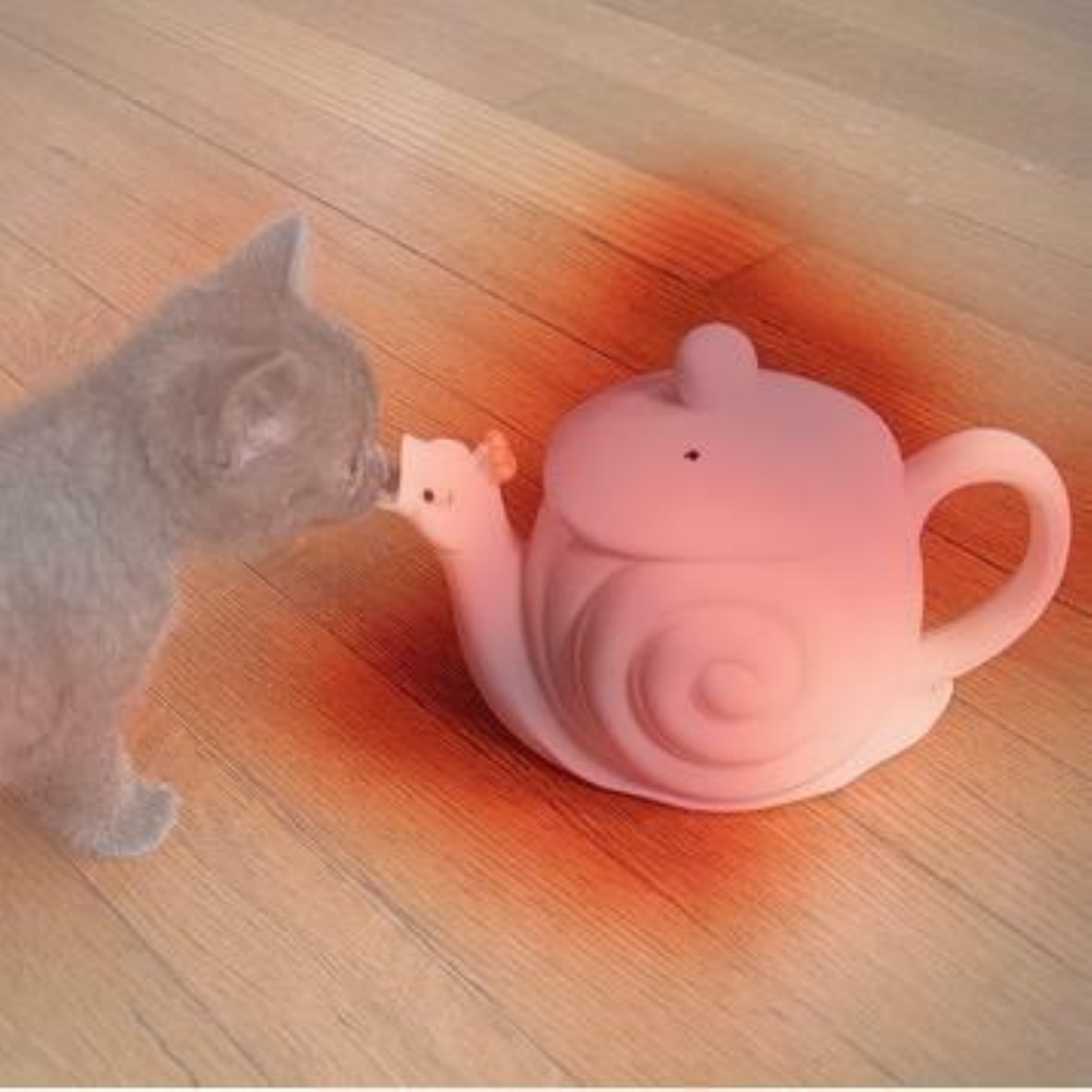}
         \caption{\textbf{ScoreCAM}\\ \cite{wang2020score}}
         \label{fig:image_vis::ScoreCAM}
     \end{subfigure}
     \hfill
     \begin{subfigure}[b]{0.23\textwidth}
         \centering
         \includegraphics[width=\textwidth]{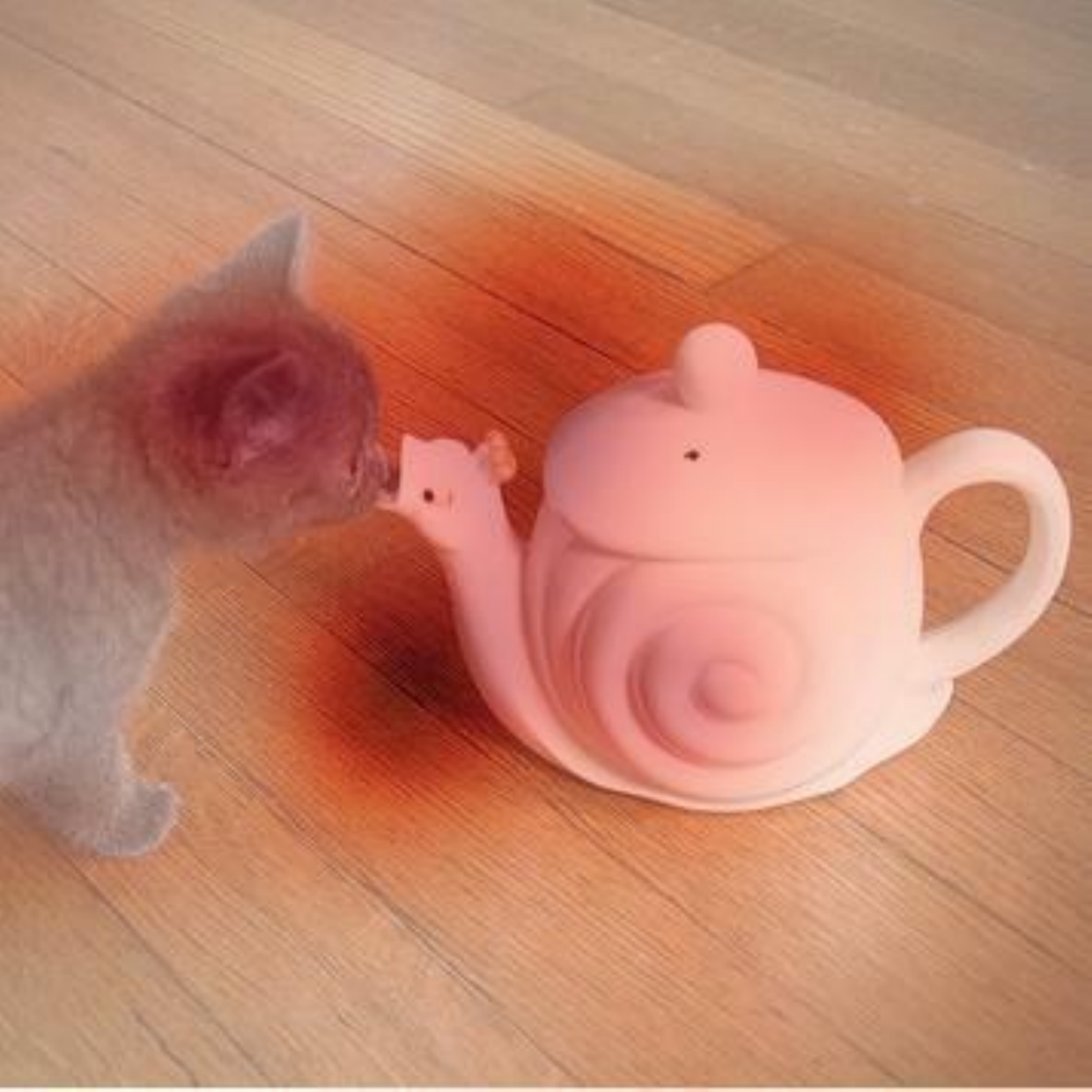}
         \caption{\textbf{AblationCAM}\\ \cite{ramaswamy2020ablation}}
         \label{fig:image_vis::ACAM}
     \end{subfigure}
     \hfill
     \begin{subfigure}[b]{0.23\textwidth}
         \centering
         \includegraphics[width=\textwidth]{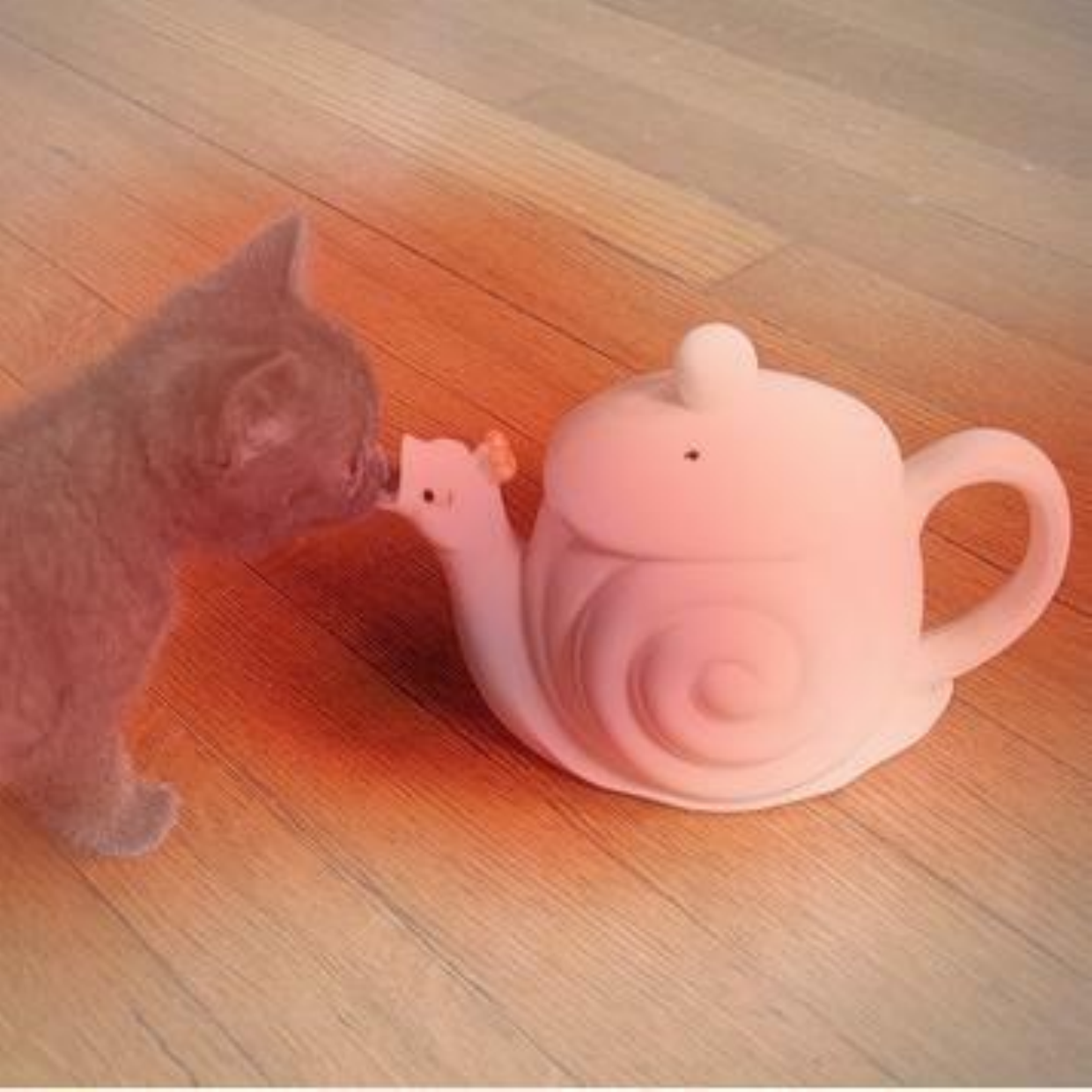}
         \caption{\textbf{ISCAM}\\ \cite{naidu2020cam}}
         \label{fig:image_vis::ISCAM}
     \end{subfigure}
     \hfill
     \begin{subfigure}[b]{0.23\textwidth}
         \centering
         \includegraphics[width=\textwidth]{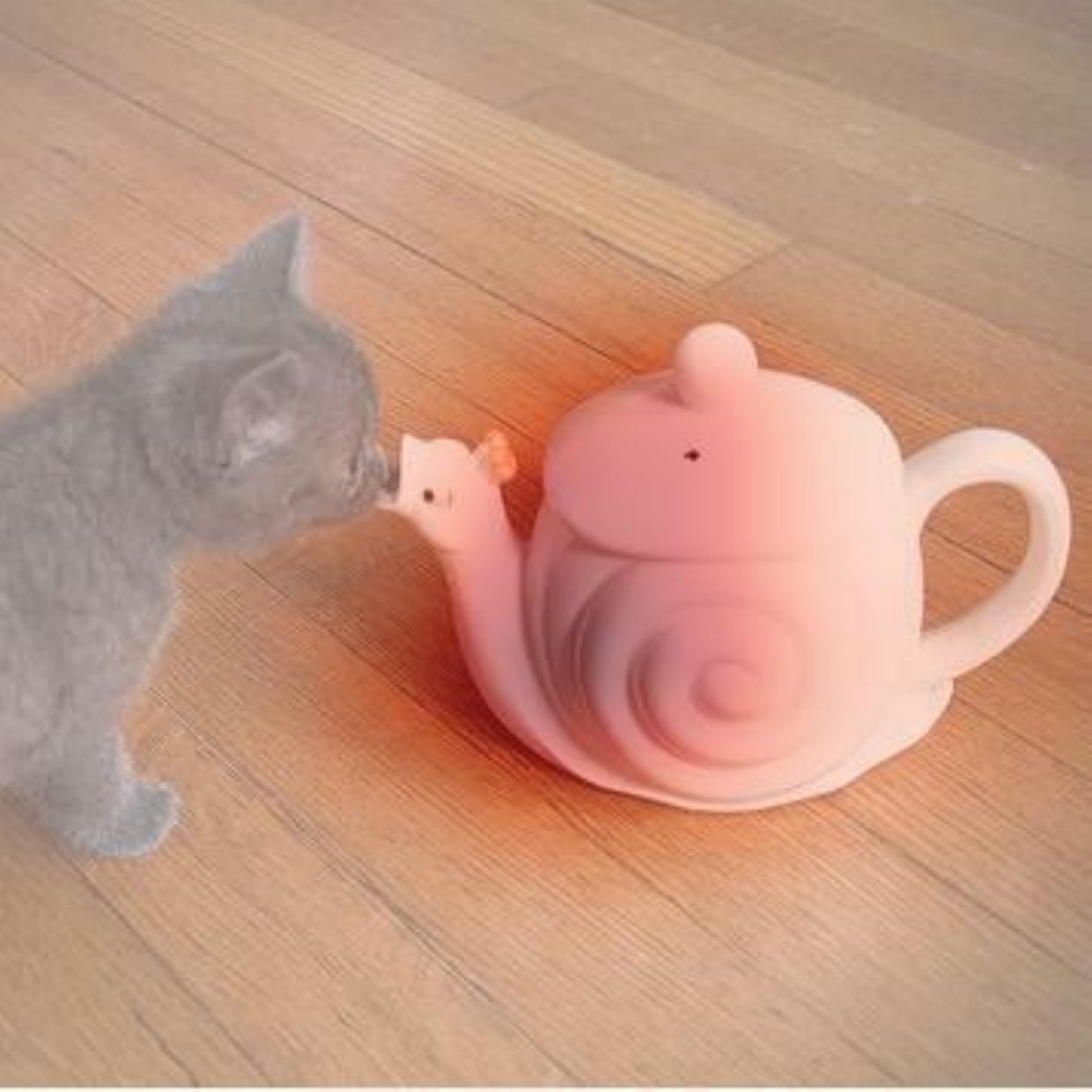}
         \caption{\textbf{SSCAM}\\ \cite{naidu2020ss}}
         \label{fig:image_vis::SSCAM}
     \end{subfigure}
     \\
     \begin{subfigure}[b]{0.23\textwidth}
         \centering
         \includegraphics[width=\textwidth]{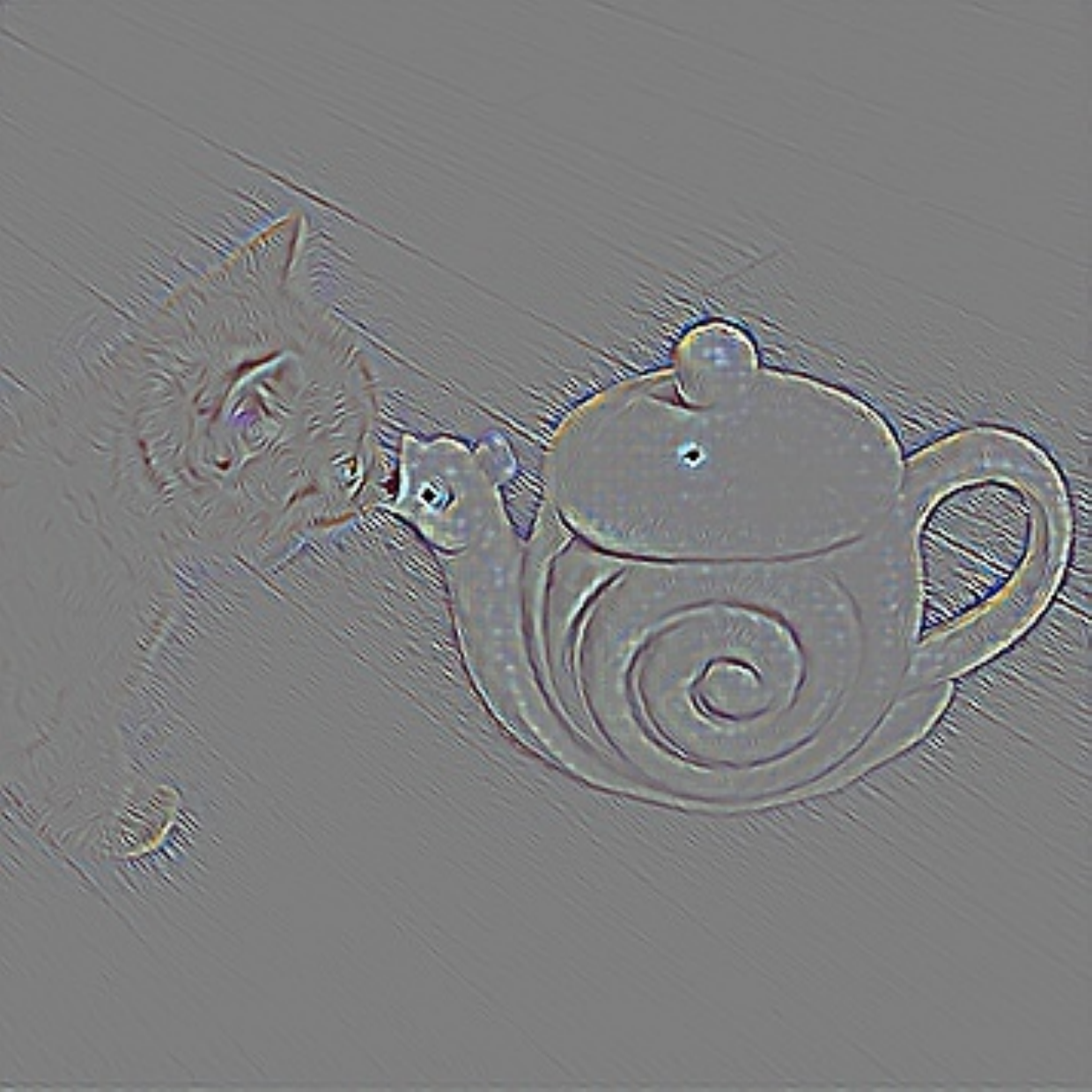}
         \caption{\textbf{GB}\\ \cite{springenberg2015striving}}
         \label{fig:image_vis::GB}
     \end{subfigure}
     \hfill
     \begin{subfigure}[b]{0.23\textwidth}
         \centering
         \includegraphics[width=\textwidth]{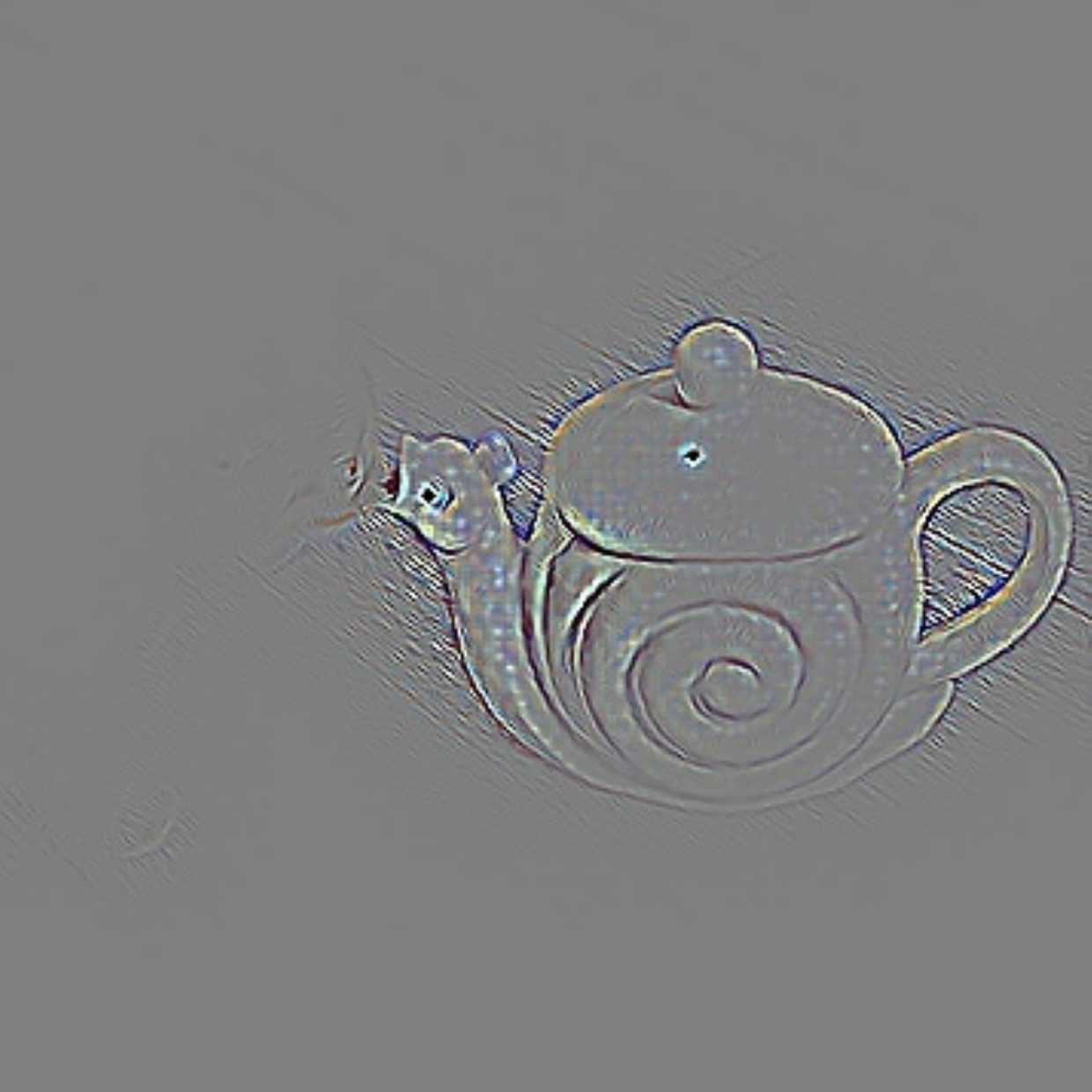}
         \caption{\textbf{GB-CAM}\\ \cite{selvaraju2017grad}}
         \label{fig:image_vis::GBCAM}
     \end{subfigure}
     \hfill
     \begin{subfigure}[b]{0.23\textwidth}
         \centering
         \includegraphics[width=\textwidth]{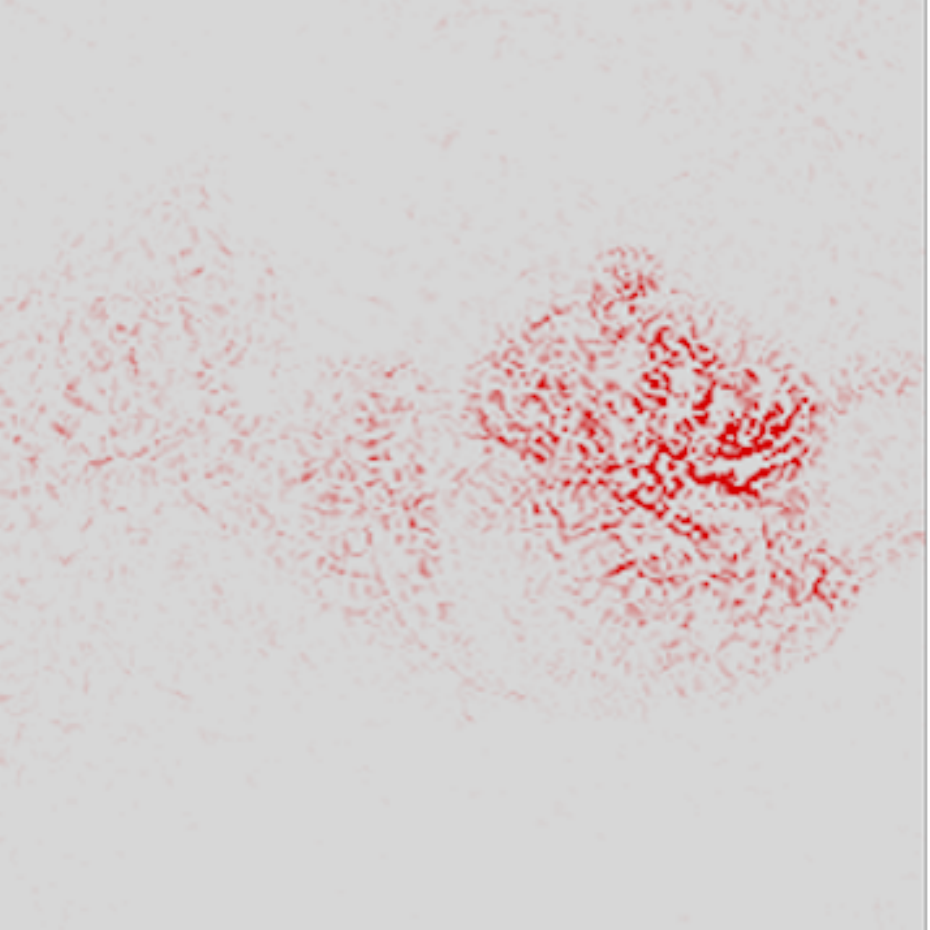}
         \caption{\textbf{IG}\\ \cite{sundararajan2017axiomatic}}
         \label{fig:image_vis::IG}
     \end{subfigure}
     \hfill
     \begin{subfigure}[b]{0.23\textwidth}
         \centering
         \includegraphics[width=\textwidth]{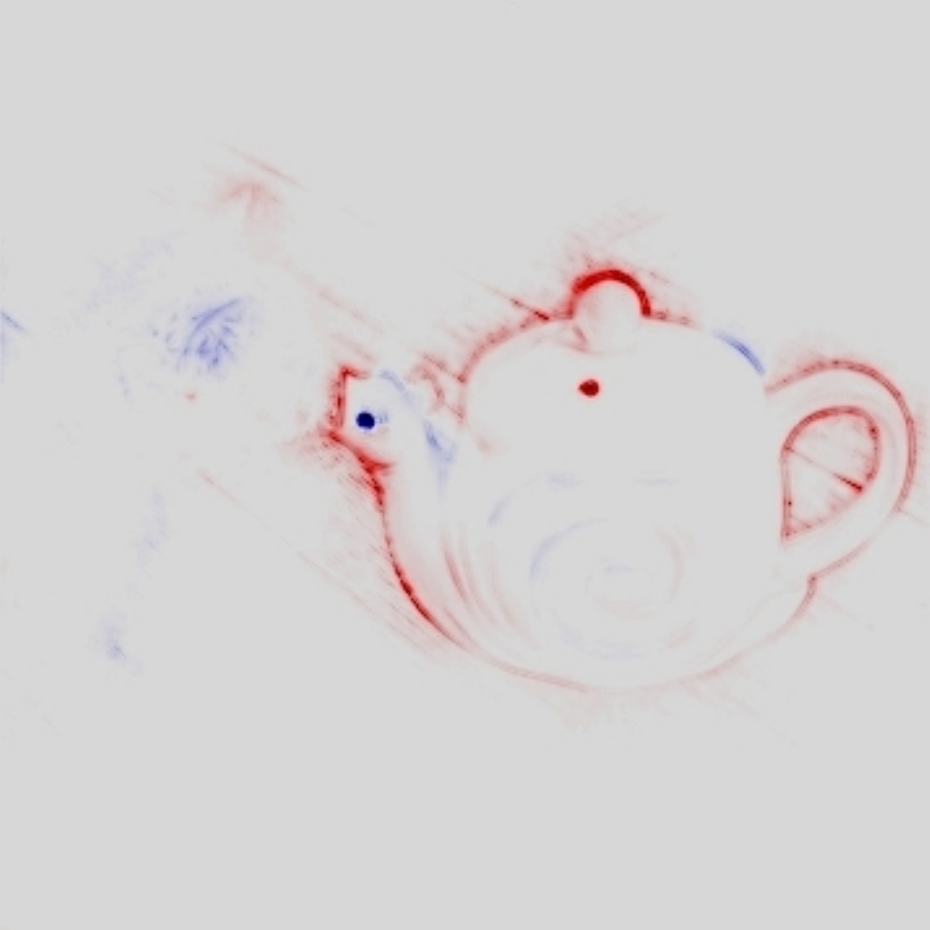}
         \caption{\textbf{LPR}\\ \cite{bach2015pixel}}
         \label{fig:image_vis::LPR}
     \end{subfigure}
     \\
     
\caption{\textbf{Feature visualisations}. Presented methods are considering activation-based visualisations (top two rows) and neural attribution methods (bottom row).}
\label{fig:image_vis}
\end{figure}

We focus on visual interpretation approaches for spatial data visualisations. Due to the high popularity of the image-based domain, spatial feature visualisation techniques have been explored in greater depth than techniques for spatio-temporal data. We therefore first overview spatial feature explanation methods. Based on the visualisation tasks, we divide them into three categories. The first set of methods is based on the utilisation of feature activations aimed at providing descriptions for the causes resulting in a certain activation. The second set is for the creation of visualisations that correspond to features based on layer and class weights. The resulting explanations demonstrate the characteristics and features that specific layer weights correspond to. The last set of methods is based on neural attribution. Attributions aim at exploring the influence between neurons of adjacent layers. Examples of activation-based and neural attribution methods are presented in \Cref{fig:image_vis}.

% activations

\subsection{Activation-based visualisations}
\label{ch:7::sec:spatial::sub:activations}

% CAM methods
One of the earliest activation-based methods, \textit{class activation maps (CAM)} \cite{zhou2016learning}, was based on localised salient regions in CNN inputs through the combination of feature regions that correspond to features of a specific class. Approaches have further been based on the creation of masks in order to provide a visual representation of a model's class focus \cite{fong2017interpretable}, while other mask-based methods such as \textit{RISE} \cite{petsiukrise2018rise} use randomly masked versions of inputs to discover an activation correspondence. Class Activation Maps have been generalised with \textit{GradCAM} \cite{selvaraju2017grad} which instead uses convolutional features that relate to a specific class. This creates a salient mask over the input image. \textit{GradCAM++} \cite{chattopadhay2018grad} introduced a pixel-wise weighting based on spatial positions in the final convolution maps. \textit{Ablation-CAM} \cite{ramaswamy2020ablation} uses a similar weighting procedure for feature maps that is based on feature map importance with respect to a specific class. Supplementary approaches have also been used to obtain feature maps during the forward information pass of the original inputs masked by feature activation maps \cite{wang2020score}. This results in a score-based weighting of the activation maps combined with the input images (Score-CAM).

% Network dissection methods
Other approaches have considered a \textit{network dissection} \cite{bau2017network} in which a measure is created between the alignment of neural units though their produced activations and concepts of objects. This is done with the use of the neural activations as a segmentation mask, measured over an intersection-over-union score for the discovered regions. Further works have also considered the decomposition of classes in terms of their features alongside their respected regions \cite{zhou2018interpretable}. Similarly, visual attention measures have also led to works that include dual modalities \cite{park2018multimodal} with the incorporation of textual justifications for the choices made based on the discovered features. Works have also focused on the creation of supplementary learnable models through sub-modular optimisation with \textit{Local Interpretable Model-agnostic Explanations (LIME)} \cite{ribeiro2016should}. Zintgraf \etal~\cite{zintgraf2017visualizing} have visualised both positive and negative feature correlations between image regions and classes.

\subsection{Extracted weight feature representation}
\label{ch:7::sec:spatial::sub:weights}

\begin{figure}[ht]
\includegraphics[width=\textwidth]{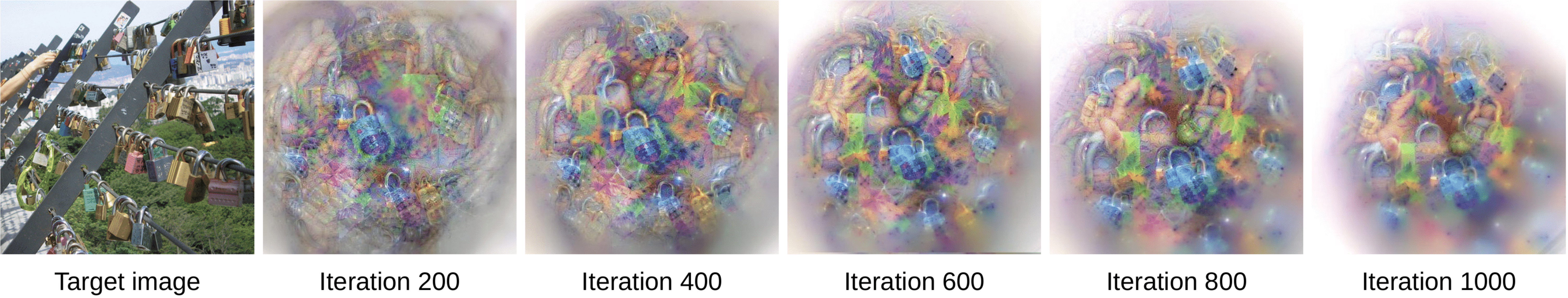}
\caption{\textbf{Top-30 feature visualisations through weight-excitation}.`layer6.12.conv3' from wide-ResNet-101 \cite{zagoruyko2016wide} is used to create the visualisations. Image sourced from \cite{stergiou2021mind}.}
\label{fig:am_ca}
\end{figure}

% Activation maximisation
Weight visualisations can provide insights in terms of the type of features that a CNN extracts and associates with a specific class or object category. A popularised approach considers the parametrisation of inputs and subsequently their updates, in order to maximise a specific neuron that corresponds to a class \cite{erhan2009visualizing}. Although this creates a visual correspondence between classes and their respective features, the representations can often be less intuitive as they highly depend on the values used during initialisation. Zeiler and Fergus \cite{zeiler2014visualizing} proposed a de-convolution approach to address this problem. Their approach aimed towards the approximation of features. Simonyan \etal~further explored maximising activations as a visualisation technique where the parameterised image optimised by capturing class features is learned in a supervised manner. Works of Nguyen \etal~\cite{nguyen2015deep} have shown how convolutional features can demonstrate high correspondence to unrealistic visual features based on the vast space of possible images and patterns that are similar to conventionally extracted patterns. Part of this phenomenon is based on \textit{feature entanglement} \cite{olah2018building} as convolutional features may not correspond to singular semantic concepts which can be visually understood. Recent work addresses the issues presented by \textit{feature entanglement} through the inclusion of Gaussian filters \cite{yosinskiunderstanding,wang2018visualizing}, jitter effect \cite{mordvintsev2015inceptionism} and creating centre-biased gradients \cite{nguyen2016multifaceted}. Other recent works have also been based on the creation of a multi-objective task \cite{stergiou2021mind} of activation maximisation and activation distance minimisation, (example shown in \Cref{fig:am_ca}) in order to reduce the search space or representations.

% GANs for feature explainability
To improve the realism in images, some approaches have considered the use of adversarial networks to synthesise visual features \cite{baumgartner2018visual}. Based on the approach of utilising adversarial training, Nguyen \etal~\cite{nguyen2016synthesizing} proposed a deep image generator network (DGN) without priors of hidden distributions as in generative models \cite{goodfellow2014generative,kingma2013auto}. Although there are improvements in the visual quality of visualisations of generative models, representations produced by generator models lack learned feature causalities \cite{holzinger2019causability} as they introduce an additional ambiguity step with the inclusion of the generator sub-network.

\subsection{Neural attributions}
\label{ch:7::sec:spatial::sub:attributions}

In order to discover the influence of neurons across layers, works of Springenberg \etal~\cite{springenberg2015striving} studied how additional signals from higher layers can be added within the propagation process. The resulting \textit{guided backpropagation} method includes zeroing-out negative gradients and thus creates a mask for the regional features that only display a positive influence. The visualisation of relevant information was also performed by decomposing the resulting class feature vectors into pixel relevance scores that reflect each pixel's contribution to the final class prediction with \textit{Layer-wise Relevance propagation (LRP)} \cite{bach2015pixel}. This approach has been extended to include non-linearities. Such non-linearities include normalisation, which provides additional difficulties in terms of standardising a technique for the backpropagation of relevant information \cite{binder2016layer}. Similarly, Shrikumar \etal~\cite{shrikumar2017learning} proposed a reference state for neurons based on their inputs, creating scores between obtained and reference states. Other approaches considered attributions that include the creation of integrated gradients through the integral of gradients along the information path \cite{sundararajan2017axiomatic}.

\section{Spatio-temporal features}
\label{ch:7::sec:spatio_temporal}

% Early works
Despite the large number of visualisation methods for image-based features, there is only a small selection of methods that addresses the video domain. Early works of Karpathy \etal~\cite{karpathy2016visualizing} were aimed at the creation of visual representations for LSTM recurrent cells. Based on this, Bargal \etal~\cite{bargal2018excitation} have explored visual explanations in action recognition with models that incorporate 2D Convolutions and LSTMs, with their main method being an extension of \textit{Excitation Backpropagation} \cite{zhang2018top}. However, both works focus on the decisions of recurrent cells within the context of action recognition with 2D convolutions being used as per-frame feature extractors.

% 3D Conv visualisations
Visualisations of spatio-temporal features in action recognition have proven challenging. Early works by Chattopadhay \etal~\cite{chattopadhay2018grad} have extended class activation maps for object recognition. Further works also included extensions of Layer-wise Relevance Propagation (LRP) \cite{srinivasan2017interpretable} by indicating the spatio-temporal locations that are deemed more important for each video class. Similar works \cite{hiley2019discriminating} have also extended Deep Taylor Decomposition (DTD) \cite{montavon2017explaining}, showing positive and negative gradient relevance in inputs. Hiley \etal~\cite{hiley2020explaining} have also demonstrated relevance through DTD in a spatial-only and temporal-only context for 3D convolutions. Li \etal~\cite{li2021towards} have utilised Extremal Perturbations \cite{fong2019understanding, fong2017interpretable} which are based on a learned mask that specifies the region that maximises the class probability. Further works of Price and Damen \cite{price2020play} have used a frame-wise Shapley value \cite{shapley1953value} to determine the contribution of each frame to the final class prediction made by the model.

% Proposed methods
In our proposed methods we study spatio-temporal convolutions in a hierarchical manner. With the proposed \textit{Saliency Tubes} as a method tailored for the representation of spatio-temporal informative regions and the proposed \textit{back-stepping} technique in \textit{Class Feature Pyramids}, we can effectively traverse multiple network layers and generate a class dependency graph based on the most informative features.

\section{Saliency Tubes feature visualisation}
\label{ch:7::sec:saliencyTubes}

\begin{figure}[t]
\centering
\includegraphics[width=.9\textwidth]{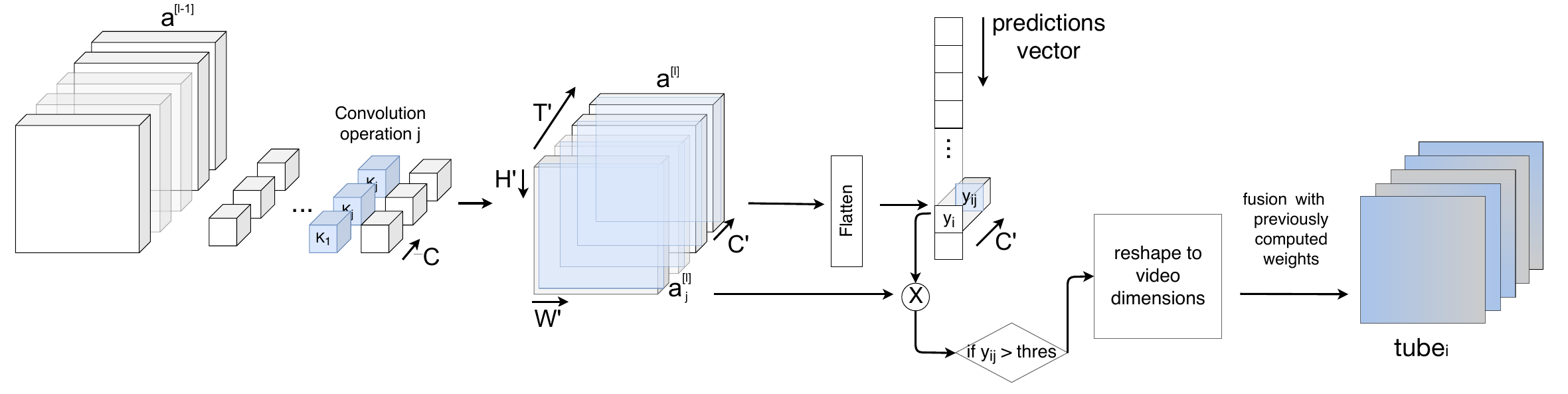}
\caption{\textbf{Saliency Tubes main process pipeline}. Feature importance of class $i$ is defined based on the corresponding class prediction values in feature vector $\textbf{y}_{i}$. Individual features are visualised through the fusion of the activation maps in the last step of the saliency tubes. The figure presents a simplified case for a single feature in the final convolution layer for easier interpretability.}
\label{fig:saliency_tubes_pipeline}
\vspace{-1em}
\end{figure}

% Class Saliency formulation
\Cref{fig:saliency_tubes_pipeline} outlines the proposed approach for spatio-temporal salient region localisation. We denote activation maps with $a$, while layers are indexed by $[l]$. The activations ($\textbf{a}^{[l]}$) for the final convolution layer ($[l]$) include activation maps of size $C' \! \times \! \textbf{R}$, where $C'$ is the number of corresponding channels, and \textbf{R} is the spatio-temporal region of $T'$ temporal extent, height $H'$ and width $W'$ (with size |$\textbf{R}| = T' \! \times \! H' \! \times \! W'$). We note that $C'$ is also based on the total number of convolutions that are performed in layer $l$. The tensor in the final fully-connected layer that is responsible for class predictions is denoted by $\textbf{y}$. Each class ($i \in \{0,..,N\}$) from the total $N$ classes can then be selected as $\textbf{y}_{i}$. Each channel ($j \in \{0,...,C'\}$) in a class's ($i$) prediction vector is formulated as $\textbf{y}_{i,j}$ and relates to a specific feature of the network's activations map $\textbf{a}^{[l]}$ and designates how informative that specific activation map is towards a correct prediction for an example of class $i$. To relate the importance of feature $i$ to spatio-temporal input regions, we propagate back to activation maps ($\textbf{a}^{[l]}_{j \times T \times H \times W}$) and multiply ($\otimes$) all of the map elements by the equivalent prediction weight vector $\textbf{y}_{i,j}$. The normalised class weighted operation is the formulated as $\textbf{z}_{i \rightarrow j}$:

\begin{equation}
\label{eq:saliency}
\textbf{z}_{i \rightarrow j} = \frac{f(\textbf{a}_{j}) - \min\limits_{\textbf{a}_{j} \in \textbf{R}} f(\textbf{a}_{j}) }{ \max\limits_{\textbf{a}_{j} \in \textbf{R}} f(\textbf{a}_{j}) - \min\limits_{\textbf{a}_{j} \in \textbf{R}} f(\textbf{a}_{j}) } \quad where \; f(\textbf{a}_{j}) = \textbf{y}_{i,j} \otimes \textbf{a}_{j}
\end{equation}

% Threshold
Considering the large number of features ($C'$) and in order to limit the effect of low-information regions, we introduce a threshold argument $\tau$. Based on it, only significantly contributing activations are selected. Values that fall below that threshold are defined as parts of set $\textbf{E}$. We thus consider that the activation weight operation ($f(a_{j})$) for channel $j$ also includes the condition that $f(\textbf{a}_{j}) = \textbf{y}_{i,j} \otimes \textbf{a}_{j} \: \forall j \in \{0,...,C'\} \notin \textbf{E}$. 

\begin{figure}[ht]
\centering
\includegraphics[width=.75\textwidth]{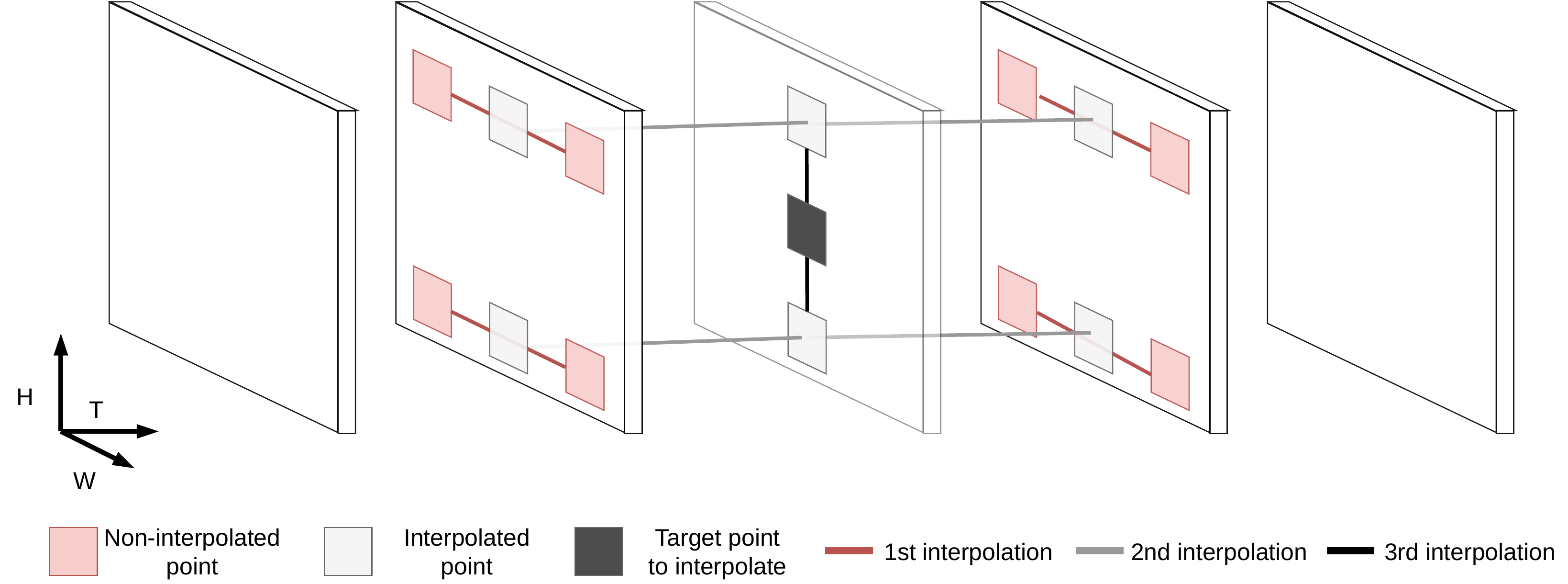}
\caption{\textbf{Spatio-temporal interpolation}. Original points are denoted with red and interpolated points are shown as light and dark grey.}
\label{fig:3d_interpolation}
\end{figure}

% Interpolation
As the produced normalised class-weighted activations ($\textbf{z}_{i \rightarrow j}$) are of spatio-temporal size $T' \! \times \! H' \! \times \! W'$, we use third-order spline interpolation in order to approximate the activation mask in an extended size of $T \! \times \! H \! \times \! W$ to fit the input. Formally, we define each 3-dimensional point of the class activations $z_{i \rightarrow j, t, h, w}$ as a point (\enquote{knot}) for a non-continuous ($g$) function. Each point is included in the function with $C^{2}$ smoothness that has a continuously differentiable first derivative. We visualise how interpolation is performed over space and time in \Cref{fig:3d_interpolation}, where original/non-interpolated points are denoted with red. The interpolation process is first done spatially with the approximation of new points (light grey between red boxes) and the subsequent increase of the spatial dimensionality. The second set of interpolation operations is performed in order to discover points across time based on the knots from the first interpolation (grey points in the middle frame). The final target point is then discovered by the third interpolation sequence.

\begin{figure}[h]
\centering
\includegraphics[width=\textwidth]{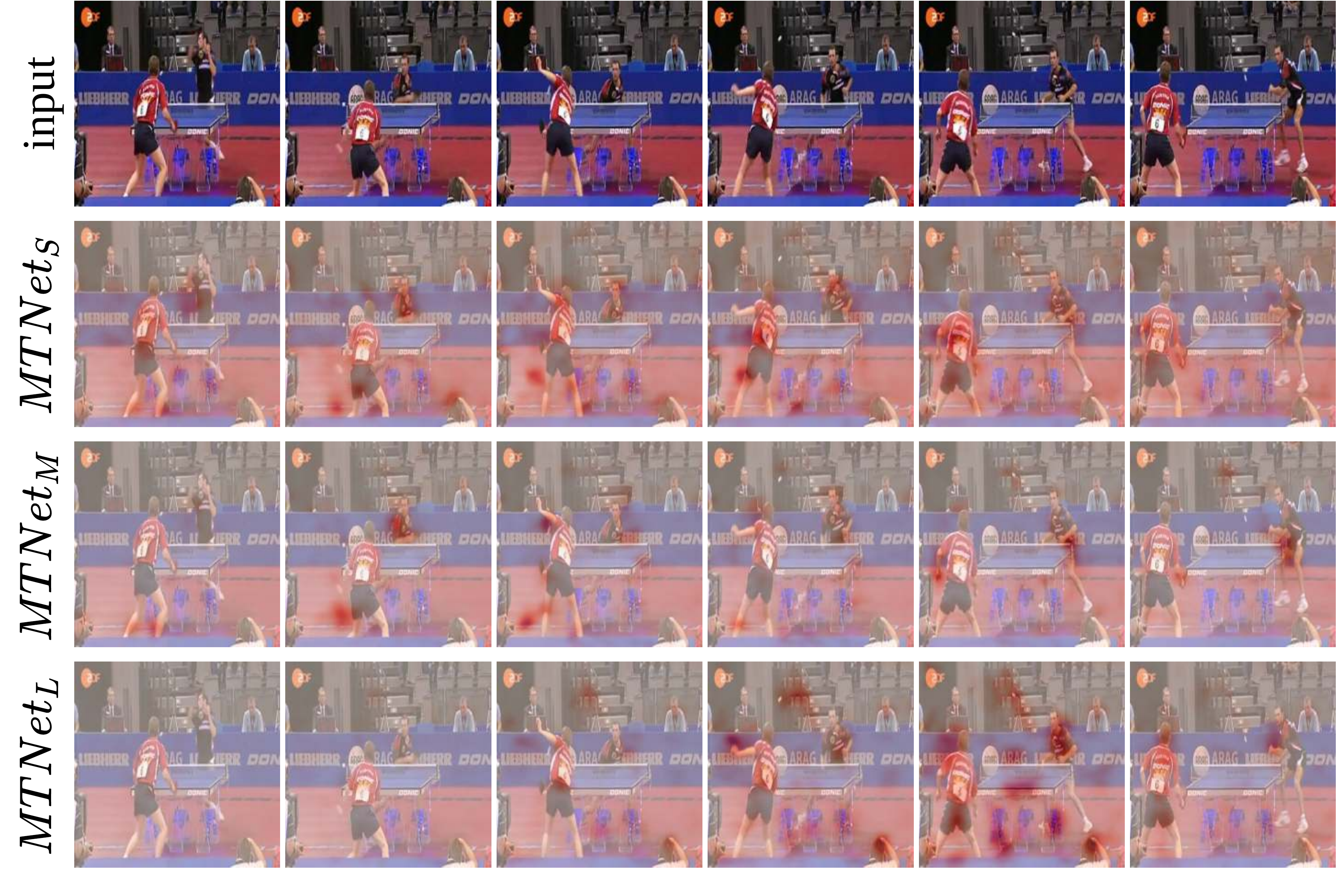}
\includegraphics[width=\textwidth]{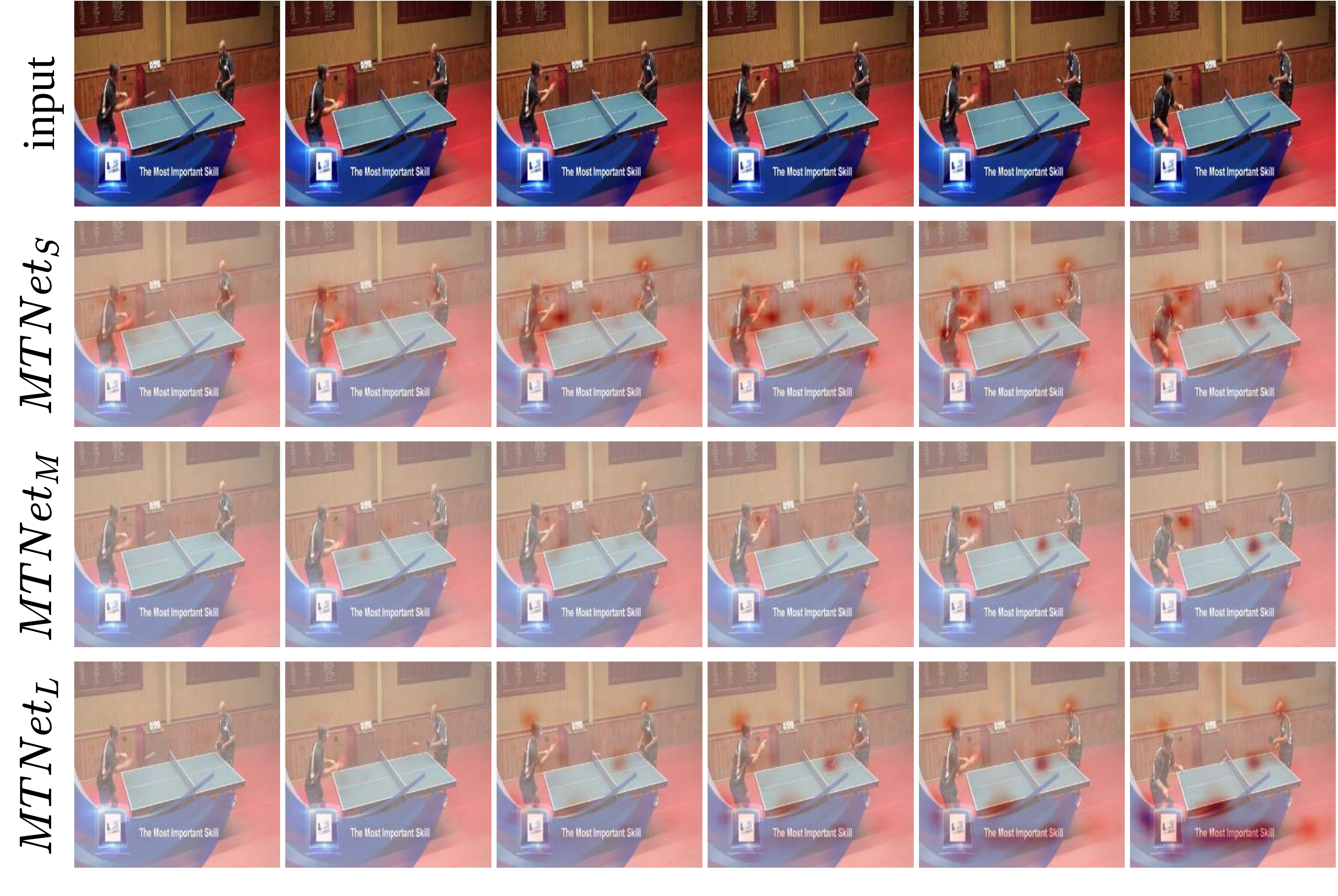}
\caption{\textbf{Saliency Tubes on MTNets \cite{stergiou2020learning} for HACS `table tennis' class}. Both examples have been randomly sampled from HACS.}
\label{fig:saliency_tubes_tabletennis}
\end{figure}

\begin{figure}[h]
\includegraphics[width=\textwidth]{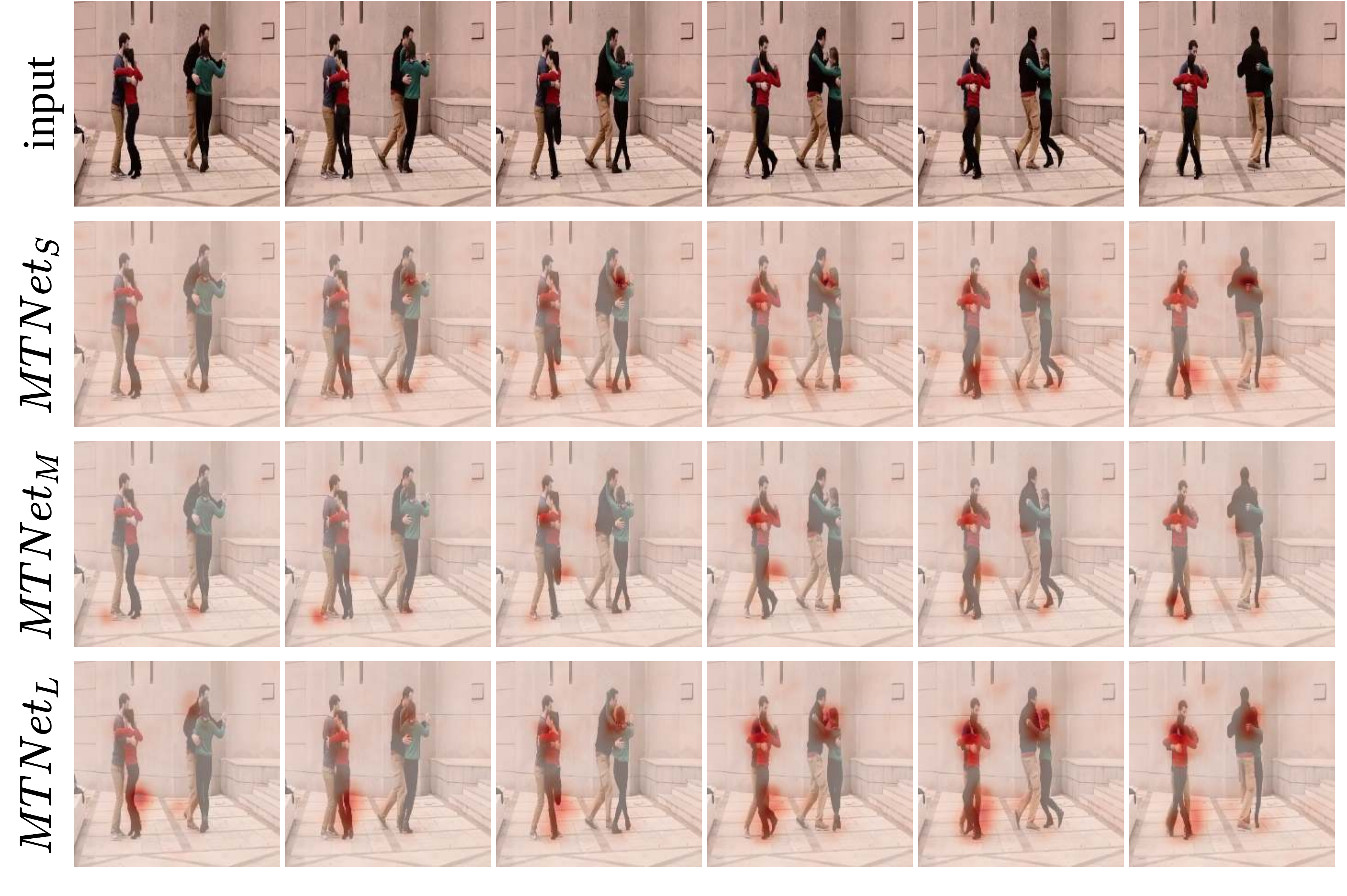}
\includegraphics[width=\textwidth]{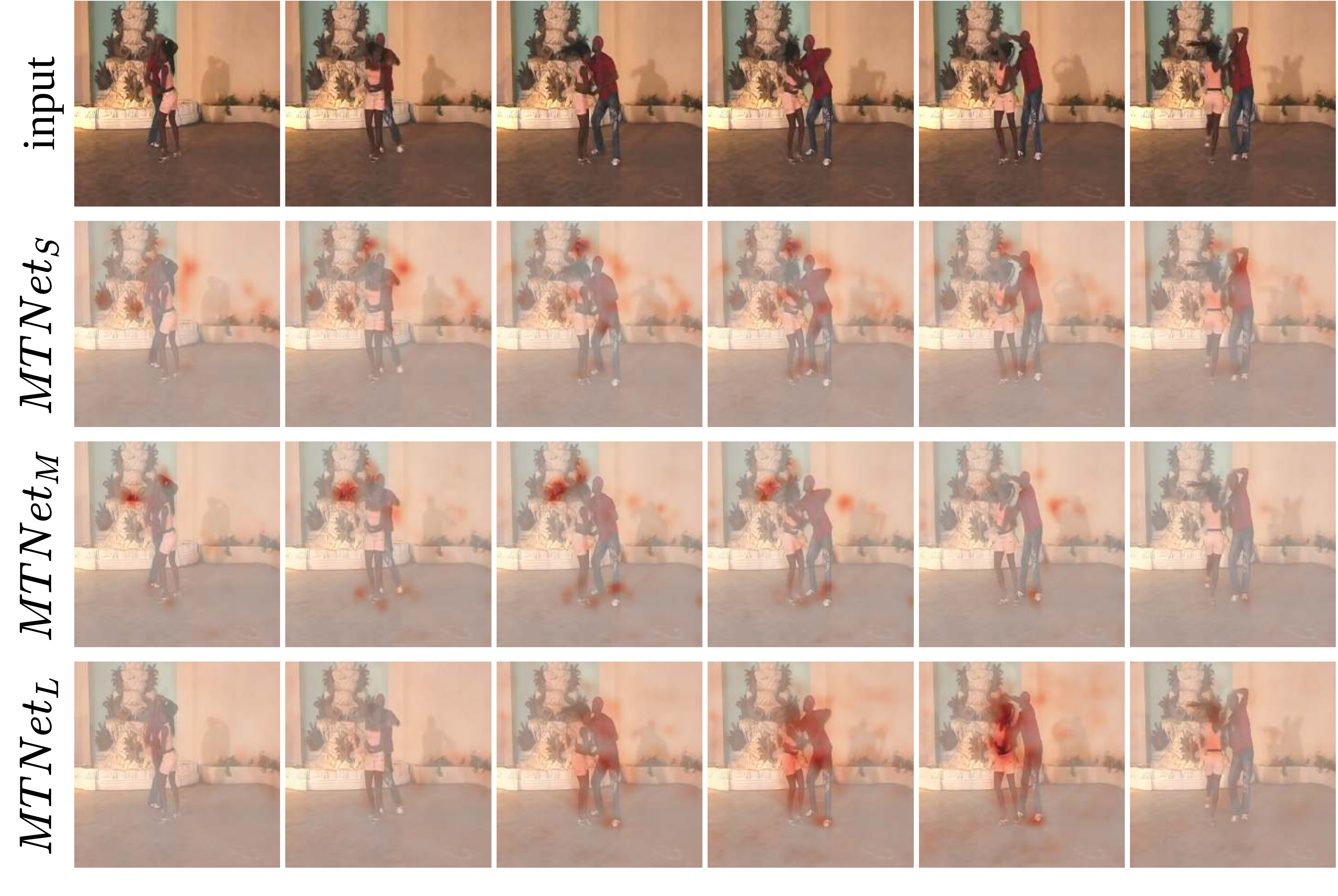}
\caption{\textbf{Saliency Tubes on MTNets \cite{stergiou2020learning} for HACS \textit{\enquote{tango}} class}. Both examples have been randomly sampled from HACS.}
\label{fig:saliency_tubes_tango}
\end{figure}

The final spatio-temporal attention volume, named Saliency Tubes, ($\textbf{tube}_{i} = \sum_{j \in C'} (z_{i \rightarrow j})^{2}$) is created by the channel-wise summation of the squared, normalised class-weighted activations ($\textbf{z}_{i \rightarrow j}$). We use the square of the class-weighted activations as a straightforward amplification of salient regions to improve visualisation quality. We present visual examples in \Cref{fig:saliency_tubes_tabletennis,fig:saliency_tubes_tango}.

\clearpage

\newpage

\newpage

\section{Class Feature Pyramids}
\label{ch:7::sec:cfp}

%with the main method formulation discussed in \Cref{ch:7::sec:cfp::formulation,ch:7::sec:cfp::layer_dependencies}. Feature-wise and layer-wise visualisations differences are considered in \Cref{ch:7::sec:cfp::layer_feat_association} and adaptations based on convolutional block structures are presented in \Cref{ch:7::sec:cfp::conv_layers}. In \Cref{ch:7::sec:cfp::inference,ch:7::sec:cfp::visualisations} examine the running time inferences alongside visualisations results.

% Limitations of Saliency Tubes
While Saliency Tubes provide a straightforward method for visualising class-informative feature spatio-temporal regions, they also include certain limitations. Their most notable restriction is their constraints to only providing visual interpretations for the final convolutional layer due to the fact that the channel dimensionality directly relates to the number of channels of the class prediction vector. This can be informative to an extent, as the features captured by later network layers include a greater degree of complexity. However, they do not provide a broader view for the entire network. The second shortfall of Saliency Tubes is the fact that visually salient classes also display similar salient regions as they share the majority of class feature activations. Distinctions between the two classes are mostly present over a small number of activations that are specific to each class. However, these cannot be visualised without a greater exploration of features across the entire network architecture. 

% Extension to Class Feature Pyramids
In our proposed extension of Saliency Tubes, we aim to address the issues of visualising class-specific features of different convolutional layers. Our Class Feature Pyramids (CFP) enable a hierarchical traversal over network layers motivated by their linear process of kernel application over activation maps. Through our method we reverse this process to enable the propagation of information over the network from the prediction layers towards earlier network layers. We term this approach back-step as it corresponds to the discovery of earlier layer kernels that produce feature maps associated with a specific class. 

\subsection{Class feature association}
\label{ch:7::sec:cfp::formulation}

\begin{figure}[t]
\centering
\includegraphics[width=\textwidth]{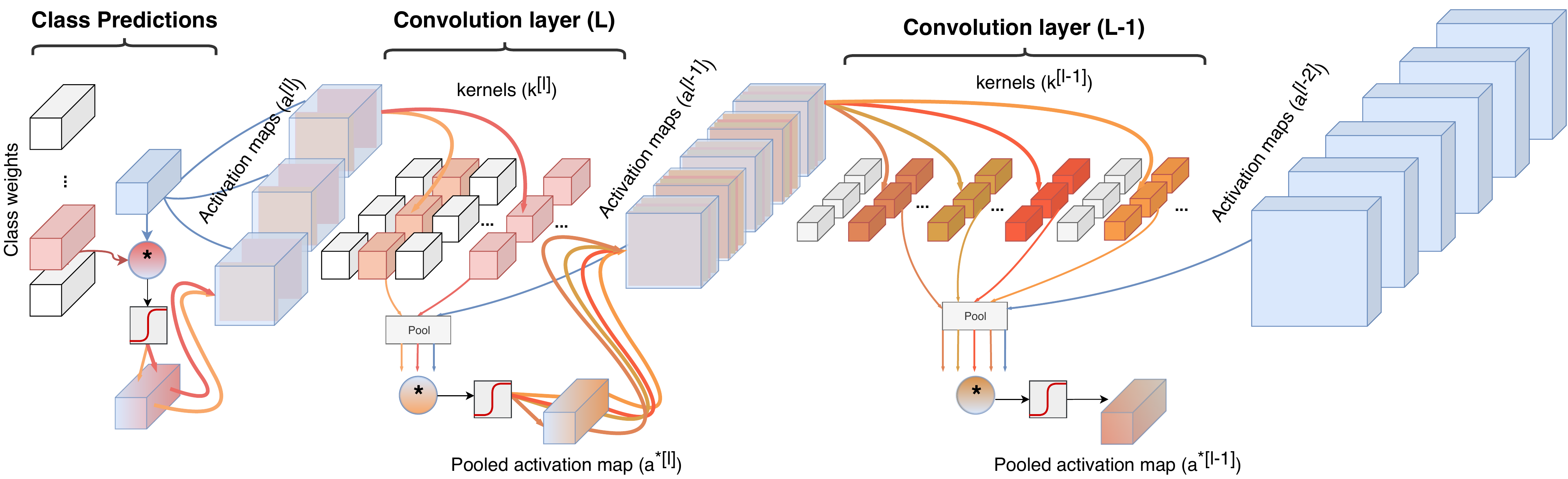}
\caption{\textbf{Back-step process.} Averaged pooled versions of layer weights ($\overline{\textbf{k'}}^{[l]}$) and their input activation maps ($\overline{\textbf{a}}^{[l-1]}$) are utilised for the channel-wise creation of global space-time vector representation. Through this, the locality of convolutions is alleviated. With the element-wise multiplication ($\otimes$) of the kernels and activations, pooled class-based activation maps ($\textbf{a}^{\star[l]}$) are created that only correspond the selected features. The highest feature activations that relate to the class are selected through a sigmoid applied over the volume. The same process is applied iteratively to discover each high activation in the current layer.}
\label{fig:backstep}
\end{figure}

% Formulation
We first consider the class weights of the final prediction layer ($\textbf{w}^{[p]}$) with ($[p]$) denoting the prediction layer. In order to identify the maximum probability class ($c$) we apply a standard softmax activation and select the corresponding class-weight vector ($\textbf{w}^{[p]}_{c}$). This can be seen as the right-side red coloured tensor in \Cref{fig:backstep}. The discovery of individual effects that each feature of the class weight vector has over the activation map ($\textbf{a}^{[l]}$) of the final convolution layer ($[l]$) is performed similarly to \Cref{eq:saliency}. This is achieved with the multiplication of each class weight vector channel and activation map. Based on this, we create a class-based activation map ($\textbf{a}^{\star[p]}_{c}$) that has the same dimensionality as the initial input activation map. However, each of the channels represents the relationship between channel $C'$ and the selected class $c$:

\begin{equation}
\label{eq:a_star}
\textbf{a}^{\star[p]}_{c} \overset{(\ref{eq:saliency})}{=\joinrel=} \{\textbf{z}_{c \rightarrow j}\} \: \forall j \in C'
\end{equation}

% Sigmoidial threshold
With the aggregation of class-based normalised information in $\textbf{a}^{\star[p]}_{c}$, we then explore the channel-wise dependencies of the class and features extracted in a specific layer. This is done to address probabilistic distributions of small and broader scales. In contrast to the one-hot selection of classes, multiple features affect a specific class, thus the selection process needs to include multiple values. To this end, we apply a monotonic shifted logistic sigmoid function with a user-defined threshold value ($\theta$):

\begin{equation}
\label{eq:thres_sigmoid}
\textbf{feats}_{i} = \{i \:: \: \textbf{F}^{[p]}_{i}>0 \}\: where \: \mathbf{F}^{[p]} = \frac{1}{1+e^{-(\textbf{a}^{\star[p]}_{c}+\theta)}}
\end{equation}

% Association of activations and previous layer kernels.
The discovered most dominant feature indices ($\textbf{feats}_{i}$) are identified from \Cref{eq:thres_sigmoid}. These indices have a direct correspondence to the feature locations from the previous layer ($[l]$) outputs, which in turn also relate to respective kernels ($\textbf{k}^{[l]}$). Through this sigmodial selection process, the relationship of each channel in the activation volume $\textbf{a}^{\star[p]}$ and kernels from previous layer ($[l]$) can be directly discovered. The influence of each of the kernels in the layer ($\textbf{k}^{[l]}$) to the final class prediction is based on the aforementioned logistic function which is visualised as the orange cross-layer connections in \Cref{fig:backstep}.

\subsection{Layer dependencies}
\label{ch:7::sec:cfp::layer_dependencies}

% Activations associations cross multiple layers
In comparison to propagating class-related information directly, the propagation of class-related information through corresponding features in the previous layer includes some difficulties. As the spatio-temporal extent of activations changes across layers alongside it's number of channels, a direct correlation measure is not straightforwardly achievable. This is associated with the curse of dimensionality problem as high-dimensional signals (e.g., deeper layer activations) cannot be linearly mapped to lower-dimensional spaces (e.g., earlier network layers) and vice versa. Even without considering the spatio-temporal size difference between activations from different layers, their convolution operations are followed by non-linearities and with that, the problem of representing cross-layer information also persists. In addition, the hierarchical architecture of CNNs includes a strict locality of the operations performed in each layer. As the input volume decreases in spatio-temporal size, the space-time patches that are used by each set of kernels increase in size, making the discovery of a cross-layer relationship difficult.

% Addressing these issues in Class Feature Pyramids
Class Feature Pyramids deal with the above issues by considering class information in a global manner, for the entirety of the activation maps. The localities of kernels, as well as their respective activations, are transformed to include accumulated information over the entire spatio-temporal regions. Specifically, given the discovered feature indices $\textbf{feats}_{i}$ from \Cref{eq:thres_sigmoid} and in order to back-step from layer $[l]$ to $[l-1]$, we first select the respective kernels ($\textbf{k}'^{[l]}$). Channels ($C$) for layer $[l]$ are denoted with $C^{[l]}$. Feature traversal is then performed based on the average pooled version of activations from the previous layer ($\overline{\textbf{a}}^{[l-1]}$) and selected layer kernels ($\overline{\textbf{k'}}^{[l]}$):

\begin{align}
\textbf{a}^{\star[l]} = \overline{\textbf{a}}^{[l-1]} \otimes \overline{\textbf{w}}^{[l]}_{j}, \: \forall \: j \in \textbf{feats}_{i} \qquad \quad \label{eq:feature_association} \\
where, \: \overline{\textbf{a}}^{[l-1]} = \mathit{Pool}(\textbf{a}^{[l-1]})\: and \:  \overline{\textbf{k'}}^{[l]} = \mathit{Pool}(\textbf{k'}^{[l]}) \label{eq:feature_weight_pool}
\end{align}

% Feature correspondence
Through \Cref{eq:feature_association} we create a relationship between class activation maps ($\textbf{a}^{\star[l]}$) of layer $[l]$ and the previous layer's ($[l-1]$) activations ($\textbf{a}^{[l-1]}$) with the absence of feature locality. Our approach differs from the standard convolutional procedure where created activations use the dot product of the previous layer activations and the current layer kernels. By replacing the dot-product with a single element-wise multiplication, we can solve the channel-dimensionality reduction problem and discover cross-layer feature dependencies. The main functions of \Cref{eq:a_star,eq:thres_sigmoid,eq:feature_association} are used for every adjacent layer pair until a user-specified \textit{back-stepping} depth that is searched. Previous layer activations ($\textbf{a}^{[l-1]}$) then follow the same normalisation as in \Cref{eq:saliency,eq:a_star} to produce a kernel-based activation map ($\textbf{a}^{\star[l-1]}$). The threshold ($\theta$) and feature indices selection ($\textbf{feats}_{i}$) processes remain as previously defined.

\subsection{Feature and layer-wise associations}
\label{ch:7::sec:cfp::layer_feat_association}

% Differences of feature and layer-wise back-stepping
An elemental part of a network's explainability is the understanding of the effect of kernels ($k^{[l]}$) at layer ($[l]$), individually or in a group, to previous layer ($[l-1]$) feature activations. These requirements become detrimental in earlier layers of lower complexity as specific activation maps have multiple strong associations with feature extractors in deeper layers. This increase in meaningful connections in earlier layers prevents the creation of an overall coherent channel-wise dependency graph. For this reason, and in order to ensure that an intuitive measure for channel selection is used during the back-step process, channel relationships in different layers can be explored either feature- or layer-wise. In the feature-wise approach, the correspondence of singular channels over the previous layer activations are explored. In the layer-wise process all discovered informative channels of a layer are grouped together and their averaged relationship to the previous layer studied.

% Feature-wise
\textbf{Feature-wise relevance back-step}. Individually, kernels from layer ($[l]$) are discovered based on the list of informative feature indices ($\textbf{feats}_{i}$) from \Cref{eq:thres_sigmoid}. Each kernel can correspond to a specific activation map from \Cref{eq:a_star} (($\textbf{a}^{\star[l]}_{k},$ where $k \in \textbf{k'}^{[l]}_{j}$)). The \textit{back-step} process is then applied individually for each of the features ($k$) and, respectively, the maximum corresponding channels are selected. A possible limitation of using a feature-wise propagation approach, especially in networks with a large number of channels, is the overall feature indices duplication. As features are explored individually, channels in the previous layers are more likely to form multiple strong connections with multiple features of the preceding layer. For this reason, feature-wise relevance is better suited for cases where the \textit{back-step} depth remains small.

% Layer-wise
\textbf{Layer-wise relevance back-step}. Similarly to the singular kernel-oriented approach, layer-wise feature exploration is performed based on the list of informative features ($\textbf{feats}_{i}$). However, instead of a per-feature activation map, a concatenated version of the activations is created. Apart from limiting the number of feature duplications on the \textit{back-step} process with the compact feature representation, layer-wise propagation includes an additional advantage. Kernel selection in the previous layer is done based on the average of multiple features, providing a more robust representation in terms of the region that a layer finds salient. In addition, as multiple features are considered at once (many-to-many), the activations in the following layer also present higher values than those discovered by singular features (one-to-many). 

\begin{figure}[ht]
\centering
\includegraphics[width=.7\textwidth]{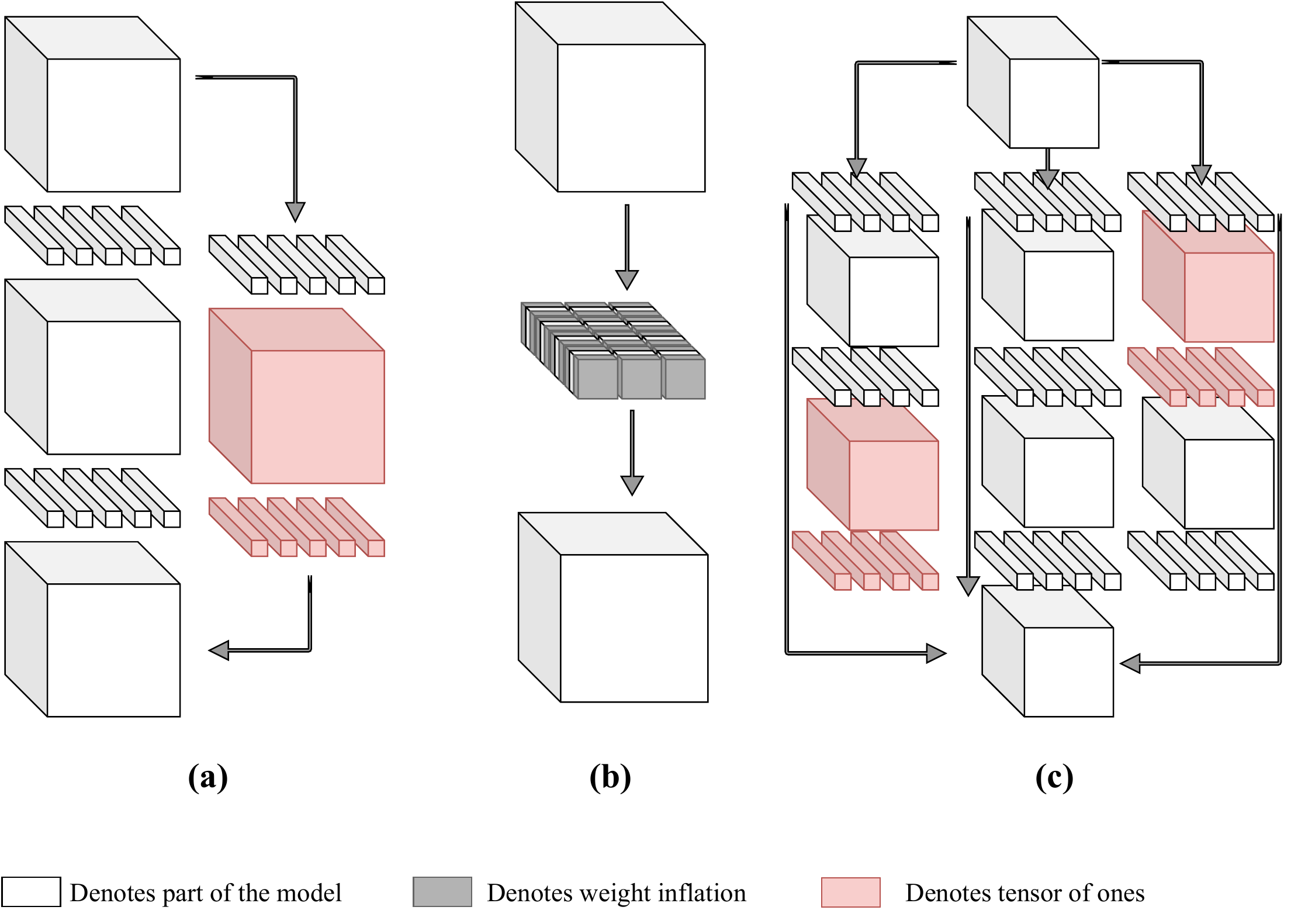}
\caption{\textbf{Back-step on different convolution block types.} (a) \textbf{Residual connections} \cite{he2016deep}, created activations and kernels of ones are in \textcolor{red}{red}. (b) \textbf{Grouped convolutions} \cite{xie2017aggregated}, inflated kernels are in \textcolor{gray}{grey}. (c) \textbf{Convolutions in branches} \cite{szegedy2015going}. The branch with the maximum number of convolutions is selected as the base with tensors of ones added to the other branches.}
\label{fig:cfp_conv_cases}
\end{figure}

\subsection{Addressing convolutional block structures}
\label{ch:7::sec:cfp::conv_layers}

% Overview
As CNNs are architecturally different in terms of their blocks, we extend our method to address the traversal within popular block variants. We discuss how back-step can be used over convolutions applied in activations in parallel and with varying kernel and channel dimensions performed at the same time. Our extension focuses towards the inclusion of these three convolution processes which exhibit a high degree of complexity in terms of how information is connected and how convolutions are performed. We provide an illustration of our approaches in \Cref{fig:cfp_conv_cases}.

% Residual connections
\textbf{Residual connections}. One of the most widely used architectural building blocks in CNNs is based on the use of residual connections \cite{he2016deep}. These blocks are also included in a large number of works by the video recognition community e.g., \cite{ diba2018spatio,feichtenhofer2019slowfast,feichtenhofer2020x3d,hara2018can,qiu2019learning,tran2018closer}. In the bottleneck variant, \textit{back-step} is more complex as information is divided between two paths. Subsequently, this does not permit a hierarchical discovery of feature dependencies across the network structure. Our adaptation to the current method for such cases is the inclusion of one-valued weight and activation tensors. Value of one is preferred given its multiplication property and our use of element-wise multiplication ($\otimes$). Through this, all discovered feature indices in the previous layer with high activations are passed to the next layer directly. Therefore, activation indices are shared between previous network layers in the residual branch in the same manner as those in the main pathway of the block.

% Cross-Channel convolutions
\textbf{Grouped convolutions}. Convolutions can also be performed over groups of features \cite{xie2017aggregated}. Works such as \textit{Channel Separated Networks} (CSN) \cite{tran2019video} use grouped 3D convolutions for action recognition. These convolutional types demonstrate significant challenges during \textit{back-step} as their kernel sizes are composed of a sub-set of channels. This does not allow for an immediate correspondence between the kernel channels and the entire activation volume from the previous layer. We deal with this by explicitly inflating each of the grouped kernels to the same dimensional space as the activation maps by using values of zero to cancel out any effect. We can thus simulate the channel-wise convolution process in the channel dimensional space of the input.

% Multi-streams of activations
\textbf{Convolutions in branches}. This block decouples information into multiple branches \cite{szegedy2015going}. The branch-based approach has also been extended with success for video classification e.g. \cite{ carreira2017quo,chen2018multifiber,feichtenhofer2019slowfast,stergiou2020learning,wang2018action}. The main mindset is the application of convolution or pooling operations within the block, over multiple pathways originating from a single activation map. Variations to the type of operations and their number add an additional degree of ambiguity for constructing such blocks in a hierarchical manner. In these cases, back-step through branches and pathways are accomplished with the addition of kernels with values of one and activation maps that act as small sub-structures that pad all branches to the same length.

\subsection{Inference calculation}
\label{ch:7::sec:cfp::inference}

As Class Feature Pyramids are based on layer traversal, we study the times required by architecturally different spatio-temporal networks to traverse over the network and identify the kernels that produce the highest activations for a specific class.

% Times obtained
We report inference times on nine architectures. Our evaluation includes the number of GFLOPs alongside the total running times for feature discovery. We summarise the results \Cref{table:cfp_inference}. The choice of threshold values ($\theta$) is empirically discovered based on the produced number of informative features for each of the networks. Apart from the network architecture and the threshold value, latency also depends on the number of layers that class features are back-stepped to.

\begin{table}[htb]
\begin{center}
\resizebox{0.75\textwidth}{!}{%
\begin{tabular}{l|c|c|c|c}
\hline
Network & GFLOPS & Back-step time (msec) & \# layers & $\theta$ \\
\hline
Multi-FiberNet \cite{chen2018multifiber} & 22.70 & 24.43 & 3 & 0.6 \\
\hline
I3D \cite{carreira2017quo} & 55.79 & 23.21 & 1 + mixed5c & 0.65 \\
\hline
ResNet50-3D \cite{hara2018can} & 80.32 & 21.39 & 3 & 0.55 \\
\hline
ResNet101-3D \cite{hara2018can} & 110.98 & 39.48 & 3 & 0.6 \\
\hline
ResNet152-3D \cite{hara2018can} & 148.91 & 31.06 & 3 & 0.6 \\
\hline
ResNeXt101-3D \cite{hara2018can} & 76.96 & 70.49 & 3 & 0.6 \\
\hline
MTNet$_S$ \cite{stergiou2020learning} & 11.30 & 26.32 & 3 & 0.67 \\
\hline
MTNet$_M$ \cite{stergiou2020learning} & 14.12 & 32.17 & 3 & 0.67 \\
\hline
MTNet$_L$ \cite{stergiou2020learning} & 20.82 & 34.65 & 3 & 0.67 \\
\hline
\end{tabular}%
}
\end{center}
\caption[]{\textbf{Inference times and GFLOPs}. Threshold value ($\theta$) is determined by the model complexity. All architectures are back-stepped for three layers. For I3D, this corresponds to the class filters and the last \textit{mixed} block.}
\label{table:cfp_inference}
\end{table}

As shown in \Cref{table:cfp_inference}, the proposed method can traverse across layers at reasonable speeds without significant additional operations over normal forward information passes. In most cases, times are similar or faster to those of simple feed-forward operations. This is primarily attributed to the speed-up gained by using a global representation instead of the dot product of local convolutional operations.

\clearpage

\begin{figure}[h]
\centering
\includegraphics[width=.85\textwidth]{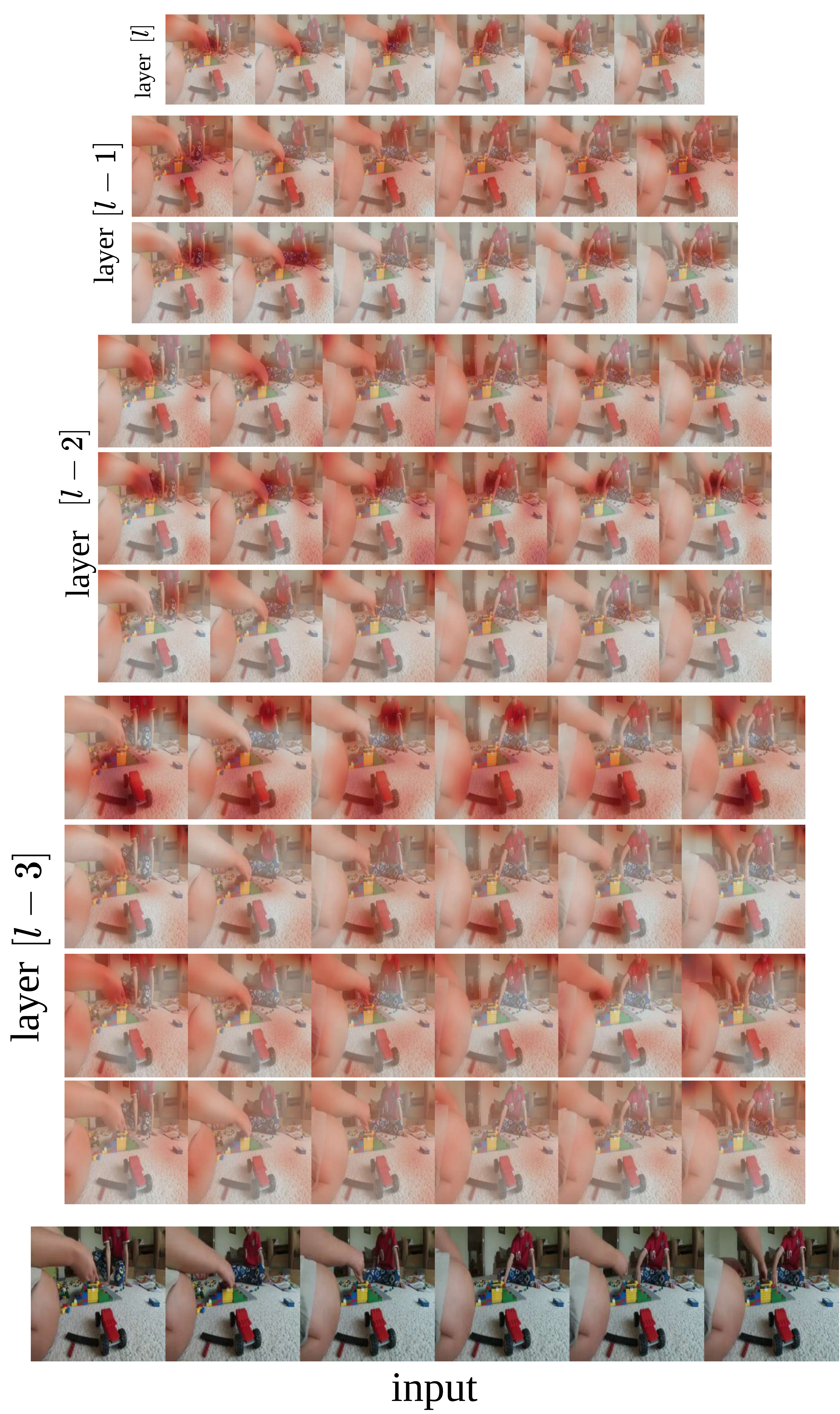}
\caption{\textbf{Kinetics \textit{\enquote{building lego}} class.} MTNet$_L$ \cite{stergiou2021refining} is used to produce the visualisations. }
\label{fig:cfp_lego}
\end{figure}

\clearpage

\begin{figure}[h]
\centering
\includegraphics[width=.85\textwidth]{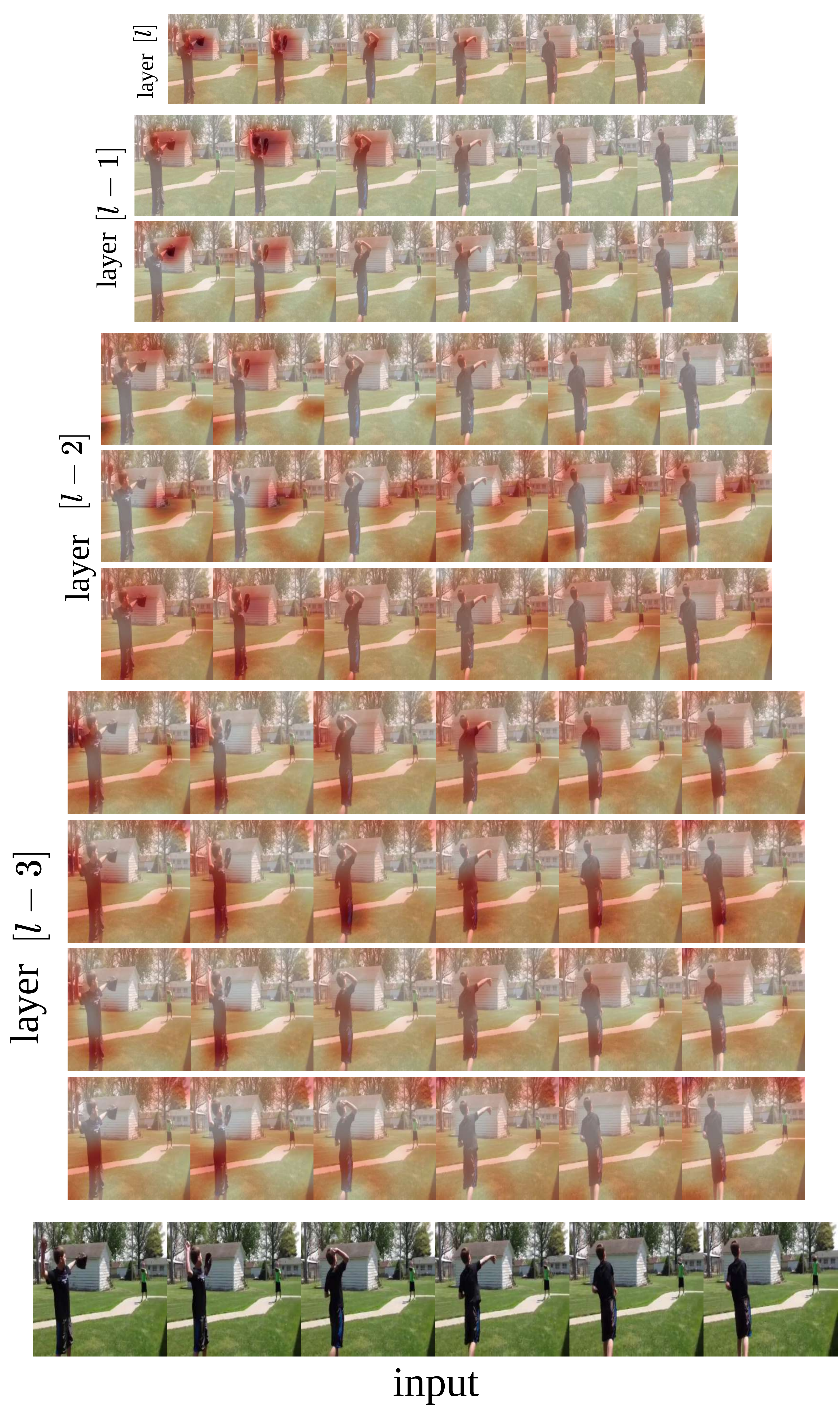}
\caption{\textbf{Kinetics \textit{\enquote{throwing or catching baseball}} class.} MTNet$_L$ \cite{stergiou2021refining} is used to produce the visualisations. }
\label{fig:cfp_baseball}
\end{figure}

\subsection{Visualisation results}
\label{ch:7::sec:cfp::visualisations}

In this section we visualise the resulting salient regions for different layer features of MTNet$_L$ based on feature-wise relevance \textit{back-step}. Visualisations in \Cref{fig:cfp_lego,fig:cfp_baseball} have been produced with the same threshold value as in \Cref{table:cfp_inference}. Additional results on layer features based on the layer-wise approach are shown in \Cref{fig:CFP_nets} where we demonstrate the concatenated salient features of the final three layers of five architectures. 

Based on the produced layer-wise feature activation visualisations in \Cref{fig:CFP_nets}, for the Kinetics-700 \textit{\enquote{playing field hockey}} class, salient regions are shown to vary significantly between different networks. Given this, layers that incorporate temporal information of different scales within their architecture, such as SlowFast \cite{feichtenhofer2019slowfast} and MTNets \cite{stergiou2020learning}, focus on more specific spatio-temporal regions. Evident by the salient regions of MTNet$_L$ layers in \Cref{fig:CFP_mtnetl}, the network demonstrates that an association is created between the field hockey ball, the ball's movement in the field and the players in the scene with the specific class. Through this hierarchical representation of information, we can further visualise the layers where associations to appearance cues are made. For example, both layers $[l-2]$ of MFNet and SlowFast in \Cref{fig:CFP_mfnet,fig:CFP_sf50} show that the field's grass also attributes to the class prediction if in comparison to the \textit{\enquote{ice hockey}} class.

\section{Discussion}
\label{ch:7::sec:discussion}

% Scope of the section
Research in the domain of human action recognition has seen tremendous progress with the introduction of spatio-temporal (3D) convolutions. Still, as the complexity of CNN architectures increases, so does the demand to understand the strength and limitations of these networks in terms of the spatio-temporal patterns that they can model. In this section, we have proposed a method that can provide visual interpretations of class-based features of 3D CNNs and overviewed its extension for features over multiple layers.

% Saliency Tubes
The representation of salient regions based on extracted features has been a widely popularised method in image-based CNNs \cite{selvaraju2017grad,zhou2016learning}. Their main role is to provide a visual description of the regions which causes a specific feature activation. In our proposed Saliency Tubes, we adapt this method in order to create such representations over space-time volumes. Through the interpolated class-based activations, the most informative spatio-temporal regions in the input can be presented over the original input clip. However, similar to other activation-based visualisation methods, the representations are only specific to the features based on which class predictions are made. This does not provide a higher understanding of the entire feature extraction process within the entire network.

% Class Feature Pyramids
We address these issues through an extension of Saliency Tubes to visualise class features over previous layers. The proposed method named Class Feature Pyramids can hierarchically traverse a network backwards through the discovery of feature dependencies across adjacent layers. This process is termed back-step and enables the exploration of features in any network depth. The proposed method is applicable to various convolutional blocks through invariance with implementations that can be used for residual connections \cite{he2016deep}, channel-separated (grouped) convolutions \cite{xie2017aggregated}, and convolutions over branches \cite{szegedy2015going}. 

% In conclusion
We believe that the current limited research on the explainability of 3D-CNNs restricts a high-level understanding of the feature types that are learned. Our proposed methods aim to uncover features, providing a level of transparency that can aid the comprehension of extracted 3D-CNN patterns. We believe that visual interpretation methods can significantly benefit the action recognition field and uncover new research directions.    

\clearpage

\begin{figure}[h]
    \centering
    \begin{subfigure}{\textwidth}
        \begin{minipage}{.15\textwidth}
            \centering
            layer $[l]$\\
            \vspace{1.8cm}
            layer $[l-1]$\\
            \vspace{1.8cm}
            layer $[l-2]$
        \end{minipage}%
        \begin{minipage}{.85\textwidth}
            \includegraphics[width=\linewidth]{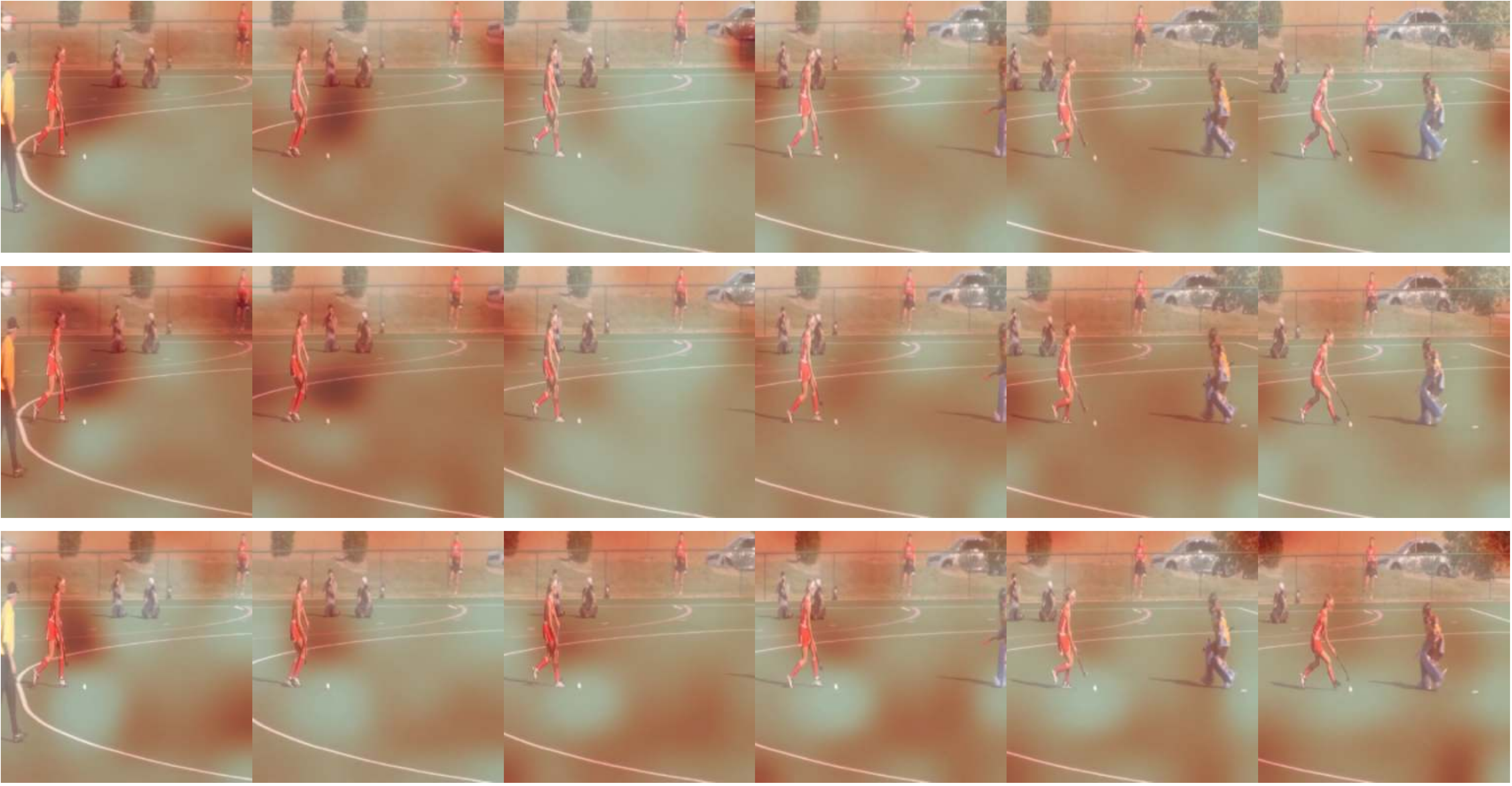}
        \end{minipage}
        \subcaption{R3D-50}
        \label{fig:CFP_r3d}
    \end{subfigure}
\medskip
    \begin{subfigure}{\textwidth}
        \begin{minipage}{.15\textwidth}
            \centering
            layer $[l]$\\
            \vspace{1.8cm}
            layer $[l-1]$\\
            \vspace{1.8cm}
            layer $[l-2]$
        \end{minipage}%
        \begin{minipage}{.85\textwidth}
            \includegraphics[width=\linewidth]{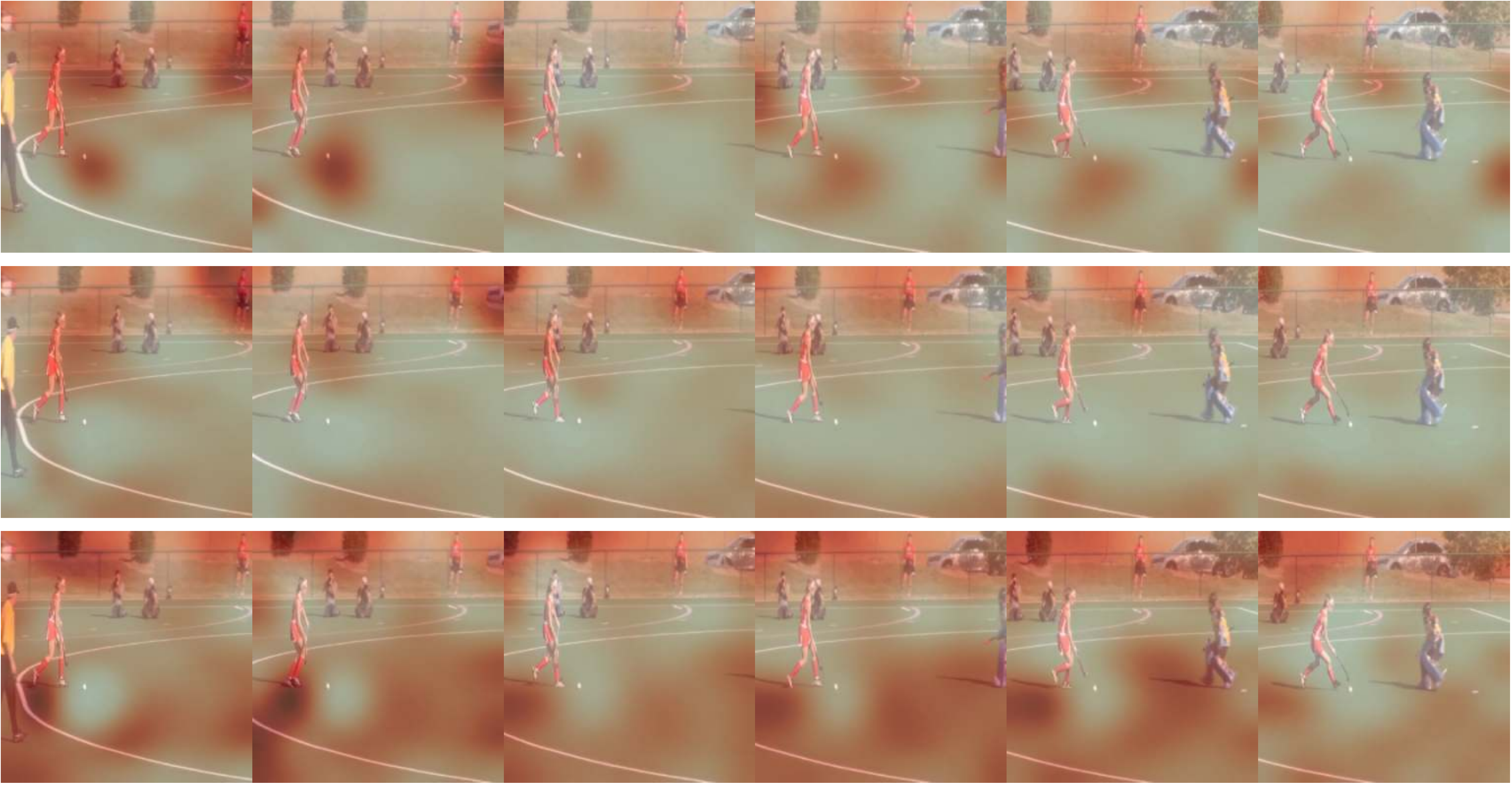}
        \end{minipage}
        \subcaption{R(2+1)D-50}
        \label{fig:CFP_2plus1d}
    \end{subfigure}
\medskip
    \begin{subfigure}{\textwidth}
        \begin{minipage}{.15\textwidth}
            \centering
            layer $[l]$\\
            \vspace{1.8cm}
            layer $[l-1]$\\
            \vspace{1.8cm}
            layer $[l-2]$
        \end{minipage}%
        \begin{minipage}{.85\textwidth}
            \includegraphics[width=\linewidth]{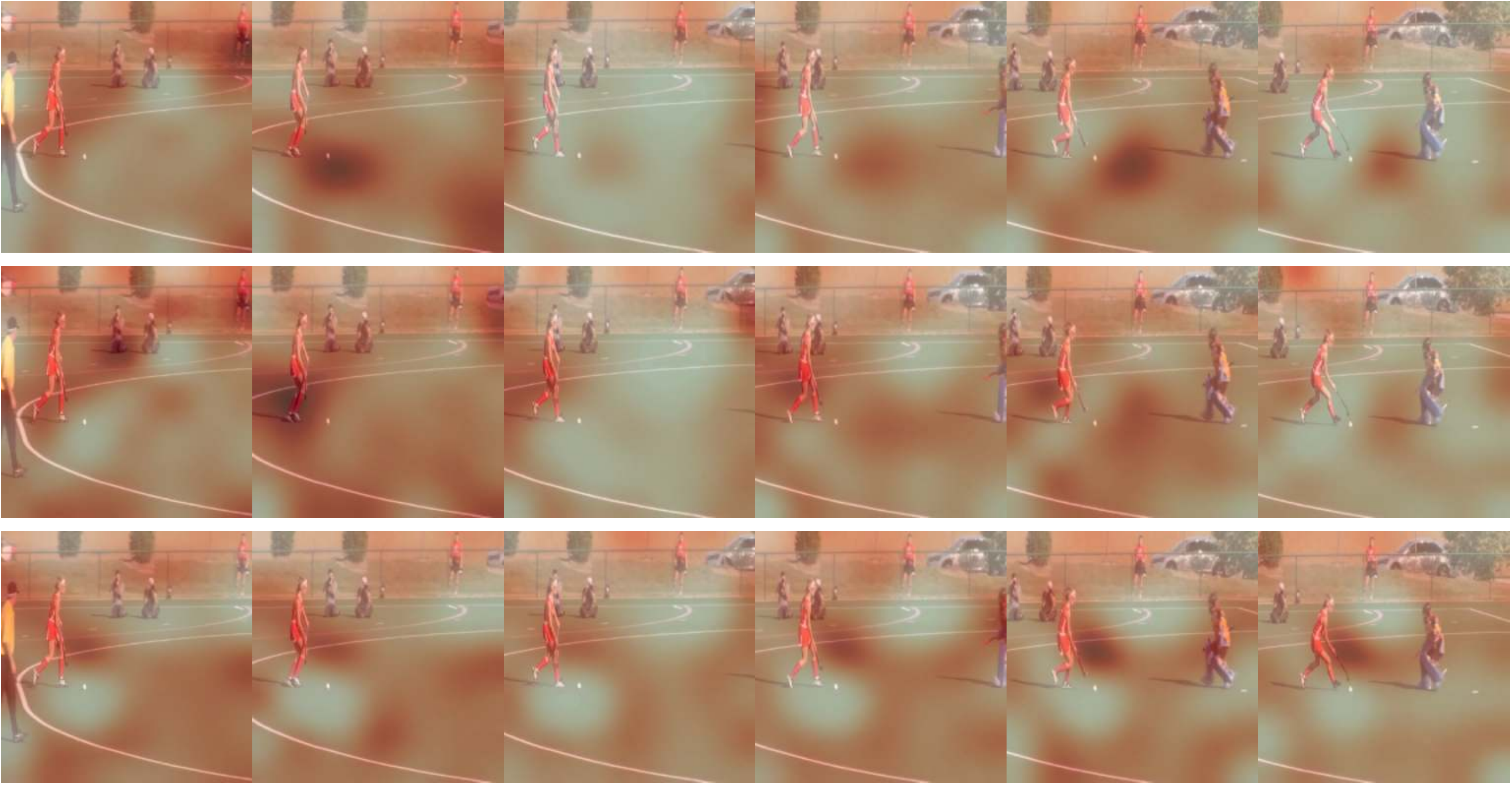}
        \end{minipage}
        \subcaption{I3D}
        \label{fig:CFP_i3d}
    \end{subfigure}
\caption{\textbf{Class Feature Pyramids visualisations}.}
\end{figure}%

\begin{figure}[h]\ContinuedFloat
    \begin{subfigure}{\textwidth}
        \begin{minipage}{.15\textwidth}
            \centering
            layer $[l]$\\
            \vspace{1.8cm}
            layer $[l-1]$\\
            \vspace{1.8cm}
            layer $[l-2]$
        \end{minipage}%
        \begin{minipage}{.85\textwidth}
            \includegraphics[width=\linewidth]{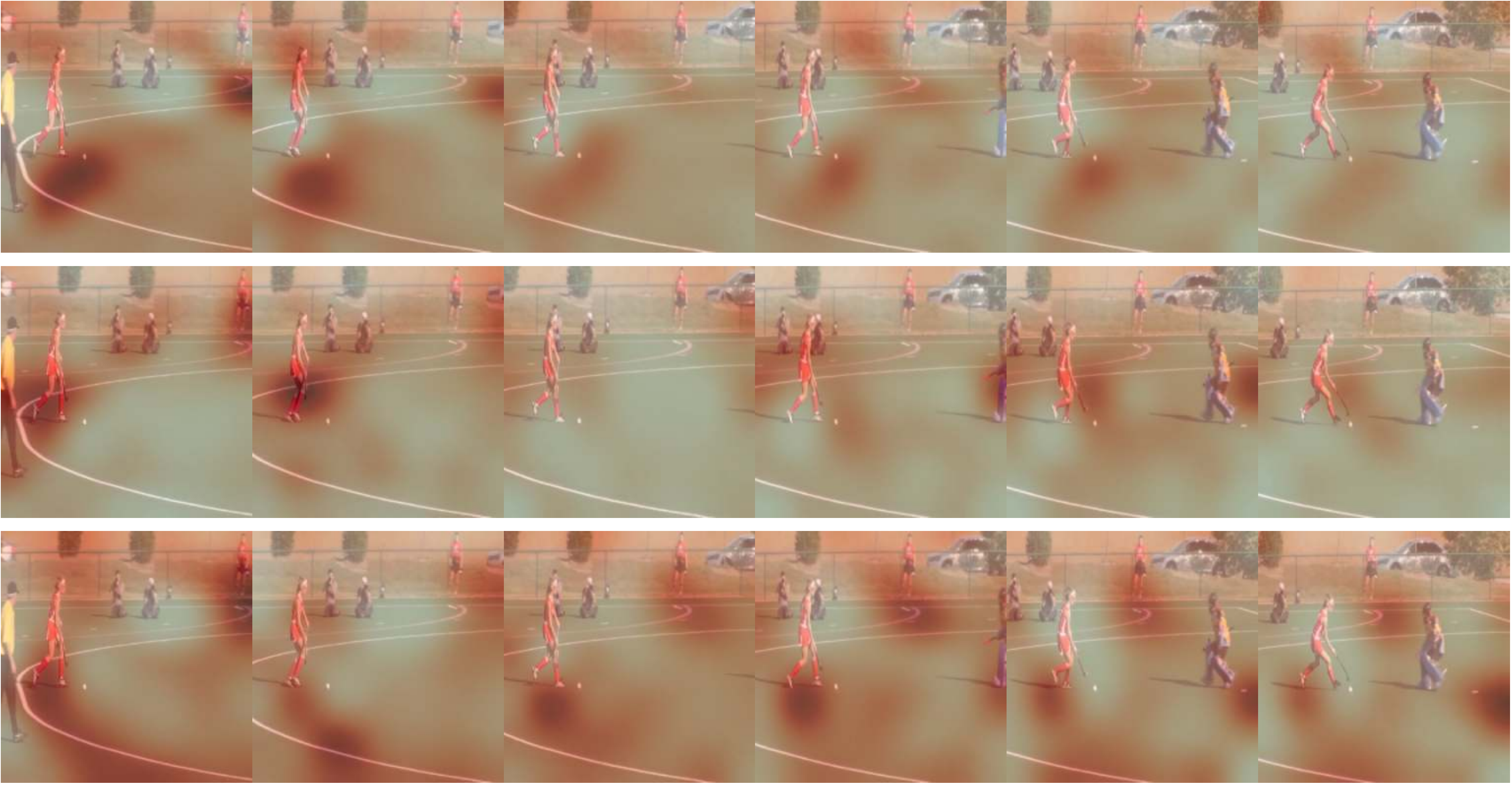}
        \end{minipage}
        \subcaption{MFNet}
        \label{fig:CFP_mfnet}
    \end{subfigure}
\medskip
    \begin{subfigure}{\textwidth}
        \begin{minipage}{.15\textwidth}
            \centering
            layer $[l]$\\
            \vspace{1.8cm}
            layer $[l-1]$\\
            \vspace{1.8cm}
            layer $[l-2]$
        \end{minipage}%
        \begin{minipage}{.85\textwidth}
            \includegraphics[width=\linewidth]{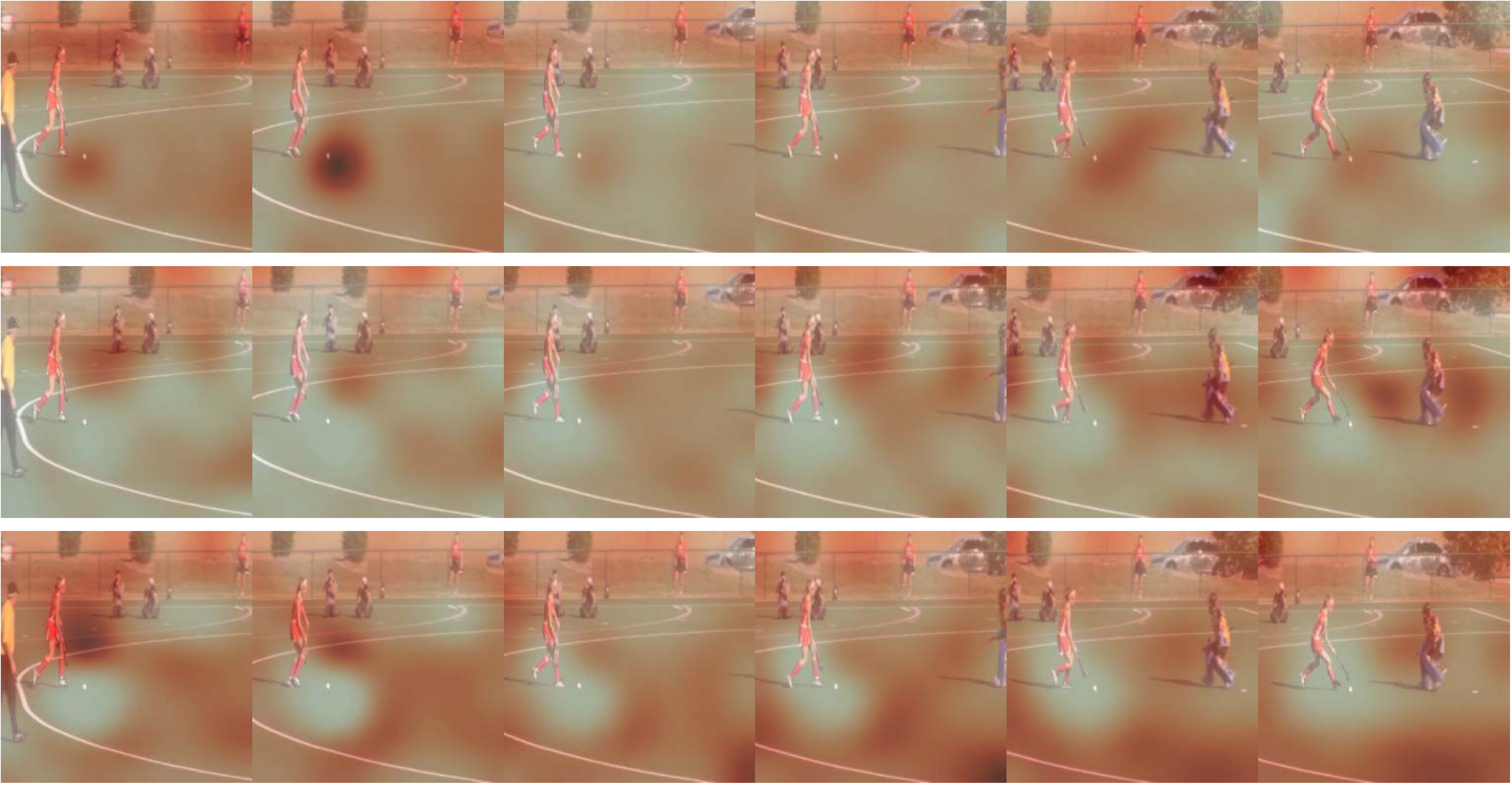}
        \end{minipage}
        \subcaption{SlowFast 50}
        \label{fig:CFP_sf50}
    \end{subfigure}
\medskip
    \begin{subfigure}{\textwidth}
        \begin{minipage}{.15\textwidth}
            \centering
            layer $[l]$\\
            \vspace{1.8cm}
            layer $[l-1]$\\
            \vspace{1.8cm}
            layer $[l-2]$
        \end{minipage}%
        \begin{minipage}{.85\textwidth}
            \includegraphics[width=\linewidth]{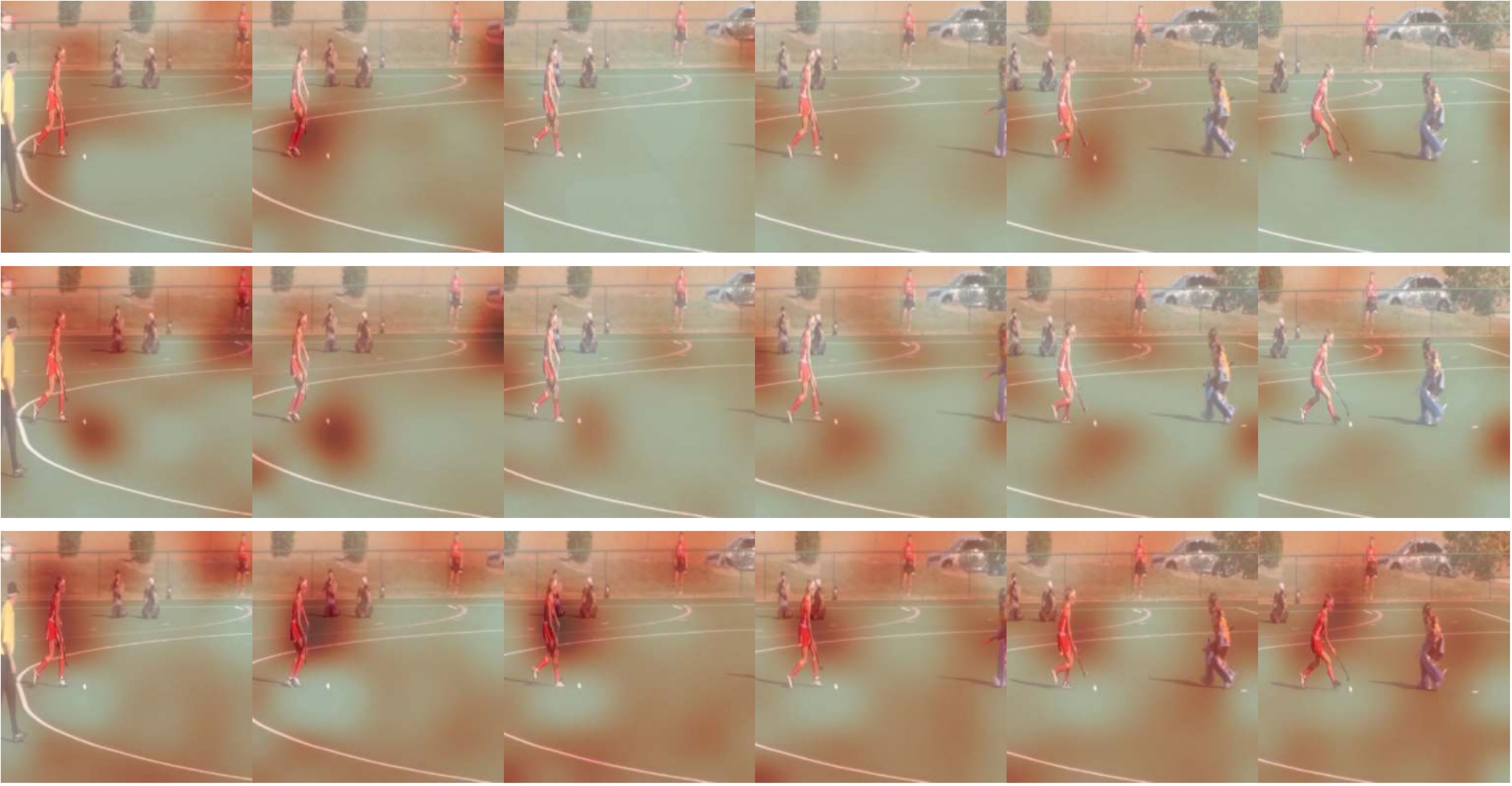}
        \end{minipage}
        \subcaption{MTNet$_L$}
        \label{fig:CFP_mtnetl}
    \end{subfigure}
\caption{\textbf{Class Feature Pyramids visualisations}.}
\end{figure}%

\begin{figure}[t]\ContinuedFloat
    \begin{subfigure}{\textwidth}
        \includegraphics[width=\linewidth]{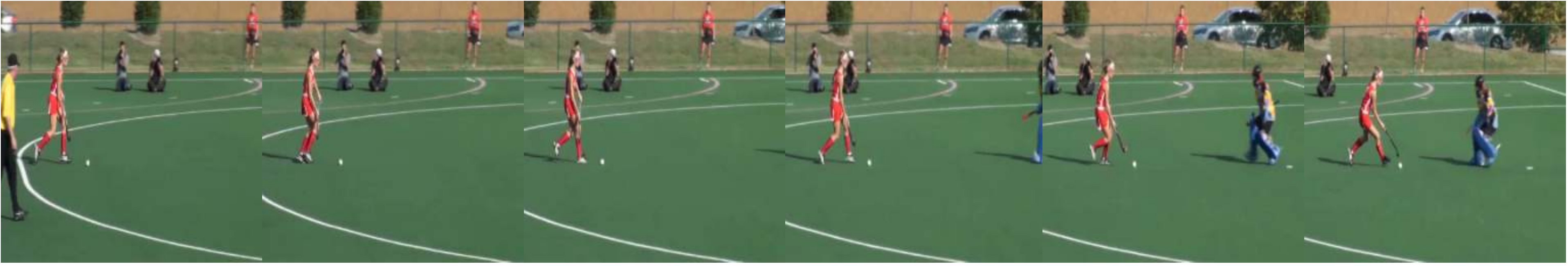}
        \subcaption{Original clip}
        \label{fig:CFP_input}
    \end{subfigure}
\caption[]{\textbf{Class Feature Pyramids visualisations}. Each model architecture is denoted by the respective sub-caption. Rows represent the layer-wise visualisations produced for each layer. The sample is drawn from \textit{\enquote{playing field hockey}} from Kinetics-700. The theta ($\theta$) values for each of the presented networks are [0.6, 0.6, 0.7, 0.65, 0.65, 0.67]. Layer $[l]$ denotes the final layer for each of the architectures.}
\label{fig:CFP_nets}
\end{figure}

\stopcontents[chapters]
    
    \author
{}
\title{Discussion and Future Research Directions}

\maketitle
\label{ch8}

% Our contribution
In this thesis, we have investigated the variance in human action performance. In particular, we have addressed the temporal variations from different performances of actions through the extraction and fusion of features over different spatio-temporal sizes (\Cref{ch4,ch5}). We additionally studied the hierarchical correlation between classes and extracted features. We have enforced this connection between features and classes, through the regularisation of features based on their correspondence to the target class in \cref{ch6}. We finally provided visual interpretation methods for a qualitative assessment of the features and the attention regions that are found to be descriptive of specific action sequences in \cref{ch7}.   

% Future work
In this final section we summarise the contributions of our works. We then address the most prominent aspects and directions that can motivate future works in the field of human action and interaction recognition.

\section{Summary}
\label{ch:8::sec:summary}

In this section we provide a summary for each of the works that we have presented in the chapters of this thesis. 

\subsection{Squeeze and Recursion Temporal Gates}
\label{ch:8::sec:summary::sub:srtg}

In \Cref{ch4} we have studied the problem of temporal variations in human actions. As there is no typical duration length or number of frames for an action, the heavy dependence of 3D kernels in CNNs to extract temporally fixed-sized features can be improved. We have created an alignment between patterns of short motions and the movements that are performed over the entire video. The proposed SRTG method can model the relevant local feature information across the entire video through the use of recurrent cells. Recursion is performed by 2-layer LSTMs over the spatially averaged vector of feature activations. The fusion to the spatio-temporal activation volume is achieved based on the temporal cyclic-back consistency between the LSTM outputs, that include global motion information, and the vectorised convolutional features that were used. This process can achieve a degree of similarity between the purely local and global attention-based activations. Results on five benchmark datasets have shown notable improvements between SRTG and non-SRTG architectures, while the effects of including SRTG in terms of the computational complexity are minimal.  

\subsection{Multi-Temporal convolutions}
\label{ch:8::sec:summary::sub:mtconv}

In \Cref{ch5} we have further worked on addressing temporal disparities in the performance of human actions. We showed how the sets of local motions do not necessarily correspond to or accurately describe the action that is performed. Through variations in the temporal size of these action sequences, better descriptions can be provided. We therefore proposed Multi-Temporal convolutions (MTConv) that discover feature descriptors of both short local patterns and spatio-temporal features of different durations. The proposed method utilises three branches. The local branch focuses on spatio-temporal patterns that are performed over short space-time windows similar to standard 3D convolutions. The prolonged branch models spatio-temporal features of extended durations and spatial sizes. This is achieved by sub-sampling the activation volume to decreased dimensions. Features of both branches are aligned in the global aggregated feature importance branch based on their temporal motion dynamics across the entire video. The branch uses Squeeze and Recursion to discover the temporal attention of features from the other two branches and capture the local feature dependencies within the global scale of the entire video. Based on these convolution blocks, we have introduced MTNets that include MTConvs in a X3D backbone. MTNets achieve comparable or, in many cases, higher classification accuracy than state-of-the-art models on widely used action recognition benchmark datasets. With MTNets a notable reduction in terms of computation costs can be achieved.

\subsection{Class Regularisation}
\label{ch:8::sec:summary::sub:classreg}

The work described in \Cref{ch6} addressed the discriminative nature of CNNs that is limited to the final fully-connected layers. We studied the effects of enforcing the propagation of class-specific activations throughout the network. We proposed a method named Class Regularisation that relates class information to extracted features through direct connections between the class prediction layer and the convolutional layers. Class Regularisation also benefits the non-linearities of the network by modelling the effects of activations. To avoid the vanishing gradient problem, and the possibility of negatively influencing activations, the weights are normalised in a range given an affection rate value. The method is evaluated on standard benchmark action recognition datasets and with six models demonstrating results with and without Class Regularisation. Models across all datasets showed improvements when including Class Regularisation with minimal additional computational costs over the original architectures. In addition, Class Regularisation can aid in the creation of explainable 3D CNNs. Qualitative visualisations reveal which spatio-temporal features are strongly correlated to specific classes. Such analysis can be made for specific layers and, as such, provide insight into the discriminative patterns that specific features represent.

\subsection{Spatio-temporal feature visualisation}
\label{ch:8::sec:summary::sub:vis}

In \Cref{ch7} we have overviewed the current limitations of research methods in explaining the spatio-temporal features of 3D CNNs. This is partly due to a scarcity of proper visualisation methods that specifically address the temporal dimension of space-time video data. For this we have proposed two methods. The first method, Saliency Tubes, visualises the feature activation of 3D CNNs with relation to a class of interest. Saliency Tubes were built upon 2D CNN methods relating to network interpretability, and extended for 3D convolution to represent salient spatio-temporal regions. These regions correspond to the most discriminative class features of the network. We then extended Saliency Tubes to Class Feature Pyramids (CFP) that capture and present hierarchically informative features over different layers. CFPs are independent of the network type and can be employed regardless of the type of 3D convolution operation. They additionally enable the visualisation of activations in layer-wise, group-wise or kernel-wise formats. Our method is therefore suitable to visualise and, consequently, to better understand what kinds of features are learned to identify a specific class.

\section{Limitations and Future directions}
\label{ch:8::sec:directions}

The past years have seen an exponential growth in the potential of video understanding systems and models for tasks, such as intelligent video indexing to smart surveillance. The inclusion of temporal information has shown great improvements in terms of performance for action recognition. However, in contrast to appearance features, movements are not specific to fixed durations or standardised sets of motions. Actions that are semantically similar may be visually very different, with the reverse of this also being possible. This means that, although the action class may remain the same, the perception and goals of the actor are bound to be different, thus producing highly varying results. Our works in \Cref{ch4,ch5} addressed these time-varying features from different executions as well as discovered relationships between the short motions that are performed. Based on this, we believe that subsequent research should aim towards three main directions. 

% Inclusion of global information in the local temporal modelling process
The first direction is to improve the modelling of temporal motion variations. As dataset sizes are growing, to improve upon the large level of diversity of real world data, the representation of complex motion patterns and sequences becomes even more crucial. One notable issue that arises from the task of increasing temporal modelling capabilities of networks is the impact in computational requirements. A key element that needs to be considered is temporal order. To encode temporal information within local patterns, information only from the preceding temporal locations affect the current frame. Causal convolutions \cite{oord2016wavenet} have developed this notion as part of their main functionality in which multiple layers with ascending dilation sizes are used to ensure temporal ordering. Similarly, non-local blocks \cite{wang2018non} were based on temporal attention to capture the temporal feature dependencies. In \Cref{ch4} we proposed an alignment of local features based on their importance. The study of local features within the global context of a video has shown to be a promising direction towards utilising feature relevance. An approach that can create spatio-temporal feature descriptors that encode information from preceding temporal locations is therefore favourable. Such approaches are also beneficial for online action recognition and action predictions, as early action predictions are feasible given that the dependency of the local features to the entire sequence is modelled explicitly.

% Temporal modelling as part of the architecture
Similarly, this dependency of fixed-sized features to the global motions in videos can be addressed architecturally. The approach in \Cref{ch5} of sacrificing part of the spatial feature resolution has shown promise. However, this requirement should not come at the cost of reduced spatial modelling capabilities but rather as a balanced approach of addressing temporal variation without a decline in spatial pattern extraction. A direction that could provide such a balancing approach could be through the use of reinforcement learning and evolutionary algorithms. As they aim at creating robust network architectures (e.g., \cite{zoph2017neural}). The reliance to appearance-based or temporally-based can be learned through optimisation within different parts of the network. The search space parameterisation can be also extended to the temporal kernel receptive fields or even the number of different temporal kernels that are used and how they are aggregated. As the temporal characteristics relating to the visual aspects are explored in a greater extent, we find such approaches promising.

% Action labelling and correspondance between features and classes
The second research direction is towards the finer-grained labelling of human actions and interactions. A key aspect in the performance of actions is how the actor(s) identify an action. Most of the current practices only consider an action as belonging to a single class. But human behaviour is often more open to subjectivity. Therefore, a less strict separation into classes could be beneficial for their generalisation. Works based on overlapping labels or behaviour hierarchies (e.g., \cite{frosst2017distilling,yeung2018every}) can facilitate the focus on distinctive patterns at different levels of granularity. The inclusion of semantic meaning of actions (e.g., \cite{damen2020rescaling,shao2020finegym}) has also shown great promise. Complex actions can be categorised to smaller and less complex sub-actions creating finer-grained labels. This can motivate the creation of larger datasets or the re-evaluation of current benchmark datasets.

Actions have been predominantly classified directly based on an input. Works have also considered semantic mid-level representations based on classes (e.g., \cite{lan2012discriminative, sefidgar2015discriminative,kong2012learning,tian2015deep}). These methods show that they can provide some invariance to the features learned and that they relate to a specific class. They can facilitate modelling of actions through spatio-temporal patterns of these mid-level features that are connected to classes. In \Cref{ch6} we have proposed a way of relating the extracted convolutional features to classes and modelling their importance as part of their activations. This attention to features that are specific to a class can be crucial to distinguishing between subtly different classes. Spatio-temporal feature representations should take into account the correlation of features to classes. The creation of either, a hierarchy of general-to-finer features in layers, or the early prediction of classes across different network depths, appear promising in that respect.  

\begin{figure}[ht]
\includegraphics[width=\textwidth]{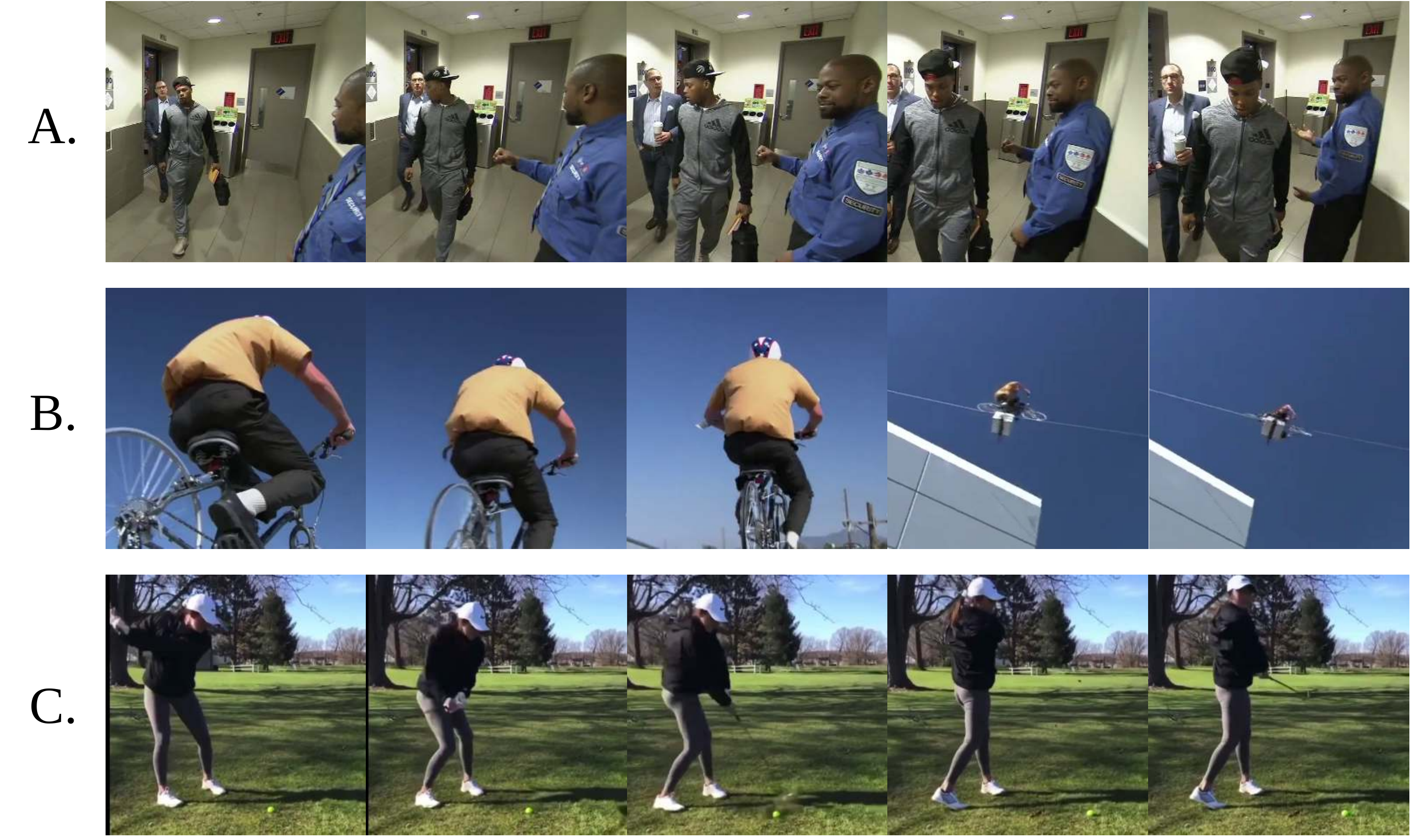}
\caption{\textbf{Ambiguous action and interaction examples}. Sequence A shows that there is no interaction between the two main actors in the scene. There is however an intent for a fist-bump. Sequence B shows action ambiguity in terms of the action label to describe the activity performed. tightroping while on a bicycle is not an action that can be (easily) standardised within a dataset. In sequence C, although a golf swing is performed, the actor misses the ball and thus the action is not performed in full.}
\label{figure:actexp}
\end{figure}

% Improved evaluation
Finally, we believe that the use of qualitative methods apart from accuracy should be adopted by the video understanding community. The use of a single score to determine the capabilities of the model limits the ability to determine the cases of features that are found to be more easy or challenging to classify. In order to understand and be able to address weaknesses of networks, uncovering the internal states of models has potential as an approach. Through the discovery of the patterns and features that are found to be descriptive, an assessment about the model's perception of the world can be made. Addressing skewed or incorrect perceptions is to be prioritised, since by understanding the moment-to-moment internal state, we are able to understand the effect of past states and possible future states that can be expected.  

% Detection
A direct application of using attention can be on the automated analysis of actions and interactions focusing on their detection. Understanding human behaviour can benefit from linking actors and the class that the action is assigned to. We believe that there is great potential on leveraging detection for the understanding of human behaviour when looking for deviations from common practices or anomalies in the sequence of movements performed. For example, sequences A and C in \cref{figure:actexp} show examples of interactions and actions in which their classification would be difficult as actor intent does not (necessarily) correspond to the performance of an action. In addition, using finite definitions of actions without incorporating the relations to people or objects can only describe a finite number of scenarios. Considering sequence B in \Cref{figure:actexp}, without a standard class specifically for tightroping whilst on a bicycle, the sequence can be easily misclassified. However, if the bicycle is detected and the action of tightroping is also discovered an association between the two can be learned instead. Based on this, a longer-term analysis of the actors and their roles or relations to other actors, as well as knowledge of the social and cultural norms, can help in providing an improved understanding of observed social behaviours. Understanding the intentions of a person can help to analyse what a person is doing rather than focusing on how that is achieved.

% closing
We have just scratched the surface when it comes to the automatic understanding of human actions and interactions. This thesis has provided targeted approaches for problems that have been identified, and proposed new promising directions of research to further address the current limitations. We believe that the presented work and the identified research directions are substantial motives for future works in the automatic understanding of human actions and interactions.
    
    \clearpage
\vspace{30em}
\printbibliography

\end{document}